\documentclass[a4paper,11pt,euler,numbered,print,index]{PhDThesisPSnPDF}

\input{preamble}

\title{Learning to compress and search visual data in large-scale systems}

\author{Sohrab Ferdowsi}

\supervisor{Prof. Slava Voloshynovskiy}

\jury{Prof C. Guillemot (INRIA - Rennes)\newline
Prof K. Eguiazarian (U. of Tampere)\newline
Dr. F. Fleuret (IDIAP - EPFL)\newline
Prof. T. Pun (U. of Geneva)\newline
Dr. S. Marchand-Maillet (U. of Geneva)
}

\degreetitle{``Docteur \`{E}s Sciences'' \\to the department of Computer Science of the University of Geneva\\Thesis number: 5295}

\begin{document}
\maketitle 
\thispagestyle{empty}
\begin{center}

\newcommand{\supervisors}{Professeur Sviatoslav Voloshynovskiy}

\begin{tabulary}{\textwidth}{@{} l p{1.0cm}  p{1.0cm} l @{}}
\large{UNIVERSIT\'E DE GEN\`EVE}  &&& \large{FACULT\'E DES SCIENCES} \\
\small{D\'epartement d'Informatique} && & \small{\supervisors} \\
\thickhline
\end{tabulary}
 \vspace{1in}

\Huge Learning to compress and search\\ visual data in large-scale systems

\vspace{1.8cm}

\Large TH\`{E}SE \\
\vspace{1.0cm}
\large pr\'{e}sent\'{e}e \`{a} la Facult\'{e} des sciences de l'Universit\'{e} de Gen\`{e}ve \\
\vspace{0.2cm}
\large pour obtenir le grade de Docteur \`{e}s sciences, mention informatique \\
\vspace{1.0cm}
par \\
\vspace{0.2cm}

\Large \textbf{Sohrab FERDOWSI} \\

\vspace{0.18cm}
\large de \\
\vspace{0.18cm}

Kermanshah (Iran) \\

\vspace{2.5cm}
Th\`{e}se N$^{\circ}$  5295

\vspace{1.5cm}

Universit\'{e} de Gen\`{e}ve \\

janvier 2019

\end{center}

\clearpage

\includepdf[page=-]{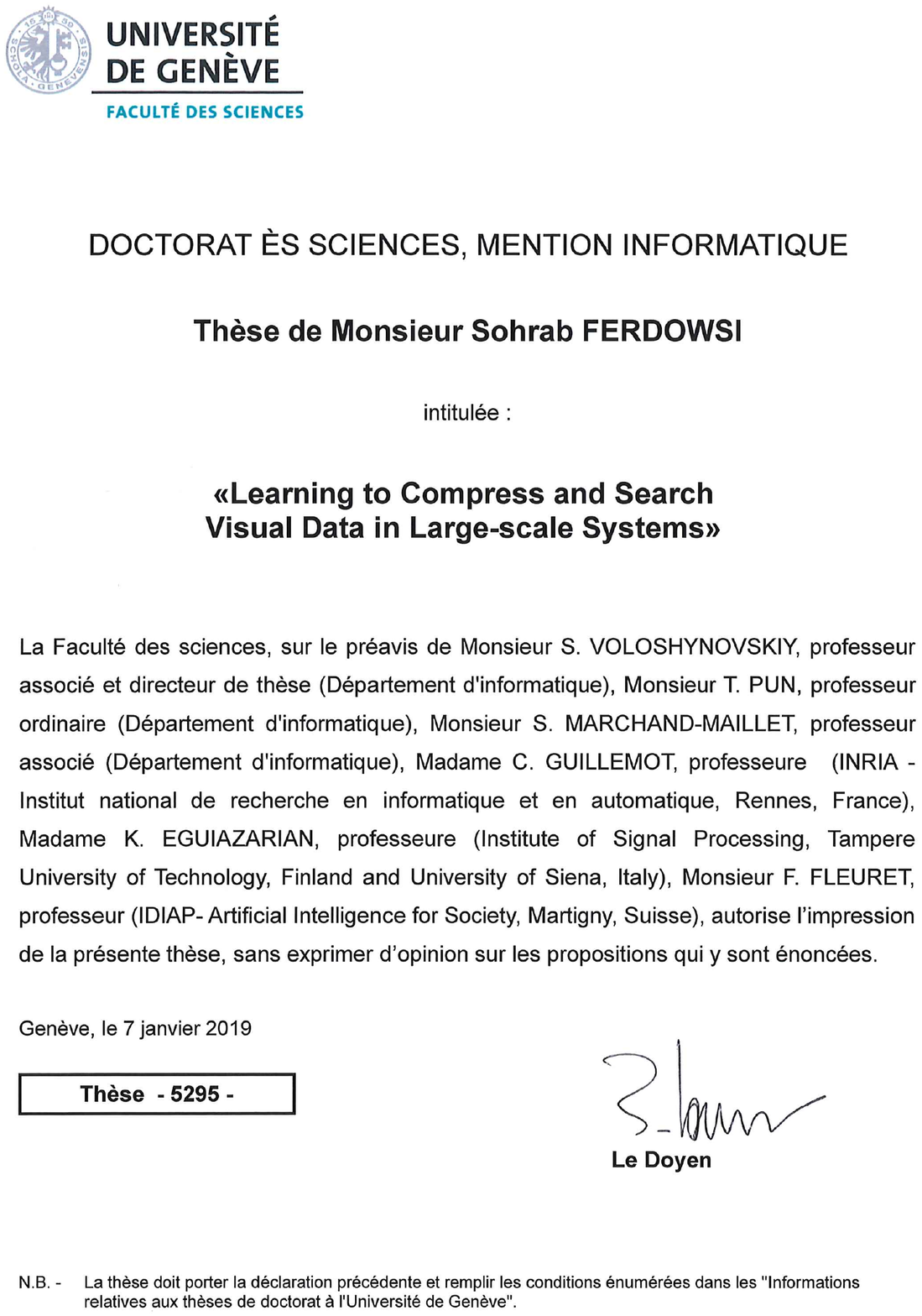}
\includepdf[page=-]{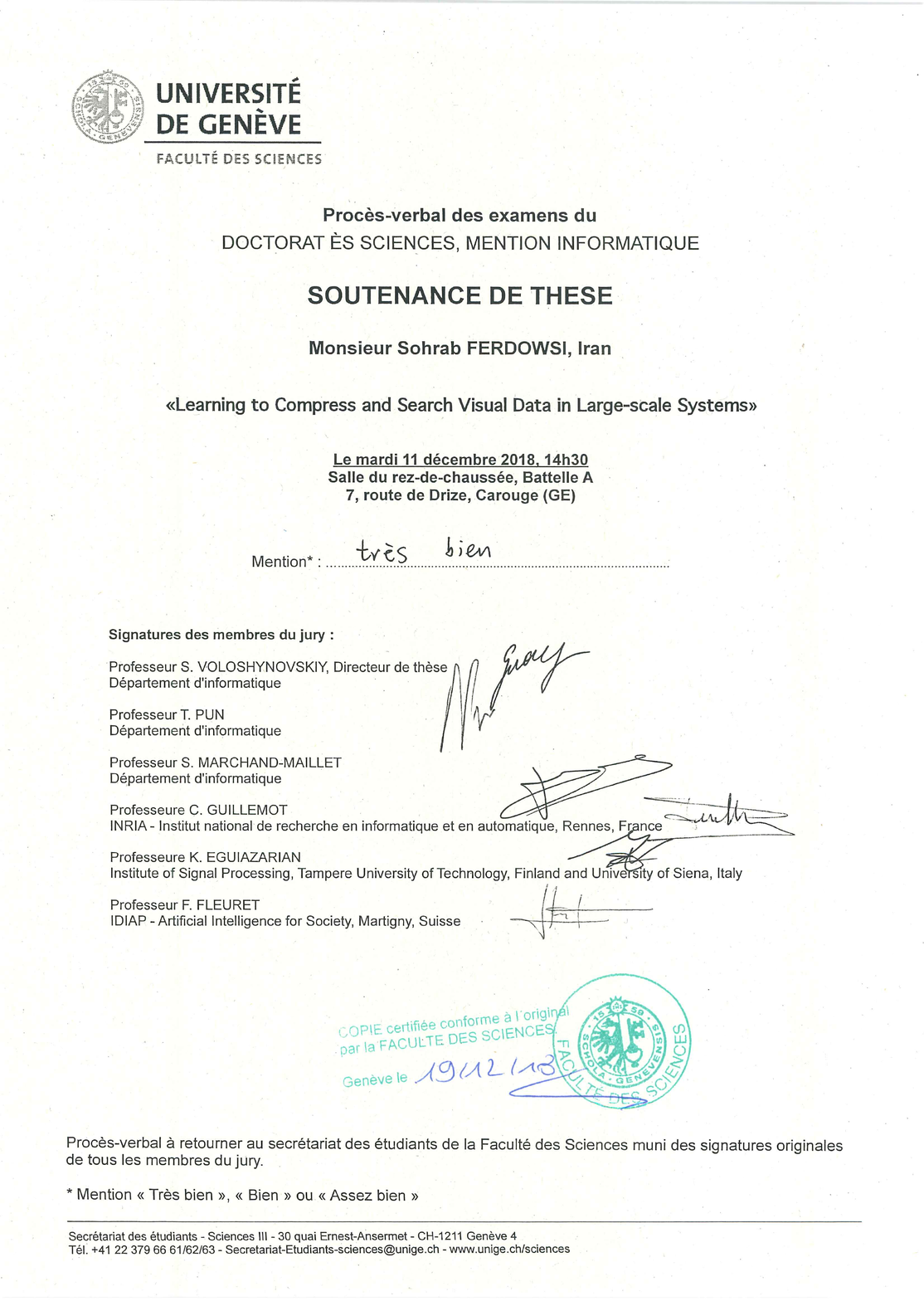}

\includepdf[page=-]{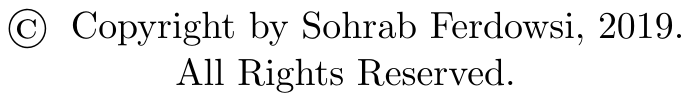}
\frontmatter


\begin{dedication} 

To my parents\\ who gave me life,\\ a frame for it,\\ and some meaning to it.

\end{dedication}


\begin{acknowledgements}  
This manuscript is the outcome of a couple of years of my work as research (and teaching) assistant at the University of Geneva, followed by a fast-paced three months of writing period trying to summarize, organize and document it. Throughout these years, countless number of ideas have been tested, a couple of topics (across several disciplines) have been considered, many simulations have been run, a lot of papers have been read, some wheels have been re-invented, mistakes have been made, things have been learned, trials and errors, many iterations and corrections; and finally it is presented as you read.

During this time, Slava, the supervisor of the thesis and the one who gave me the opportunity to work on it, was a motivating engine for its continuation. He was always there, available and involved. He frequently provided me with his ideas and viewpoints. We had so long discussions, meetings, and brain-storms for so many years, where I learned a lot from him. I would like to acknowledge his remarkable passion as a researcher, his patience as a teacher, and the dedication he has to the work he does.



I am thankful to professors Christine Guillemot, Karen Eguiazarian, Fran\c{c}ois Fleuret, Thierry Pun and St\'{e}phane Marchand-Maillet, for the time they spent evaluating my thesis as members of the jury committee and providing me with their useful comments and feedback. Having got my work approved by these prominent experts, I feel more confident to follow up on my ideas in this thesis in the future.

I would like to thank Thierry, Slava and St\'{e}phane for their role in the formation of the CVML lab, providing a minimal, but very useful basis for the PhD students and researchers to present their ideas and receive feedbacks during regular meetings.

Located in the outskirts of the town, the department of computer science of the University of Geneva is a very small and somehow isolated place. Spending time, usually 6 or 7 days per week at its building called Battelle has the risk of losing your contacts with the rest of the world. However, while being at Battelle during these years, I had the chance of meeting with so many wonderful people from there, some of them I still call friends:

Dimche, with whom I shared the office for all these years and during the most unusual office hours. We spent all those tough periods together, encouraging each other to resist some more hours of work to reach the next deadline. I learned a lot from him, many ideas, optimization rules, and math tricks. His dedication to work was indeed extraordinary.

Farzad, a very supportive friend from whom I learned a variety of things. From how to make my first \LaTeX\ script, to concepts from information theory, all the way to tutorials on how to handle complicated bureaucratic issues as a foreign PhD student in Switzerland. 

Meghdad, a fantastic friend who taught me wise life-lessons and hacks I wouldn't learn otherwise, anytime soon.

SUN, a bright researcher, a high-class scholar and a great friend with whom we had regular coffee-breaks sharing thoughts and our experiences. 

Majid, an original thinker, and a generalist. He taught me some machine learning, basics of computer science, portfolio optimization and psycho-sociology. More importantly, he taught me critical thinking whose outcomes I continue to benefit, as of today.

I would like to mention other friends from the building: Paco whose positive vibes changed the group spirits in many good ways, Michal who managed to keep up a smiley face all these years, Behrooz who helped me in several occasions with sincerity and prudence, Thomas, Soheil, Theodoros, April, Oscar, Fokko, Olga, Denis, Chen, Allen, Alex, Coralie, Phil and yet several more.

I would like to thank my friends Shideh and Amina for their benevolence and help. I acknowledge help from Taras and Maurits while carefully reviewing some of my writings and presentations.

My research was funded by the department of computer science of the University of Geneva. Apart from the stipend, the institution provided me with a desk and a desktop, some basic building facilities, the right to apply for a yearly-valid student residence permit within Switzerland which, if happened to be valid, also allowed me to travel within the Schengen area, some human resources facilities, as well as a thesis supervisor.

I should also mention the helpful hints I received from online Q\&A platforms like \textit{Stack Exchange}, \textit{Stack Overflow} and \textit{PyTorch Forums} and by people I never met. Similarly, thanks to this less-than-ideal peer-reviewing we have, I received comments from anonymous reviewers, some of them happened to be useful. More generally and stating the obvious, I should mention this overwhelming amount of educational material out there and for free, which I sometimes benefited from. Noticing their sheer scale, one might then wonder why to go to school at all? 

During my PhD, I still reserved myself some life outside of the Battelle building. I would like to mention some people and friends I have met that had influences in my life in one way or another. In particular, I would like to mention Khadidja, Rishabh, Ioana, Seva, Sepideh, Soheil, Sina, Ali, Golzar, Bahareh, Keyan, Stephanie, Johannes, Setareh, Neda and La\"{e}titia. Also, I notice the positive role of my swimming habits at the GN1885 club, without which I would be much less productive.

Last, but not least, I would like to mention Mohammad whose kind of help and goodwill you would expect only from a brother.
\bigbreak \bigbreak

Geneva, Switzerland, January 2019 \qquad \qquad \qquad \qquad \qquad \qquad \qquad \qquad  Sohrab Ferdowsi

\end{acknowledgements}

\begin{abstract}

This thesis studies high-dimensional vectorial data, in particular, visual information, e.g., digital images or image descriptors and addresses several of the issues dealt with these data, particularly under large-scale setups.

Attempts for general signal and image modeling in the literature are first reviewed where they are framed under the Bayesian paradigm. These are categorized roughly as basic and composite models, where the former benefits from low sample-complexity, as well as sound theoretical bases, while the latter achieves better performance benefiting from larger data samples. 

This thesis pursues the algorithmic development of its models from basic to composite ones. The basic models are developed under two families of synthesis and analysis priors. Our synthesis model introduces the rate-allocation criterion as a regularization to the K-means algorithm and hence is termed the VR-Kmeans. We show that this is very successful in avoiding over-fitting of K-means at high-dimensional settings and particularly under low-sample setups.

Our analysis-like formulation leads to the framework of Sparse Ternary Codes (STC). This starts with the characterization of its information-theoretic properties and follows by investigating ways to maintain rate-distortion optimal encoding and decoding. We then notice the limitations of these models in achieving high-fidelity and low-complexity encoding and point out the need to opt for more intricate models.

The evolution from basic and single-layer architectures to composite and multi-layer models is done using the principle of successive refinement in information theory. In particular, the VR-Kmeans and the STC are extended to the RRQ and the ML-STC using additive residual-based encoding, respectively. These models are analyzed algorithmically and their rate-distortion performances are shown to be superior compared to their existing alternatives.

The ML-STC, and its more data-dependent version the ML-STC-Procrustean admit yet another evolution. This is the joint parameter update using the back-propagation framework which resembles that of artificial neural networks and hence we term it as the STNets. However, our model has certain particularities as compared to the common deep learning frameworks. Instead of starting from random weights, the STNets is first pre-trained layer-by-layer and according to the STC. This is then fine-tuned using back-propagation along with other standard recipes of training neural networks. Technically, this is possible thanks to the properties of ternary encoding which allows us to replace the non-differentiable discrete non-linearity with its smooth counterpart and without incurring approximation errors. Consequently, we are able to learn discrete and compact representations for data and under a wide range of data-availability and operational rate-regimes.

Having developed our algorithmic infrastructure, we next tailor them to three important practical applications. First, the problem of large-scale similarity search in retrieval systems is addressed, where complexity and memory constraints limit the na\"{i}ve exhaustive scan of the database for a given query. We develop a complete search system based on the STC framework and show its superiority in terms of the triple complexity-memory-performance trade-offs as compared to the two main-stream solutions from the literature, namely the binary hashing and the vector-quantization based family of methods.

We next target the problem of learned image compression. We argue the benefits of learning to compress w.r.t. the conventional codecs and show that it is possible to compress high-resolution natural images using our algorithms trained on a limited number of images and achieve comparable results to the JPEG2000, even without performing different stages of the compression pipeline. More particularly and for a class of domain-specific images, we show that it is possible to benefit from the extra structural redundancy present in these images to compress them further. We also show that learning to compress can be beneficial beyond the task of compression itself.

Finally, we show that compression can be used to solve inverse problems. This is achieved by imposing the compressibility of data under a certain trained model as an effective prior to regularize the solution of ill-posed inverse problems, which is invoked in an iterative algorithm. In particular, we show that it is possible to deonise images using the JPEG2000, or recover under-sampled and noisy auto-regressive data using the ML-STC and through our proposed algorithm.

The thesis is concluded by pointing out some open problems and issues. This paves the way for certain potential directions of very promising future research to be pursued in the interplay between signal processing and machine learning and under the theme of learning compact representations.

\end{abstract}

\begin{resume}

Cette th\`{e}se \'{e}tudie des donn\'{e}es vectorielles de grande dimension, en particulier des informations visuelles, telles que des images num\'{e}riques ou des descripteurs d’images, et aborde plusieurs des probl\`{e}mes trait\'{e}s avec ces donn\'{e}es, en particulier dans des configurations \`{a} grande \'{e}chelle.

Les tentatives de mod\'{e}lisation g\'{e}n\'{e}rale du signal et de l’image dans la litt\'{e}rature sont d’abord examin\'{e}es l\`{a} où elles sont encadr\'{e}es par le paradigme bay\'{e}sien. Celles-ci sont class\'{e}es grossi\`{e}rement en deux parties: les mod\`{e}les de base et les  mod\`{e}les composites, où le premier b\'{e}n\'{e}ficie d'une faible complexit\'{e} d'\'{e}chantillon, ainsi que de bases th\'{e}oriques solides, tandis que le second offre de meilleures performances en tirant parti d'\'{e}chantillons de donn\'{e}es plus volumineux.

Cette th\`{e}se poursuit son d\'{e}veloppement algorithmique des mod\`{e}les de base aux mod\`{e}les composites. Les mod\`{e}les de base sont poursuivis sous deux familles d’a priori de synth\`{e}se et d’analyse. Notre mod\`{e}le de synth\`{e}se introduit le crit\`{e}re d’allocation de d\'{e}bit en tant que r\'{e}gularisation de l’algorithme K-means et est donc appel\'{e} VR-Kmeans. Nous montrons que ceci est tr\`{e}s efficace pour \'{e}viter le sur-ajustement de K-means dans des configurations de grandes dimensions et en particulier dans des configurations de faible \'{e}chantillon.

Notre formulation de type analyse conduit au cadre des Sparse Ternary Codes (STC). Cela commence par la caract\'{e}risation de ses propri\'{e}t\'{e}s th\'{e}oriques de l'information, puis par la recherche de moyens permettant de maintenir un codage et un d\'{e}codage optimaux en d\'{e}bit-distorsion. Nous remarquons ensuite les limites de ces mod\`{e}les dans l’obtention d’un codage haute fid\'{e}lit\'{e} et peu complexe, et soulignons la n\'{e}cessit\'{e} d’opter pour des mod\`{e}les plus complexes.

L'\'{e}volution des architectures de base et \`{a} couche unique vers les mod\`{e}les composites et \`{a} couches multiples se fait selon le principe du raffinement successif de la th\'{e}orie de l'information. En particulier, le VR-Kmeans et le STC sont \'{e}tendus au RRQ et au ML-STC en utilisant un codage additif bas\'{e} sur les r\'{e}sidus, respectivement. Ces mod\`{e}les sont analys\'{e}s algorithmiquement et leurs performances d\'{e}bit-distorsion sont sup\'{e}rieures \`{a} leurs alternatives existantes.

Le ML-STC et sa version plus d\'{e}pendante des donn\'{e}es, le ML-STC-Procrustean, admettent encore une autre \'{e}volution. Il s’agit de la mise \`{a} jour conjointe des param\`{e}tres \`{a} l’aide du cadre de back-propagation, qui ressemble \`{a} celui des r\'{e}seaux de neurones artificiels. Nous l’appelons donc STNets. Cependant, notre mod\`{e}le pr\'{e}sente certaines particularit\'{e}s par rapport aux cadres d'apprentissage profondis communs. Au lieu de commencer par des poids al\'{e}atoires, les STNets sont d'abord pr\'{e}-form\'{e}s couche par couche et conform\'{e}ment au STC. Ceci est ensuite ajust\'{e} en utilisant la back-propagation avec d'autres recettes standard de r\'{e}seaux de neurones d'apprentissage. Techniquement, cela est possible grâce aux propri\'{e}t\'{e}s du codage ternaire qui nous permet de remplacer la non-lin\'{e}arit\'{e} discr\`{e}te non diff\'{e}rentiable par son \'{e}quivalent lisse et sans encourir d’erreurs d’approximation. Par cons\'{e}quent, nous pouvons apprendre des repr\'{e}sentations discr\`{e}tes et compactes pour les donn\'{e}es et sous un large \'{e}ventail de r\'{e}gimes de disponibilit\'{e} d'echantillon et de d\'{e}bit op\'{e}rationnel.

Apr\`{e}s avoir d\'{e}velopp\'{e} notre infrastructure algorithmique, nous les adaptons ensuite \`{a} trois applications pratiques importantes. Nous abordons d’abord le probl\`{e}me de la recherche de similarit\'{e} \`{a} grande \'{e}chelle dans les syst\`{e}mes de recherche d'information, où la complexit\'{e} et les contraintes de m\'{e}moire limitent l’analyse exhaustive de la base de donn\'{e}es pour une requête donn\'{e}e. Nous d\'{e}veloppons un syst\`{e}me de recherche complet bas\'{e} sur le cadre STC et montrons sa sup\'{e}riorit\'{e} en termes de triple compromis complexit\'{e}-m\'{e}moire-performance par rapport aux deux solutions principales existantes dans la litt\'{e}rature, \`{a} savoir le hachage binaire et la famille de m\'{e}thodes bas\'{e}es sur la quantification vectorielle.

Nous ciblons ensuite le probl\`{e}me de la compression d’image apprise. Nous expliquons les avantages de l’apprentissage de la compression par rapport aux codecs classiques et montrons qu’il est possible de compresser des images naturelles haute r\'{e}solution \`{a} l’aide de nos algorithmes entraîn\'{e}s sur un nombre limit\'{e} d’images et d’obtenir des r\'{e}sultats comparables \`{a} ceux du JPEG2000, même sans effectuer diff\'{e}rentes \'{e}tapes communs du pipeline de compression. Plus particuli\`{e}rement, et pour une classe d’images sp\'{e}cifiques \`{a} un domaine, nous montrons qu’il est possible de tirer parti de la redondance structurelle suppl\'{e}mentaire pr\'{e}sente dans ces images pour les compresser davantage. Nous montrons \'{e}galement qu'apprendre \`{a} compresser peut être b\'{e}n\'{e}fique au-del\`{a} de la tâche de compression elle-même.

Enfin, nous montrons que la compression peut être utilis\'{e}e pour r\'{e}soudre des probl\`{e}mes inverses. Ceci est r\'{e}alis\'{e} en imposant la compressibilit\'{e} des donn\'{e}es sous un certain mod\`{e}le appris en tant q'un moyen efficace de r\'{e}gularisation de la solution des probl\`{e}mes inverses mal pos\'{e}s, en utilisant un algorithme it\'{e}ratif. En particulier, nous montrons qu'il est possible de d\'{e}bruiter des images \`{a} l'aide du JPEG2000 ou de r\'{e}cup\'{e}rer des donn\'{e}es auto-r\'{e}gressives sous-\'{e}chantillonn\'{e}es et bruyantes \`{a} l'aide du ML-STC et grace \`{a} notre algorithme propos\'{e}.

La th\`{e}se se termine en soulignant certains probl\`{e}mes et questions en suspens. Cela ouvre la voie \`{a} certaines pistes de recherche future tr\`{e}s prometteuses \`{a} poursuivre dans l'interaction entre le traitement du signal et l'apprentissage automatique et sous le th\`{e}me de l'apprentissage des repr\'{e}sentations compactes.

\end{resume}


\tableofcontents




\printnomenclature

\mainmatter
\part{Preliminaries}

\chapter{Introduction}  

The story of big data, data-driven analytics and machine learning has widely been narrated as a phenomenal revolution affecting, changing or even entirely redefining aspects of humans lives, as well as bringing new frontiers to it. 

But before being able to tell how much and to what extent these narratives are valid, or whether this is  yet another hype that will be forgotten at some point, it is very obvious that a careful, rigorous and extensive understanding of what the data is, how it should be processed and the different aspects and challenges of reasoning based on the data should be regarded as a central question meriting due attention.

Within the academic societies, this has started from a century ago, continued to grow till today and perhaps due to the unprecedented attention and support it received from outside of academia, has accelerated particularly within the last one or two decades. 

In spite of the large body of research work published ever since, in view of the new challenges that are faced, the potentials of new possibilities and the highly raised expectations, one may conclude that our understanding of the domain today is rather explorative than well-established.

This thesis\footnote{Some of the results presented in this thesis can be reproduced easily from \href{https://github.com/sssohrab/PhDthesis}{https://github.com/sssohrab/PhDthesis}. This repository gets gradually updated to include more reproducible results and implementations.} is an effort to study some of the aspects regarding data processing, address several of its issues, as well as provide some solutions and strategies for several data-driven applications.

This being a very broad explanation relating to a veriety of fields and research communities, we next focus our scope and discuss the main themes of this thesis.

\section{Scope of the thesis} \label{sec:intro_scope}
We next clarify what we particularly mean by data in this thesis and then bring some of the important themes of this thesis into attention. We then briefly mention some of the application areas that our thesis will focus on and the communities that we relate to.
\subsection{What data do we use?} \label{subsec:intro_scope_what}
The sources of the data are endless. Measurements of physical sensors, large collections of web data, texts, speech recordings, images and user activities are just several examples. After the data has been captured and stored in digital form, it can be represented in different ways, e.g., graphs, sequences or vectors.

In this thesis, we consider mainly vectorial data, i.e., we assume every given data example is an $n$-dimensional point in the space of $\Re^n$, where $n$ is the number of features of that description. The first challenge faced here is that most of the intuitions we developed for the space of $\Re^3$, i.e., the 3D physical world we live in, are unlikely to be valid and applicable to this $n$-dimensional space of data-points. This means that we should stick into formal mathematical rules rather than some limiting intuitions.

While a lot of the materials we present are valid for the larger family of correlated and non-sequential data, we focus our attention on digital images and image descriptors. These are two-dimensional arrays of pixel values that can be unraveled as vectors.

An alternative term that we use is a ``signal'', which was traditionally used to refer to sensor measurements. Nowadays, thanks to the applicability of the concepts developed within signal processing to domains beyond sensors, its definition has entirely changed. In this thesis, we use the terms data and signal interchangeably to refer to the same concept.
\subsection{Main themes} \label{subsec:intro_scope_themes}
Here are a couple of themes that we will encounter in this thesis.
\subsubsection*{Unsupervised learning} Thanks to the ever-increasing availability of data and the democratized computational powers at hand, a central philosophy behind all data-driven approaches is that, rather than the conventional ways of reasoning, which are based on digging into the internal structures of a system or a phenomenon and studying their relations, we can now shift our paradigm of analysis by relying more on the study of its exemplary behavior under different conditions, i.e., by processing of the data that the system produces.

One notable instance of this idea is through supervised learning, where semantic labels are associated to vectorial data examples and the mapping of these data-label pairs are to be learned from the data. While this is perhaps the most popular and successful instance of learning from data, this thesis focuses on unsupervised learning, where instead of the data-label mapping, internal structures of the data itself are to be revealed.
\subsubsection*{Data versus priors}
As it has been publicized extensively within the context of artificial intelligence, it sounds very attractive to perform all the reasoning and the related tasks entirely from the data. In practice, however, this is rarely the case. For the learning algorithm to be successful, along with the data, one has to condition the reasoning on some domain knowledge or prior information about the phenomenon under study.

In fact, there is a fundamental trade-off between the amount of required data to perform a certain task and the quality and strength of the prior information incorporated. Although for many applications there is no shortage of available data, it can be shown for some tasks that without proper regularization using priors, it is impossible to achieve satisfactory results. This is the case particularly for high dimensional data, since as a rule of thumb, the linear increase of data dimensionality requires exponential growth of data examples.

In this thesis, we consider a continuum of scenarios in this respect. Starting from the hypothetical case where we know the exact probabilistic rule from which the data is generated and hence do not need the data itself, we gradually shift towards more reliance on the data and less on the assumptions and propose the solutions accordingly.   
\subsubsection*{Large-scale scenarios}
A particularly challenging aspect in a lot of data-oriented domain areas is in large-scale scenarios, where the number of data-points to be processed as well as the data dimensionality is larger than what our computational resources needed for the processing can accommodate. While the rate of growth of data is exponential, the increase of processing power, memory resources as well as communication bandwidth is logarithmic\footnote{It has been concluded that Moore's law which was guaranteeing a linear increase in computational power for decades has stopped somewhere in the mid-2010's.}, or at best linear. 

In this thesis, we focus only on algorithms whose computational and storage requirements grow no more than linearly with the number of data samples, and no more than polynomially with the dimensionality.

Our experiments are performed usually on the scale of several thousand to a million data samples. While we were restricted by our available computational resources and did not perform experiments on billion-scale data, we believe that the developed algorithms are applicable to such scale, as well.
\subsubsection*{Representation/dictionary learning}
As captured from physical sensors or sampled from some underlying phenomena, it is unlikely to be able to perform meaningful analyses directly on the data in its ambient form. To make sense out of it and to make it amenable for analyses, the data has to be decomposed in some way and represented within that decomposition framework. Within the last 4 or 5 decades, this has been the central topic of a large body of extensive research among the signal processing and more recently machine learning communities.

With the addition of compactness criteria to signal representation that we describe next, this is the central most topic of this thesis.  
\subsubsection*{Compact data representations}
Restricted by computational, storage and particularly bandwidth constraints, starting from the early days of information and communication, the need for compression of the data has been pointed out as one of its fundamental aspects. Ever since the theoretical foundations have been laid, the quest for better compression solutions has not stopped.

Apart from this motivation, compression, and more generally finding compact representations for data is closely linked to the primary objective of unsupervised learning, i.e., trying to capture the underlying probabilistic rules that generate the data.

We pay particular attention to this notion in this thesis and provide several solutions for compactly representing the data. Apart from the task of compression itself, we show promising possibilities of such representation in performing other tasks, including inverse problems in signal processing. 

\subsection{Example applications} \label{subsec:intro_scope_applications}
In our computer-simulated experiments, we focus our attention and provide numerical results for the following application areas:
\subsubsection*{Large-scale similarity search}
A central idea behind a lot of data-driven paradigms is to reflect semantic meanings from our real world into vectorial representations. This means that the neighborhood within vectors, i.e., the geometrical closeness of data points in the space of $\Re^n$ would ideally reflect semantic similarity in the real world. Researchers have already found ways to somehow achieve this idea in practice; to some limited, but sometimes acceptable extent. 

An immediate application of this data is in content-based retrieval systems like reverse image search engines, where users want to search for similar images from within a large database. For example, imagine we have seen a photo on the web containing a monument or a city landmark, but we are not sure which monument this was or where it is located. We expect to be able to find visually similar images from the search engine's database.

This, however, poses serious and limiting challenges with regard to the computational and storage requirements. Assuming meaningful feature vectors are present, this thesis studies different aspects of this problem by addressing the fundamental triple trade-off between memory, complexity and search performance, as well as providing practical solutions for it.
\subsubsection*{Learning to compress images}
Classical compression schemes like JPEG and JPEG2000 use data-independent image decomposition bases, i.e., the DCT and the DWT. Although these codecs are highly engineered and tailored, the use of such data-independent decompositions has limited their efficiency. In fact, it has been shown in many tasks that adapting the decompositions or the transforms to the data can bring noticeable improvements. Image compression, too, can benefit from this.

Moreover, apart from the general class of natural images, the set of images encountered in some practical applications can be highly structured. It is then expected to be able to benefit from the excess redundancies that the general-purpose image codecs cannot capture. 

In this thesis, without going into the details of typical image compression pipelines, e.g., chrominance sub-sampling or entropy coding, we apply our algorithms, both to general natural images and also to a domain-specific example, where they learn to compress by adapting to the statistics of the training images.

\subsubsection*{Compression as regularization}
To go beyond the task of compression, we showcase some examples that learning to compress can be useful for other tasks. We show that compressibility of data under a certain (trained) model can be used as a very strong regularizer to solve inverse problems. In particular, we provide some preliminary results on image compression and compressive sensing showing some advantage. We leave this line of work as an exciting and promising direction for future research and further analysis.

\subsection{Target communities}
Our emphasis on large-scale and high-dimensional setups made us think of basic information-theoretic concepts as the guiding principle. It is interesting to note that while high dimensionality in many domains is considered as a curse, on the contrary in information theory, this is a blessing. In fact, the promises made in information theory are achieved in asymptotic cases, and higher dimensionality only helps to approach them.

From the other hand, these theoretical results are not directly applicable to practical scenarios, since they are based on several restricting assumptions. Starting from these assumptions in information theory, in this thesis, we gradually depart from such hypothetical cases and take more practical approaches. Therefore, rather than theoretical treatments like achievability and converse arguments, our contributions are only practical application instances for the domain of information theory.

Our initial target audience in this thesis is within signal and image processing communities since most of our derivations have the touch and feel of signal processing style treatments.

Nevertheless, nowadays, the boundaries between signal processing and machine learning research are shrinking. Both communities borrow concepts from one another. A lot of the materials in this thesis are inter-disciplinary between these two domains. We made some effort to benefit from the advantages of these two disciplines together.

We believe there is still a lot more to be discovered. In fact, the best is yet to come, as the synergy between the two paradigms can radically transform existing learning-based methods.

\section{Basic setup and notations} \label{sec:intro_basics}
As it was mentioned earlier, we assume data-points are vectors in $\Re^n$, the space of $n$-dimensional real numbers. We depict vectors with bold letters, e.g., $\mathbf{f} = [f_1, \cdots, f_j, \cdots_, f_n]^T \in \Re^n$, consisting of scalars $f_j$'s.

We concatenate a set of $N$ data-points in columns of a matrix, e.g., $\mathrm{F} = [\mathbf{f}_1, \cdots, \mathbf{f}_i, \cdots, \mathbf{f}_N]$. All matrices are depicted as capital letters with up-right fonts\footnote{We use the \textsf{mathrm} command in Latex typesetting.}, e.g., $\mathrm{F}, \mathrm{A}, \mathrm{C}, \cdots$. 

All vectors are column vectors. So if we want to decompose a matrix by its rows (instead of columns as above), we write $\mathrm{A} = [\mathbf{a}(1), \cdots, \mathbf{a}(m'), \cdots, \mathbf{a}(m)]^T$, where $\mathbf{a}(m') = [\mathrm{A}(m',1), \cdots, \mathrm{A}(m',n)]^T$ is a column vector that represents the $m^{'^{\text{th}}}$ row of $\mathrm{A}$.

When we have a collection of data-points, we sometimes need to specify a probabilistic rule from which these data-points have been generated. We use the notion of the random variable (for scalar samples) and random vectors (for vectorial samples). We use capital letters to denote scalar random variables, e.g., $F$, and capital bold letters to denote random vectors, e.g., $\mathbf{F} = [F_1, \cdots, F_n]^T$. So $\mathbf{F}$ is the underlying rule whose instances are the data-points, i.e.,  $\mathbf{f}_1, \cdots, \mathbf{f}_N$.

For the random vector $\mathbf{F}$, we assign a (joint) probability distribution as $\mathbf{F} \sim p(\mathbf{f})$. A recurring example is the multivariate Gaussian distribution with zero mean and covariance matrix $\mathrm{C}_{\mathbf{F}}$, i.e.,  $\mathbf{F} \sim \mathcal{N}(\mathbf{0}, \mathrm{C}_{\mathbf{F}})$. When dimensions of $\mathbf{F}$ are independent and identically distributes (\textit{i.i.d.}), this is depicted as  $\mathbf{F} \sim \mathcal{N}(\mathbf{0}, \sigma^2 \mathrm{I}_{n})$, where $\mathrm{I}_n$ is the $n$-dimensional identity matrix, or simply as $F \sim \mathcal{N}(0,\sigma^2)$. Another important case is the independent, but not identically distributed (\textit{i.n.i.d.}) Gaussian data, which we depict as $\mathbf{F} \sim \mathcal{N}(\mathbf{0}, \text{diag}([\sigma_1^2, \cdots, \sigma_n^2]^T)$. 

We frequently use the ternarizing operator with a threshold $\lambda$, which is an element-wise function\footnote{We use the threshold-based ternarizing function in this thesis since it is more straightforward for analysis. However, in our experiments, we also use the ``k-best'' operator which picks the $k$ elements with $k$ largest magnitudes. In this case, the non-linearity is not element-wise and the threshold is adapted to all the elements.} and is defined as:

\begin{equation}  \label{eq:math_TernarizingFunc}
\phi_{\lambda}(t) = \text{sign} (t) \cdot \mathbbm{1}_{\lbrace |t| > \lambda \rbrace},
\end{equation}
where the indicator function $\mathbbm{1}_{\lbrace |t| > \lambda \rbrace}$ indicates whether the event $|t| > \lambda$ has occurred. This is depicted in Fig. \ref{subfig:SingleLayer_STC_thresholding_phi}.

Closely related to the discrete ternarizing function is the hard-thresholding function, which is defined as:
\begin{equation}  \label{eq:math_HardThresholdingFunc}
\psi_{\lambda}(t) = t \cdot \mathbbm{1}_{\lbrace |t| > \lambda \rbrace},
\end{equation}
and whose output is not discretized. This is depicted in Fig. \ref{subfig:SingleLayer_STC_thresholding_psi}. Another related operator is the soft-thresholding function, which shrinks its input and  is defined as:
\begin{equation}  \label{eq:math_SoftThresholdingFunc}
\eta_{\lambda}(t) = (t - \text{sign}(t) \cdot \lambda) \cdot \mathbbm{1}_{\lbrace |t| > \lambda \rbrace} = \text{sign}(t) \cdot (|t| - \lambda)^+.
\end{equation}

We encapsulate all our algorithms under the concept of encoding and decoding. An encoder $\mathbb{Q}[\cdot] \colon \Re^n \to \mathcal{X}^m$ maps a data-point to a more compact and perhaps discrete space and produces the code $\mathbf{x} = \mathbb{Q}[\mathbf{f}]$. This compact encoded representation, apart from reducing the storage cost, may be useful for other purposes, e.g., performing fast search, reducing over-fitting in supervised tasks or regularization in inverse-problems. 

This code may then be decoded with the decoder $\mathbb{Q}^{-1}[\cdot] \colon \mathcal{X}^m \to \Re^n$ to produce an approximation to the original data-point as $\hat{\mathbf{f}} = \mathbb{Q}^{-1}[\mathbf{x}]$. 

A fundamental aspect of the behavior of the encoder-decoder pair is its rate-distortion performance. This is first formulated in the rate-distortion theory as we briefly mention next. 

\subsection{Shannon source coding theorem} \label{subsec:fundamentals_SourceCoding_RD}
Following the characterization of \cite{shannon1959coding}, for a source $F \sim p(f)$ emitting \textit{i.i.d.} sequences $\mathbf{F} = [F_1, \cdots, F_n ]^T$, a rate-distortion pair $(R,D)$ is said to be achievable under the encoder-decoder pair $\mathbb{Q}[\cdot]: \mathcal{F}^n \to \{1,\cdots, 2^{nR}\}$ and $\mathbb{Q}^{-1}[\cdot]: \{1,\cdots, 2^{nR}\} \to \hat{\mathcal{F}}^n$ and a distortion measure $d(\cdot,\cdot):\mathcal{F}^n \times \hat{\mathcal{F}}^n \to [0,\infty) $, if $\mathbb{E}[d(\mathbf{F}, \mathbb{Q}^{-1}[\mathbb{Q}[\mathbf{F}]])] \leqslant D$.

The fundamental result of this theory states that, for a given $D$, all rates $R \geqslant\mathcal{R}(D)$ are achievable, if $n \rightarrow \infty$, and such an $\mathcal{R}(D)$ function is calculated as in Eq. \ref{eq:fundamentals_RD}:

\begin{equation} \label{eq:fundamentals_RD}
\begin{aligned}
&\mathcal{R}(D) &= \underset{{p(\hat{f}|f)}}{\text{ min }}   I(F;\hat{F}) &\\
&\text{s.t.}   &\mathbb{E}_{p(f)p(\hat{f}|f)}[d(F,\hat{F})] &\leqslant D.
\end{aligned}
\end{equation}

Two important instances of the rate-distortion theory that we will encounter extensively in this thesis are the Gaussian \textit{i.i.d.} and the Gaussian \textit{i.n.i.d.} sources as we mention next.

\subsubsection{\textit{i.i.d.} Gaussian sources} \label{subsubsec:fundamentals_SourceCoding_RD_iid}
For a Gaussian source $F \sim \mathcal{N}(0, \sigma^2)$, the rate-distortion function is derived as:
\begin{equation} \label{eq:fundamentals_RDGaussian}
\mathcal{R}(D) \geqslant \frac{1}{2} \log_2{\frac{\sigma^2}{D}}.
\end{equation}

This can provide a description for an \textit{i.i.d.} Gaussian vector $\mathbf{F} \sim \mathcal{N}(\mathbf{0}, \sigma^2 \mathrm{I}_n)$, as to how much and with which quality it can be compressed in its asymptotic limit.
\subsubsection{Independent, but not identically distributed (\textit{i.n.i.d.}) Gaussian sources} \label{subsubsec:fundamentals_SourceCoding_RD_inid}
The joint description  of $n$ Gaussian sources with different variances, i.e., the random vector $\mathbf{F} \sim \mathcal{N}(\mathbf{0}, \text{diag}([\sigma_1^2, \cdots, \sigma_n^2]^T))$, is asymptotically limited by the following rate-distortion function:
\begin{equation}  \label{eq:fundamentals_OptRate}
\mathcal{R}(\mathcal{D}) = \sum_{j = 1}^n R_j = \sum_{i = 1}^n \frac{1}{2} \log_2 \Big(\frac{\sigma_j^2}{D_j}\Big),
\end{equation}
where $D_j$ for $j=1, \cdots, n$ is the distortion of the $j^\text{th}$ source after rate allocation, and whose value is the solution to the following convex problem:

\begin{equation} \label{eq:fundamentals_revWFObj}
\begin{aligned}
& \underset{D_j}{\text{min}}  &  \sum_{j = 1}^n \text{max}[0,\frac{1}{2} \log_2 \frac{\sigma_j^2}{D_j}] \\
& \text{s.t.}  & \sum_{j = 1}^n D_j = D, \\
\end{aligned}
\end{equation}
which is solved in closed-form as:

\begin{equation} \label{eq:fundamentals_revWFDist}
D_j =
\begin{cases}
   \gamma ,& \text{if   } \sigma_j^2 \geqslant \gamma \\
    \sigma_j^2, & \text{if   } \sigma_j^2 < \gamma,
\end{cases}
\end{equation} 
where $\gamma$ is a constant, which should be chosen to guarantee that $\sum_{j=1}^n D_j = D$.

This problem is often referred to as the reverse water-filling paradigm, as described in Ch. 10 of \cite{CoverThomas200607}.

The Shannon's setup of rate-distortion theory described above, however, has serious limitations. For example, the need for storage and search within an exponentially-large codebook and the asymptotic and probabilistic assumptions related to them limit their application in practical settings. A central objective of this thesis is, therefore, to design practical encoding schemes for our applications. 

\section{Outline of the thesis} \label{sec:intro_outline}
This manuscript is divided into 3 parts:

\textbf{Part I} discusses the preliminary materials including this introductory chapter, as well as \textbf{chapter \ref{chapter:ModelingLit}}, which presents an overview of signal and image modeling techniques in the literature. We regard the Bayesian framework as a central connecting point under which a lot of efforts in the literature can be explained. This consists of an ``observation'' part (or data part), which differentiates between certain methods and a ``prior part'' whose variations, the way it is injected into the objective and how the overall optimization routine is treated, is the main factor to distinguish between different methods.

We then argue that the straightforwardness, along with the richness of models within their space of possible parameters are key properties of priors that indicate how algorithms behave under different sample size regimes. We believe that the existing models in the literature fill the spectrum of such possible trade-offs, only partially. A systematic way to choose models and parameters based on the sample size regime seems to be missing. This thesis makes some effort in this direction.

\textbf{Part II} develops our core algorithms that will be used in the rest of the thesis. In \textbf{chapter \ref{chapter:SingleLayer}}, we formalize a general objective function for these algorithms, which is to target rate-distortion optimality for high-dimensional data with discretized representations. We start with simple models from the literature: the synthesis and the analysis prior models. Our synthesis-like formulation leads us to the VR-Kmeans algorithm, which regularizes the famous K-means and enables it to represent high-dimensional data with a small number of training samples available.

Our analysis-like formulation results in the Sparse Ternary Codes (STC). This will be the building block for many of our further developments. We derive its information-theoretic properties and then propose 3 solutions how to decode it in a rate-distortion optimal fashion.

Having faced the limitations of these basic models under the so-called single-layer architectures, in \textbf{chapter \ref{chapter:MultiLayer}}, we target more intricate and composite models based on these building blocks. We use the concept of successive refinement from information theory as the foundational principle to build composite models out of simpler ones. Under our setups, successive refinement simplifies as a simple residual encoding which is done in multiple layers. We use this rule to upgrade both of our models.

Our multi-layer extension of VR-Kmeans results in the RRQ framework, where a rate allocation is performed for each layer before encoding that prevents from over-fitting, reduces the number of distance computations and results in sparse dictionaries.  

Using the same multi-layer residual-based principle, the basic STC model extends to the ML-STC and the more data-dependent Procrustean approach extends to the ML-STC-Procrustean. While this is a highly trained model, the algorithm is performing layer-wise training. In case a lot of training samples are available, an end-to-end extension is needed to fully use the high capacity of the model.

The analysis-like structure of ML-STC resembles that of neural networks. Therefore, for the end-to-end training of ML-STC, we can use the recipes from the deep learning literature, e.g., the back-propagation, Adam or dropout. This leads us to the STNets model, which is pre-trained with ML-STC or ML-STC-Procrustean models. We then discuss its advantages and its differences with the standard practice of end-to-end learning within the deep-learning literature.

\textbf{Part III} applies these algorithms to practical applications. In \textbf{chapter \ref{chapter:Search}}, we address the problem of similarity search in large-scale scenarios. We observe that the existing solutions follow two main strategies: The family of binary hashing that manages to perform very efficient search in the space of codes but fails to use the information in the un-encoded query which is available. Moreover, its coding efficiency in terms of distance preservation and rate-distortion trade-off is very limited, resulting in a low-performance search. The second family is based on VQ and hence maintains a very well rate-distortion balance. However, it does not benefit from fast search in the space of codes, and hence it is slower.

Our STC framework can naturally propose a middle-ground approach for similarity search. Following an information-theoretic analysis, where we demonstrate the coding efficiency of the STC, we first perform a fast search in the space of ternary codes that gives an initial approximative list. Thanks to the excellent rate-distortion performance of our ML-STC discussed in chapter \ref{chapter:MultiLayer}, this is then followed by a list-refinement step where the initial short-list of candidates is further improved after reconstructing from the codes and performing the search within the original vector space. Several million-scale experiments are performed to validate this idea.

We next target image compression. In {chapter \ref{chapter:ImCompression}}, we advocate the recent trend in machine learning that aims at improving traditional image codecs by learning. We show how this idea can be beneficial, both for the class of general natural images and also on domain-specific data with structured redundancy.  

While compression is essential on its own, chapter \ref{chapter:CompPrior} shows some preliminary, but promising results suggesting the benefits of learning to compress beyond compression. We propose a very simple iterative algorithm that regularizes inverse problems by promoting compressible solutions. We show some interesting results on image compression, as well as compressive sensing.

\section{Our main contributions} \label{sec:intro_contributions}
Here we summarize the contributions that we claim for this thesis. First, the contributions of each chapter are discussed and finally the overall achievements of the thesis are highlighted.
\subsection{Chapters} \label{subsec:intro_contributions_chapters}

\subsubsection*{Chapter 2}  
A surveying chapter, particularly one that does not make any effort to be exhaustive, may not claim many contributions. Yet, we believe that the presentation of materials in this chapter is not usual and can be useful for a lot of readers.

In particular, our emphasis on the necessity to understand algorithms based on their sample complexity requirements is important. While more rigorous arguments exist within the probability and statistics literature to analyze priors within the Bayesian framework, our informal argument on the quality of the priors and the subsequent categorization of approaches in signal processing and machine learning communities under ``basic'' and ``composite'' priors is useful and constructive and lets us locate our efforts in this within a proper ground.

\subsubsection*{Chapter 3}  
Few methods have addressed data representation in compressed and discretized domains. We propose two families of solutions based on synthesis and analysis models, respectively.

Within the family of synthesis models, one popular solution is the family of VQ models based on the K-means algorithm. We introduce an important extension to K-means, i.e., variance regularization by rate allocation. While rate allocation is a very old concept in signal processing, the existing solutions back in the 1970's to 90's were based mostly on heuristics. Surprisingly enough, this concept seems to be absent in modern learning-based approaches, perhaps assuming it can be learned from the data. However, we show that one has to pay high prices in terms of sample complexity to be able to learn it from the data.

Our solution, on the other hand, systematically uses mathematical optimization to inject rate limitations as prior, for which we provide a straightforward iterative algorithm with the analysis of the solution. When regularization is imposed with infinite weight, i.e., when we generate codebooks from random distributions with data-adjusted variances, we achieve surprisingly good solutions. This can be useful for methods requiring very fast learning or methods that require randomness, possibly for privacy preservation reasons.

Our analysis model solution, i.e., the STC framework, is one of the most important contributions of this thesis. As far as this chapter is concerned, this consists of its original formulation\footnote{The original motivation behind the STC framework was for the problem of fast similarity search and as an alternative for binary encoding. This was motivated by the famous $\ell_0$-based sparse coding of analysis dictionaries whose solution based on hard-thresholding can be approximated as the ternarizing operator. When its decoder was later designed, however, we realized it can be applied to a broad set of problems beyond similarity search.}, the derivation of its information-theoretic properties, as well as designing 2 types of decoders\footnote{Here we ignore the non-linear decoder since we did not perform experiments for it.}, for which we characterized rate and distortion. 

For many methods, particularly in binary quantization, the decaying nature of variance after projections is an undesirable phenomenon since the significance of dimensions will be lost. Therefore, these methods try to avoid such projections and spread out the variance in some way. For ternary encoding, on the other hand, this is always desired since it increases sparsity and hence smaller rate. As a result, we can use analytical solutions like PCA to initialize our algorithms (if more data-oriented and heavier training is needed). This is possible thanks to the very important weighting vector mechanism that we introduce, which links discretized (ternary) values with floating-point numbers.
 
\subsubsection*{Chapter 4}  
Our reference to the successive refinement principle from information theory is important. Although the exact prerequisites of these theories are not met in practical settings, they can provide rough analytical hints and can be very inspiring for design. Our emphasis on residual encoding and its application to both families of our algorithms stems from this root. 

While residual encoding has been around for decades, its popularity diminished during the 1990's, perhaps as a result of getting over-trained after a couple of layers. Both of our versions of residual encoding we develop in this chapter, i.e., the RRQ and the ML-STC (and its neural network extension STNets) can be trained for very high dimensions and for arbitrarily large numbers of layers. 

Thanks to the advantageous structure of the ML-STC compared to the RRQ, we can benefit from end-to-end training by benefitting from the large set of practical know-how and intuitive insights developed within the deep learning communities. So while we keep the residual structure of the network (thanks to which we can analyze the role of each layer), we can further perform end-to-end fine-tuning using the back-propagation techniques. We term this evolution as STNets.

A challenging issue with training neural networks with quantization layers is the non-differentiability of quantization functions like the sign function. Existing methods to address this issue are essentially a set of tricks to approximate the quantizers with some smooth counterparts, therefore incurring approximation errors. Our ternarizing operator, however, has a favorable property in this respect. We show that as the sparsity increases, most of the information content will be concentrated in the positions of the non-zeros rather than their values. Therefore, we can safely replace the ternarizing operator with the differentiable hard-thresholding function during back-propagation. In the test time, we use again the ternary values with proper re-weighting. Provided that the sparsity is high, this imposes virtually no approximation loss. This important contribution is not limited to our architecture and may also be useful for a lot of other networks. 

Comparing the STNets with usual deep learning solutions, apart from some architectural differences, e.g., the presence of the weighting vector, the ternarizing operator, or the fact that the final objective is fed with outputs from all layers instead of only the last layer, our proposition has the following fundamental difference:

We do not start from random weights. Instead, we pre-train the model using the ML-STC or ML-STC-Procrustean. Apart from the much faster convergence gain we achieve, this technique also helps with over-fitting as we will show. So, in the end, this leaves us with a range of choices for learning which we can choose w.r.t. the sample size available. In case training data is limited, we keep the pre-trained ML-STC and we do not back-propagate at all. In the opposite case when samples are abundant, we gain considerable speed-up from pre-training.


\subsubsection*{Chapter 5}  
This thesis advocates an alternative solution to the similarity search, i.e., the Sparse Ternary Codes (STC) framework, instead of the binary hashing or the VQ family from the literature. 

We introduce the notion of coding gain for similarity search to assess the effectiveness of an encoding scheme in terms of achieving the triple trade-off of performance-memory-complexity and with information-theoretic concepts. We assess the coding gain of the STC and demonstrate its superiority w.r.t. the binary codes, showing a vast space of possibilities to design the STC, in order to target different aspects of the triple trade-off.

Having shown its rate-distortion trade-offs in earlier chapters, along with its superior coding gain, we propose a practical pipeline for similarity search, which is a middle-ground between the family of binary hashing and VQ. Our proposition can benefit from fast search in the space of ternary codes. This is not achieved for the VQ family and is inferior for binary hashing, compared to the STC. On the other hand, similar to the VQ-based methods, we can benefit from the optimal rate-distortion behavior of our STC to approximate distances; something which is wasteful for binary hashing.

So our pipeline for search is a double-stage procedure, where the first stage is performing a fast search within binary codes and produces a short, but approximative list, while the second stage is refining the list by distance approximation. We then prototyped this pipeline and achieved promising results on million-scale databases. These results are similar or superior to the state-of-the-art package for similarity search such as FAISS from Facebook AI Research, but with much lower complexity, as we argue.
\subsubsection*{Chapter 6} 
As required by our algorithms developed in the previous chapters, an efficient pre-processing is introduced that whitens large images in a database in a very low complexity and by keeping its global structure.   

We then investigated two scenarios for learned image compression. In the first scenario, we aimed at compressing high-resolution RGB images that do not contain a particular structure beyond the class of general natural images. We showed that, even without applying the entire necessary pipeline of image compression, we can learn from only 200 such images and apply the network for the compression of test images showing a performance comparable to the JPEG2000 codec and much superior to the JPEG, as measured with PSNR. This performance is also similar to the state-of-the-art deep networks that are trained on millions of images and whose training time on parallel GPUs exceeds several days to produce only one operational point. The training of our network takes around one hour on a desktop to learn all its 20 operational points. 

In another experiment, we measure the effect of the block-wise division of images into patches on the quality of image compression and conclude that it does seriously degrade the performance, so more effort should be put to avoid patches in the pixel domain.


In the second scenario for image compression, we considered the case where the images are highly structured and beyond the usual structures of natural images. We picked up the case of facial images as an example to show that it is possible to learn to compress the extra redundancy in these images, as we did so by our RRQ structure achieving superior results to the JPEG2000. To the best of our knowledge, state-of-the-art deep networks for compression have not targeted such scenarios, perhaps since training data is not sufficient.

Finally, we considered another aspect of learning to compress and beyond the task of compression itself. We showed that by capturing the statistics of a collection of images through compression, we can help the task of image denoising as well. Provided that the training set consists of clean images, the test set, if contaminated with noise, is denoised as it gets reconstructed. Surprisingly, we showed that this solution can outperform the BM3D in the very noisy regimes. 




\subsubsection*{Chapter 7}  
While the idea of regularization by compression, or more generally, model selection by minimum description length principle exists (and is somehow neglected) in the literature, our proposal is different in three ways: First, we decouple the optimization of compression from the optimization to solve the inverse problem. This brings much more flexibility and is more practical. So a compression network is trained once, is kept fixed and can always be invoked in the solution of inverse problems. This is thanks to the black-box interpretation of the network that we do not need to know its internal structures and can hence use it in a gradient descent-based optimization. The second difference is that our solution is learning-based and data-adaptive. In contrast, since the existing works date back to several decades ago, the modern interpretation of learning is absent in this line of work, and they solve single-example problems. The third difference is that the compression engine can be changed. For example, one could choose from JPEG2000, the ML-STC, RRQ or autoencoder-based compressors as the underlying engine of our iterative algorithm and benefit from their compression properties.
\subsection{Highlights} \label{subsec:intro_contributions_highlights}
Here we highlight some of the contributions of this thesis in the order of their appearance:

\begin{enumerate}
\item Introducing the rate-allocation regularization into the formulation of K-means, solving the optimization problem, analyzing its solution and showing its efficiency in avoiding over-fitting in high-dimensional scenarios.
\item Introducing the framework of Sparse Ternary Codes (STC) as a universal encoding-decoding mechanism for lossy compression of correlated vectors, solving for optimal parameters and characterizing its rate-distortion performance.
\item Developing the introduced synthesis-based and analysis-based prior models under the successive refinement principle as the RRQ and the ML-STC frameworks, and expanding their operational rate-distortion regime to arbitrary values.
\item Proposing a range of effective choices to target different trade-offs of sample complexity and model capacity for both families of methods. In particular, the RRQ fills the spectrum of possibilities by changing the regularization from infinity to zero, while the ML-STC becomes more data-dependent rather than prior-based, by shirting to ML-STC-Procrustean, STNets and STNets-Procrustean frameworks. 
\item Introducing a novel neural network architecture, which is developed from basic components and is capable of significantly reducing the training time and sample-complexity by benefiting from pre-training the basic components, as well as fine-tuning using back-propagation.
\item Proposing a systematic way to be able to back-propagate in the presence of the non-differentiable quantizer function used in the network, i.e., the ternarizing operator, as the result of studying the information concentration properties of this function.
\item Defining the notion of coding gain for similarity search using information-theoretic measures and as a systematic way to measure the efficiency of achieving the triple trade-off between memory, complexity, and performance in the problem of similarity search, as well as showing the superiority of the STC as a viable alternative to binary hashing. 
\item Proposing a decoding mechanism for STC that significantly reduces the computational complexity w.r.t. the exhaustive search, as well as its extension to multiple layers.
\item As compared with the two families of existing solutions in the literature of similarity search, proposing a middle-ground solution between these two families that benefits from efficient search in the space of codes, while at the same time refines the results with accurate estimates of distances, thanks to its excellent rate-distortion performance.
\item Performing learning-based image compression on the two scenarios of high-resolution natural images and domain-specific facial images, and showing promising results for compression and advocating the idea of learned compression, as an alternative to data-agnostic solutions.
\item Injecting the effective prior of learned compression on the problem of denoising of domain-specific data and showing superior performance w.r.t. the state-of-the-art under very noisy conditions.
\item Investigating compressibility as a prior to solve inverse problems and proposing an effective iterative algorithm to achieve it. The algorithm decouples the optimization of compression network from the optimization to solve the inverse problem and is hence very flexible and can be used in many practical scenarios while being able to invoke any compression paradigm as its underlying engine. 


\end{enumerate}

%


\chapter{Image models: literature overview} \label{chapter:ModelingLit}
To achieve different objectives and to target different applications, a multitude of processing tasks should be performed on signals. These tasks, within their application context, try to make sense out of signals in one way or another. Focusing on images in particular, famous examples of these tasks are ``image restoration'', ``image compression'', ``compressive sensing'', ``image recognition'' and ``(content-based) image retrieval''.

Image restoration involves cases, where a physical phenomenon has degraded the quality of the given image, e.g. as in ``image denoising'', where noise has contaminated the image, ``image de-blurring'', where the image at hand is blurred, ``image inpainting'', where parts of the image has been lost or degraded, ``image super-resolution'', where the resolution of the given image is lower than desired. In all these tasks, the objective is to undo the degradation process, perhaps approximatively. Therefore, these tasks are also related to as ``inverse-problems''.

Image compression involves finding a more compact representation for an image than what the direct representation of its pixel numerical values would take from memory. Finding compressive representations is an important focus of this thesis for which we develop different solutions in the chapters \ref{chapter:SingleLayer} and \ref{chapter:MultiLayer}. Focusing on images in particular, we provide image compression solutions later in chapter \ref{chapter:ImCompression}. 

Compressive sensing tries to reduce the number of times image sensors are being used to reproduce an image with a certain quality. This is important, e.g., for applications like medical tomography, where the acquisition process is very slow, expensive and exposes the patient to radiations. So given the under-sampled observations, recovering the original image is the primary objective of this application.

Image recognition involves assigning a semantic label to an image. The procedure is based on a training phase, where different images with assigned labels are present to the algorithm, and then a test phase, where the labels of some other similar images are to be predicted based on the seen examples from the training phase. 

In content-based image retrieval, usually without the availability of categorical labels or keywords, an image is presented to the retrieval system and queries its similar images from a (usually large) collection of images within the system's database. Chapter \ref{chapter:Search} is dedicated to the similarity searching within these databases by reviewing fundamental concepts and providing our contributions.  

These tasks seem to be very diverse, take different forms, and are studied even in different communities. Obviously, not all of them fit within the scope of this thesis. However, it is important to point out that they all use somehow similar principles for their solutions. So the understanding of these common principles might turn out to be mutually beneficial for these applications. Next in section \ref{sec:ModelingLit_Bayesian}, we use the Bayesian framework to somehow conceptualize and unify such efforts. This is a useful start to understand how these problems are posed.

In sections \ref{sec:ModelingLit_SP} and \ref{sec:ModelingLit_ML}, we then provide a generalist literature overview of signal modeling efforts within the signal processing and machine learning communities. Finally, section \ref{sec:ModelingLit_Proposed} positions the ideas used in this thesis with respect to the literature.

\section{Bayesian framework} \label{sec:ModelingLit_Bayesian}

Almost all attempts in signal modeling can somehow be interpreted under the Bayesian framework, either explicitly, or through some of its variations like the empirical Bayes, where signal priors are learned from the data.


The general Bayesian principle involves incorporating and merging two components: First, the evidence or observations, i.e., the given data; and second the prior beliefs, i.e., the signal models or signal decompositions.

Suppose we are given the observation $\mathbf{q}$ that we want to infer an underlying phenomenon $\mathbf{f}$ from it. Within a probabilistic setup involving randomness, this task can be posed as finding the posterior probability distribution $p(\mathbf{f}|\mathbf{q})$. While this might be impractical to calculate directly, the Bayes rule provides us with an alternative:
\begin{equation*}
p(\mathbf{f}|\mathbf{q}) = \frac{p(\mathbf{f}) p(\mathbf{q}|\mathbf{f}) }{p(\mathbf{q})},
\end{equation*}
where $p(\mathbf{f})$ is a (subjective) prior belief about that underlying phenomenon and is injected to our observations along with the likelihood $p(\mathbf{q}|\mathbf{f})$, which is usually much easier to handle than the direct $p(\mathbf{f}|\mathbf{q})$. The estimation of $\mathbf{f}$ can then be formulated as:
\begin{equation*}
\mathbf{f} = \underset{{\mathbf{f}}}{\text{argmax }} p(\mathbf{f}|\mathbf{q}) = \underset{{\mathbf{f}}}{\text{argmax }} \frac{p(\mathbf{f}) p(\mathbf{q}|\mathbf{f}) }{p(\mathbf{q})}.
\end{equation*}

We next try to elaborate on these two components of the Bayesian framework through a very generalist and non-exhaustive narrative of signal modeling literature.

\subsection{Observation setups} \label{subsec:ModelingLit_Bayesian_Obs}
The first component of the Bayesian framework is trying to integrate the evidence provided by the observed data into the problem objective, which requires to somehow take into account the observation setup. For example, for the inverse problems to estimate $\mathbf{f}$ from the degraded observation with known stochastic model $\mathbf{q} = \mathbf{f} + \mathbf{p}$, where $\mathbf{p}$ is a white additive Gaussian noise, the desired  $\mathbf{f}$ is bounded to the observation with $\ell_2$ penalty, i.e., $||\mathbf{f} - \mathbf{q}||_2^2$.

This, however, should be set up, in view of the second element, i.e., the prior, as well. For example in single image denoising, where a prior about the desired image, e.g., sparsity under some bases is to be imposed into the formulation, it might not be applicable to use full-frame high-dimensional image directly. Instead, the image is divided into rectangular divisions, known as patches, and the data fidelity term $||\mathbf{f} - \mathbf{q}||_2^2$, as well as the prior are added to the ensemble of patches. 

Other than computational constraints, where working directly with high-dimensional images may not be feasible, an important reason behind patch-wise division of images is due to the need for an ensemble of training samples for some methods. While a full-frame image can be interpreted as a single vector in $\Re^n$, it can be divided into $p$ patches in $\Re^{\frac{n}{p}}$. This way, it becomes possible to train a dictionary on these $p$ samples or to form the notion of similarity between these patches. This division, however, breaks the global structure of the image and implies a trade-off as whether to keep the global structure, or to have more training samples. This trade-off is characterized by the size of the patches and the stride of the overlap between them.

\subsection{Priors} \label{subsec:ModelingLit_Bayesian_Priors}
For a very broad range of data and signal sources, there are certain structures and priors more or less valid universally. These structures demonstrate themselves in certain ways, e.g.: intrinsic dimension being much less than the ambient dimension, low-entropy nature of signals, inter- and intra-correlations and similarities across signal dimensions and signal realizations, data being spread mostly on some manifold rather than filling the entire space, sparsity of representation under some bases, low-rankedness of the stacked representation, smoothness of neighboring pixel values and still several more. 

While these priors are related in essence and are somehow different manifestations of the same underlying concept, the way they are imposed on the problem formulation and how it changes the optimization procedure makes the algorithms very different. In fact, within the Bayesian framework, rather than the observation component, most of the differences between methods arise from the way they impose priors.

We next provide an intuitive argument and point out different characteristics that make different priors suitable for different regimes of data availability.





\subsubsection{Priors and sample complexity} \label{subsubsec:ModelingLit_Bayesian_Priors_Specificity}
Prior beliefs are highly subjective. They are enforced to the problem formulation based on the knowledge of the designer from that domain and some mathematical convenience needed to actually solve the problem. But how do they influence the quality of the solution?

One significantly important aspect of this question regards the priors and their required sample complexity. This can be studied through the asymptotic consensus theorem: 

Consider two different priors $p_1(\mathbf{f})$ and $p_2(\mathbf{f})$. Assuming they agree on the set of possible values for $\mathbf{f}$, i.e., $\{ \mathbf{f} : p_1(\mathbf{f}) > 0 \} = \{ \mathbf{f} : p_2(\mathbf{f}) > 0 \}$, in the asymptotic case when the number of training samples $N$ is unlimited, we have that:
\begin{equation*}
\lim_{N \rightarrow \infty} p_1(\mathbf{f}|\mathbf{q}) = \lim_{N \rightarrow \infty} p_2(\mathbf{f}|\mathbf{q}). 
\end{equation*}

So given enough samples, a ``good prior'' and a ``bad prior'' will converge to the same posterior in the end. But what if the amount of training samples is limited? Obviously the good prior here is the one that achieves a better posterior with less amount of data. This may require different factors, e.g., the mathematical optimization routine involved and how easily it can be solved, the interpretability of the prior, amenability to analysis, or some other factors.

Now what if the set of possible values of the two priors are different? In particular, assume that $\{ \mathbf{f} : p_1(\mathbf{f}) > 0 \} \in \{ \mathbf{f} : p_2(\mathbf{f}) > 0 \}$, i.e., the prior $p_1$ is more specific than $p_2$ which does not limit the space of possibilities. From the other hand, assume that $p_1$ has very convenient optimization solution and is more straightforward and intuitive for analysis.

Furthermore, assume that the set of ``true answers'' $\mathbf{f}^*$ (for some application) overlaps to some reasonable extent with the domain of $p_1$, i.e., $\mathbf{f}^* \bigcup \{ \mathbf{f} : p_1(\mathbf{f}) > 0 \} \neq \emptyset$, but is entirely contained in the domain of $p_2$, i.e., $\mathbf{f}^* \in \{ \mathbf{f} : p_2(\mathbf{f}) > 0 \}$.

Then which of the $p_1$ or $p_2$ is the good prior? The answer, in fact, depends on the application and the underlying setup. In case the training data is limited, perhaps the prior $p_1$ can achieve some mediocre answer quickly while $p_2$ may not be able to achieve any good-enough answer. From the other hand, when training data is abundant and the computational resources are cheap, prior $p_2$ achieves very good answers in the end, while $p_1$ is stuck with the same mediocre answer.

The above (informal) reasoning is, in fact, the underlying explanation for the behavior of many algorithms in practice. Simpler and more specific priors give rise to interpretable results with rigorous solutions and probabilistic performance guarantees, as well as convergence analyses. However, their simplicity might be too limiting in some cases, leading to only mediocre solutions in practice. For example, it might be too restricting to expect that all data samples are exactly sparse under some over-complete dictionary. 

From the other hand, less specific, off-the-shelf and very high-capacity prior models resist interpretation and rigorous analyses. However, given a large training set, good computational resources and using a lot of practical know-how, they can outperform the first group in some cases, sometimes significantly.

Roughly speaking, the first group of priors has been common mostly in classical signal processing, while the second group is practiced during the last several years and most notably in machine learning and deep learning communities and they managed to attract much attention.

Following this reasoning and by a very rough division, we categorize prior models to two families: The basic models in section \ref{sec:ModelingLit_SP}, and the composite models in section \ref{sec:ModelingLit_ML}.

\section{Basic models} \label{sec:ModelingLit_SP}
Within the last three or four decades, many attempts to model signals and images within the signal processing community can roughly be divided into 2 basic categories, namely the ``synthesis models'' and the ``analysis models''\footnote{Although highly related to these two models, a third category can be considered as an important basic model. This is the category of ``low-rank'' models, which we do not allude to in this thesis. We refer the interested reader to a useful recent review on this topic in \cite{8399563}}. These models have their own advantages and shortcomings and their popularities have changed during the decades. 


Synthesis models try to synthesis signals from a (perhaps sparse) combination of basis vectors while analysis models analyze the behavior of signals in some projected domains.

Let us take the case of image restoration problem\footnote{Although lesser of our attention in this thesis, this problem is perhaps more intuitive for such general treatment.} and see how these two models approach this problem:

\textbf{General degradation model}\footnote{A notation remark: In the literature of signal processing, it is very common to use $\mathbf{Y}$ as a random vector describing the degraded signal and $\mathbf{X}$ as the desired signal (or vice-versa). However, to be coherent with the rest of the thesis, instead, we use $\mathbf{Q}$ and $\mathbf{F}$ for this purpose and use $\mathbf{Y}$ and $\mathbf{X}$ as their coded representations, respectively.} which encapsulates a lot of different applications can be formulated as the following:\footnote{A recent interesting exception is the ``image rendition'' \cite{rendition_milanfar} problem, where the degradation is assumed to be a black-box rather than an explicit model.} 

\begin{equation} \label{eq:ModelingLit_SigDegradation}
\mathbf{Q} = \mathbb{T}[\mathbf{F}] + \mathbf{P}, 
\end{equation}
where $\mathbf{F} \in \Re^n$ is the original signal, $\mathbb{T}[\cdot]\colon \Re^n \to \Re^l $ is a (usually) known degradation operation, $\mathbf{P} \sim \mathcal{N}(\mathbf{0}, \sigma_P^2 \mathbf{I}_l)$ is an additive random perturbation which is assumed to realize \textit{i.i.d.} noise and $\mathbf{Q} \in \Re^l$ realizes the observed degraded signal.

The inverse problem to recover $\mathbf{F}$ from the observed $\mathbf{Q}$ is formulated within the Bayesian Maximum A Posteriori (MAP) estimation as:
\begin{equation}  \label{eq:ModelingLit_MAP}
\begin{aligned}
\hat{\mathbf{f}} &= \underset{{\mathbf{f}}}{\text{argmax }} p(\mathbf{f}|\mathbf{q}) = \underset{{\mathbf{f}}}{\text{argmax }} \frac{p(\mathbf{q}|\mathbf{f}) p(\mathbf{f})}{p(\mathbf{q})}\\
                 &= \underset{{\mathbf{f}}}{\text{argmax }} \log{p(\mathbf{q}|\mathbf{f})} + \log{p(\mathbf{f})}, \\
\end{aligned}
\end{equation}
where, $p(\mathbf{q})$ (a.k.a. the evidence) is assumed to be fixed\footnote{Note that this assumption makes the MAP only a point-wise estimation of the posterior. In the machine learning literature, sometimes the Bayesian estimation may involve the estimation of the full posterior (by calculating $p(\mathbf{q}) = \int{p(\mathbf{q},\mathbf{f}) d\mathbf{f}}$), and then MAP may not be considered exactly as a fully Bayesian approach. In this thesis, however, as is also common in the signal processing literature, we regard MAP as Bayesian, since it anyway merges prior with observations.}, $\log{p(\mathbf{q}|\mathbf{f})}$ is the log-likelihood serving as the data-fidelity which reduces to the $\ell_2$ norm in our setup, and $\log{p(\mathbf{f})}$ is the prior information about the data.

Let us next see how different methods model the prior. 
\subsection{Synthesis model} \label{subsec:ModelingLit_SP_Synthesis}
Synthesis model assumes that random signal $\mathbf{F}$ is synthesized from the column-space of dictionary or codebook $\mathrm{C} \in \Re^m \times \Re^n$ using the code vector $\mathbf{X}$, i.e., $\mathbf{F} \simeq \mathrm{C} \mathbf{X}$. When using such a construction, a general Boltzmann-like\footnote{This is motivated by the maximum entropy principle and the fact that the family of Boltzmann distributions possesses several such properties.} probability distribution can be assumed for $\mathbf{F}$, as the following:

\begin{equation} \label{eq:ModelingLit_SynthesisPrior}
p(\mathbf{f}) = 
     \begin{cases}
       \kappa \exp[-\tau || \mathbf{x}(\mathbf{f})||_p^q] &\quad\text{if } \textbf{f} \in \text{col}(\mathrm{C}) \\
       0 &\quad\text{otherwise, } \\
     \end{cases}
\end{equation}
where $ \mathbf{x}(\mathbf{f})$ is the appropriate code to synthesize a realization $\mathbf{f}$, $\tau$ and $\kappa$ are constants, $||\cdot||_p^q$ is based on the $\ell_p$-norm and can be interpreted as an energy functional and $\text{col}(\mathrm{C})$ designates the column-space of $\mathrm{C}$.

Based on this setup, the MAP-estimation procedure applied to the general restoration problem described in Eq. \ref{eq:ModelingLit_SigDegradation} is derived as:

\begin{equation} \label{eq:ModelingLit_MAPSynthesis}
\begin{aligned}
\hat{\mathbf{f}} &= \underset{{\mathbf{f}}}{\text{argmax }} \log{p(\mathbf{q}|\mathbf{f})} + \log{p(\mathbf{f})} \\
                 &= \mathrm{C} \cdot \underset{\mathbf{x}}{\text{argmin}} \Big[ ||\mathbf{q} - \mathbb{T}[\mathrm{C}\mathbf{x}]||_2^2 + 
\mu|| \mathbf{x}||_p^q \Big],
\end{aligned}
\end{equation}
where we used the Bayes rule, monotonicity of the $\log(\cdot)$ function, and took into account the Gaussianity of noise in the degradation, the synthesis model and the prior Eq. \ref{eq:ModelingLit_SynthesisPrior}. 


Let us take a particular case of Eq. \ref{eq:ModelingLit_SynthesisPrior} as prior, which is when $p = 0$ and $q = 1$. The pseudo-norm $\ell_0$ in $||\mathbf{x}||_0$ counts the number of non-zeros of $\mathbf{x}$. In fact, this particular case is conceptually very important since $\ell_0$ norm quantifies the sparsity of $\mathbf{x}$. In fact, sparsity and parsimonious representations are of fundamental significance in signal modeling. 

Setting the degradation as the special case $\mathbf{q} = \mathbf{f} + \mathbf{p}$, the MAP estimation problem to find the code $\mathbf{x}$ becomes:

\begin{equation} \label{eq:ModelingLit_SparseCodingL0}
\hat{\mathbf{x}} = \underset{\mathbf{x}}{\text{argmin}}  ||\mathbf{q} - \mathrm{C}\mathbf{x}||_2^2 + 
\mu|| \mathbf{x}||_0 ,
\end{equation}
which is referred to as the ``sparse coding problem'', an active field of research in the late 1990's and the 2000's.

If $\mathrm{C}$ is a square matrix whose columns are linearly independent (or equivalently its rank is complete), basic linear algebra tells us that any vector $\mathbf{f} \in \Re^n$ can be exactly described under $\mathrm{C}$. However, this does not provide interesting solutions, i.e., sparse representation and learning is not feasible in this case since it does not guarantee sparse description of $\mathbf{f}$ under $\mathrm{C}$.

In order for $\mathbf{x}$ to be sparse, informally speaking, the dictionary should resonate well with the signal. For example, the DCT matrix produces sparse representation for periodic signals since its basis vectors are cosine waves with varied frequencies. However, in more realistic cases, where $\mathbf{f}$  is more complicated than that, the choice of orthonormal DCT as the dictionary does not provide both sparse and accurate representations.

Therefore, the only way to have a sparse representation under a synthesis model that is accurate enough in terms of approximation is to have $\mathrm{C}$ as an over-complete matrix (that moreover resonates well with $\mathbf{f}$).

This, however, brings many challenges and is the main theme of thousands of research papers during two to three decades. In order to realize a bit of such challenges, notice that solving Eq. \ref{eq:ModelingLit_SparseCodingL0} by restricting the sparsity to $||\mathbf{x}||_0 = k$ will amount to $m \choose k$ possibilities, only to determine the position of non-zeros of $\mathbf{x}$. This problem, as is, is shown to be NP-hard due to its combinatorial nature. 


Next, let us see how this problem is approached in the literature.

\subsubsection{Sparse coding algorithms} 
Motivated by the potential applications that it can serve, e.g., in compressive sensing, medical tomography, seismic signal processing, neuroscience, and dictionary learning, along with its theoretical elegance, this problem has attracted much attention within engineering and applied math.

Without any attempt to be exhaustive, here we mention some of the key approaches to this problem.

A group of methods relaxes the original $\ell_0$ pseudo-norm in Eq. \ref{eq:ModelingLit_SparseCodingL0} with the convex $\ell_1$ norm. This relaxed version of the problem, a.k.a., the basis-pursuit problem, has an important advantage that it is now a convex program. This means that it can be solved for global optimum with polynomial complexity. So thanks to the $\ell_1$ term, the whole problem can be cast as a linear program and solved using interior-point methods.

How much is the relaxed problem related to the original problem? It has been shown in several works, e.g., \cite{donoho2003optimally} that for very low sparsity values, the basis-pursuit problem is equivalent to the original sparse coding problem.

A limiting difficulty with these methods, however, is due to the complexity of the optimization procedure. In the case of linear programming, e.g., thousands of constraints with millions of variables are involved to solve a medium-size problem. This limits their application to low-dimensional problems.

A group of methods follows the greedy approach.  The idea is to abandon the exhaustive search for the best solution for some locally optimal solutions. While many variants are possible, a notable example is the OMP \cite{1337101} that iteratively improves the estimate of the signal by choosing the column of $\mathrm{C}$ that has the highest correlation with the residual. 

These methods are widely used, yet their performance is limited. In fact, their performance guarantees are very pessimistic and limited to low sparsities. Moreover, they are not successful for high-dimensional applications.

An important family of methods iterates by applying hard or soft thresholding functions on top of a set of basic algebraic operations. These are the instances of the famous proximal methods in optimization, where, despite their easy structure consisting of the repetition of basic operations, they are very successful in practice. In fact, when it comes to high-dimensional problems, these methods are of significant importance.

A basic intuition behind these algorithms is to iteratively move towards the direction of the gradient of the observation term, while at each step a thresholding operator is applied to satisfy the sparsity-based prior. In the case of $\ell_0$-based sparsity, the sparsifying operator is the hard-thresholding function $\psi_{\lambda}(\cdot)$, and in the case of $\ell_1$-based sparsity, the soft-thresholding function $\eta_{\lambda}(\cdot)$ is applied. Notable examples of this family are the IHT \cite{IHT:BLUMENSATH2009265} algorithm and the ISTA \cite{ISTA:661182}, the basic forms of which are depicted in the iterative procedures in Eq. \ref{eq:ModelingLit_IHT} and Eq. \ref{eq:ModelingLit_ISTA}, respectively:
\begin{subequations}
\begin{align}
\mathbf{x}^{(t+1)}  &= \psi_{\lambda} \Big(  \mathbf{x}^{(t)} + \mathrm{C}^T[\mathbf{q} - \mathrm{C} \mathbf{x}^{(t)}  ] \Big), \label{eq:ModelingLit_IHT}\\
\mathbf{x}^{(t+1)}  &= \eta_{\lambda} \Big(  \mathbf{x}^{(t)} + \mathrm{C}^T[\mathbf{q} - \mathrm{C} \mathbf{x}^{(t)}  ] \Big). \label{eq:ModelingLit_ISTA}
\end{align}
\end{subequations}

Without going into further details of this interesting topic, we direct the reader to useful resources like \cite{Bach:2012:OSP:2185817.2185818,maleki2010approximate,Elad:2010:SRR:1895005}.

\subsubsection{Dictionary learning} \label{subsubsec:ModelingLit_SP_Synthesis_DicLearning}
The promise of synthesis-based Bayesian estimation of Eq. \ref{eq:ModelingLit_MAPSynthesis} was based on the availability of a good dictionary $\mathrm{C}$. For example, in the important case of $\ell_0$-based optimization of Eq. \ref{eq:ModelingLit_SparseCodingL0}, an accurate and sparse solution can be achieved if the (perhaps over-complete) dictionary $\mathrm{C}$ resonates well with $\mathrm{f}$. 

Apart from the famous orthogonal alternatives like the DCT, DFT, and DWT, several attempts have targeted the design of good over-complete dictionaries, e.g., the Curvelets \cite{candes2000curvelets} or the Contourlets \cite{1532309} that design dictionaries that can better capture the statistical properties of natural images.

While this is an important step towards a better approximation of natural images, a more radical solution is to learn the dictionary entirely from the data. So while for every $\mathbf{f}_i$ a sparse representation is sought, the collection of the data samples, say a matrix $\mathrm{F} = [\mathbf{f}_1, \cdots, \mathbf{f}_N]$ is used to train dictionaries.

To see this idea, let us relax the problem of restoration to the special case of approximation (i.e., putting $\mathbf{q} = \mathbf{f}$ in Eq. \ref{eq:ModelingLit_SigDegradation}). The joint optimization of the dictionary, as well as the sparse coding problem can be formulated as:

\begin{equation} \label{eq:ModelingLit_K-SVD}
\begin{aligned}
& \underset{{\mathrm{C},\mathrm{X}}}{\text{minimize}}
& & \frac{1}{2Nn}|| \mathrm{F} - \mathrm{C} \mathrm{X} ||_{\mathcal{F}}^2\\
& \text{s.t.} & & ||\mathbf{x}_i||_0 \leqslant k,  \\
&             & & \text{for } i = 1, \cdots, N,
\end{aligned}
\end{equation} 
where the code vectors $\mathbf{x}_i$'s (corresponding to $\mathbf{f}_i$'s) are gathered in $\mathrm{X} = [\mathbf{x}_1, \cdots, \mathbf{x}_N]$.

Eq. \ref{eq:ModelingLit_K-SVD} is referred to as sparse (synthesis) dictionary learning problem and, after the success of the sparse coding problem, has been an active field of research, particularly since mid-2000's.

A notable example of this line of work is the K-SVD algorithm \cite{1710377}, which has been very successful. While the formulation of Eq. \ref{eq:ModelingLit_K-SVD} is the generalization of the famous K-means algorithm that we will encounter later in Eq. \ref{eq:SingleLayer_Kmeans}, the solution of K-SVD bears some similarity with that of K-means by replacing the code update step with a sparse coding algorithm like the OMP and the codebook update step with the Singular Value Decomposition (SVD). 

Now that we have seen a glimpse of the basic synthesis prior models, let us next consider the other alternative, i.e., the analysis prior model.  
\subsection{Analysis model} \label{subsec:ModelingLit_SP_Analysis}
While the synthesis model ``synthesizes'' the signal using columns of a dictionary, analysis model ``analyses'' the data within a projection domain, i.e., $\tilde{\mathbf{f}} = \mathrm{A} \mathbf{f}$, where $\mathrm{A} \in \Re^m \times \Re^n$ is a projection matrix. Examples of $\mathrm{A}$ are derivative-based operators like the Laplacian to encourage smoothness (e.g., as in the Total Variation \cite{RUDIN1992259} approach), or wavelet-based projection whose rows contain spatial derivatives of varying scale.

A general Boltzman-like distribution can then be assumed for $\mathbf{f}$ as :
\begin{equation}  \label{eq:ModelingLit_Analysis_BoltzmanDist}
p(\mathbf{f}) = \kappa \exp[-\tau || \mathrm{A} \mathbf{f}||_p^q],
\end{equation}
which assigns higher probability to e.g., smooth images.

Based on this analysis prior, the MAP estimation then becomes:

\begin{equation}  \label{eq:ModelingLit_MAPAnalysis}
\hat{\mathbf{f}} = \underset{\hat{\mathbf{f}}}{\text{argmin}} ||\mathbf{q} - \mathbb{T}[\mathbf{f}]||_2^2 + \mu || \mathrm{A} \mathbf{f} ||_p^q
\end{equation}

An important special case is then in the sparse encoding of $\mathbf{f}$ and reduces as: 

\begin{equation} \label{eq:ModelingLit_AnalysisL0Code}
\begin{aligned}
\mathbf{x} &= \underset{{\mathbf{x}}}{\text{argmin}}
& & ||\mathrm{A}\mathbf{f} - \mathbf{x}||_2^2\\
& \text{s.t.} & & ||\mathbf{x}||_0 = k. 
\end{aligned}
\end{equation}

Eq. \ref{eq:ModelingLit_AnalysisL0Code}, however, unlike its synthesis counterpart of Eq. \ref{eq:ModelingLit_SparseCodingL0} is solved very easily and in closed-form as:

\begin{equation}  \label{eq:ModelingLit_AnalysisL0CodeSolution}
\mathbf{x}^* = \psi_{\lambda} \big( \mathrm{A} \mathbf{f} \big), 
\end{equation}
for which the hard-thresholding function of Eq. \ref{eq:math_HardThresholdingFunc} chooses its threshold $\lambda$ such that $||\mathbf{x}||_0 = k$. 

In general, thresholding signal in the projected domain using data-independent and analytical transforms is one of the most classical and successful practices in signal processing. Although their popularity was under-shadowed by learning-based synthesis dictionary learning for some time, there was a resurgence of analysis models in the 2010's, but this time using data-adaptive projections as we describe next.
\subsubsection{Transform learning} \label{subsubsec:ModelingLit_SP_Analysis_TransformLearning}
Following the success of synthesis-based dictionary learning, the idea of going beyond analytically-generated transforms by adapting them to the data appeared, also for the analysis model.

So, given a stack of data-points $\mathrm{F} = [\mathbf{f}_1, \cdots, \mathbf{f}_N]$, the sparsifying transform learning problem can be posed as:
\begin{equation} \label{eq:ModelingLit_TransformLearning}
\begin{aligned}
& \underset{{\mathrm{A},\mathrm{X}}}{\text{minimize}}
& & \frac{1}{2Nn}|| \mathrm{A} \mathrm{F} - \mathrm{X}  ||_{\mathcal{F}}^2\\
& \text{s.t.} & & ||\mathbf{x}_i||_0 \leqslant k,  \\
&             & & \text{for } i = 1, \cdots, N,
\end{aligned}
\end{equation} 
where the codes are stacked as $\mathrm{X} = [\mathbf{x}_1, \cdots, \mathbf{x}_N]$.

An early attempt to pose such kind of problems was in \cite{7074220}, where the authors considered the noiseless case.

An important work was in \cite{6466951,6339108}, where they introduced the penalty:
\begin{equation*}
\Omega\{ \mathrm{A}  \} \colon -\mu_1 \log \det \mathrm{A} + \mu_2 || \mathrm{A} ||_{\mathcal{F}}^2,
\end{equation*}
to regularize the square transform for full-rank and better-conditioned solutions (as a less restrictive alternative to positive definite matrices), and avoid trivial solutions like repeated rows. They then proposed a conjugate gradient-based algorithm and showed that it outperforms image denoising using K-SVD while being much faster.

This line of work is then extended, e.g., in  \cite{6638690,7045534} by proposing a closed-form solution (after slightly changing the objective function), or in \cite{6638226} by extending the concept to over-complete transforms.

%
%
%
%
%
%
%
%
%

\subsection{Synthesis versus analysis priors} \label{subsec:ModelingLit_SP_versus}
Now that we have reviewed the two basic prior models, a natural question arises as to how do these models compare. While they have been shown to be related, and even equivalent under some (restricting) conditions, there is no clear answer to this question in general terms. A useful work shedding light to the relation between analysis and synthesis priors is \cite{0266-5611-23-3-007}, which leaves the general answer as open. 

So finally which of these two basic prior models do we choose? The choice is very subjective and should be decided depending on the setup and the application at hand.

In this thesis, we build on top of both of these models and adapt them to our general objective of having compressed and discrete representations. While we do not give a decisive answer to this question, we point out that the analysis model is more compatible with neural network architectures for which excellent practical knowledge has been built during the last several years. Therefore, when it comes to building composite models out of basic ones, we tend to be more in favor of analysis models, since they can benefit from back-propagation.\footnote{On the other hand, one can argue that synthesis models are more compatible with Expectation-Maximization framework and can benefit from EM-like solutions.}

\section{Composite models} \label{sec:ModelingLit_ML}
Not all efforts to address our problems of interest follow the structures we have seen in the previous section. In fact, some of the state-of-the-art results reported during the last several years deviate from the category of basic synthesis or analysis models. These models can be considered as composite structures made up of basic ones, and for which the inference procedure is more involved.

As discussed earlier in section \ref{subsubsec:ModelingLit_Bayesian_Priors_Specificity}, it is expected that more complex priors lift the limitations of basic ones by providing a richer space of parameters. On the other hand, their required sample complexity is not minimal, moreover, their analysis and interpretation is not straightforward.  

In section \ref{subsec:ModelingLit_ML_PriorDecomp}, we make an effort to understand these models by decomposing them into simpler parts. Then in section \ref{subsec:ModelingLit_ML_Literature}, without making the effort to provide a structured review, we recount some highlights from the literature, and in particular, the application of complex models to solve inverse problems. 
\subsection{Decomposition of priors} \label{subsec:ModelingLit_ML_PriorDecomp}
Priors need not be limited to the basic forms we saw in the previous section. For example, the image distribution may be different from the Boltzmann-like distribution of Eq. \ref{eq:ModelingLit_Analysis_BoltzmanDist}, or may be composed in a different way, perhaps from some simpler elements.

This can consist of different levels of interaction with the data and conditioning on the results of the previous stages or conditioning on different parameter sets. While it is not straightforward to model the structure of such composite priors in general, let us next see a very abstract way to explain this. 

Denote the set of parameters of the model as $\theta \in \Theta$. For example, the analysis prior of Eq. \ref{eq:ModelingLit_Analysis_BoltzmanDist} was parametrized by the projection $\mathrm{A}$, or the synthesis model of Eq. \ref{eq:ModelingLit_SynthesisPrior} was parametrized by the dictionary $\mathrm{C}$. So, for both models, we can decompose the prior symbolically as $p(\mathbf{f}; \theta)$.\footnote{When learning these parameters, these models are referred to as parametric models in the statistical machine learning literature since they have a clear parameter set. This is in contrast with non-parametric models whose parameter set can adapt to the data. However, we prefer not to use this terminology here and do not make an explicit distinction between them, as they happen to be vague for some cases.}

One can think of going beyond this rigid prior and think of composite models. For example, we can consider an $L$-stage decomposition of the prior, where each stage is conditioned on previous stages of prior modeling as:
\begin{equation*}
p(\mathbf{f};\theta) = p(\mathbf{f}^{[0]};\theta^{[1]}) p(\mathbf{f}^{[1]}|\mathbf{f}^{[0]},\theta^{[1]};\theta^{[2]}) \cdots p(\mathbf{f}^{[L-1]}|\mathbf{f}^{[L-2]}, \cdots, \mathbf{f}^{[0]}, \theta^{[L-1]}, \cdots, \theta^{[1]};\theta^{[L]}),
\end{equation*}
where $\mathbf{f}^{[0]} = \mathbf{f}$ is the given data, and $\mathbf{f}^{[1]}, \cdots, \mathbf{f}^{[L-1]}$ are the inputs to the second to the last stages of processing. These layers are parameterized each by the parameter sets $\theta^{[1]}, \cdots, \theta^{[L]}$, respectively.

This is a very general and intricate decomposition for the priors that may lead to over-complicated structures. Simplifications should be imposed by relaxing the conditioning within the layers.

One possible simplification can be considered by assuming a sort of Markovian assumption. This leads to the decomposition of the prior as:
\begin{equation}  \label{eq:ModelingLit_GeneralMarkovPrior}
p(\mathbf{f};\theta) = p(\mathbf{f}^{[0]};\theta^{[1]}) p(\mathbf{f}^{[1]}|\mathbf{f}^{[0]};\theta^{[2]}) \cdots p(\mathbf{f}^{[L-1]}|\mathbf{f}^{[L-2]};\theta^{[L]}),
\end{equation}  
which can be realized in various ways as we see next.
\subsubsection*{Feed-forward neural networks} 
A prominent instance of the prior decomposition of Eq. \ref{eq:ModelingLit_GeneralMarkovPrior} is the family of feed-forward neural networks. These networks are characterized as:
\begin{equation} \label{eq:ModelingLit_FeedForward}
\begin{aligned}
\mathbf{f}^{[1]}              &= \sigma^{[1]} \Big( \mathrm{A}^{[1]} \mathbf{f}^{[0]} + \mathbf{b}^{[1]}\Big), \\ 
                 &\vdots             \\ 
\mathbf{f}^{[L]}              &= \sigma^{[L]}\Big(\mathrm{A}^{[L]} \mathbf{f}^{[L-1]} + \mathbf{b}^{[L]} \Big),        
\end{aligned}
\end{equation}
where $\mathrm{A}^{[1]}, \cdots, \mathrm{A}^{[L]}$ are projection matrices for the first to the last layers, $\mathbf{b}^{[1]}, \cdots, \mathbf{b}^{[L]}$ are bias vectors, and $ \sigma^{[1]}(\cdot), \cdots  \sigma^{[L]}(\cdot)$ are non-linear but differentiable functions applied to the projections, respectively.

These networks are trained by first forming a cost function composed of $\mathrm{f}^{[L]}$ and, perhaps a set of labels in supervised scenarios, or $\mathrm{f}^{[0]}$ in unsupervised cases and in autoencoder networks, along with optional regularization consisting of norms on the projection matrices. The cost function is then minimized usually using variants of gradient descent by differentiating w.r.t. the parameters of the network, i.e., $\mathrm{A}^{[1]}, \cdots, \mathrm{A}^{[L]}$ and $\mathbf{b}^{[1]}, \cdots, \mathbf{b}^{[L]}$. The differentiation is performed using the back-propagation technique which is perhaps the most important element behind the success of neural networks. This is essentially the chain rule in multivariate calculus which is applicable, thanks to the structure of the prior in Eq. \ref{eq:ModelingLit_GeneralMarkovPrior}.  

The configuration of the network is very flexible. Depending on the choice of the projection matrices and the non-linearities, the feed-forward structure can take different forms. For example, $\mathrm{A}^{[l]}$, can be an unstructured matrix, which leads to the so-called perceptron layer, or cyclic convolution matrix, which leads to a convolutional layer. 

The non-linearities also play crucial roles in the network. Famous choices for $\sigma(\cdot)$ are the sigmoid, tanh and Re-Lu functions.
\subsubsection*{Residual networks} 
As the number of layers increases, training of feed-forward networks becomes increasingly involved. This is due to a set of factors leading to the vanishing (or exploding) gradient phenomenon. A remedy was proposed in \cite{ResNet:7780459} by introducing the so-called skip-connections that, along with the output of layer $l$, redirects the output of layer $l-1$ to the input of layer $l+1$ of the network. This is equivalent to relaxing the Markovian simplification of Eq. \ref{eq:ModelingLit_FeedForward} to conditioning on the outputs of other layers, apart from the previous layer.

This evolution has been very successful and made the training of networks with many layers feasible.

The literature of neural networks and deep learning is filled with different recipes and practical insights on how to better train these complex learning machines. This is because of their very complicated structure for which there does not exist enough theoretical understanding and explanation. 

This field has received an unprecedented amount of attention from academia and industry. So the research and practice in this field has explosively expanded during the last decade. We refer the reader to \cite{Goodfellow-et-al-2016} that reviews the key achievements in this field.
\subsection{Literature review} \label{subsec:ModelingLit_ML_Literature}
In section \ref{sec:ModelingLit_SP}, we have seen the basic formulation of MAP estimation and how the basic prior models can be added to the objective function, e.g., as in Eq. \ref{eq:ModelingLit_MAPSynthesis} or Eq. \ref{eq:ModelingLit_MAPAnalysis}. 

For the composite models, on the other hand, this can be done in many different ways. Since these models are very flexible, and also the fact that they can benefit more from the availability of data than the basic models, one can consider many different scenarios.

Denote the equivalent network architecture of Eq. \ref{eq:ModelingLit_FeedForward} as $\mathbf{f}^{[L]} = \mathbb{N}_{\theta}(\mathbf{f}^{[0]})$, where $\mathbb{N}_{\theta}(\cdot)$ is a network with paramaters symbolized as $\theta$. One can generate a large number of degraded-clean pairs of images and learn to map the degraded images to clean ones by learning a network $\mathbf{f} = \mathbb{N}_{\theta}(\mathbf{q})$ on the training pairs. This, however, needs to learn one network for any degradation level. Attempts to learn one network for all degradations requires learning very complex networks. One such effort is in \cite{NIPS2016Mao}, where they train a 30-layer network to learn to denoise several contamination levels, simultaneously. Another effort is in \cite{Tai-MemNet-2017}, where they train a very complex model consisting of 80 layers with memory units. 

Another possibility is to train $\hat{\mathbf{q}} = \mathbb{N}_{\theta_1}(\mathbf{q})$ on the set of degraded images and then train $\hat{\mathbf{f}} = \mathbb{N}_{\theta_2}(\mathbf{f})$ on the set of clean images. A third network can then be learned to map the set of parameter pairs $\theta_1$ and $\theta_2$, i.e., $\theta_2 = \mathbb{N}_{\theta_3}(\theta_1)$. An example of such effort is in \cite{7339460} and for the task of single-image super-resolution.

One other possibility is to learn the network parameters $\theta$ on a large set of clean images, e.g., using an autoencoder network, i.e., a network trained with reconstruction distortion as cost function, and use the trained $\hat{\mathbf{f}} = \mathbb{N}_{\theta}(\mathbf{f})$ as the prior term $p(\mathbf{f};\theta)$ in Eq. \ref{eq:ModelingLit_MAP}. Examples of these efforts can be found in \cite{6126278,7410393}.

Another very interesting line of work tries to benefit from the learning capabilities of composite models in enriching the solution of the basic models. The idea is to take the solution structure of basic models and implement it using a neural network. The parameters of the model are trained using input-output pairs provided from the basic model. This is done using the idea of ``unfolding'', which expands the iterative solutions in several iteration steps. Notable examples of this line of work are in \cite{gregor2010learning} that unfolds the solution of ISTA of Eq. \ref{eq:ModelingLit_ISTA} into several time-steps, and in \cite{xin2016maximal} that unfolds the IHT of Eq. \ref{eq:ModelingLit_IHT} and solves it with a neural network, lifting some limitations of IHT w.r.t. dictionary coherence.

The recent work of \cite{ulyanov2018deep} reveals a very interesting fact about neural structures used in image processing. They show that, contrary to the common understanding, the success of these structures and in particular convolutional networks, is not due solely to their ability to train useful priors from the data, but also from the structure itself. In other words, the structure itself is a very powerful prior, even when not trained on large collections of images. After all, these structures have been evolving constantly over the years and it already resonates with the data very well. They then show that operating on a single image, the network prior can solve a range of inverse problems.


%
%
%
%
%
%
%
\section{This thesis: the general picture} \label{sec:ModelingLit_Proposed}
Now that we have seen some ideas on signal modeling from the literature, we should also mention the strategies of this thesis regarding the choice of the priors. So among the basic models and composite models, which ones do we choose to serve our applications? 

\subsection{Basic vs. composite models} \label{subsec:ModelingLit_Proposed_BasicComplex}

Let us first review some of our understandings of these models in comparison to each other.

In general, basic models are minimal in terms of sample complexity while composite models require more samples, sometimes much more than necessary. This is further related to the computational complexity, at least during the training phase, which can be very exaggeratedly high for certain complex models. However, even in the most modern setups, one can always consider applications, where samples are not abundant or computational resources are scarce, or when quick decisions should be made, perhaps since the data statistics is rapidly changing. Therefore, this is a shortcoming in general for the complex models.

On the other hand, however, composite models have shown to be capable of benefiting from the presence of larger amounts of data. It has been observed in multiple tasks that incorporating million-scale data into learning can noticeably boost the performance of complex models w.r.t., say thousand-scale data. Basic models, however, may saturate in performance at a much lower regime, which, in many cases, is considered as a serious and limiting disadvantage.

Moreover, composite models are very flexible and can be incorporated in many different setups, some of which we have mentioned in section \ref{subsec:ModelingLit_ML_Literature}. In general, one can decompose the whole learning or inference task at hand into many smaller components, each of which conditioned on some others and try to learn the components using an appropriate neural network structure.  

\subsubsection*{Our strategy} 

Reviewing different ideas from the literature and seeing different instances of both of these extremes, we can come into an important conclusion:

Although it is expected, ideally, that there should exist a spectrum of different choices in the ground of models between being basic and composite, this, however, is not exactly the case. In fact, the literature seems to be quite disjoint and these two families do not seem to be able to benefit from one another using well-established ways. 

For example, while within the signal processing literature, the concept of sparsity is very well established and whose importance is very well investigated and proven, deep learning does not seem to benefit from this concept as much. In fact, most important results within deep learning are achieved with models without sparsity or related structures like low-rankedness. Similarly, a lot of the key achievements from deep learning do not seem to be transferable to the domain of sparse representations.

This thesis, of course within the limitations of its scope, makes some effort to try to somehow bridge this gap.\footnote{We should also mention some efforts with similar objectives from the literature, e.g., \cite{mallat2013deep,papyan2016convolutional}.} We aim at using basic models initially, and then building on top of them gradually as there appear more samples and computational resources. This way, ideally, we transfer all the insights of the basic models to their composite versions while benefiting from the high capacity of composite models and the well-established techniques to train them.

We pursue this strategy through the framework of successive refinement from information theory, as we briefly point out next.
\subsection{The successive refinement framework as a prior model} \label{subsec:ModelingLit_Proposed_SR}
Other than the feed-forward model of Eq. \ref{eq:ModelingLit_FeedForward}, the prior decomposition of Eq. \ref{eq:ModelingLit_GeneralMarkovPrior} can be instantiated in other ways. One particular such possibility is the residual structure summarized as:
\begin{equation} \label{eq:ModelingLit_RecursionSR}
\begin{aligned}
\mathbf{f}^{[1]} &= \mathbf{f}^{[0]} - \mathbb{Q}^{{[1]}^{-1}}\big[  \mathbb{Q}^{[1]} \big[\mathbf{f}^{[0]} \big]  \big], \\
\vdots \\
\mathbf{f}^{[L]} &= \mathbf{f}^{[L-1]} - \mathbb{Q}^{{[L]}^{-1}}\big[  \mathbb{Q}^{[L]} \big[\mathbf{f}^{[L-1]} \big]  \big], \\
\end{aligned} 
\end{equation}
where $\mathbb{Q}^{[l]}[\cdot]$ and $\mathbb{Q}^{{[l]}^{-1}}[\cdot]$ are the encoder and the decoder of the layer $l$, respectively.\footnote{Notice that this structure is different from the Residual Network framework of \cite{ResNet:7780459}, in several ways.}

This, in fact, is a foundational skeleton for composite models that we develop throughout this thesis and whose idea comes from the framework of successive refinement from information theory. While the building-block of encoder-decoder pairs that we use will be developed in chapter \ref{chapter:SingleLayer}, the concept of successive refinement, as well as the residual structure above will be clarified in chapter \ref{chapter:MultiLayer}.

As an intuitive explanation, our general assumption behind the adoption of this prior is that the data, however complicated in its internal statistical structure or its underlying manifold, can be explained in several layers of processing, where each layer describes the data only very roughly. This is validated in all our experiments, where we observe that after multiple layers, the data to be encoded, effectively follows the \textit{i.i.d.} Gaussian noise structure. 



\section{Conclusions} \label{sec:ModelingLit_Conclusions}

In this chapter, we reviewed some fundamental concepts from signal and image modeling and provided several important instances of the existing solutions from the literature. 

We framed all attempts under the Bayesian framework that tries to systematically merge two sources of knowledge: the data and the prior. 

While the data can be incorporated to the Bayesian objective in various ways, we saw that most variations that differentiate between algorithms come from the way the prior knowledge is incorporated into the formulation and its consequences on how the solution is actually achieved through mathematical optimization.
 
We argued that the choice of the prior directly influences the quality of the solution w.r.t. the training samples available. In particular, we noticed that methods from the literature can roughly be divided into two broad categories: the basic and the composite models, as we termed them. While the basic models are more intuitive to understand and analyze, faster to train and perhaps require less number of samples, the composite models, on the other hand, can benefit from the availability of larger amounts of training data and can provide with better solutions under this regime.

Basic models are more common in signal processing communities and can roughly be divided into two families of synthesis and analysis priors. While each of them has their own particularities, excellent theoretical treatments already exist for both. On the other hand, the composite models have developed rather within the machine learning and deep learning communities. This has provided us with excellent practical know-how leading to an advanced technology capable of achieving auspicious results.

We observed, however, that there does exist a noticeable gap between the two. In particular, we do not seem to be able to develop composite models by building on top of the basic ones in a systematic way. In fact, the performance of basic models seems to have somehow saturated, while our understanding of composite models seems to be only practical.


This thesis adopts its strategy for signal modeling as trying to develop composite models by repeatedly invoking basic ones. We realize this idea using the additive residual structure which is rooted in the concept of successive refinement of information. We pursue this idea in the next part of the thesis.


%
%
%

\part{Algorithms}
\chapter{Single-layer architectures} \label{chapter:SingleLayer}

In chapter \ref{chapter:ModelingLit}, we saw a general literature overview of signal modeling and how for a broad variety of tasks, similar ideas for signal decomposition and prior modeling can be framed under the Bayesian paradigm. We further sketched a general picture of the strategies of this thesis in using priors and how they relate to the considered literature.

In this chapter, we first conceptualize a general framework in section \ref{sec:SingleLayer_General} to encompass most of the objectives and ideas followed in this thesis. Later in the third part of the thesis, the different flavors of these ideas show up when addressing several applications. Inspired by signal processing literature, we next pursue the solutions to these general problems by making them more concrete within two general strategies, the synthesis and the analysis prior models. Our synthesis model treatment leads us to the Variance-Regularized K-means (\textit{VR-Kmeans}) algorithm in section \ref{sec:SingleLayer_Synthesis} and our analysis formulation leads to the Sparse Ternary Codes (STC) of section \ref{sec:SingleLayer_Analysis}. We start the development of these algorithms by assuming an underlying probability distribution for the data. We then lift these assumptions and gradually shift to more data-dependent solutions.

While the algorithms developed in this chapter follow a structure that we refer to as a ``single-layer architecture'', we will see their limitations and make them more intricate and powerful in chapter \ref{chapter:MultiLayer}, where we discuss ``multi-layer architectures''.
\section{General objective: encoder-decoder pair} \label{sec:SingleLayer_General}

For a lot of purposes relevant to this thesis, it can be very useful to encapsulate different objectives under the ``encoder-decoder'' split. This is defined as follows:

Consider an encoder $\mathbb{Q}[\cdot]: \Re^n \to \mathcal{X}^m$ that assigns a code $\mathbf{x} = \mathbb{Q}[\mathbf{f}]$ to a vector $\mathbf{f} \in \Re^n$. The idea is to limit the entropy of representation from $\Re^n$ to a lower-entropic space $\mathcal{X}^m$, which is not necessarily a Hilbert space, perhaps for some coding or mapping efficiency. Furthermore, for some applications, we might be interested in efficiently storing and indexing this representation in memory. Therefore, we may also choose a discretized alphabet for $\mathbf{x}$.

For this basic setup, a general optimization objective would be to minimize some cost function $c(\cdot): \Re^n \to \Re$, that measures the deviation of the data w.r.t. some (perturbed) observation $\mathbf{q}$ as:

\begin{equation} \label{eq:SingleLayer_GeneralEnc}
\begin{aligned}
& \underset{{\mathbb{Q}[\cdot]}}{\text{minimize}}
& & c(\mathbf{f},\mathbf{q})\\
& \text{s.t.} & & \Omega \{\mathbf{x}\},   \\
&             & & \Omega\big\{\mathbb{Q}[\cdot] \big\},
\end{aligned}
\end{equation} 
where $\Omega \{\mathbf{x}\}$ and $\Omega\big\{\mathbb{Q}[\cdot] \big\}$ represent a set of constraints on the code and the encoder respectively.

Given the code $\mathbf{x}$, for a certain set of tasks like compression, we are interested in reconstructing the original $\mathbf{f}$, either exactly or approximatively. Therefore, we define accordingly a decoder $\mathbb{Q}^{-1}[\cdot]: \mathcal{X}^m \to \Re^n$ that reconstructs $\mathbf{f}$ by decoding $\mathbf{x}$, denoted as $\hat{\mathbf{f}} = \mathbb{Q}^{-1}[\mathbf{x}]$.

We may then focus on the quality of reconstruction within a trade-off with a set of constraints $\Omega\big\{\mathbb{Q}[\cdot], \mathbb{Q}^{-1}[\cdot] \big\}$ on both the encoder and decoder. This idea can be formalized as:
\begin{equation} \label{eq:SingleLayer_GeneralEncDec}
\begin{aligned}
& \underset{{\mathbb{Q}[\cdot], \mathbb{Q}^{-1}[\cdot] }}{\text{minimize}}
& & d_{\mathcal{E}}(\mathbf{f}, \hat{\mathbf{f}})\\
& \text{s.t.} & & \Omega \{\mathbf{x}\},  \\
&             & & \Omega\big\{\mathbb{Q}[\cdot], \mathbb{Q}^{-1}[\cdot] \big\},
\end{aligned}
\end{equation} 
where $d_{\mathcal{E}}$ is the Euclidean distortion measure between two vectors $\mathbf{f}$ and $\hat{\mathbf{f}}$, and is defined as:
\begin{equation} \label{eq:EuclideanDistortion}
d_{\mathcal{E}}(\mathbf{f},\hat{\mathbf{f}}) \triangleq \frac{1}{n} ||\mathbf{f} - \hat{\mathbf{f}}||_2^2,
\end{equation}
and whose expected value is a fundamental property of an encoding and is referred to as the distortion, which is defined as in Eq. \ref{eq:D}, if the distribution is known; or as in Eq. \ref{eq:D_hat}, if training samples are available instead:
\begin{subequations}
\begin{align}
\mathcal{D} &= \mathbb{E}[d_{\mathcal{E}}(\mathbf{F},\hat{\mathbf{F}})]. \label{eq:D} \\
\hat{\mathcal{D}} &= \frac{1}{N} \sum_{i=1}^N d_{\mathcal{E}} (\mathbf{f}_i, \hat{\mathbf{f}}_i).  \label{eq:D_hat} \\
\mathcal{R} &= \frac{1}{n} \mathbb{E}[\text{\# bits used for encoding } ]. \label{eq:R}
\end{align}
\end{subequations}

Depending on the code constraints, i.e., $\Omega \{\mathbf{x}\}$, the codes need different number of bits to represent them. In other words, $\Omega \{\mathbf{x}\}$ specifies the rate of encoding, another fundamental property of an encoding scheme which is defined as in Eq. \ref{eq:R}. 

While it is desired to reduce both the rate of encoding, i.e., to have more compact codes, and the distortion of reconstruction, i.e., more fidelity to the data, for any source of information and under any encoding scheme, these are in fact conflicting requirements. Therefore, we always encounter a rate-distortion trade-off. A large body of this thesis, as we will see later, will be dedicated to designing encoder-decoder pairs that provide efficient rate-distortion trade-offs under different models.

Let us next consider some instances of Eqs. \ref{eq:SingleLayer_GeneralEnc} and \ref{eq:SingleLayer_GeneralEncDec} in the following two sections.
\section{The Variance-Regularized K-means problem} \label{sec:SingleLayer_Synthesis}
Here we start elaborating on Eq. \ref{eq:SingleLayer_GeneralEncDec} using the synthesis model described in section \ref{subsec:ModelingLit_SP_Synthesis}. We let $\hat{\mathbf{f}} = \mathrm{C}\mathbf{x}$, i.e., the vector $\mathbf{f}$ is approximated from the column-space of the codebook $\mathrm{C}$. 

We should also specify $\mathcal{X}$ by putting some constraints on the encoding procedure. So what structure do we impose on the code? We follow the strategy that we mentioned in section \ref{sec:ModelingLit_Proposed}, i.e., keeping the single-layer architecture simple and with low representation power, then compensating using a multi-layer architecture.

So we set $\Omega\{\mathbf{x}\}: ||\mathbf{x}||_0 = ||\mathbf{x}||_1 = 1$, which means that we allow only one column of the codebook $\mathrm{C}$, and with a unit coefficient to approximate $\mathbf{f}$. The problem then becomes:
\begin{equation} \label{eq:SingleLayer_Synthesis}
\begin{aligned}
& \underset{{\mathrm{C},\mathbf{x}}}{\text{minimize}}
& & d_{\mathcal{E}}(\mathbf{f}, \mathrm{C}\mathbf{x})\\
& \text{s.t.} & & ||\mathbf{x}||_0 = ||\mathbf{x}||_1 = 1,  \\
&             & & \Omega\big\{\mathbb{Q}[\cdot] \big\}.
\end{aligned}
\end{equation} 

Let us first assume that the codebook $\mathrm{C}$ is given. The problem of finding the code $\mathbf{x}$ becomes an extreme case of the sparse coding problem of section \ref{subsec:ModelingLit_SP_Synthesis}, where sparsity is maximal. Fortunately, this extreme case, unlike the general case of NP-hard sparse coding problem, can be solved exactly and easily. In fact, it is very easy to show that finding such $\mathbf{x}$ is equivalent to finding an index $m'$ from: 
\begin{equation} \label{eq:SingleLayer_kmeansCode}
\begin{aligned}
m' &=  \underset{{1 \leqslant m' \leqslant m}}{\text{argmin  }}   \frac{1}{n}||\mathbf{f} - \mathbf{c}_{m'}||_2^2\\
& \text{s.t. } \mathbf{c}_{m'} \text{ is a column of } \mathrm{C}.  
\end{aligned}
\end{equation}
In other words, the activity of the code $\mathbf{x}$ simply happens at the position corresponding to the nearest column of $\mathrm{C}$ to any given $\mathbf{f}$.

We next focus on finding the optimal codebook $\mathrm{C}$ under several different setups.
\subsection{The K-means algorithm} \label{subsec:SingleLayer_Synthesis_Kmeans}
The simplest instance of Eq. \ref{eq:SingleLayer_Synthesis} is when there is no particular structure imposed on the codebook $\mathrm{C}$. So let us solve this problem using a set of training samples $\mathrm{F} = [\mathbf{f}_1, \cdots, \mathbf{f}_N]$. The problem can be written as:

\begin{equation} \label{eq:SingleLayer_Kmeans}
\begin{aligned}
& \underset{{\mathrm{C},\mathrm{X}}}{\text{minimize}}
& & \frac{1}{2Nn}|| \mathrm{F} - \mathrm{C} \mathrm{X} ||_{\mathcal{F}}^2\\
& \text{s.t.} & & ||\mathbf{x}_i||_0 = ||\mathbf{x}_i||_1 = 1,  \\
&             & & \text{for } i = 1, \cdots, N,
\end{aligned}
\end{equation}
where the code vectors $\mathbf{x}_i$'s (corresponding to $\mathbf{f}_i$'s) are gathered as $\mathrm{X} = [\mathbf{x}_1, \cdots, \mathbf{x}_N]$.

This problem is equivalent to the famous K-means which can be solved by iterating between two steps: \textbf{i.} Fixing $\mathrm{C}$ and updating each $\mathbf{x}_i$ from $\mathrm{X}$ as in Eq. \ref{eq:SingleLayer_kmeansCode}, and \textbf{ii.} Fixing $\mathrm{X}$ and updating columns of $\mathrm{C}$ by finding the mean of all training samples that fall within the Voronoi cell of any $\mathbf{c}_{m'}$. A simple description of this procedure is detailed in Algorithm \ref{alg:SingleLayer_Kmeans}.

\begin{algorithm} \caption{\textit{K-means}} \label{alg:SingleLayer_Kmeans}
\begin{algorithmic}[0]
    \INPUT  Training set $\mathrm{F}$, $\#$ of codewords $m$
    \OUTPUT Codebook $\mathrm{C}$ and codes $\mathrm{X}$
    
\end{algorithmic}
\begin{algorithmic}[1]
\State $\mathrm{C} \gets$  $m$ random samples from $\mathrm{F}$.
\While { $\hat{\mathcal{D}}$ not converged, }
\Statex \textbf{i.} Fix $\mathrm{C}$, update $\mathrm{X}$:
\State $\mathrm{X} \gets \mathrm{0}_{n \times N}$  \Comment Initialize with an all-zero matrix
\For{$i = 1,\cdots, N$}
\State $m' \gets \underset{1 \leqslant m' \leqslant m}{\text{argmin}} || \mathbf{f}(i) - \mathbf{c}_{m'}||_2^2 $
\State $\mathbf{x}_i(m') \gets 1$
\EndFor
\Statex \textbf{ii.} Fix $\mathrm{X}$, update $\mathrm{C}$:
\State $\mathrm{C} \gets \mathrm{F}\mathrm{X}^{\dagger}$ 
\State $\hat{\mathcal{D}} \gets \frac{1}{nN} ||\mathrm{F} - \mathrm{C}\mathrm{X}||_{\mathcal{F}}^2$
\EndWhile
\end{algorithmic}
\end{algorithm}


The K-means algorithm, although conceptually very simple, is widely used for many applications in vector quantization and clustering. In fact, it is a very successful instance of unsupervised learning based on which further supervised tasks can successfully be addressed, e.g., as in \cite{coates2011analysis,coates2012learning}.

Despite its success, however, the number of training samples required to provide a good solution might be substantial, particularly at higher dimensions. In fact, K-means is unstructured since it does not impose any constraint on the codebook. That is why we try to address this issue in two complementary ways: First, keeping its single-layer architecture but imposing a useful prior as in section \ref{subsec:SingleLayer_Synthesis_VRKmeans}. Second, using a multi-layer structure as we will see later in section \ref{sec:MultiLayer_RRQ}. 

In order to see the lack of structure in K-means, we will next draw a probabilistic picture of the basic problem of Eq. \ref{eq:SingleLayer_Synthesis} and discuss the solutions under two different assumptions.  
\subsection{Probabilistic viewpoint} \label{subsec:SingleLayer_Synthesis_probabilist}

Instead of approaching the problem of Eq. \ref{eq:SingleLayer_Synthesis} with training samples, let us first study the solution by assuming that the data follows simple but known probability distributions. This will provide useful insights for the evolution of K-means to the \textit{VR-Kmeans} algorithm. 
\subsubsection{\textit{i.i.d.} Gaussian sources} \label{subsubsec:SingleLayer_Synthesis_probabilist_iid}
We assume first that the data follows simply an \textit{i.i.d.} Gaussian distribution with variance $\sigma^2$, i.e., $\mathbf{F} \sim \mathcal{N}(\mathbf{0}, \sigma^2 \mathrm{I}_n)$. For this setup, provided that the dimension $n$ is high enough, it is easy to show that the data will be uniformly distributed on the outer shell of an $n$-sphere of radius $\sqrt{n \sigma^2}$, with high probability.\footnote{To verify this, it suffices to study the behavior of $||\mathbf{F}||_2$ in asymptotic.}

It is then desired to find the optimal codebook $\mathrm{C} = [\mathbf{c}_1, \cdots, \mathbf{c}_m]$ that optimizes Eq. \ref{eq:SingleLayer_Synthesis}. The best achievable distortion for this setup is known from its rate-distortion function which was described in section \ref{subsec:fundamentals_SourceCoding_RD}. In particular, from the inverse of Eq. \ref{eq:fundamentals_RDGaussian}, we know that the distortion is lower-bounded by $\mathcal{D}(R) \geqslant \sigma^2 2^{-2R}$, for a given rate-budget $R$ which is simply $R = \frac{1}{n} \log_2{m}$ for this setup, as was defined in Eq. \ref{eq:R}.

Therefore, we can write: 
\begin{equation*} 
\begin{aligned}
\mathcal{D} &= \frac{1}{n}\mathbb{E}\Big[|| \mathbf{F} - \mathbf{C} ||_2^2 \Big] \\
            &= \sigma^2 2^{-2R} = \sigma^2   (m^{\frac{-2}{n}}),  
\end{aligned}
\end{equation*}
where $\mathbf{C}$ is a random vector corresponding to the closest column of $\mathrm{C}$ to the given $\mathbf{F}$. It can be shown that the optimal distribution to generate the codebook columns from is the \textit{i.i.d.} Gaussian distribution. Thus, we have that $\mathbf{C} \sim \mathcal{N}(\mathbf{0}, \sigma_C^2 \mathrm{I}_n)$ and we only need to find $\sigma_C^2$.

In order for $\mathbf{C}$ to be optimal, according to the principle of orthogonality, the residual error $\mathbf{E} \triangleq \mathbf{F} - \mathbf{C}$ should be orthogonal to $\mathbf{C}$. Therefore, we have that $||\mathbf{C}||_2^2 = ||\mathbf{F}||_2^2  -  ||\mathbf{E}||_2^2$. Therefore, the variance of the distribution from which the codebooks are generated is easily calculated as:

\begin{equation*}  
\sigma_C^2 = \sigma^2 - \sigma^22^{-2R} = \sigma^2(1 - 2^{-2R}).
\end{equation*}

So to summarize, under the above setup and for a fixed rate-budget $R = \frac{1}{n} \log_2{m}$, the $m$ columns of $\mathrm{C}$ should be generated from $\mathbf{C} \sim \mathcal{N}(\mathbf{0}, \sigma^2(1 - 2^{-2R}) \mathrm{I}_n)$. This provides us with a geometric picture for the optimal codebook. In fact, like the data itself, the codebook columns are also uniformly distributed on the outer shell of an $n$-sphere, but with a smaller radius $\sqrt{n\sigma^2(1- 2^{-2R})}$. 

\subsubsection{\textit{i.n.i.d.} Gaussian sources} \label{subsubsec:SingleLayer_Synthesis_probabilist_inid}
The previous setup might be too restrictive for many practical scenarios. In fact, a much more flexible assumption on the data is the independent but not identically distributed Gaussian case where different dimensions are allowed to have different variances, i.e., $\mathbf{F} \sim \mathcal{N}(\mathbf{0}, \text{diag}([\sigma^2_1, \cdots, \sigma^2_n]^T))$. This can, at least loosely, correspond to a wide range of correlated signals after a whitening stage, where data becomes Gaussian after projection and the variance profile is exponentially decreasing or is sparse.

We also pointed out the rate-distortion behavior of this setup back in section \ref{subsec:fundamentals_SourceCoding_RD} and mentioned that rate-allocation for this setup should be performed according to the so-called ``reverse water-filling'' paradigm. In particular, we had that $D_j$, the distortion at dimension $j$ should optimally be assigned according to Eq. \ref{eq:fundamentals_revWFDist}, with $\gamma$ a constant to ensure that $\sum_{j=1}^n D_j = D$, for a total distortion level $D$. 

Now imagine we want to assign optimal codewords for this setup. Obviously, unlike the \textit{i.i.d.} case, where $\sigma_C^2$ turned out to be the same for all dimensions, the optimal variances should be different. Denote $\sigma_{C_j}^2$ as the variance of the codewords corresponding to dimension $j$. Again due to the principle of orthogonality and also the fact that different dimensions are assumed to be independent from each other, we have that $\sigma_{C_j}^2 = \text{min} \big[ 0,\sigma_j^2 - D_j \big]$. Therefore, according to Eq. \ref{eq:fundamentals_revWFDist}, the optimal assignment of the codeword variances will be a soft-thresholding of $\sigma_j^2$ with $\gamma$:
\begin{equation} \label{eq:SingleLayer_revWFSig2C}
\sigma_{C_j}^2 = \eta_{\gamma}(\sigma_j^2)  = 
\begin{cases}
   \sigma_j^2 - \gamma ,& \text{if   } \sigma_j^2 \geqslant \gamma \\
    0, & \text{if   } \sigma_j^2 < \gamma.
\end{cases}
\end{equation} 

This means that the optimal rate-allocation requires that the dimensions with variances less than $\gamma$ should not be assigned any rate at all. Notice also that for the special case, where $\sigma_j^2 = \sigma^2$ for all dimensions,  Eq. \ref{eq:SingleLayer_revWFSig2C} will correspond to the \textit{i.i.d.} case above.

To summarize, the optimal codewords for the \textit{i.n.i.d.} case should be generated according to $\mathbf{C} \sim \mathcal{N}(\mathbf{0}, \mathrm{S})$, where:
\begin{equation} \label{eq:SingleLayer_revWFS}
\mathrm{S} \triangleq \text{diag}([\sigma_{C_1}^2, \cdots, \sigma_{C_j}^2, \cdots, \sigma_{C_n}^2]^T),
\end{equation} 
and $\sigma_{C_j}^2$'s are calculated according to Eq. \ref{eq:SingleLayer_revWFSig2C}.

Finally, the geometric picture for this setup is obviously more general than in the \textit{i.i.d.} case. In fact, while the data is uniformly distributed on the outer shell of an $n$-dimensional ellipsoid, depending on its elongation along different axes, the optimal codebook is distributed on the outer shell of an ellipsoid whose dimension can be smaller than the ambient dimension $n$. In fact, for highly correlated data which correspond to quickly decaying variance profiles, the effective dimensionality is much smaller than the ambient dimension and hence much fewer degrees of freedom should be foreseen in the design of the codebook than the na\"{i}ve non-structured design. 
\subsection{The \textit{VR-Kmeans} algorithm} \label{subsec:SingleLayer_Synthesis_VRKmeans}
Now let us focus again on addressing the problem of Eq. \ref{eq:SingleLayer_Synthesis}. Our objective here is to impose some structure on the codebook that is more realistic or specific than the K-means algorithm and hopefully can reduce the sample complexity.

Although knowing the exact probability distribution of the data is an unrealistic assumption and in practice we are given only a bunch of training samples, the probabilistic picture sketched in section \ref{subsec:SingleLayer_Synthesis_probabilist} can suggest us some useful insights on the design of codebooks. Under those probabilistic assumptions, we saw that the optimal codebook turns out to have a very clear structure. More preceisely and for the \textit{i.n.i.d.} Gaussian data, we saw that for the random codebook $\mathbf{C}$:
\begin{equation*}
\Omega \big\{ \mathbb{Q}[\cdot] \big\} \colon \mathbf{C} \sim \mathcal{N}(\mathbf{0}, \mathrm{S}),
\end{equation*}
where $\mathrm{S}$ is the optimal covariance matrix of the codebook and is derived from Eqs. \ref{eq:SingleLayer_revWFS} and \ref{eq:SingleLayer_revWFSig2C} and $\Omega \big\{ \mathbb{Q}[\cdot] \big\}$ constraint on the encoding procedure within the language of Eq. \ref{eq:SingleLayer_Synthesis}.

In practice, however, the \textit{i.n.i.d.} assumption on the data can only be partially achieved. For example, the PCA algorithm\footnote{In chapter \ref{chapter:ImCompression}, we will introduce a better alternative of PCA for images.} can only de-correlate dimensions which is a much weaker (second-order) sort of independence rather than the full statistical independence required by the setup of \textit{i.n.i.d.} data. Moreover, the results of section \ref{subsec:SingleLayer_Synthesis_probabilist} rely on rate-distortion theory, which are more realistic for the asymptotic case when $n \rightarrow \infty$. In practice though, data dimensionality is finite.

To relax matters, therefore, we introduce the structure above as a regularization to Eq. \ref{eq:SingleLayer_Synthesis} rather than an explicit constraint. 

This can still be too restrictive. In general, assuming a full covariance matrix that has $\frac{n(n+1)}{2}$ parameters is a too strong assumption. More particularly, the structure above suggested by covariance matrix $\mathrm{S}$ forces all dimensions of the codebook to be independent of each other, which is not realistic for whitened data. Therefore, during optimization, we would instead only penalize the diagonal elements of the covariance matrix of the codebook and let the off-diagonals vary freely during optimization.  

\subsubsection{Problem formulation}
Assuming that the training samples $\mathrm{F} = [\mathbf{f}_1, \cdots, \mathbf{f}_N]$ are whitened such that the data dimensions are un-correlated (if not independent), we follow our treatement of Eq. \ref{eq:SingleLayer_Synthesis} by formulating our main optimization problem of this section in Eq. \ref{eq:SingleLayer_VRKmeans} below and term it as the Variance-Regularized K-means (\textit{VR-Kmeans}) formulation:

\begin{equation} \label{eq:SingleLayer_VRKmeans}
\begin{aligned}
& \underset{{\mathrm{C},\mathrm{X}}}{\text{minimize}}
& & \frac{1}{2Nn}|| \mathrm{F} - \mathrm{C} \mathrm{X} ||_{\mathcal{F}}^2 + \frac{\mu}{n} ||\sum_{j=1}^n \mathrm{P}_j \mathrm{C} \mathrm{C}^T \mathrm{P}_j^T - m \mathrm{S}||_{\mathcal{F}}^2 \\
& \text{s.t.} & & ||\mathbf{x}_i||_0 = ||\mathbf{x}_i||_1 = 1,  \\
&             & & \text{for } i = 1, \cdots, N,
\end{aligned}
\end{equation}
where $\mathrm{P}_j$ is defined here as an $n \times n$ matrix with all elements equal to zero, except at the position $(j,j)$, where $\mathrm{P}_{(j,j)} = 1$. This is to choose only the diagonal elements of $\mathrm{C} \mathrm{C}^T$ at the selected $j^{\text{th}}$ position which corresponds to $\sigma_{C_j}^2$ of Eq. \ref{eq:SingleLayer_revWFSig2C}. As discussed, this regularization tries to push the variances of codebook dimensions to follow the reverse water-filling paradigm and does not impose any structure on its off-diagonals.

\subsubsection{The solution}

We solve this problem like the K-means algorithm by iterating between two steps: 

\textbf{i. Fix $\mathrm{C}$, update $\mathrm{X}$:} This is exactly like K-means and follows the recipe of Eq. \ref{eq:SingleLayer_kmeansCode}.

\textbf{ii. Fix $\mathrm{X}$, update $\mathrm{C}$:} Eq. \ref{eq:SingleLayer_VRKmeans} is reduced to
\begin{equation}  \label{eq:SingleLayer_VRKmeansCUpdate}
\begin{aligned}
& \underset{{\mathrm{C}}}{\text{minimize}}
& &  \frac{1}{2nN} ||\mathrm{F} - \mathrm{C} \mathrm{X}||_{\mathcal{F}}^2 + \frac{\mu}{2n} || \sum_{j = 1}^n \mathrm{P}_j \mathrm{C} \mathrm{C}^T \mathrm{P}_j^T - m\mathrm{S} ||_{\mathcal{F}}^2.\\
\end{aligned}
\end{equation}

The first term can be re-written using the trace operator $\text{Tr}\big[\cdot\big]$ as:
\begin{align*}
\frac{1}{2nN}||\mathrm{F} - \mathrm{C} \mathrm{X}||_{\mathcal{F}}^2 =& \frac{1}{2nN}\text{Tr}\Big[(\mathrm{F} - \mathrm{C} \mathrm{X})^T(\mathrm{F} - \mathrm{C} \mathrm{X}) \Big]\\
                                       =& \frac{1}{2nN}\text{Tr}\Big[ \mathrm{F}^T\mathrm{F} - 2\mathrm{F}\mathrm{X}^T\mathrm{C}^T + \mathrm{X}^T\mathrm{C}^T\mathrm{C}\mathrm{X}  \Big],
\end{align*}
and the second term can be written as:
\begin{align*}
\frac{\mu}{2n}||\sum_{j = 1}^n \mathrm{P}_j \mathrm{C} \mathrm{C}^T \mathrm{P}_j - m\mathrm{S}||_{\mathcal{F}}^2
= & \frac{\mu}{2n}\text{Tr}\Big[(\sum_{j = 1}^n \mathrm{P}_j \mathrm{C} \mathrm{C}^T \mathrm{P}_j - m\mathrm{S})^T(\sum_{j = 1}^n \mathrm{P}_j \mathrm{C} \mathrm{C}^T \mathrm{P}_j - m\mathrm{S}) \Big]\\
= & \frac{\mu}{2n}\text{Tr}\Big[ (\sum_{j = 1}^n \mathrm{P}_j \mathrm{C} \mathrm{C}^T \mathrm{P}_j) (\mathrm{C} \mathrm{C}^T - 2m\mathrm{S}) +  m^2\mathrm{S}^T\mathrm{S} \Big],
\end{align*}
where we used the cyclic shift property of trace, the fact that $\mathrm{P}_j^T = \mathrm{P}_j$, and also:
\[   
\mathrm{P}_j \mathrm{P}_{j'} = 
     \begin{cases}
       \mathrm{P}_j = \mathrm{P}_{j'} &\quad\text{if } j = j'\\
        \mathrm{0}                    &\quad\text{if } j \neq j'.\\
     \end{cases}
\]

Therefore, after dropping the constant terms, Eq. \ref{eq:SingleLayer_VRKmeansCUpdate} can be simplified as:

\begin{equation}  \label{eq:SingleLayer_VRKmeansCUpdate2}
\begin{aligned}
 \underset{{\mathrm{C}}}{\text{minimize Tr}}
 & \Big[-\frac{1}{N} \mathrm{F}\mathrm{X}^T\mathrm{C}^T + \frac{1}{2N}\mathrm{C}\mathrm{X}\mathrm{X}^T\mathrm{C}^T \\ +& \frac{\mu}{2} (\sum_{j = 1}^n \mathrm{P}_j \mathrm{C} \mathrm{C}^T \mathrm{P}_j) (\mathrm{C} \mathrm{C}^T - 2m\mathrm{S}) \Big].\\
\end{aligned}
\end{equation}

The 1-sparse encoding structure we imposed on $\mathrm{X}$ means that every example is activated at only one position and hence the inner product of any two rows of $\mathrm{X}$ is zero. Therefore, $\mathrm{X}\mathrm{X}^T$ is, in fact, a diagonal matrix. Also, $\sum_{j = 1}^n \mathrm{P}_j \mathrm{C} \mathrm{C}^T \mathrm{P}_j$ is a diagonal matrix by construction. Moreover, the trace operator only considers the diagonal elements of its argument. Therefore, rows of $\mathrm{C}$ can be optimized independently. As a result, the problem of minimizing Eq. \ref{eq:SingleLayer_VRKmeansCUpdate2} will be reduced to minimizing $n$ independent sub-problems.\footnote{Note that apart from the ease of analysis, this means that these sub-problems can be processed in parallel since they are independent.}
\subsubsection{Solving independent sub-problems}
Denote for simplicity: 
\begin{equation*}
Z \triangleq \mathrm{F} \mathrm{X}^T = [\mathbf{z}(1), \cdots,\mathbf{z}(j), \cdots, \mathbf{z}(n)]^T, 
\end{equation*}
and $\mathrm{X}\mathrm{X}^T = \frac{1}{N} \text{diag}(\boldsymbol{\zeta})$ with $\boldsymbol{\zeta} = [\zeta_1, \cdots, \zeta_{m'}, \cdots, \zeta_m]^T$, where $\zeta_{m'}$ is the ratio of examples clustered into the codeword $\mathbf{c}_{m'}$. Also, unlike before where we considered $\mathrm{C} = [\mathbf{c}_1, \cdots, \mathbf{c}_m]$ in terms of its columns, consider this time $\mathrm{C} = [\mathbf{c}(1), \cdots, \mathbf{c}(j), \cdots, \mathbf{c}(n)]^T$ in terms of its rows, $\mathbf{c}(j) = [c_1(j), \cdots, c_m(j))]^T$. Optimization of Eq. \ref{eq:SingleLayer_VRKmeansCUpdate2} then simplifies as:

\begin{equation}  \label{eq:SingleLayer_VR-Kmeans-subProblem}
\begin{aligned}
& \text{for } j = 1, \cdots, n \colon\\ 
 \underset{{\mathbf{c}(j)}}{\text{minimize}}
 & \Bigg[ -\frac{1}{N} \mathbf{z}(j)^T\mathbf{c}(j) + \frac{1}{2}\big[ \boldsymbol{\zeta} \odot \mathbf{c}(j) \big]^T \mathbf{c}(j) + \frac{\mu}{2} ||\mathbf{c}(j)||^2 \Big(||\mathbf{c}(j)||^2 -2m\sigma_{C_j}^2\Big) \Bigg],\\
\end{aligned}
\end{equation}
where $\odot$ specifies the element-wise (Hadamard) product between two vectors.

The independent sub-problems of Eq. \ref{eq:SingleLayer_VR-Kmeans-subProblem} do not seem to have a closed-form solution.\footnote{Notice that putting $\mu = 0$ reduces to the standard K-means algorithm. It is easy to verify that in this case, we have the same closed-form solution as the codebook update step of K-means.} However, we can easily compute the gradient vector and the Hessian matrix w.r.t. each of $\mathbf{c}(j)$'s as in Eq. \ref{eq:SingleLayer_VR-Kmeans_gradient-Hessian} and use Newton-like procedures to solve them:

\begin{subequations} \label{eq:SingleLayer_VR-Kmeans_gradient-Hessian}
\begin{equation} \label{eq:SingleLayer_VR-Kmeans_gradient}
g_{\mathbf{c}(j)} = -\frac{1}{N} \mathbf{z}(j) + \mathbf{c}(j) \odot \Big[\boldsymbol{\zeta} + \mu \big(||\mathbf{c}(j)||^2 - m\sigma_{C_j}^2 \big) \mathbf{1}_m \Big],
\end{equation}

\begin{equation}  \label{eq:SingleLayer_VR-Kmeans_Hessian}
H_{\mathbf{c}(j)} = \text{diag}\Big[\boldsymbol{\zeta} + 2\mu\big(||\mathbf{c}(j)||^2 - m\sigma_{C_j}^2\big) \mathbf{1}_m \Big] + 2 \mu \mathbf{c}(j) \mathbf{c}(j)^T.
\end{equation}
\end{subequations}

Fortunately, the Hessian matrix of Eq. \ref{eq:SingleLayer_VR-Kmeans_Hessian} has a favorable form. In fact, it is the summation of a diagonal and a rank-1 matrix. This brings computational advantages while inverting the Hessian matrix for Newton's procedure . For example, we can use the Sherman–Morrison formula for fast inversion\footnote{This breaks the inversion complexity of Hessian matrix from $\mathcal{O}(m^3)$ to $\mathcal{O}(m)$.} as:
\begin{equation}  \label{eq:SingleLayer_VR-Kmeans_HessianInv}
H_{\mathbf{c}(j)}^{-1} = \text{diag} [\frac{1}{\boldsymbol{\zeta}'}] - \frac{2\mu \text{diag} [\frac{1}{\boldsymbol{\zeta}'}] \mathbf{c}(j) \mathbf{c}(j)^T \text{diag} [\frac{1}{\boldsymbol{\zeta}'}]}{1 + 2\mu\mathbf{c}(j)^T \text{diag} [\frac{1}{\boldsymbol{\zeta}'}] \mathbf{c}(j) },
\end{equation}
where we define $\boldsymbol{\zeta}' \triangleq  \boldsymbol{\zeta} + 2\mu\big(||\mathbf{c}(j)||^2  - m\sigma_{C_j}^2 \big) \mathbf{1}_m$ and $\frac{1}{\boldsymbol{\zeta}}$ is its element-wise inversion.

To guarantee that a solution returned by the Newton's method is not a saddle point and is a (local) minimum, during optimization, we should ensure that $\text{diag} [\boldsymbol{\zeta}]$ is invertible. Furthermore, to ensure that Eq. $\mathbf{c}(j)^T \text{diag} [\frac{1}{\boldsymbol{\zeta}'}] \mathbf{c}(j) \neq \frac{-1}{2 \mu}$ always holds and \ref{eq:SingleLayer_VR-Kmeans_HessianInv} does not become degenerate, it is sufficient to ensure for every $\mathbf{c}(j)$ that:
\begin{equation} \label{eq:SingleLayer_VR-Kmeans_HessianPD}
\mathbf{c}(j)^T \mathbf{c}(j) - m\sigma_{C_j}^2 > \frac{-1}{2 \mu} \boldsymbol{\zeta}, 
\end{equation}
which can be achieved easily in practice, e.g., by choosing $\mu$ not to be too large.\footnote{Moreover, like the standard K-means, it is beneficial to ensure a strategy to avoid empty clusters, i.e., to avoid zero (or small) elements for $\boldsymbol{\zeta}$.} This avoids all numerical issues and the procedure converges quadratically to a local minimum. In practice, we observe very smooth convergence between 3 to 10 iterations within the codebook update step.

Finally, we can reduce the number of sub-problems of Eq. \ref{eq:SingleLayer_VR-Kmeans-subProblem} substantially. In practice and particularly for high-dimensional problems, it turns out that after whitening, data takes a very harsh variance decaying profile and hence a lot of $\sigma_j^2$'s and therefore, according to Eq. \ref{eq:SingleLayer_revWFSig2C}, a larger number of the $\sigma_{C_j}^2$'s will be zero. This means that we can skip a lot of the optimizations of the sub-problems corresponding to dimensions for which $\sigma_j^2$ is very small and simply put $\mathbf{c}(j) = \mathbf{0}$, since we know that their optimization will anyway give $\mathbf{c}(j)$'s that have $||\mathbf{c}(j)||^2 \simeq 0$.

We formalize this idea by setting a global ratio $\gamma'$ as a hyper-parameter for which we solve the sub-problems of Eq. \ref{eq:SingleLayer_VR-Kmeans-subProblem}, only if they fall into the active-set of dimensions defined in Eq. \ref{eq:SingleLayer_VR-Kmeans_HessianPD} below and set them to zero vectors otherwise:
\begin{equation}  \label{eq:SingleLayer_VR-Kmeans_ActiveSet}
\mathcal{A}_{\gamma} \triangleq \lbrace j: 1 \leqslant j \leqslant n | \sigma_j^2 \geqslant \gamma' \gamma^* \rbrace,
\end{equation}
where $\gamma^*$ is the so-called water-level, i.e., the threshold of Eq. \ref{eq:SingleLayer_revWFSig2C} which we choose as the solution to the following optimization problem:
\begin{equation}  \label{eq:SingleLayer_VR-Kmeans_Gamma}
\gamma^* =  \underset{{\gamma}}{\text{argmin}}  \Big|\frac{\log_2{m}}{n} - \frac{1}{2n} \sum_{j \colon \sigma_j^2 \geqslant \gamma} \log_2{ \big( \frac{\sigma_j^2}{\gamma}\big) }  \Big|,
\end{equation}
which can be solved easily using either a grid-search or simple 1D numerical root finders.

This can be seen as a further regularization; as if we set $\mu \rightarrow \infty$ for $j \in \bar{\mathcal{A}_{\gamma}}$. Not only this regularization reduces the number of optimizations, but it also results in sparsity in the codebook $\mathrm{C}$\footnote{Not to be mistaken with the sparsity of codes.}, which provides both storage and computational advantages.  

Note that instead of setting a fixed $\mu$ for $\mathcal{A}_{\gamma}$ and setting $\mu \rightarrow \infty$ for $\bar{\mathcal{A}_{\gamma}}$ as above, a smoother way of regularization would be to vary $\mu$ for $j \in \mathcal{A}_{\gamma}$ as well. In practice, this can be applied using some heuristics on $\sigma_j$'s, such that larger $\sigma_j$'s get lower regularization, since their observation is more reliable. 
\subsubsection{The algorithm summary}
The above steps for rate allocation, as well as the VR-Kmeans optimization can be summarized in Algorithm \ref{alg:SingleLayer_WFiller} and Algorithm \ref{alg:SingleLayer_VR-Kmeans}, respectively.
\begin{algorithm} \caption{\textit{Rate allocation using reverse-water-filling (Rev-WFiller)}} \label{alg:SingleLayer_WFiller}
\begin{algorithmic}[0]
    \INPUT  Data dimension variances $[\sigma_1^2, \cdots, \sigma_n^2]^T$, target rate $R$, hyper-parameter $\gamma'$
    \OUTPUT Target codebook covariance matrix  $\mathrm{S}$, the active set $\mathcal{A}_{\gamma}$

\end{algorithmic}
\begin{algorithmic}[1]
\State $\gamma^* \gets  \underset{{\gamma}}{\text{argmin}}  \Big|R - \frac{1}{2n} \sum_{j \colon \sigma_j^2 \geqslant \gamma} \log_2{ \big( \frac{\sigma_j^2}{\gamma}\big) }  \Big|$. \Comment As in Eq. \ref{eq:SingleLayer_VR-Kmeans_Gamma}
\For{$j = 1,\cdots, n$}
\State $\sigma_{C_j}^2 \gets \big( \sigma_j^2 - \gamma \big)^+$ \Comment Equivalent to soft-thesholding $\sigma_j^2$ with $\gamma$.
\EndFor
\State $\mathrm{S} \gets \text{diag} \big( [\sigma_{C_1}^2, \cdots, \sigma_{C_n}^2 ]^T \big) $
\State $\mathcal{A}_{\gamma} \gets \lbrace j: 1 \leqslant j \leqslant n | \sigma_j^2 \geqslant \gamma' \gamma^* \rbrace$
\end{algorithmic}
\end{algorithm}


\begin{algorithm} \caption{\textit{VR-Kmeans}} \label{alg:SingleLayer_VR-Kmeans}
\begin{algorithmic}[0]
    \INPUT  Training set $\mathrm{F}$ (whitened), $\#$ of codewords $m$, $\mathrm{S}$ the target covariance matrix for $\mathrm{C}$, active set $\mathcal{A}_{\gamma}$, regularization hyper-parameter $\mu$, Newton's step size $\eta$
    \OUTPUT Codebook $\mathrm{C}$ and codes $\mathrm{X}$

\end{algorithmic}
\begin{algorithmic}[1]
\State $\mathrm{C} \gets$  $m$ random samples from $\mathcal{N}(\mathbf{0},\mathrm{S})$.
\While {$\mathcal{J}(\mathrm{C},\mathrm{X})$ (cost function of Eq. \ref{eq:SingleLayer_VRKmeans}) not converged, }
\Statex \textbf{i.} Fix $\mathrm{C}$, update $\mathrm{X}$:
\State $\mathrm{X} \gets \mathrm{0}_{n \times N}$  \Comment Initialize with an all-zero matrix
\For{$i = 1,\cdots, N$}
\State $m' \gets \underset{1 \leqslant m' \leqslant m}{\text{argmin}} || \mathbf{f}(i) - \mathbf{c}_{m'}||_2^2 $
\State $\mathbf{x}_i(m') \gets 1$
\EndFor
\Statex \textbf{ii.} Fix $\mathrm{X}$, update $\mathrm{C}$:
\For{$j \in \mathcal{A}_{\gamma}$}
\While { objective of Eq. \ref{eq:SingleLayer_VR-Kmeans-subProblem} not converged, }
\State $\mathbf{c}(j) \gets \mathbf{c}(j) - \eta H_{\mathbf{c}(j)}^{-1} g_{\mathbf{c}(j)}$  \Comment Use Eqs. \ref{eq:SingleLayer_VR-Kmeans_gradient} and \ref{eq:SingleLayer_VR-Kmeans_HessianInv} for calculations. 
\EndWhile  \Comment Also ensure Eq. \ref{eq:SingleLayer_VR-Kmeans_HessianPD}, otherwise re-initialize.
\EndFor
\State $\mathcal{J}(\mathrm{C},\mathrm{X}) \gets \frac{1}{2Nn}|| \mathrm{F} - \mathrm{C} \mathrm{X} ||_{\mathcal{F}}^2 + \frac{\mu}{n} ||\sum_{j=1}^n \mathrm{P}_j \mathrm{C} \mathrm{C}^T \mathrm{P}_j - \mathrm{S}||_{\mathcal{F}}^2  $
\EndWhile  
\end{algorithmic}
\end{algorithm}


\subsection{Rate-distortion behavior} \label{subsec:SingleLayer_Synthesis_RD}
We perform several experiments to validate the usefulness of our proposal, i.e., to inject the rate-allocation prior to the formulation of standard K-means which results in the VR-Kmeans algorithm. 

We train these algorithms under several setups and study their rate-distortion performance on the corresponding unseen test sets.

\subsubsection*{Comparison with asymptotic values}  
\begin{figure}  [H]
   \begin{center} 
\includegraphics[width=0.99\textwidth]{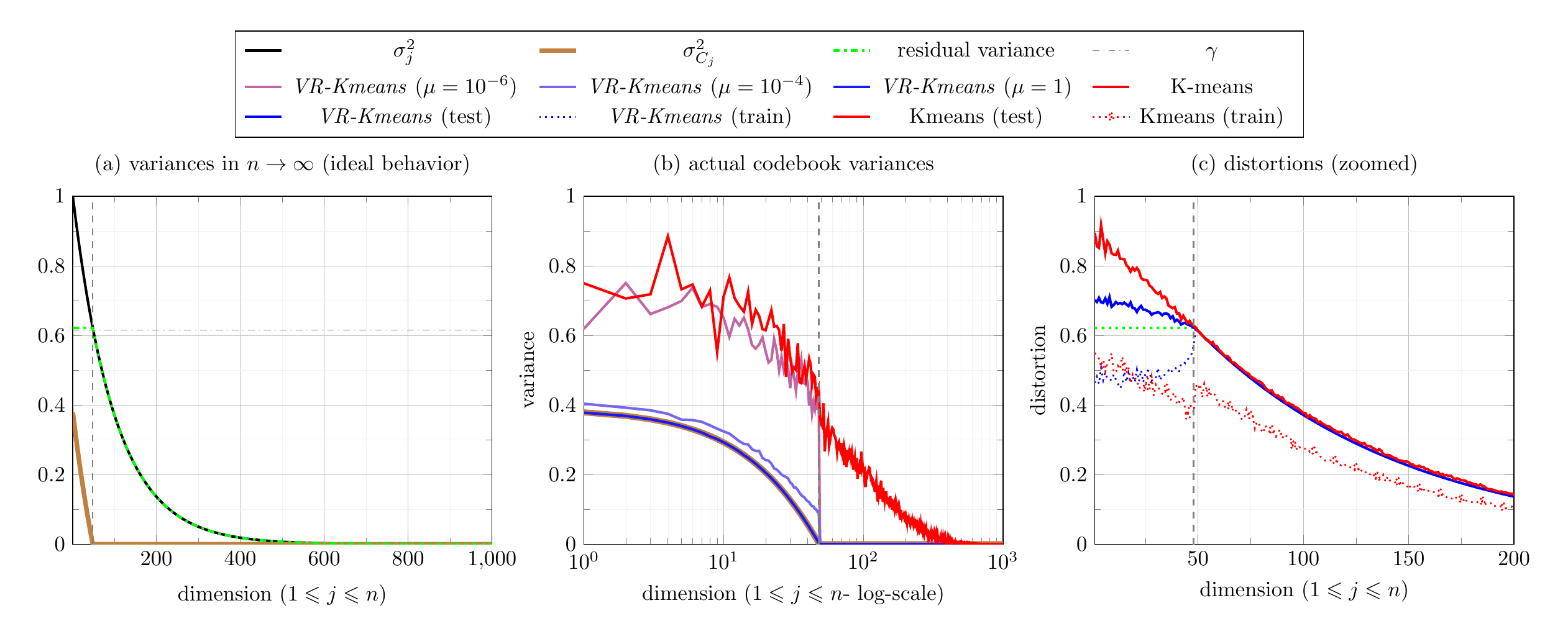}
   \end{center}
\vspace{-0.5cm}    
   \caption{The VR-Kmeans performs rate-allocation by regularizing the codebook dimensions. (a) Optimal variance values (for $n \rightarrow \infty$) per dimension. (b) Variances of trained codebooks for different regularization levels across dimensions. Note that the standard K-means corresponds to $\mu = 0$. (c) The variances of residuals (distortions) per dimension for train and test sets. Note that regularization avoids over-fitting. }
   \label{fig:SingleLayer_VRKmeans_DimDEMO}
   \end{figure}

Fig. \ref{fig:SingleLayer_VRKmeans_DimDEMO} shows how the VR-Kmeans pushes $\sigma_{\mathbf{C}_j}^2$'s to the asymptotic values. It is clear from the figure how this idea helps with over-fitting since it avoids very low distortions for the training set while achieving the distortion on the test set lower than the K-means. This experiment was done for the \textit{Var-Decay} database with $n = 1000$ dimensions, $m = 256$ codewords and $N = 1000$ training samples while the decay profile was generated according to $\sigma_j^2 = \exp{(-0.01j)}$.

\subsubsection*{Train-test gap on various data sources}
In another experiment, we measure the train and test distortions of the K-means and the VR-Kmeans on 3 different synthetic sources, i.e., the \textit{i.i.d.}, the \textit{Var-Decay} and the \textit{AR(1)} (with $\rho=0.99$); as well as two real public databases, i.e., the \textit{MNIST} digits and the \textit{C-Yale} facial images. For the synthetic databases we set $n=1000$ and for all databases we use $N=1000$ training examples and test on $10000$ test examples (except for \textit{C-Yale} that has a test set of size $1207$). We also use $m=256$ centroids for all experiments.

Before compression, the \textit{AR(1)} and \textit{MNIST} are whitened using simple PCA. \textit{C-Yale} is whitened using a procedure that we will explain later in section \ref{subsec:ImCompression_STNets_Whitening}. The other two databases do not need whitening. 

Table \ref{Table:SingleLayer_VRKmeans_RD} illustrates the results of these experiments, validating the fact that the standard K-means suffers from over-training of high-dimensional data. Notice e.g., that for the \textit{i.i.d.} or \textit{Var-Decay}, K-means cannot compress at all. 

 \begin{table}  [H]
\centering 
\includegraphics[width=0.99\textwidth]{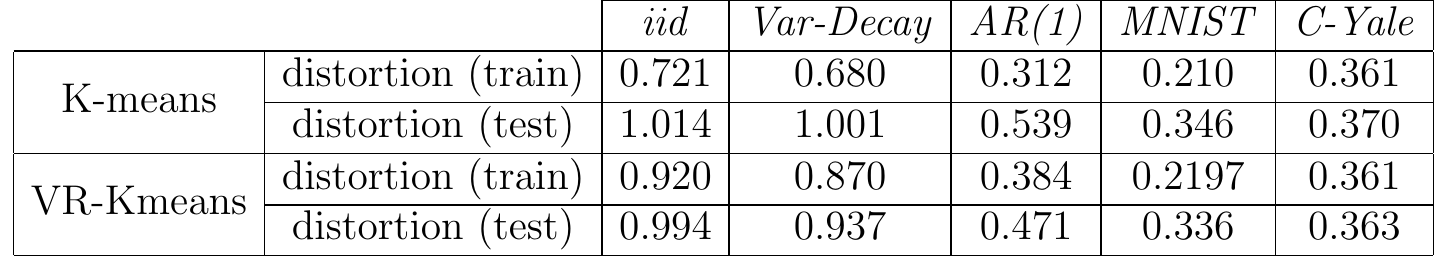}
\vspace{0.5cm}
  \caption{The gap between the normalized training set distortion calculated as $\hat{\mathcal{D}} = \frac{|| \mathrm{F} - \hat{\mathrm{F}}||_{\mathcal{F}}^2}{ || \mathrm{F} ||_{\mathcal{F}}^2 } $ on a train set $\mathrm{F}$,  and the test set distortion (ideally approaching $\mathbb{E}[\frac{ ||\mathbf{F} - \hat{\mathbf{F}} ||^2}{ ||\mathbf{F} ||^2 }]$). The VR-Kmeans reduces this gap with respect to K-means. Results are averaged over 5 independent experiments.}
   \label{Table:SingleLayer_VRKmeans_RD}
   \end{table}

It should be mentioned that by increasing the number of training samples $N$, K-means naturally improves performance and reduces the train-test gap. However, as we will see in the next experiment, the required number of training samples to achieve a certain performance increases (exponentially) with the rate. Moreover, for a lot of practical scenarios, the number of training samples is limited. For example, the compression of the \textit{C-Yale} database, as we will see later in chapter \ref{chapter:ImCompression}, is not feasible using K-means (without dividing the images into patches which reduces efficiency), while the VR-Kmeans can successfully compress the full-frame images. This experiment on the \textit{C-Yale} will be extended later in section \ref{sec:ImCompression_RRQ}.

\subsubsection*{Rate limitation}
We now study the impact of the rate on the performance of the algorithms and how it deviates from the theoretical Shannon bounds. For this experiment, to simulate different levels of correlation, we use the \textit{i.i.d.} database, as well as the \textit{AR(1)} database with 2 different correlation factors, i.e., $\rho = 0,5,0,99$. We change the rate by increasing the codebook size $m$, and measure the train and test distortions for both K-means and VR-Kmeans, as in the previous experiment. The number of training samples was increased w.r.t. the increase of $m$. To give an idea about optimality, we also measure the Shannon Lower Bound (SLB), i.e., the asymptotic limit for these sources.

\begin{figure}  [H]
   \begin{center} 
\includegraphics[width=0.99\textwidth]{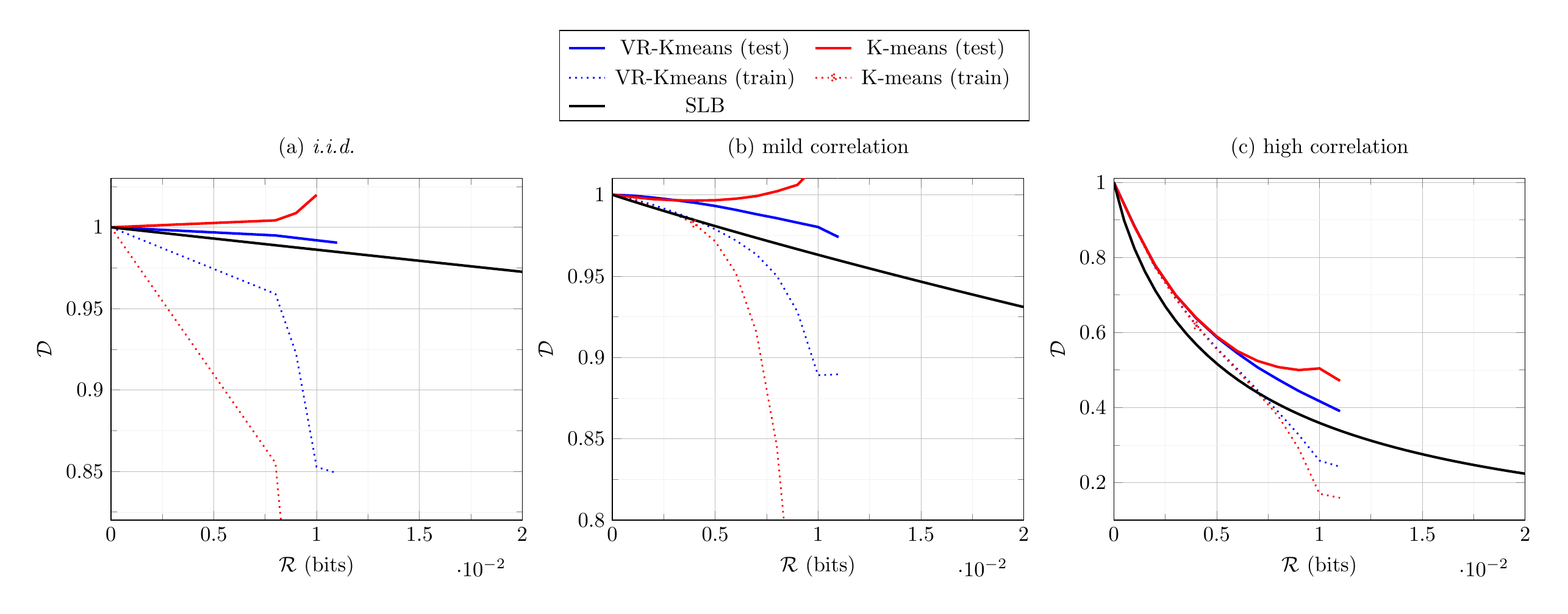}
   \end{center}
\vspace{-0.5cm}    
   \caption{The effect of rate on the encoding performance of K-means and VR-Kmeans for AR(1) Gaussian sources with $\rho = 0,0.5,0.99$, corresponding to (a) \textit{i.i.d.}, (b) mildly correlated and (c) highly correlated sources. Results are averaged over 5 independent experiments.}
   \label{fig:SingleLayer_VRKmeans_RateLimit}
   \end{figure}

Fig. \ref{fig:SingleLayer_VRKmeans_RateLimit} illustrates the results of this experiment. As in the previous example, we can clearly observe the tendency of K-means to over-train. Moreover, the increase in rate seems to aggravates this phenomenon.

Notice again that for the \textit{i.i.d.} data or mildly-correlated data, compression is not possible with K-means, and that for the highly correlated data, VR-Kmeans reduces the test distortion around $15\%$.

VR-Kmeans on the other hand seems to have a wider operational range.  While for the sake of simplicity we kept $\mu$ fixed for all rates, to account for the lack of training samples, $\mu$ can increase with the rate. So for higher rates, $\mu \rightarrow \infty$ will rely more and more on the prior rather than the data. Therefore, in principle, VR-Kmeans will never get over-trained. However, apart from the fact that more reliable prior will be needed for higher rates, there exists a serious issue with this idea.

In fact, independent from the training algorithm used, the underlying architecture, i.e., the synthesis dictionary with 1-sparsity encoding suffers from a fundamental limitation: The codebook size increases exponentially with the rate. Furthermore, the required number of training samples should increase, at least, at the same speed.

For the above experiment, we could choose the maximum number of codewords to be $m = 2048$ which required $N = 5000$. With dimensionality $n = 1000$, this corresponds only to $R \simeq 0.01$ bits. What if we wanted to target higher rates, say $R = 4$ bits?

We answer this question later in section \ref{sec:MultiLayer_RRQ}, by extending the VR-Kmeans to the RRQ.

\section{The Sparse Ternary Codes} \label{sec:SingleLayer_Analysis}
\begin{figure}  
   \begin{center} 
\subcaptionbox{$x = \psi_{\lambda}(\tilde{f})$\label{subfig:SingleLayer_STC_thresholding_psi}} {\includegraphics[width=0.18\textwidth]{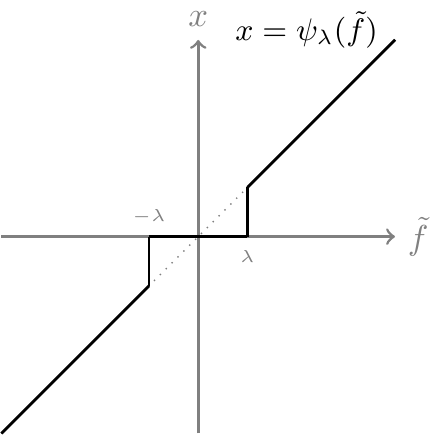}}
\subcaptionbox{$p(x)$ from $\psi$\label{subfig:SingleLayer_STC_thresholding_psiGauss}} {\includegraphics[width=0.18\textwidth]{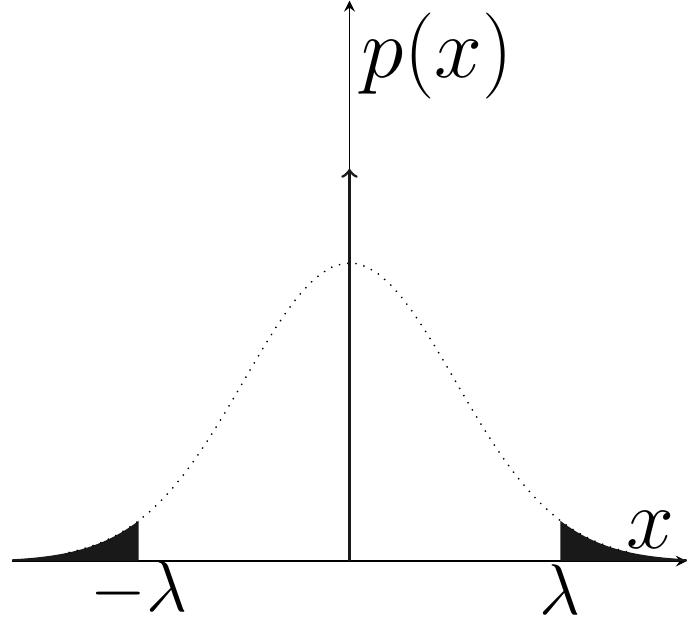}}
\subcaptionbox{$x = \phi_{\lambda}(\tilde{f})$\label{subfig:SingleLayer_STC_thresholding_phi}} {\includegraphics[width=0.18\textwidth]{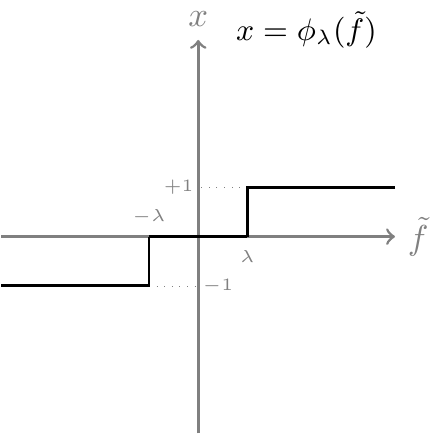}}
\subcaptionbox{$p(x)$ from $\phi$\label{subfig:SingleLayer_STC_thresholding_phiGauss}} {\includegraphics[width=0.18\textwidth]{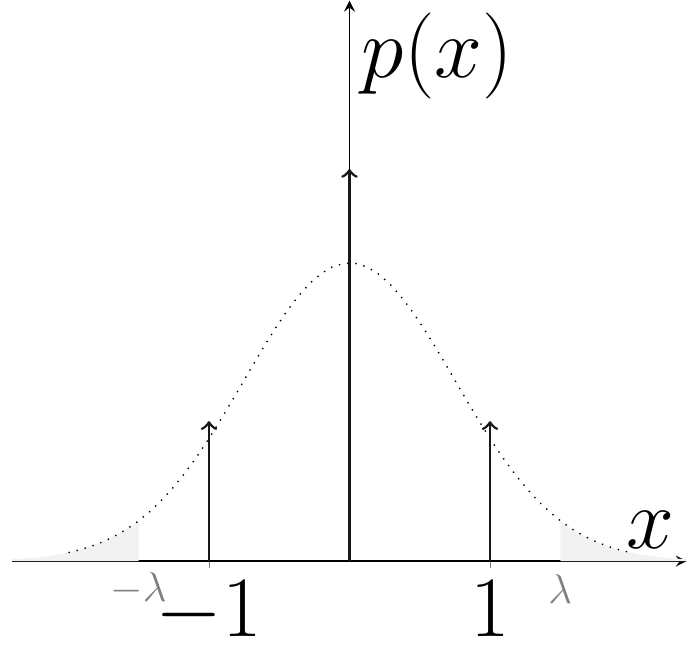}}  
\subcaptionbox{$p(x)$ from $\phi$ (weighted STC of Eq. \ref{eq:SingleLayer_STC_EncodingWeighted})\label{subfig:SingleLayer_STC_thresholding_phiBeta}} {\includegraphics[width=0.18\textwidth]{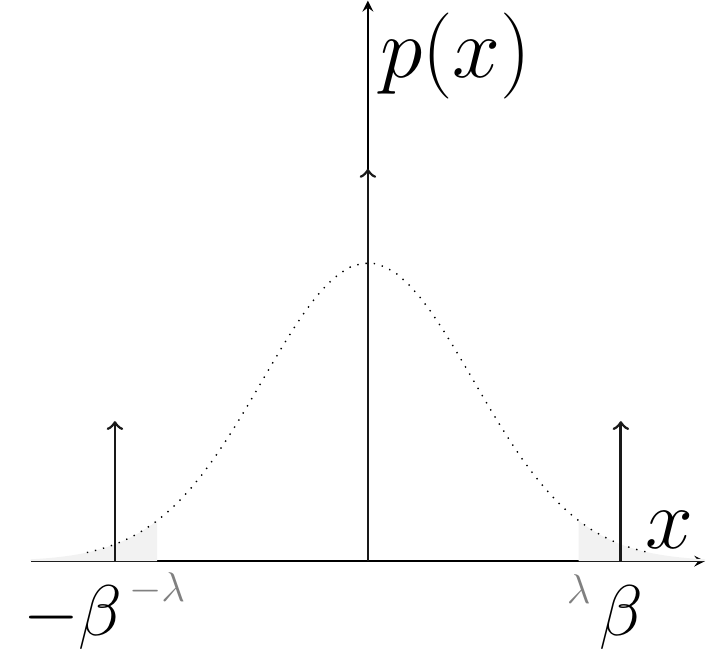}}  
   \end{center}
\vspace{-0.5cm}    
   \caption{The hard-thresholding and ternarizing operators and their impact on a Gaussian input distributed as $\tilde{F} \sim \mathcal{N}(0,\sigma^2)$.}
   \label{fig:SingleLayer_STC_thresholding}
   \end{figure}

Let us now take the other popular model used in the literature of signal processing, i.e., the analysis model described in section \ref{subsec:ModelingLit_SP_Analysis}. Let $\mathbf{x} \simeq \mathrm{A} \mathbf{f}$. Under this model, recall that a famous instance of Eq. \ref{eq:SingleLayer_GeneralEnc} was studied in Eq. \ref{eq:ModelingLit_AnalysisL0Code}, for which a simple AWGN contamination of $\mathbf{q} = \mathbf{f} + \mathbf{p}$ was assumed. Fortunately, we saw that by choosing sparsity as the code constraint, i.e., $\Omega \{ \mathbf{x} \} \colon ||\mathbf{x}||_0 = k$, a closed-form solution exists in this case as in Eq.\ref{eq:ModelingLit_AnalysisL0CodeSolution}, which involved the hard-thresholding function of Fig. \ref{subfig:SingleLayer_STC_thresholding_psi}.

For our purposes in this thesis, however, as was described earlier, we prefer to have a discretized alphabet. The simplest such alphabet that satisfies sparsity, is discrete and moreover can have zero mean (for some practical purposes) is the ternary alphabet $\{ +1, 0, -1 \}$, i.e., to choose the code from:
\begin{equation*}
\mathcal{X}_t^m = \Big\{ \mathbf{x} \in \{ +1, 0, -1 \}^m \text{ s.t. } ||\mathbf{x}||_0 = k \Big\}.
\end{equation*}

Therefore, by setting $\Omega \{ \mathbf{x} \} \colon \mathbf{x} \in \mathcal{X}_t^m$, the solution of Eq. \ref{eq:SingleLayer_GeneralEnc} becomes $\mathbf{x} = \phi_{\lambda} (\mathrm{A} \mathbf{f})$, where the ternarizing operator $\phi_{\lambda}(\cdot)$ was defined in Eq. \ref{eq:math_TernarizingFunc} and is depicted in Fig. \ref{subfig:SingleLayer_STC_thresholding_phi}. In general and for any $\mathbf{f} \in \Re^n$, in order to take this solution into consideration, we choose our encoding as:

\begin{equation}  \label{eq:SingleLayer_STC_EncodingSimple}
\mathbf{x} = \mathbb{Q}[\mathbf{f}] =  \phi_{\lambda} (\mathrm{A} \mathbf{f}).
\end{equation}

Eq. \ref{eq:SingleLayer_STC_EncodingSimple} (and its weighted version Eq. \ref{eq:SingleLayer_STC_EncodingWeighted}), will serve us as a building block for this thesis and will lead to the Sparse Ternary Codes (STC) framework as we will see later.

However, before developing it further, let us first see a probabilistic analysis of the current STC in section \ref{subsec:SingleLayer_Analysis_IT} and under simple setups. This leads to the characterization of the information measures for the elements of ternary encoding and will help us gain a better understanding.
\subsection{Information measures of STC encoding for noisy data} \label{subsec:SingleLayer_Analysis_IT}
Now let us see what happens to the data $\mathbf{f}$ and its noisy version $\mathbf{q}$ when we encode them as in Eq. \ref{eq:SingleLayer_STC_EncodingSimple}. For the sake of analysis, let us again assume a probability distribution for the data and noise.

Concretely, assume a simple AWGN model $\mathbf{Q} = \mathbf{F} + \mathbf{P}$, where $\mathbf{P} \sim \mathcal{N}(\mathbf{0}, \sigma_P^2 \mathrm{I}_n)$ is the white Gaussian noise which is added to the data. 

We also assume a random structure for the projection matrix $\mathrm{A}$ ($m$ by $n$), by assuming its elements to be \textit{i.i.d.} realizations of the random variable $A$. This random choice for projection matrix is justified for many applications, perhaps because of its performance guarantees, e.g., distance preservation as in the famous Johnson-Lindenstrauss lemma \cite{johnson1984extensions}, and also for the sake of analysis which is useful for our theoretical treatment of this part.

A popular choice for random projections is the Gaussian distribution. However, when the projections dimensionality is high, there also exists performance guarantees for sparse random matrices, e.g., as in \cite{li2006very}. This way the complexity of projections can be reduced by the sparsity factor.

So suppose we choose 
$A \sim
\begin{cases}
   \pm \sqrt{\frac{s}{2m}},   &\text{w.p.       }  \frac{2}{s}, \\
   0,    &\text{w.p.       }  1- \frac{2}{s}, 
\end{cases}$
to generate the unit-norm $\mathrm{A}$. Denote the projected data as $\tilde{F} = \mathrm{A} \mathbf{F}$ and $\tilde{Q} = \mathrm{A} \mathbf{Q}$, where $s$ is a parameter to specify sparsity. 


As for the data, we assume the \textit{i.i.d.} Gaussian case of $\mathbf{F} \sim \mathcal{N}(\mathbf{0}, \sigma^2 \mathrm{I}_n)$, which has the same distribution in the projected domain as $\tilde{\mathbf{F}} \sim \mathcal{N}(\mathbf{0}, \sigma^2 \mathrm{I}_n)$. Since the dimensions are independent we can simplify the analysis by considering the scalar case $\tilde{F}_j = \tilde{F}$ from $\tilde{\mathbf{F}} = [\tilde{F}_1, \cdots, \tilde{F}_n]^T$, so we can assume $\tilde{F} \sim \mathcal{N}(0,\sigma^2)$ and the corresponding noisy projected data will be distributed as $\tilde{Q} \sim \mathcal{N}(0, \sigma^2 + \sigma_P^2)$.

This gives the joint distribution of the projected clean and noisy data as a bivariate Gaussian with $\rho = \frac{\sigma}{\sqrt{\sigma^2 + \sigma_P^2}}$, i.e.,:

\begin{equation}  \label{eq:SingleLayer_STC_JointDist}
p(\tilde{f},\tilde{q}) = \mathcal{N} \Bigg( \begin{bmatrix} 0\\0 \end{bmatrix},\begin{bmatrix} \sigma^2 & \sigma^2\\ \sigma^2 & \sigma^2 + \sigma_P^2 \end{bmatrix} \Bigg)
\end{equation}

Here we are interested in studying the behavior of encoded $F$ and $Q$ and characterizing their information measures under different noise levels and different thresholds. 

We denote the encoded clean and noisy data as $X = \phi_{\lambda_X}(\tilde{F})$ with threshold $\lambda_X$, and $Y = \phi_{\lambda_Y}(\tilde{Q})$ with threshold $\lambda_Y$, respectively.

The relation of $\tilde{F}$ and $X$ (and similarly the relation of $\tilde{Q}$ and $Y$) is depicted in Fig. \ref{subfig:SingleLayer_STC_thresholding_phi}, where $X$ follows a ternary distribution as:

\begin{equation*}
p(x) =
\begin{cases}
   +1, &\text{w.p. }   \quad  \alpha_X, \\
   0,  &\text{w.p. }   \quad  1 - 2 \alpha_X, \\
   -1, &\text{w.p. }   \quad  \alpha_X, 
\end{cases}
\end{equation*}
where we define $\alpha_X$ (and similarly $\alpha_Y$) as the code sparsity using the Q-function as:

\begin{align}   \label{eq:SingleLayer_STC_alpha}
\begin{split} 
\alpha_X &=  \frac{1}{2} \int_{-\infty}^{-\lambda_X} p(\tilde{f}) d\tilde{f} + \frac{1}{2} \int_{\lambda_X}^{\infty} p(\tilde{f}) d\tilde{f} = \mathcal{Q} \Big(\frac{\lambda_X}{\sigma} \Big), \\ 
\alpha_Y &=  \frac{1}{2} \int_{-\infty}^{-\lambda_Y} p(\tilde{q}) d\tilde{q} + \frac{1}{2} \int_{\lambda_Y}^{\infty} p(\tilde{q}) d\tilde{q} = \mathcal{Q} \Bigg(\frac{\lambda_Y}{\sqrt{\sigma^2 + \sigma_P^2}} \Bigg).
\end{split} 
\end{align}

We can then easily characterize the entropy of the ternary code w.r.t. its sparsity, from its definition as:

\begin{equation} \label{eq:SingleLayer_STC_H-ternary}
H(X) = -2\alpha_X log(\alpha_X) - (1-2\alpha_X)log(1-2\alpha_X).
\end{equation}

\begin{figure} 
   \begin{center} 
\includegraphics[width=0.9\textwidth]{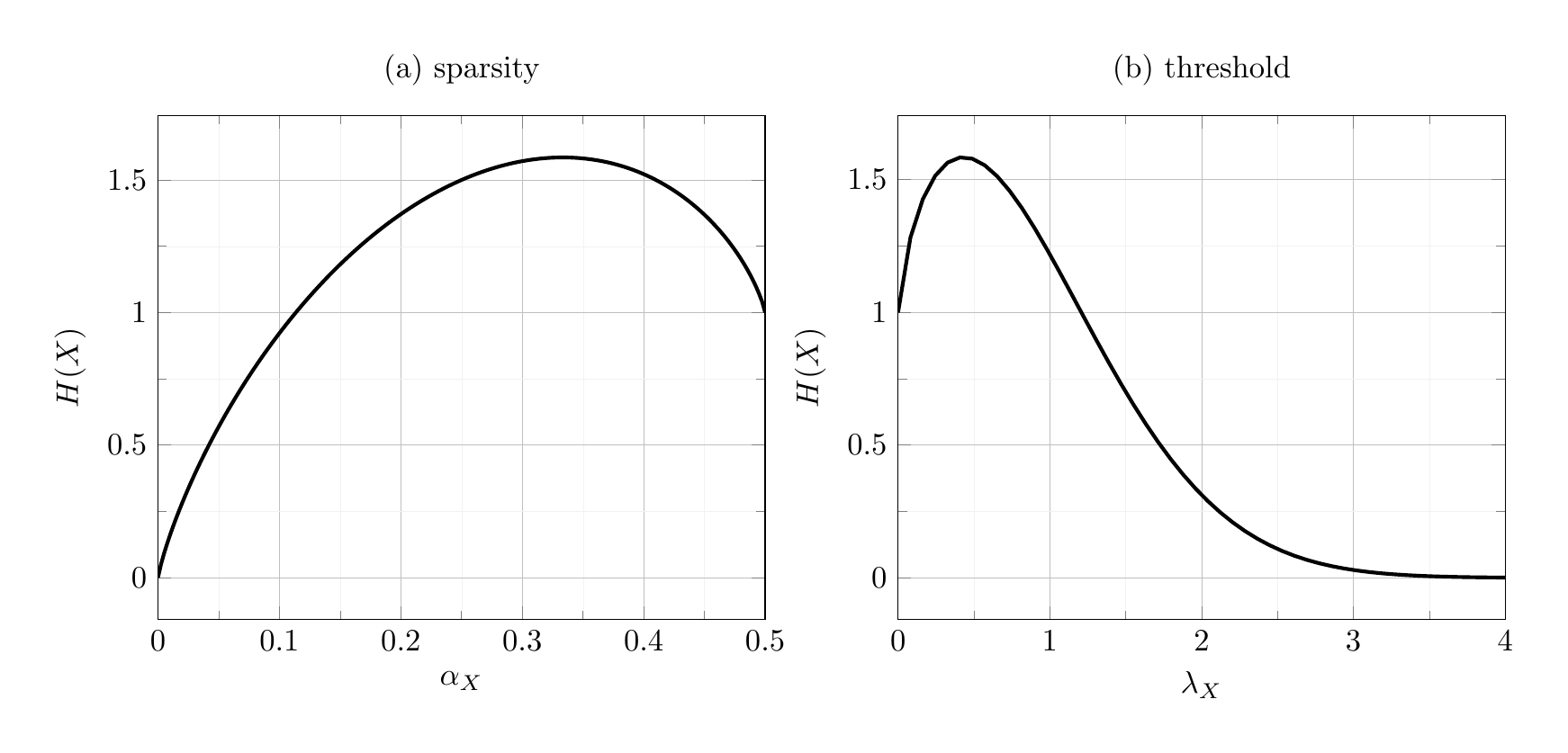} 
  \end{center}

\caption{Entropy of the ternary code w.r.t. (a) $\alpha_X$, and (b) $\lambda_X$.}
\label{fig:SingleLayer_STC_H}
 \end{figure}

Since we will extensively use $H(X)$ in this thesis, it is useful to see it in terms of both $\alpha_X$ and $\lambda_X$, as in Fig. \ref{fig:SingleLayer_STC_H}.

\begin{figure}  
   \begin{center} 
\subcaptionbox{joint distribution\label{subfig:SingleLayer_STC_basics_ellpsoid}} {\includegraphics[width=0.3\textwidth]{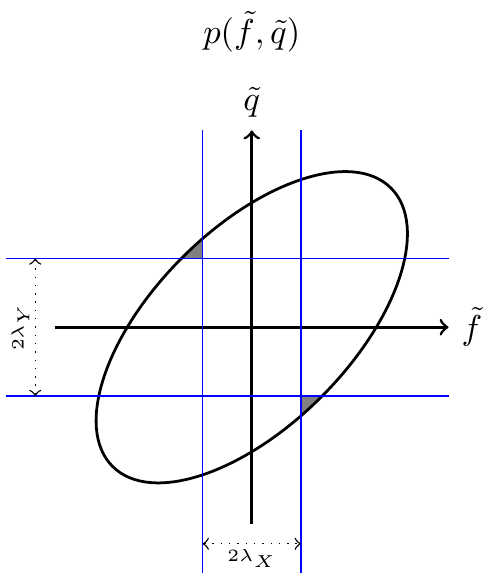}  \hspace{0.2\columnwidth}}  
 \subcaptionbox{equivalent ternary channel\label{subfig:SingleLayer_STC_basics_channelTernary}} {\includegraphics[width=0.3\textwidth]{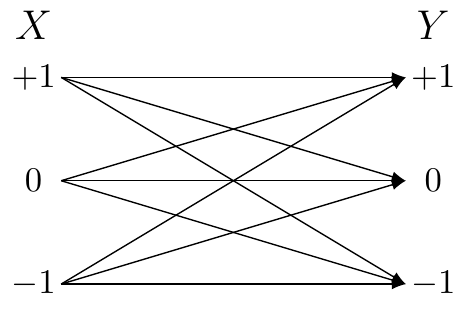}\vspace{+1.3cm}}  
   \end{center}
   \caption{(a) A countour of the joint distribution $p(\tilde{f}, \tilde{q})$. (b) Integration of $p(\tilde{q}|p(\tilde{f}))$ characterizes the ternary channel between $X$ and $Y$.}
   \label{fig:SingleLayer_STC_basics}
   \end{figure}

Now we should see the relation between $X$ and $Y$, which is a function of $\lambda_X$, $\lambda_Y$, and also the signal-to-noise-ratio defined as $\text{SNR} = 10\log_{10}{\frac{\sigma^2}{\sigma_P^2}}$.  This relation can be characterized by studying the effect of ternarization on the ellipse-shaped joint distribution $p(\tilde{f}, \tilde{q})$, which is sketched in Fig. \ref{subfig:SingleLayer_STC_basics_ellpsoid}.

While the AWGN we assumed is equivalent to a Gaussian channel between $F$ and $Q$, we can model the relation of $X$ and $Y$ as a ternary channel like in Fig. \ref{subfig:SingleLayer_STC_basics_channelTernary}. For this channel, we can consider the following transition probabilities:

\begin{equation}
 \label{eq:SingleLayer_STC_transition} 
\mathrm{P}_t = p(y|x)=
  \begin{bmatrix}
    p_{+1|+1} & p_{0|+1} & p_{-1|+1} \\
    p_{+1|0} & p_{0|0} & p_{-1|0} \\
    p_{+1|-1} & p_{0|-1} & p_{-1|-1} \\
  \end{bmatrix}.
\end{equation}

The elements of $\mathrm{P}_t$ cannot be expressed in closed-form, however, they can easily be calculated by numerical integration on the conditional distribution $p(\tilde{Q}|\tilde{F}) = \frac{p(\tilde{F},\tilde{Q})}{p(\tilde{F})}  $, with proper integral limits based on $\lambda_X$ and $\lambda_Y$. For example, $p_{+1|+1} = p_{\tilde{Q}|\tilde{F}}(+1|+1) = \frac{\int_{\lambda_Y}^{\infty} \int_{\lambda_X}^{\infty} p(\tilde{f},\tilde{q}) d\tilde{f} d\tilde{q}} {\int_{\lambda_X}^{\infty} p(\tilde{f}) d\tilde{f}}$ and $p_{0|+1} = p_{\tilde{Q}|\tilde{F}}(0|+1) = \frac{\int_{-\lambda_Y}^{\lambda_Y} \int_{\lambda_X}^{\infty} p(\tilde{f},\tilde{q}) d\tilde{f} d\tilde{q}} {\int_{\lambda_X}^{\infty} p(\tilde{f}) d\tilde{f}}$. Out of these 9 transition probabilities, 5 are independent and the rest are replicated due to symmetry. This can be easily verified from Fig. \ref{subfig:SingleLayer_STC_basics_ellpsoid}.

Now that we have computed $p(x)$ and $p(y|x)$, we are ready to characterize $I(X;Y)$, the mutual information between a code $X$ and its noisy $Y$, which can easily be decomposed as:
\begin{equation}   \label{eq:SingleLayer_STC_I} 
I(X;Y) = H(X) + H(Y) - H(X,Y),
\end{equation}
where $H(X)$ and similarly $H(Y)$ are computed from Eq. \ref{eq:SingleLayer_STC_H-ternary}. The joint entropy can also be calculated from the elements of $\mathrm{P}_t$ as:

\begin{align} \label{eq:SingleLayer_STC_HJoint}
\hspace{-0.5cm}
\begin{split}
H(X,Y) &=  -2 \alpha_{X} \mathrm{P}_t(1,1)log(\alpha_{X} \mathrm{P}_t(1,1)) \\
&-2 \alpha_{X} \mathrm{P}_t(1,2)log(\alpha_{X} \mathrm{P}_t(1,2)) -2 \alpha_{X} \mathrm{P}_t(1,3)log(\alpha_{X} \mathrm{P}_t(1,3))\\ 
&-2 (1-2\alpha_{X}) \mathrm{P}_t(2,1)log((1-2\alpha_{X}) \mathrm{P}_t(2,1))\\
& -(1-2\alpha_{X}) \mathrm{P}_t(2,2)log((1-2\alpha_{X}) \mathrm{P}_t(2,2)).
\end{split}
\end{align}

So far, we have characterized all the information measures for the \textit{i.i.d.} Gaussian setup. Later we will use these elements in chapter 8 and for the analysis of STC under the concept of coding gain.


\subsubsection{Maximum-likelihood decoding of STC} \label{subsubsec:SingleLayer_Analysis_IT_MLDecoding}
Let us assume the case where the (ternary) input to the channel is to be estimated from its (ternary) output. This task, as we will see later in chapter \ref{chapter:Search}, can be useful particularly for similarity search applications, where memory and complexity requiremens necessitate the need for inference only from encoded data.   

Now that we have characterized all measures for STC and its equivalent ternary channel, we can consider the maximum-likelihood optimal estimation for this task and under our current \textit{i.i.d.} Gaussian setup. So for this section and without going into the details of similarity search for now, suppose a database of $N$ ternary codes $\mathrm{X} = [\mathbf{x}_1, \cdots, \mathbf{x}_N]$ is available. At the output of the ternary channel, the code $\mathbf{y}$ is observed. The objective is to find an index $i$, corresponding to a $\mathbf{x}_i$ from the database, based on its similarity to $\mathbf{y}$.

This estimation can best be calculated from the maximum-likelihood rule of Eq. \ref{eq:SingleLayer_Analysis_MLDecoder} by maximizing the log-likelihood:

\begin{equation} \label{eq:SingleLayer_Analysis_MLDecoder}
\hat{i} = \underset{{1 \leqslant i \leqslant N}}{\text{argmax }} \log{p (\mathbf{y}|\mathbf{x}_i)} = \underset{{1 \leqslant i \leqslant N}}{\text{argmax }} \sum_{m'=1}^m \log{p \Big( y_{m'}|\mathrm{X}(m',i) \Big)},
\end{equation}
where $p \Big( y_{m'}|\mathrm{X}(m',i) \Big)$ is calculated for the element $m'$ of the received $\mathbf{y}$ and of $\mathbf{x}_i$\footnote{Here we indexed the elements of $\mathbf{x}_i$, from the code matrix $\mathrm{X}$, as $\mathrm{X}(m',i)$.}, and from the corresponding element of the transition matrix $\mathrm{P}_t$ of Eq. \ref{eq:SingleLayer_STC_transition}. Note that this can be generalized to sort the similarities of the database items by sorting the values of $\sum_{m'=1}^m \log{p \Big( y_{m'}|\mathrm{X}(m',i) \Big)}$ for all of them. 

This estimation, while optimal,  however, may be slow to compute for large values of $N$ and $m$. Adapted to the search problem, we will introduce a sub-optimal but sub-linear and hence fast rule, later in section \ref{sec:Search_MLSTC}.

\subsection{Reconstructing from ternary encoded data} \label{subsec:SingleLayer_Analysis_Reconstruction}
So far, our development of Eq. \ref{eq:SingleLayer_GeneralEnc} under the analysis model has led us to the framework of Sparse Ternary Codes. Similar to what was done for synthesis model in section \ref{sec:SingleLayer_Synthesis}, now let us try to take our STC framework and further develop it under the formulation of Eq. \ref{eq:SingleLayer_GeneralEncDec}. This leads us to the formulation below:

\begin{equation} \label{eq:SingleLayer_STC_Reconstruction}
\begin{aligned}
& \underset{{\mathbb{Q}[\cdot], \mathbb{Q}^{-1}[\cdot] }}{\text{minimize}}
& & d_{\mathcal{E}}(\mathbf{f}, \hat{\mathbf{f}})\\
& \text{s.t.} & & \mathbf{x} = \mathbb{Q}[\mathbf{f}] = \phi_{\lambda}(\mathrm{A}\mathbf{f}),  \\
&             & & \hat{\mathbf{f}} = \mathbb{Q}^{-1}[\mathbf{x}].
\end{aligned}
\end{equation}

We pursue the solution of Eq. \ref{eq:SingleLayer_STC_Reconstruction} in 2 ways. In section \ref{subsubsec:SingleLayer_Analysis_Reconstruction_nonlinear} we assume a general non-linear decoder to reconstruct from STC. This formulation, however, is less practical for our purposes in this thesis, e.g., the fast similarity search that requires online decisions for received queries. We, therefore, abandon this formulation and opt for linear decoding that can be implemented very fast. In section \ref{subsubsec:SingleLayer_Analysis_Reconstruction_DSW18}, based on the stationarity assumption for the data, and that it can be de-correlated as marginally Gaussian dimensions, we design a linear decoder that can successfully reconstruct the ternary encoded data. We also see the intrinsic limitations of this formulation that later will lead us to its multi-layer evolution in section \ref{sec:MultiLayer_MLSTC}. 

Finally, in section \ref{subsubsec:SingleLayer_Analysis_Reconstruction_Procrustean}, we relax the assumptions and learn the encoder-decoder pipeline entirely from the data using an alternating minimization approach. 


\subsubsection{Non-linear decoding} \label{subsubsec:SingleLayer_Analysis_Reconstruction_nonlinear}
Let us assume that $\mathbf{f}$ is Gaussian with a known covariance matrix $\mathbf{C}_F$, i.e.,  $\mathbf{F} \sim \mathcal{N}(\mathbf{0}, \mathbf{C}_F)$, and we do not have access to a training set. Our objective here is to reconstruct the unknown $\mathbf{f}$, only from its corresponding STC, i.e., $\mathbf{x} = \phi_{\lambda}(\mathrm{A\mathbf{f}})$.

We use a Maximum A Posteriori kind of formulation to merge the a priori distribution $p( \mathbf{f} )$ with our observation $\mathbf{x}$ that has undergone the STC. This can be formulated as:
\begin{equation*}
\begin{aligned}
\hat{\mathbf{f}}_{\text{MAP}} =& \underset{{\mathbf{f}}}{\text{  argmax}} &\log{p(\mathbf{f})},\\
                 & \text{s.t.}  & \mathbf{x} = \phi_{\lambda}(\mathrm{A}\mathbf{f}).
\end{aligned}
\end{equation*}

For the a priori part, we can easily have that $\log{p(\mathbf{f})} \propto - \mathbf{f}^T \mathrm{C}_{F}^{-1} \mathbf{f}$. How do we take into account the observation term? We do this by analysing the encoding procedure as follows.

Let us write $\mathrm{A}$ ($m$ by $n$) in terms of its rows $\mathbf{a}(m')$'s as $\mathrm{A} = [\mathbf{a}(1), \cdots, \mathbf{a}(m)]^T$. Denote the 3 sub-matrices of $\mathrm{A}$ as:

\begin{equation*}
\begin{aligned}
\mathrm{A}^+ &= [\mathbf{a}(m_1^+), \mathbf{a}(m_2^+), \cdots]^T \colon &m_i^+ \in \big\{ 1 \leqslant m_i^+ \leqslant m &| x_{m_i^+} = 1  \big\},\\
\mathrm{A}^\circ &= [\mathbf{a}(m_1^\circ), \mathbf{a}(m_2^\circ), \cdots]^T \colon &m_i^\circ \in \big\{ 1 \leqslant m_i^\circ \leqslant m &| x_{m_i^\circ} = 0  \big\}, \\
\mathrm{A}^- &= [\mathbf{a}(m_1^-), \mathbf{a}(m_2^-), \cdots]^T \colon &m_i^- \in \big\{ 1 \leqslant m_i^- \leqslant m &| x_{m_i^-} = -1  \big\},
\end{aligned}
\end{equation*}
where $\mathrm{A}^+$, $\mathrm{A}^\circ$ and $\mathrm{A}^-$ correspond to rows of $\mathrm{A}$ that provide $+1$, $0$ and $-1$ codes, respectively.

Our knowledge of the ternary encoding of Eq. \ref{eq:SingleLayer_STC_EncodingSimple}, which was based on the ternarizing operator $\phi_{\lambda}(\cdot)$ of Eq. \ref{eq:math_TernarizingFunc} tells us that the projected data, before sparsification, should have had values bigger than $\lambda$ if projected with $\mathrm{A}^+$, between $\lambda$ and $-\lambda$ if projected with $\mathrm{A}^\circ$, and finally smaller than $-\lambda$ if projected with $\mathrm{A}^-$. This completes the formulation of our MAP estimation as:

\begin{equation}  \label{eq:SingleLayer_STC_MAP}
\begin{aligned}
\hat{\mathbf{f}}_{\text{MAP}} =& \underset{{\mathbf{f}}}{\text{  argmin }} &\mathbf{f}^T \mathrm{C}_{F}^{-1} \mathbf{f} ,\\
                 & \text{s.t.}  & \mathrm{A}^+\mathbf{f} \geqslant +\lambda,\\
                 &              & |\mathrm{A}^\circ\mathbf{f}| < \lambda,\\
                 &              & \mathrm{A}^-\mathbf{f} \leqslant -\lambda,
\end{aligned}
\end{equation}
where inequalities are applied element-wise to vectors.

Eq. \ref{eq:SingleLayer_STC_Reconstruction} is a quadratic program with linear inequality constraints. Moreover, since, $\mathrm{C}_F^{-1}$ is positive-definite, it is also convex and can be solved numerically using standard convex solvers like \textit{CVX} (\cite{cvx,gb08}).

However, unfortunately, it does not have a closed-form solution. This means that for every given $\mathbf{f}$, a quadratic program should be solved. This is not suitable for our applications where most often the decoding should be done very fast while the encoding may have a training phase to tune its parameters. Therefore, we give up on this approach\footnote{Notice however, that this formulation is interesting since, in a way, it is the opposite of compressed sensing. While in compressed sensing, recovery of high-dimensional sparse signals from low-dimensional dense measurements is desired, Eq. \ref{eq:SingleLayer_STC_Reconstruction} tries to reconstruct a dense signal from the sparsity pattern of a high-dimensional projection. Therefore, it may be considered as a future direction.} and seek solutions that can benefit from a training phase and instead, can be implemented fast when training is done.

\subsubsection{Linear decoding: known covariance matrix} \label{subsubsec:SingleLayer_Analysis_Reconstruction_DSW18}
Here we restrict our decoding to be linear, i.e., consisting only of a projection step. Let us then try to learn this projection matrix from the data, i.e., we are given a training set $\mathrm{F} = [\mathbf{f}_1, \cdots, \mathbf{f}_N]$. Furthermore, we assume that the data is (second-order) stationary and for which we can estimate a covariance matrix from $\mathrm{F}$. For high-dimensional data, e.g., images, and particularly when training set size is limited, estimation of a covariance matrix that is also applicable to the test set is very difficult. However, as we will see later, since we will use only the first several eigenvectors which are more robust to estimate, our assumption of availability of the covariance matrix is safe in practice.


Therefore, concretely, assume that $\mathrm{C}_F \triangleq \frac{1}{n}\mathbb{E}[\mathbf{F} \mathbf{F}^T]$ is the covariance matrix of $\mathbf{F}$. We are interested in finding an optimal decoder that provides the best reconstruction given the ternary codes.

\textbf{Projection:} We want the codes $\mathbf{x}$ to be as informative as possible, which requires independence (or at least no correlation) among the dimensions. Thus, we perform the PCA transform by choosing $\mathrm{A} = \mathrm{U}_F^T$, where $\mathrm{C}_F = \mathrm{U}_F \mathrm{\Sigma}_F \mathrm{U}_F^T$ is the eigenvalue decomposition of $\mathrm{C}_F$. 

Therefore, the projected data $\tilde{\mathbf{f}} \triangleq \mathrm{A} \mathbf{f}$  is de-correlated and its marginal distributions converge to Gaussian\footnote{Note that, according to CLT, the Gaussianity of the marginals of $\tilde{\mathbf{F}}$ does not require the Gaussianity of $\mathbf{F}$, due to the projection step performed. However, the joint distribution $p(\tilde{\mathbf{f}})$ is not a (multivariate) Gaussian in General.} for sufficiently large $n$ as $\tilde{F}_j \sim \mathcal{N}(0, \sigma_j^2)$, where $\mathrm{\Sigma}_F = \text{diag} \big([\sigma_1^2, \cdots, \sigma_n^2]^T\big)$ with $\sigma_i^2$'s being the eigenvalues of $\mathrm{C}_F$, which are decaying in value for the correlated $\mathbf{F}$.

\textbf{Encoding:} When thinking about reconstruction from codes, we should take into account the fact that the dimensions of $\tilde{\mathbf{f}}$ have a decaying variance profile and hence different contributions in the reconstruction of $\mathbf{f}$. However, when encoded as in Eq. \ref{eq:SingleLayer_STC_EncodingSimple}, these differences will not be taken into account for reconstruction. Therefore, at this point, it makes sense to improve our encoding by a slight change as in Eq. \ref{eq:SingleLayer_STC_EncodingWeighted}, i.e., by weighting the codes with a weighting vector $\boldsymbol{\beta} \triangleq [\beta_1, \cdots, \beta_m]^T$\footnote{Note that in our general formulation, we have $\mathrm{A}$ as an $m$ by $n$ projection and hence $m$-dimensional codes. However, during section \ref{subsubsec:SingleLayer_Analysis_Reconstruction_DSW18} and for simplicity of analysis, we assume $m = n$.}, whose optimal values should be calculated:  
 
\begin{equation}  \label{eq:SingleLayer_STC_EncodingWeighted}
\mathbf{x} = \mathbb{Q}[\mathbf{f}] =  \phi_{\lambda} (\mathrm{A} \mathbf{f}) \odot \boldsymbol{\beta}.
\end{equation}

\textbf{Decoding (reconstruction):} As it was mentioned in section \ref{subsubsec:SingleLayer_Analysis_Reconstruction_nonlinear}, we prefer fast and hence linear decoding for reconstruction $\hat{\mathbf{f}} = \mathbb{Q}^{-1}[\mathbf{x}]$. This is done simply using a (back-) projection matrix $\mathrm{B}$ as:

\begin{equation} \label{eq:SingleLayer_STC_Reconstruction}
\hat{\mathbf{f}} = \mathrm{B} \mathbf{x} = \mathrm{B} \phi_{\lambda}(\mathrm{A}\mathbf{f}) \odot \boldsymbol{\beta}.
\end{equation}

The optimal value of $\mathrm{B}$ can be learned from the data. In order to facilitate this, we can provide $\mathrm{B}$ with the knowledge of the forward projection $\mathrm{A}$. Therefore, for a general (full-rank) $\mathrm{A}$ and without loss of generality, we decompose $\mathrm{B}$ as $\mathrm{B} = (\mathrm{A}^T \mathrm{A})^{-1}\mathrm{A}^T \mathrm{B'}$, i.e., with the pseudo-inverse of $\mathrm{A}$, and $\mathrm{B}'$ that leaves the degrees of freedom for training.\footnote{This decomposition serves us with the facility of presentation, but in general and within iterative procedures, such decompositions can help convergence.} Therefore, the optimization of $\mathrm{B}$ is equivalent to the optimization of $\mathrm{B}'$ as:

\begin{equation} \label{eq:SingleLayer_STC_learningPInv}
\mathrm{B'} = \argmin_{\mathrm{B'}} ||\mathrm{F} - (\mathrm{A}^T \mathrm{A})^{-1}\mathrm{A}^T \mathrm{B'} \mathrm{X}||_{\mathcal{F}}^2. 
\end{equation}
This can easily be re-expressed as:
\begin{align*}
\mathrm{B'} &= \argmin_{\mathrm{B'}}  || (\mathrm{A}^T \mathrm{A}) \mathrm{F} - \mathrm{A}^T \mathrm{B'} \mathrm{X}  ||_{\mathcal{F}}^2\\
&= \argmin_{\mathrm{B'}}  \text{Tr}\Big[ (\mathrm{A} \mathrm{F} -  \mathrm{B'}\mathrm{X})^T \mathrm{A}\mathrm{A}^T (\mathrm{A} \mathrm{F} -  \mathrm{B'}\mathrm{X}) \Big]\\ 
 &= \argmin_{\mathrm{B'}}  \text{Tr}\Big[ -2\mathrm{A}\mathrm{A}^T\mathrm{A}\mathrm{F}\mathrm{X}^T\mathrm{B'}^T + \mathrm{B'}\mathrm{X}\mathrm{X}^T\mathrm{B'}^T\mathrm{A}\mathrm{A}^T \Big].
\end{align*}
Derivation w.r.t. $\mathrm{B'}$ and equating to zero gives:
\begin{equation} \label{eq:SingleLayer_STC_OptimalB'}
\mathrm{B'}^* = \mathrm{A} \mathrm{F} \mathrm{X}^T (\mathrm{X}\mathrm{X}^T)^{-1}.
\end{equation}

Noting that $\mathrm{A}\mathrm{F} = \tilde{\mathrm{F}}$ and $\mathrm{X} = \phi(\tilde{\mathrm{F}}) \odot (\boldsymbol{\beta} \mathbf{1}_N^T)$, and by expressing matrix multiplications as the sum of rank-1 matrices, $\mathrm{B'}^*$ can be re-written as:

\begin{align*}
\mathrm{B'} &= \Big[\tilde{\mathbf{F}} [\phi(\tilde{\mathrm{F}}) \odot (\boldsymbol{\beta} \mathbf{1}_N^T)]^T\Big]  \Big[[\phi(\tilde{\mathrm{F}}) \odot (\boldsymbol{\beta} \mathbf{1}_N^T)][\phi(\tilde{\mathrm{F}}) \odot (\boldsymbol{\beta} \mathbf{1}_N^T)]^T \Big]^{-1}\\
            &= \Big[ \sum_{i=1}^N \tilde{\mathbf{f}}_i  [\phi(\tilde{\mathrm{f}}_i) \odot \boldsymbol{\beta} ]^T  \Big]
               \Big[ [\boldsymbol{\beta} \odot \boldsymbol{\beta}^T]  \sum_{i=1}^N \phi(\tilde{\mathbf{f}}_i) \phi(\tilde{\mathbf{f}}_i^T) \Big]^{-1}.
\end{align*}

As was mentioned earlier, now let us choose $\mathrm{A} = \mathrm{U}_F^T$, i.e., the PCA.  Due to the de-correlating property of $\mathrm{A}$, in the projected domain and hence in the code domain, the covariance matrix $\phi(\tilde{\mathrm{F}}) \phi(\tilde{\mathrm{F}})^T$ and the cross-covariance matrix $\tilde{\mathrm{F}} \phi(\tilde{\mathrm{F}})^T$ are diagonals. Therefore, the expression of $\mathrm{B'}^*$ above is a product of two diagonals and hence is a diagonal itself. This is a function only of the weighting vector $\boldsymbol{\beta}$. This means that, in our encoding-decoding formulation and their optimization to minimize the reconstruction
distortion, without loss of generality or loss of encoding-decoding power, we can set $\mathrm{B'} = \mathrm{I}_n$ and optimize for $\boldsymbol{\beta}$ instead. This means that, our optimal decoder can in fact be considered as $\mathrm{B} = (\mathrm{A}^T \mathrm{A})^{-1} \mathrm{A}^T$, and since $\mathrm{A}$ in this case is orthonormal (due to PCA), we can simply have that $\mathrm{B} = \mathrm{A}^T$.

So now that $\mathrm{B}$ is also determined, we are left only to optimize $\boldsymbol{\beta}$. For this, we should first calculate distortion as follows.

\textbf{Distortion:} By noting that $\hat{\mathbf{f}} = \mathrm{A}^T \mathbf{x}$, the expected reconstruction distortion can probabilistically be characterized as:

\begin{subequations}  \label{eq:SingleLayer_STC_DmultiLine}
\begin{align}
\mathcal{D} &= \mathbb{E} \big[ d_{\mathcal{E}}(\mathbf{F}, \hat{\mathbf{F}}) \big] \nonumber\\
&= \frac{1}{n}\mathbb{E} \big[ ||\mathbf{F} - \mathrm{A}^T \mathbf{X}||_2^2  \big] \nonumber \\ 
&=  \frac{1}{n} \mathbb{E} \big[ ||\mathrm{A}\mathbf{F} - \mathbf{X}||_2^2  \big]  \label{subeq:SingleLayer_STC_DmultiLine_a}\\  
&= \frac{1}{n}\mathbb{E} \big[ ||\mathrm{A}\mathbf{F} - \phi_{\lambda}(\tilde{\mathbf{F}}) \odot \boldsymbol{\beta}||_2^2 \big] \label{subeq:SingleLayer_STC_DmultiLine_b} \\  
&= \frac{1}{n}\mathbb{E} \big[ ||\tilde{\mathbf{F}} - \phi_{\lambda}(\tilde{\mathbf{F}}) \odot \boldsymbol{\beta}||_2^2 \big], \nonumber
\end{align}
\end{subequations}
where Eq. \ref{subeq:SingleLayer_STC_DmultiLine_a} follows from the orthonormality assumption of $\mathrm{A}$.

This expression is useful since it links the distortion of the original domain with that of the projection domain.

The total distortion $\mathcal{D}$ can be expressed as the sum of the distortions at each of the dimensions, i.e., $\mathcal{D} = \sum_{j = 1}^n D_j$. Noting that $\tilde{F}_j$, i.e., the elements of $\tilde{\mathbf{F}} = [\tilde{F}_1, \cdots, \tilde{F}_n]^T$ is distributed as $\tilde{F}_j \sim p(\tilde{f}_j) =\mathcal{N}(0,\sigma_j^2)$, each $D_j$ can be expressed as:

\begin{align*}
D_j &= \mathbb{E} \big[ (\tilde{F}_j - \beta_j \phi_{\lambda}(\tilde{F}_j))^2 \big]  \\
    &=  \int_{-\infty}^{-\lambda} (\tilde{f}_j + \beta_j)^2 p(\tilde{f}_j) d\tilde{f}_j + \int_{-\lambda}^{+\lambda} \tilde{f}_j^2 p(\tilde{f}_j) d\tilde{f}_j + \int_{+\lambda}^{+\infty} (\tilde{f}_j - \beta_j)^2 p(\tilde{f}_j) d\tilde{f}_j.
\end{align*}
This integration leads to the expression of distortion as:
\begin{equation}
D_j = \sigma_j^2 + 2\beta_j^2 \mathcal{Q}\Big(\frac{\lambda}{\sigma_j}\Big) - \frac{4\beta_j \sigma_j}{\sqrt{2\pi}} \exp\Big(\frac{-\lambda^2}{2\sigma_j^2}\Big).\label{eq:SingleLayer_STC_Dist-per-dim}  
\end{equation}

\textbf{Weighting vector:} Now we can find optimal $\beta_j$ that minimizes $D_j$. Fortunately, this can be expressed in a closed-form as: 
\begin{equation}  \label{eq:SingleLayer_STC_beta*-per-dim}
\beta_j^* = \argmin_{\beta_j} D_j = \frac{\sigma_j \exp\Big(\frac{-\lambda^2}{2\sigma_j^2}\Big)}{\sqrt{2\pi} \mathcal{Q}\Big(\frac{\lambda}{\sigma_j}\Big)}.  
\end{equation}

Fig. \ref{subfig:SingleLayer_STC_thresholding_phiBeta} sketches $\beta_j$ in the ternary encoding of a Gaussian distribution. Note that in the special case whe $\lambda=0$, the ternary encoding reduces to binary encoding and Eq. \ref{eq:SingleLayer_STC_beta*-per-dim} reduces to the well-known binary quantization formula $\Delta = \pm \sqrt{\frac{2}{\pi}} \sigma$, e.g., as in (10.1) of \cite{CoverThomas200607}. 

It should be mentioned that for the storage of codes, since $\boldsymbol{\beta}$ is the same for all codes, $\mathbf{x}$'s can be stored as fixed-point $\{+1,0,-1\}$ values in memory. In fact, $\boldsymbol{\beta}$ is necessary to convert these fixed-point codes to floating-point values that can reconstruct the $\mathbf{f}$'s. 

\textbf{Rate:} Now that we have fully characterized the distortion of ternary encoding, we should also characterize its rate. In fact, this is similar to entropy calculation of section \ref{subsec:SingleLayer_Analysis_IT}, where we had independent ternary variables with different sparsities. In this setup, however, we can only guarantee that these ternary variables are un-correlated. Therefore, by assuming independence, we can provide an upper-bound for rate as:

\begin{equation}
\begin{aligned}
&\mathcal{R} \leqslant  \frac{1}{n} H(\mathbf{X}) = \frac{1}{n} \sum_{j = 1}^n H(X_j) = \\ 
 &-\frac{1}{n} \sum_{j = 1}^n \Big(2\alpha_j \log_2(\alpha_j) + (1 - 2\alpha_j) \log_2 (1 - 2\alpha_j) \Big),   
\end{aligned} \label{eq:SingleLayer_STC_Rate}  
\end{equation}
where we used the ternary entropy of Eq. \ref{eq:SingleLayer_STC_H-ternary}.

\begin{figure} 
   \begin{center} 
\includegraphics[width=0.8\textwidth]{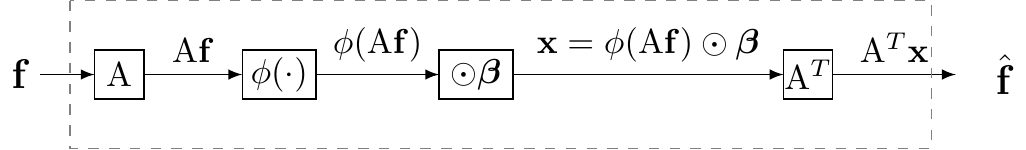} 
   \end{center}
   \caption{A unit of STC encoding and decoding for reconstruction.}
   \label{fig:SingleLayer_STC_Diagram}
   \end{figure}

\textbf{Summary:} Here we summarize the STC encoding of this section \ref{subsubsec:SingleLayer_Analysis_Reconstruction_DSW18}. Assuming that the data in the projection domain marginally follows the Gaussian distribution, we first apply the PCA transform to de-correlate the data. The ternarization is then performed according to Eq. \ref{eq:SingleLayer_STC_EncodingWeighted} for which values of $\boldsymbol{\beta}$ are calculated from Eq. \ref{eq:SingleLayer_STC_beta*-per-dim}. The decoding step for reconstruction is then performed using Eq. \ref{eq:SingleLayer_STC_Reconstruction} for which we showed that $\mathrm{B} = \mathrm{A}^T$ is the optimal choice. These steps are sketched in Fig. \ref{fig:SingleLayer_STC_Diagram}.

 \subsubsection{Linear decoding: a Procrustean approach} \label{subsubsec:SingleLayer_Analysis_Reconstruction_Procrustean}
 In section \ref{subsubsec:SingleLayer_Analysis_Reconstruction_DSW18}, we introduced a linear decoding scheme based on two assumptions. First, the data is stationary and admits a covariance matrix and hence we can de-correlate the data with PCA. Second, since we needed marginal Gaussianity in the projected domain, although the original data need not be a multivariate Gaussian itself, it should, however, belong to a restricted family of distributions in order for the CLT to be effective.
 
While these assumptions are not too restrictive, in practice, however, one might consider more complicated distributions. To put one step closer towards practicality (and hence getting one step farther from model-based derivations), we develop a learning framework that relies more on the data, and less on assumptions.\footnote{Note that in section \ref{subsubsec:SingleLayer_Analysis_Reconstruction_DSW18}, essentially, the only training part was the PCA. For the multi-layer version of those derivations, i.e., the ML-STC framework that we will introduce later in section \ref{sec:MultiLayer_MLSTC}, however, the situation is more complicated. In fact, the ML-STC uses successive PCAs on the residuals which makes it much more intricate than its single-layer version. For this section, anyway, we focus on a more complicated training algorithm for the single-layer case.}  
 
Coming back to Eq. \ref{eq:SingleLayer_STC_DmultiLine}, unlike in the previous section \ref{subsubsec:SingleLayer_Analysis_Reconstruction_DSW18}, let us not explicitly assume a Gaussian distribution for the projected $\tilde{\mathbf{F}}$, for this section.

Concretely, considering a training set $\mathrm{F} = [\mathbf{f}_1, \cdots, \mathbf{f}_N]$ (instead of a hypothetic random variable $\mathbf{F}$), Eq. \ref{subeq:SingleLayer_STC_DmultiLine_b} can be casted as the following  optimization problem:

 \begin{equation} \label{eq:SingleLayer_STC_Procrustean_main}
 \begin{aligned}
 & \underset{\mathrm{A},\boldsymbol{\beta}}{\text{minimize }}
 & &  \frac{1}{2nN} || \mathrm{A}\mathrm{F} -   \phi(\mathrm{A}\mathrm{F}) \odot \boldsymbol{\beta} ||_{\mathcal{F}}^2, \\
 & \text{s.t.}
 & & \mathrm{A} \mathrm{A}^T = \mathrm{I}_n. \\
 \end{aligned}
 \end{equation}

We solve this problem using the alternating minimization technique and in 2 steps: 

\begin{enumerate}[label=(\roman*)]
\item Fix $\mathrm{A}$, update $\boldsymbol{\beta}$:
We find the weighting vector $\boldsymbol{\beta}$ according to Eq. \ref{eq:SingleLayer_STC_beta*-per-dim}.
\item Fix $\phi(\mathrm{A}\mathrm{F})$ and  $\boldsymbol{\beta}$, update $\mathrm{A}$:
This takes the form of the famous orthogonal Procrustean problem  \cite{Procrustean:schonemann1966generalized}, for which a closed-form solution can be expressed for iteration $t+1$ of the algorithm as:

\begin{equation} \label{eq:SingleLayer_STC_Procrustean_solutionSVD}
\begin{aligned}
\mathrm{U}' \mathrm{\Sigma}' \mathrm{V}'^T &= \Big(\phi(\mathrm{A}_{[t]}\mathrm{F}) \odot \boldsymbol{\beta}_{[t]} \Big) \mathrm{F}^T, \\
\mathrm{A}_{[t+1]} &= \mathrm{U}'\mathrm{V}'^T,
\end{aligned}
\end{equation}
where $\mathrm{U}' \mathrm{\Sigma}' \mathrm{V}'^T $ is the Singular Value Decomposition (SVD) based on the current value of $\mathrm{A}$ at the iteration $t$.
\end{enumerate}

Note that (somehow) similar forms to the optimization problem of Eq. \ref{eq:SingleLayer_STC_Procrustean_main} appear in many applications. The works of \cite{6339108,ITQ:6296665} are successful examples where the technique of alternating minimization is used followed by the Procrustean solution as a powerful learning algorithm.

Algorithm \ref{alg:SingleLayer_ProcrusteanSTC} summarizes the steps of this approach.
\begin{algorithm} \caption{\texttt{Procrustean-STC}} \label{alg:SingleLayer_ProcrusteanSTC}
\begin{algorithmic}[0]
    \INPUT  Training set $\mathrm{F}$, ternary threshold $\lambda$ 
    \OUTPUT Projection matrix $\mathrm{A}$, weighting vector $\boldsymbol{\beta}$, database codes $\mathrm{X}$
    
\end{algorithmic}
\begin{algorithmic}[1]
\State Estimate the covariance matrix $\mathrm{C}_{\mathbf{F}}$ of the data and decompose it using Eigenvalue decomposition as $\mathrm{C}_{\mathbf{F}} = \mathrm{U}\mathrm{\Sigma}\mathrm{U}^T$.
\State $\mathrm{A} \gets \mathrm{U}^T$  \Comment Initialization of the projection matrix
\While { $\hat{\mathcal{D}}$ not converged, }
\Statex \textbf{i.} Fix $\mathrm{A}$, update $\boldsymbol{\beta}$:
\For{$j = 1,\cdots, n$} \Comment $\boldsymbol{\beta} = [\beta_1, \cdots, \beta_n]^T$.
\State $\beta_j \gets  \frac{\sigma_j \exp\Big(\frac{-\lambda^2}{2\sigma_j^2}\Big)}{\sqrt{2\pi} \mathcal{Q}\Big(\frac{\lambda}{\sigma_j}\Big)} $  \Comment $\mathrm{\Sigma} = \text{diag}([\sigma_1^2, \cdots, \sigma_n^2]^T)$
\EndFor 
\Statex \textbf{ii.} Fix $\boldsymbol{\beta}$, update $\mathrm{A}$:
\State $\mathrm{U}', \mathrm{\Sigma}', \mathrm{V}'  \gets \text{SVD}\Big[ \big(\phi(\mathrm{A}_{[t]}\mathrm{F}) \odot \boldsymbol{\beta}_{[t]} \big) \mathrm{F}^T \Big]$ 
\State $\mathrm{A} \gets \mathrm{U}'\mathrm{V}'^T$
\State $\hat{\mathcal{D}} \gets \frac{1}{2nN} || \mathrm{A}\mathrm{F} -   \phi(\mathrm{A}\mathrm{F}) \odot \boldsymbol{\beta} ||_{\mathcal{F}}^2 $
\EndWhile
\end{algorithmic}
\end{algorithm}

In chapter \ref{chapter:MultiLayer}, we extend this algorithm to multi-layers and develop the ``ML-STC-Procrustean'' algorithm as a powerful learning approach.
\subsubsection{Rate-distortion behavior} \label{subsubsec:SingleLayer_Analysis_Reconstruction_RD}
We are now ready to study the rate-distortion behavior of the ternary encoding. Let us focus on the linear decoding case and see how it behaves in compression of correlated sources.

\begin{figure} 
   \begin{center} 
\includegraphics[width=0.99\textwidth]{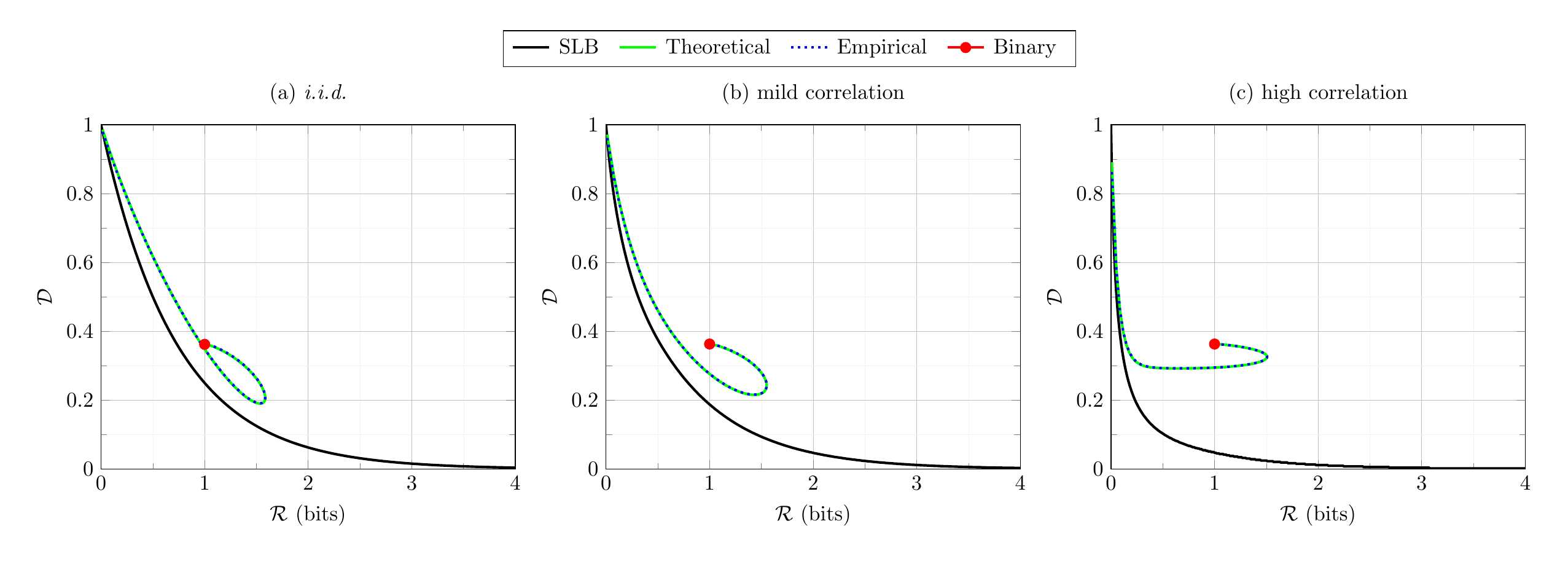} 
   \end{center}
   \caption{Distortion-rate curves of (single-layer) STC under AR(1) Gaussian source with varying correlation factors: (a) $\rho=0$, (b) $\rho=0.5$ and (c) $\rho=0.9$.}
   \label{fig:SingleLayer_STC_DR}
   \end{figure}

To model correlation, we use the AR(1) Gaussian source, i.e., $\mathbf{F} \sim \mathcal{N}(\mathbf{0}, \text{Toeplitz}[\rho^0, \cdots, \rho^{n-1}])$, which models a lot of signals in practice. Fig. \ref{fig:SingleLayer_STC_DR} shows the distortion-rate curves under 3 different correlation levels. For every figure, three curves are shown: the Shannon Lower Bound (SLB) which is the theoretical lower bound achieved only in the asymptotic case of $n \rightarrow \infty$\footnote{Note that tighter bouds and more realistic results can be achieved within the finite-blocklength regimes, e.g., as in \cite{6145679}.} for any lossy source-coding scheme, the theoretical characterization of the STC distortion derived from Eq. \ref{eq:SingleLayer_STC_Dist-per-dim} ($\mathcal{D} = \sum_{j=1}^nD_j$), and the empirical distortion calculated from simulations performed on $N = 10,000$ randomly generated vectors of dimension $n = 512$. Also the extreme case of binary encoding, i.e., zero sparsity, corresponding to $\lambda = 0$ and hence $\mathcal{R} = 1$ is marked. These curves are calculated by varying $\lambda$ from a very high value to $\lambda=0$.\footnote{Note that the special curvature of rate of Fig. \ref{fig:SingleLayer_STC_DR} is due to the ternary entropy vs. $\lambda$ of Fig. \ref{fig:SingleLayer_STC_H}.}

It is clear from Fig. \ref{fig:SingleLayer_STC_DR} that the rate-distortion behavior, particularly for the highly correlated sources, is good only at the very low-rate regimes. This performance deviates from optimal behavior as the rate increases. In particular, for our $m=n$ assumption here, this structure cannot target rates higher than $\mathcal{R} \simeq 1.58$.

In order to explain this behavior, we should study how the rate allocation is done and how it deviates from the optimal behavior. In fact, in our current setup we have a \textit{i.n.i.d.} Gaussian distribution for $\tilde{\mathbf{F}}$, for which the optimal rate allocation was specified in Eq. \ref{eq:fundamentals_OptRate}.\footnote{Remember we studied the rate-allocation of \textit{i.n.i.d.} Gaussian sources in section \ref{subsubsec:fundamentals_SourceCoding_RD_inid} and later we used its results in section  \ref{subsec:SingleLayer_Synthesis_VRKmeans} to come up with the \textit{VR-Kmeans} prior.}

Fig. compares the rate allocation of STC from Eq. \ref{eq:SingleLayer_STC_Rate} with optimal allocation of Eq. \ref{eq:fundamentals_OptRate} for 3 different rate-regimes and for the same sources of Fig. \ref{fig:SingleLayer_STC_DR}. Obvious from the figure, as the rate increases, the mismatch between optimal and actual allocations deviate. In fact, this phenomenon describes the poor performance of STC at high rates.

 \begin{figure} 
   \begin{center} 
\includegraphics[width=0.9\textwidth]{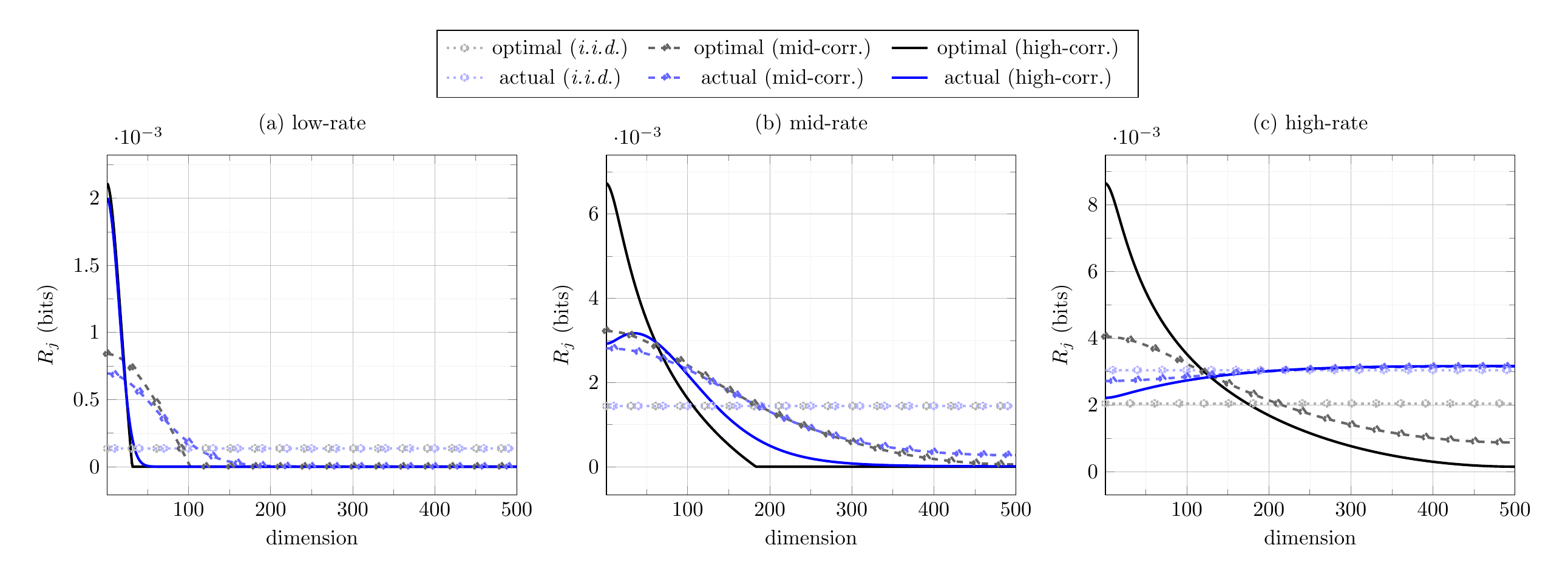} 
   \end{center}
   \caption{Mismatch of rate allocation of (single-layer) STC with optimal rate allocation. Higher rate regimes show more mismatch.}
   \label{fig:SingleLayer_STC_Mismatch}
   \end{figure}

This phenomenon has an important practical consequence; that the ternary encoding should be operated only at very low rates, or equivalently, high sparsity levels. \footnote{This means that the extreme case of binary encoding, in fact, leads to inferior RD performance. We elaborate on this later in chapter 8 and for the comparison of binary hashing and STC.} What do we do to target higher rates? We answer this question in the next chapter and under the multi-layer architecture.

\section{Conclusions} \label{sec:SingleLayer_Conclusions}
This chapter is the starting point for the development of our algorithms in this thesis. Driven by the applications we will encounter in the 3rd part of the thesis, we started by formalizing our objectives, albeit in a very general and anstract way, within Eqs. \ref{eq:SingleLayer_GeneralEnc} and \ref{eq:SingleLayer_GeneralEncDec}.

Motivated by the literature of signal processing we reviewed in chapter \ref{chapter:ModelingLit}, we concretized our objectives within two forms, i.e., the synthesis and the analysis models. From the other hand, our large-scale applications require fast and compact encoding and particularly decoding. So we developed these two models under such constraints.

Within the family of synthesis models, to avoid slow decoding and complications due to generally NP-hard nature of synthesis dictionaries, we limited encoding to the more constraint family of Vector Quantizers. We then analyzed this structure from a probabilistic point of view and noticed that the basic form of VQ, i.e., the K-means algorithm, is too unstructured and does not take into account such probabilistic recipes. In the form of regularization to the K-means, we injected an important prior to the reverse water-filling paradigm of rate-distortion theory and formulated the VR-Kmeans of Eq. \ref{eq:SingleLayer_VRKmeans}. We then solved this optimization problem and proposed an iterative procedure in Algorithm \ref{alg:SingleLayer_VR-Kmeans} and analyzed its solution and convergence. We then demonstrated the usefulness of this regularization and showed that the VR-Kmeans successfully avoids over-fitting of the K-means in high-dimensional scenarios, both for synthesized and real data.

As for the family of analysis models, inspired by the famous transform sparse coding problem of Eq. \ref{eq:ModelingLit_AnalysisL0Code} for which the closed-form solution of Eq. \ref{eq:ModelingLit_AnalysisL0CodeSolution} exists, and from the other hand limited by our constraints to have discretized alphabet, we use the ternarizing function instead of the hard-thresholding function and hence, come up with ternary alphabets. This initiates our Sparse Ternary Codes (STC) framework which will be developed in several levels throughout the thesis.

After a brief information-theoretic analysis of the STC  which will be used later in chapter \ref{chapter:Search}, we consider reconstructing from such codes and provide 3 types of solutions: First, a non-linear decoder whose solution turns out to be a quadratic program with linear constraints. This, however, is slow to solve and not useful for our large-scale applications. Second, a linear decoder based on a set of assumptions that leads to a very convenient solution based on the PCA. Finally, the third solution is relaxing the assumptions and provides an alternating optimization procedure based on the orthogonal Procrustes problem.

For both these family of algorithms, however, we showed that the desirable rate-distortion performance is achievable only within a low-rate operational regime. This being a fundamental property of these single-layer architectures, motivates us to extend these models to multiple layers and while benefiting from their conveniences, increase their operational range to arbitrarily high rate-regimes. This is discussed next in chapter \ref{chapter:MultiLayer}.

\chapter{Multi-layer architectures} \label{chapter:MultiLayer}
We developed the theoretical and algorithmic bases for this thesis in chapter \ref{chapter:SingleLayer}, where we started with basic signal decompositions under 2 main models, i.e., the synthesis and the analysis prior models. Constrained by the kind of applications we will encounter later, we then developed each of these models to intricate frameworks, i.e., the \textit{VR-Kmeans} algorithm under the synthesis model, and the STC under the analysis model. We also analyzed their behaviors and shortcomings.

In particular, we saw that both of these models suffer from a severe limitation. Independent from the learning algorithm used or the over-training issues, they cannot operate at high rates. This, in fact, is the fundamental limitation of what we referred to as single-layer architectures.  

In practical applications, however, it is often required to provide a very high-fidelity reconstruction within a reasonable rate budget. An obvious example is in image compression where often high-quality image content is desired while storage is limited. 

What do we do to target higher rates? As was pointed out earlier in section \ref{sec:ModelingLit_Proposed}, a general strategy of this thesis, which will be justified in this chapter, is to use low capacity algorithms, but in succession, while each time we improve upon the previous result. 

Equivalently stated, the idea is to use multiple layers of low-rate algorithms to provide an equivalent high total rate to ensure a high-quality reconstruction. Therefore, while the algorithms of chapter \ref{chapter:SingleLayer}, i.e., the \textit{VR-Kmeans} and STC had single-layer architectures, in this chapter, they will be evolved to multi-layer architectures, i.e., the RRQ and the STNets, respectively.

Section \ref{sec:MultiLayer_SR} provides some general insights into the idea of successive approximation and in particular the framework of ``successive refinement'' in information theory, which inspires us for the development of the main idea of this chapter, i.e., the multi-layer processing. Section \ref{sec:MultiLayer_RRQ} discusses the evolution of \textit{VR-Kmeans} to RRQ. We first review the concept of compositional VQ in the literature and then focus on the technique of Residual Quantization (RQ) based on which we develop the RRQ. Section \ref{sec:MultiLayer_MLSTC} uses very similar ideas to develop STC to its multi-layer evolution, i.e., the ML-STC. 

To further advance matters, the favorable form of the analysis model (as opposed to the synthesis model) enables us to use the ``back-propagation'' technique to further tune the ML-STC. We introduce this evolution in section \ref{sec:MultiLayer_STNets} and term it as ``STNets'', an architecture whose roots are the analysis model concepts in signal processing while it has common traits with neural networks and can now benefit from some of the practical recipes developed within the deep learning communities. 
\section{Successive Refinement of information} \label{sec:MultiLayer_SR}
The idea of successive refinement involves the repeated approximation of a source of information while each stage incrementally improves the approximation quality of the previous stage. Numerous applications can benefit from this idea. Suppose for example the compression of images where due to bandwidth constraints, a coarse description of the image is initially sent to the users. Within the same bit-stream, users that have higher bandwidth can receive finer image details while others can stop the communication. Another example can be in image retrieval systems with computational restrictions, where the query can initially be matched with only very short descriptions of the entire database, and the quality of matching can improve with longer descriptions, but only within a limited number of relevant database candidates, and hence an overall speed-up.


This natural and intuitive idea has of course been considered long ago and in different forms. In information theory, perhaps inspired by earlier works and practical insights in image coding, this idea appears with information-theoretic formalizations in \cite{SR:koshelev1980hierarchical,SR-equitz1989successive}. In the famous work of \cite{SR-Cover:75242}, the necessary and sufficient conditions for a source to be successive-refinable were derived. Then \cite{SR-Rimoldi:272493} provided a complete characterization of the achievable rate region. It is then proved in \cite{SR:866419} that the conditions of being successively refinable are not very restrictive and all sources are nearly successively refinable.

To gain an insight into this idea, first remember the fundamental result of the rate-distortion theory described earlier in section \ref{subsec:fundamentals_SourceCoding_RD}, where for $\mathbf{F} = [F_1, \cdots, F_n]^T$ and under $n \rightarrow \infty$, for a given distortion value $D$, all rates above the $\mathcal{R}(D)$ of Eq. \ref{eq:fundamentals_RD} are achievable.

The successive refinement of the description of $\mathbf{F}$ involves a hierarchy of approximations $\hat{\mathbf{F}}_1$, $\hat{\mathbf{F}}_2$, $\cdots$, $\hat{\mathbf{F}}_L$ with a particular structure. The first approximation $\hat{\mathbf{F}}_1$ incurs a distortion $D_1$ with a description rate of $R_1$. The second description complements the first description with rate $R_2$, i.e., adds a complementary description to the first bit-stream totaling $R_1 + R_2$ bits, and provides the approximation $\hat{\mathbf{F}}_2$ with distortion $D_2$. Similarly, all the other descriptions incrementally add to the bit-stream and finally provide $\hat{\mathbf{F}}_L$. Obviously, we should have that $D_L  < \cdots < D_2 < D_1$ for this construction to be useful. Moreover, it is desired that each of these descriptions be both independently and collectively optimal, i.e., as good as if we did a single-stage optimal description. 

Thus, formally, for a source $F$ to be successively refinable, for any choice of $R_l$ with $1 \leqslant l \leqslant L$, we should be able to maintain: 
\begin{equation} \label{eq:MultiLayer_SR_Definition}
\begin{aligned}
R_1              &= \mathcal{R}(D_1), \\ 
R_1 + R_2        &= \mathcal{R}(D_2),  \\
                 &\vdots             \\ 
\sum_{l=1}^L R_l &= \mathcal{R}(D_L).        
\end{aligned}
\end{equation}

It is shown in \cite{SR-Cover:75242} that the necessary and sufficient condition for the above criteria to be achieved is to be able to find $p(\hat{f}_l|f)$'s, such that the sequence of Eq. \ref{eq:MultiLayer_SR_Markov} be a Markov chain.
\begin{equation}  \label{eq:MultiLayer_SR_Markov}
F \rightarrow \hat{F}_L \rightarrow \cdots \rightarrow \hat{F}_2 \rightarrow \hat{F}_1
\end{equation}

While the Markovianity condition cannot formally be maintained for all sources, fortunately, it was further shown that the Gaussian source above, provided that $n \rightarrow \infty$, satisfies Eq. \ref{eq:MultiLayer_SR_Markov} and hence is successively refinable.
\subsection{Additive structure}  \label{subsec:MultiLayer_SR_Additive}
For general sources, the promise of successive refinability described above is realizable under general tree-like hierarchies. When $L$ is desired to be large, however, the memory constraints become exponentially critical for trees. Fortunately though, it was further shown in \cite{SR-Cover:75242} that for the case of Gaussian sources, which is the interest of our thesis and many other practical applications, this tree-like structure can be simplified to a very ``nice'' particular case, i.e., the additive structure whose storage cost and also decoding complexity increases only linearly with $L$.

The additive structure for successive refinement was formalized in \cite{SR:1214076}, where its rate-distortion was studied and the achievable rate-regions where characterized. Furthermore, necessary and sufficient conditions for the optimality of the additive successive refinement were derived which are valid for many sources.

An effort to realize these ideas as coding strategies is the framework of Sparse Regression Codes (SPARC) for lossy compression \cite{SPARC:6777349,SPARC:6781602,SPARC:6284210}, which takes the linear regression codes previously used in channel coding \cite{SPARC:5513330} and suggests a computationally efficient source coding scheme achieving Shannon limits. While their structure is essentially very similar to what had earlier been proposed informally in \cite{SR-equitz1989successive}, the authors provide much more rigorous characterizations including the error exponent of excess distortion. Moreover, they show that the scheme achieves the rate-distortion performance of \textit{i.i.d.} Gaussian, for any ergodic source, and with a favorable error-exponent.

To see why this structure is advantageous in terms of memory and complexity, recall the encoder considered in the basic rate-distortion setup of section \ref{subsec:fundamentals_SourceCoding_RD}, i.e., $\mathbb{Q}[\cdot]: \mathcal{F}^n \to \{1,\cdots, 2^{nR}\}$. If the targeted rate is high, the number of code indices, i.e., $2^{nR}$ should accordingly be exponentially high. This is impossible to be stored in memory, even for moderate values, e.g., let $n=100$ and $R=1$, the required index size is more than the number of atoms in the observable universe! Moreover, on the decoder side, to provide the approximation $\hat{\mathbf{F}}$ with the corresponding distortion $D$, the same scale is required to perform the exhaustive search on the corresponding codewords. 

Suppose instead we have $L$ such additive encoders, i.e., $\mathbb{Q}_1[\cdot], \cdots, \mathbb{Q}_L[\cdot]$, each operating at rate $\frac{R}{L}$. According to the definition of successive refinability of Eq. \ref{eq:MultiLayer_SR_Definition}, to maintain the same distortion $D$ as in the one stage-encoding, i.e., to have $\hat{\mathbf{F}}_L = \hat{\mathbf{F}}$, the equivalent rate of this setup is $R$ which is the same as the one-stage encoding above. However, due to the additive structure, the total index size is $L \times 2^{\frac{nR}{L}}$. Suppose we set $L=100$, the memory and complexity requirement of one-stage encoding, i.e.,  $2^{100}$ becomes only $200$ in this additive structure, a huge reduction indeed!

To get a clearer understanding of the additive construction and its comparison with single-stage encoding, perhaps it is more intuitive to realize the idea of additive successive refinement through a synthesis codebook with sparsity one, i.e., our setup of section \ref{sec:SingleLayer_Synthesis}. In fact, this is also the structure of Shannon random codes used in the proof of his rate-distortion theorem \cite{shannon1959coding}.


%
%
%
So let us come back to the idealized setup of section \ref{subsubsec:SingleLayer_Synthesis_probabilist_iid}, i.e., assume $\hat{\mathbf{F}} \sim \mathcal{N}(\mathbf{0},\sigma^2 \mathrm{I}_n)$, while $n \rightarrow \infty$. 

The aim is to do the encoding in several stages as above. So imagine an initial stage of encoding at rate $R_1$. We showed that the optimal codebook $\mathrm{C}_1$ should be generated at $\sigma_{C_1}^2 = \sigma^2 (1 - 2^{-2R_1})$, which leaves the distortion of $D_1 = \sigma^2 2^{-2R_1}$. The additive structure implies that the next stage should encode the residual $\mathbf{E} \triangleq \mathbf{F} - \hat{\mathbf{F}}$ with variance $D_1$, using the codebook $\mathrm{C}_2$ with $\sigma_{C_2}^2 = D_1 (1 - 2^{-2R_2}) $ and with rate $R_2$, while leaving the distortion $D_2 = D_1 2^{-2R_2} = \sigma^2 2^{-2(R_1 + R_2)}$.

So this structure can accordingly be generalized using the following simple recursion rule:

\begin{equation}  \label{eq:MultiLayer_SR_Spheres}
\begin{aligned} 
D_0 &= \sigma^2 \\
D_l &= D_{l-1} 2^{-2R_l} = \sigma^2 2^{-2(R_1 + \cdots + R_l)}  \\
\sigma_{C_l}^2 &= D_{l-1} - D_l,
\end{aligned}
\end{equation}
for $l = 1, \cdots, L$.

\begin{figure}  
   \begin{center} 
\includegraphics[width=0.9\textwidth]{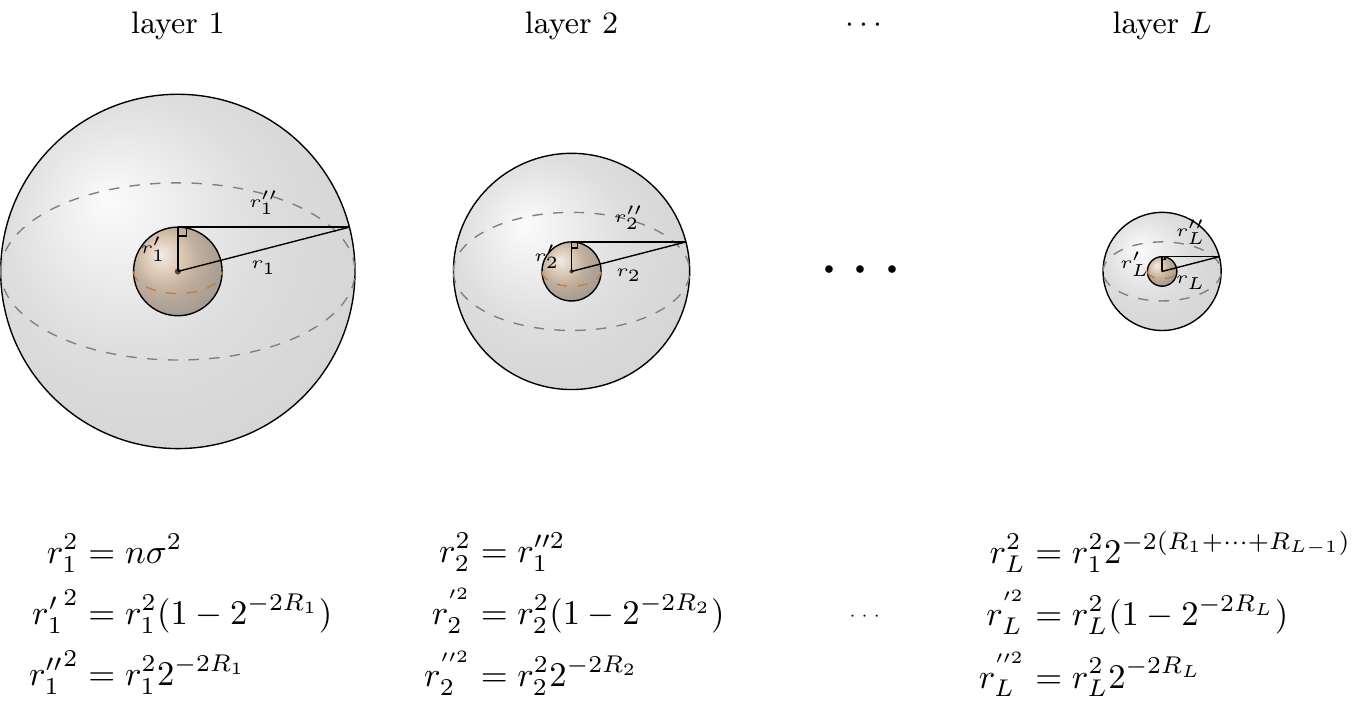}

   \end{center}
   \caption{Spheres of ambiguity diminishing as more layers describe the data.}
   \label{fig:MultiLayer_SR_Spheres}
   \end{figure}

Therefore, the sphere of ambiguity with relative entropy $h(F) =  \frac{n}{2} \log_2{\Big(2 \pi e \sigma^2\Big)}$, where we started from, gets smaller and smaller. Finally, after $L$ stages, the ambiguity is minimized as $ h(\mathbf{F} - \hat{\mathbf{F}}_{L-1}) = \frac{n}{2} \log_2{\Big(2 \pi e \sigma^2 2^{-2(R_1 + \cdots + R_L)}\Big) }$ bits. This picture is schematically depicted in Fig. \ref{fig:MultiLayer_SR_Spheres}.  

In this section, the provided arguments were based mostly on information-theoretic concepts. From the practical side, on the other hand, solutions similar to the above additive structure have extensively been used in speech and image coding. We come back to this idea shortly in section \ref{subsubsec:MultiLayer_RRQ_SOTA_RQ} and propose our solution in section \ref{subsec:MultiLayer_RRQ_RRQ}.  

\section{Regularized Residual Quantization (RRQ)} \label{sec:MultiLayer_RRQ}
In the previous section, we saw that the basic single layer structure (with sparsity $k=1$) has serious limitations in terms of achieving arbitrary high operational rates. In particular, memory and complexity constraints limit their use to very low rates, since the required index set, and hence codebook size will be exponentially huge. 

We then saw a remedy using the concept of additive successive refinement, where additive structures can virtually create the required exponentially large index size, but with actual cost linear in the number of layers. We also mentioned some works that provide theoretical guarantees for this idea in terms of achieving Shannon limits.

While this need not be the only possible solution\footnote{For example, one possible way can be to increase the sparsity $k$. This way we will have an equivalent combinatorially big index set, i.e., ${m \choose k}$, while keeping the codebook size $m$ within moderate values. However, we saw in section \ref{subsec:ModelingLit_SP_Synthesis} that this will lead to the sparse coding problem, which is NP-hard in nature and even its practical remedies are still very costly for our applications in this thesis.}, due to its theoretical foundations, its practicality and its compatibility with our algorithms, we adopt the additive successive structure in the development of our algorithms in this thesis.

However, before developing this idea further, in section \ref{subsec:MultiLayer_RRQ_SOTA} we briefly review some of the solutions from the literature beyond the successive refinement concept. In particular, we take the existing Residual Quantization (RQ) framework, see its limitations and develop it in section \ref{subsec:MultiLayer_RRQ_RRQ} to what we term the Regularized Residual Quantization (RRQ) framework. We then show its advantages in terms of rate-distortion performance in section \ref{subsec:MultiLayer_RRQ_DR}. 
\subsection{Multi-layer VQ: literature review} \label{subsec:MultiLayer_RRQ_SOTA}
We saw earlier in section \ref{sec:SingleLayer_Synthesis} that within the synthesis model, putting the 1-sparsity constraint into Eq. \ref{eq:SingleLayer_GeneralEncDec}, results in the formulation of the K-means algorithm of Eq. \ref{eq:SingleLayer_Kmeans}. We then injected the so-called reverse water-filling prior into K-means and formulated the VR-Kmeans algorithm of Eq. \ref{eq:SingleLayer_VRKmeans}, and saw how it helps with over-fitting.

However, both the K-means and VR-Kmeans algorithms are synthesis models with 1-sparsity. Therefore, as we explained above, and as was validated before in the experiments of section \ref{subsec:SingleLayer_Synthesis_RD}, they both suffer from the rate-limitation issues.\footnote{Note that the information-theoretic arguments above were under a probabilistic assumption and for $n \rightarrow \infty$, while K-means is a learning algorithm without any particular assumptions. Nevertheless, these are architectural limitations that also apply to K-means, as well as the VR-Kmeans.}

Let us see some of the remedies to these limitations from the literature. In particular, let us focus on the algorithms that provide discrete code spaces, i.e., the family of Vector Quantizers (VQs).

Within the setup of K-means, we assume again that we are given a set of training samples $\mathrm{F} = [\mathbf{f}_1, \cdots, \mathbf{f}_N]$. To encapsulate the  family of VQs under one formulation, let us remove the 1-sparsity constraint of Eq. \ref{eq:SingleLayer_Kmeans} and put a generic constraint on both the code and the codebook. This general formulation becomes:
\begin{equation} \label{eq:MultiLayer_VQGeneral}
\begin{aligned}
& \underset{{\mathrm{C},\mathrm{X}}}{\text{minimize}}
& & \frac{1}{2Nn}|| \mathrm{F} - \mathrm{C} \mathrm{X} ||_{\mathcal{F}}^2\\
& \text{s.t.} & & \Omega \{\mathrm{X} \}, \Omega \{\mathrm{C}\},
\end{aligned}
\end{equation}
where $\Omega \{\mathrm{X} \}$ is a constraint on the code and $\Omega \{\mathrm{C}\}$ is a constraint on the codebook. We see next how different VQ methods set these constraints.

As a historical remark, the VQ-based methods were very popular, both for practice and research, during the 1970's to 1990's\footnote{Perhaps this popularity was later overshadowed by the rise of (synthesis) dictionary learning methods.} and mostly within the signal processing communities, in particular, image and speech coding (See for example \cite{gersho1991vector}). However, during the 2010's, starting with \cite{PQ:5946540}, there was a resurgence of the popularity of VQ within the computer vision community and for the similarity search problem. We note that our emphasis in this review of VQ is mainly on similarity search.
\subsubsection{Product Quantization (PQ)} \label{subsubsec:MultiLayer_RRQ_SOTA_PQ} 
The main idea behind Product Quantization (a.k.a. product codes) \cite{gersho1991vector} (and lots of earlier works, e.g., \cite{VQ:sabin1984product}) is to consider a long vector in $\Re^n$ as $p$ sub-vectors in $\Re^{\frac{n}{p}}$. These sub-vectors, having lower dimensions, are easier to encode. Notice that the single-layer alphabet size of $2^{nR}$ breaks down to $p$ much smaller alphabets of size $2^{(\frac{n}{p})R}$, providing virtually the same equivalent size. This trick is very similar to the additive quantization and provides similar storage and complexity savings. The codebook and the encoding construction adopted by PQ can be formalized as follows. 

Suppose the code vector $\mathbf{x} \in \Re^n$ is constructed as the concatenation of $p$ sub-codes $\mathbf{x}^{[p']}$'s for $1 \leqslant p' \leqslant p$ and $\mathbf{x}^{[p']} \in \Re^{\frac{n}{p}}$. PQ maintains the following structure:

\begin{equation} \label{eq:MultiLayer_PQ}
\begin{aligned}
\Omega^{\text{PQ}} \{\mathrm{X}\}:& &||\mathbf{x}^{[p']}||_0 = ||\mathbf{x}^{[p']}||_1 = 1, \\
\Omega^{\text{PQ}} \{\mathrm{C}\}:& &\mathrm{C} =
\begin{bmatrix}
\mathrm{C}^{[1]} & \mathrm{0} & \cdots & \mathrm{0} \\
\mathrm{0} & \mathrm{C}^{[2]} & \cdots & \mathrm{0} \\
\vdots  & \vdots & \ddots & \vdots \\
\mathrm{0} & \mathrm{0} & \cdots & \mathrm{C}^{[p]} 
\end{bmatrix}, 
\end{aligned}
\end{equation}
for $p' = 1, \cdots, p$, where $\mathrm{0}$ is an $(\frac{n}{p} \times \frac{n}{p})$ all-zero matrix. 

Therefore, the equivalent codebook is the Cartesian product of each sub-codebook providing $m^p$ equivalent codewords. In fact, this structure has an underlying assumption that each of the $p$ signal chunks is independent of each other and hence can be trained independently using K-means. 

While the PQ codes lack theoretical guarantees, in practice, they prove to be very successful and widely adopted, particularly for search applications. We will come back to the PQ codes in search in chapter \ref{chapter:Search}. 

\subsubsection{Optimized Product Quantization (OPQ)} \label{subsubsec:MultiLayer_RRQ_SOTA_OPQ}  
The independence assumption of PQ is not realistic in practice. This is why the OPQ \cite{OPQ:6619223} (or equivalently the Cartesian K-means of \cite{OPQ:Nowrouzi6619232}) optimizes the PQ encoding by rotating the data before performing the PQ encoding. The OPQ is essentially solving the following optimization problem.

\begin{equation} \label{eq:MultiLayer_OPQ}
\begin{aligned}
& \underset{{\mathrm{C},\mathrm{X},\mathrm{R}}}{\text{minimize}}
& & ||\mathrm{F} - \mathrm{R} \mathrm{C} \mathrm{X}||_{\mathcal{F}}^2 \\
& \text{subject to}
& & \Omega^{\text{PQ}} \{\mathrm{X}\}, \Omega^{\text{PQ}} \{\mathrm{C}\}, \\
& &  &\mathrm{R}^T\mathrm{R}^T = \mathrm{I}_n,
\end{aligned}
\end{equation}
where $\mathrm{R}$ is a rotation matrix.

Authors of \cite{OPQ:6619223} propose a parametric solution to Eq. \ref{eq:MultiLayer_OPQ}, assuming an independent variance-decaying Gaussian distribution on the data dimensions, as well as a general-purpose non-parametric solution that essentially iterates between a Procrustean formulation for $\mathrm{R}$ and the usual K-means. Before the rotation and as a pre-processing, the dimensions are balanced in terms of variance. This makes the quantization more efficient, as long as $p$, the number of partitions is small. 

While the OPQ achieves a performance improvement w.r.t. the PQ, as we will show later, this improvement is limited to a particular rate regime.

Other than the PQ and the OPQ, there exists a couple of solutions trying somehow to put constraints on the codebooks and the encoding procedure to address the limitations of the single-layer K-means. For example, the Additive Quantization (AQ) \cite{AQ:babenko2014additive} uses similar encoding to PQ but does not limit to orthogonal sub-codebooks and hence works with full-dimensional sub-codebooks. Thanks to its higher expressive power, AQ achieves superior performance to PQ, but this comes with the cost of a much slower combinatorial optimization than the K-means of PQ.

%

\subsubsection{Residual Quantization (RQ)} \label{subsubsec:MultiLayer_RRQ_SOTA_RQ}
While the family of PQ methods provides effective solutions to the limitations of the basic single-layer structures in terms of achieving arbitrary high rates, they do not have the benefits of successive approximations. For example, a PQ code learned to target a rate $R$, cannot help with the design of a PQ operating at rate $R'>R$. In fact, all the sub-codebooks of PQ at rate $R'$ should be re-trained from the beginning.

A structure which admits the idea of successive refinement of section \ref{sec:MultiLayer_SR} is the framework of Residual Quantization (RQ) or Residual Vector Quantization (RVQ). Unlike PQ, RQ can provide complementary descriptions of length $R'-R$ bits to an existing description at rate $R$, totaling a rate $R'>R$.

Apart from the above idea, since RQ has further an additive structure, it also solves the rate limitations of single-layer structures as we described in section \ref{subsec:MultiLayer_SR_Additive}. 

The idea of RQ has been around for many years in signal processing. In particular, speech coding was addressed with RQ first in \cite{RQ:1171604}. Many solutions for image coding have also been proposed based on RQ. Good reviews of these methods appear in \cite{RQ:480761,RQ:3776,RQ:212286}.

RQ imposes a hierarchical structure on its codebooks. This comes from the training procedure, which is based on encoding the residual of the previous layer. This idea is described in Algorithm \ref{alg:MultiLayer_RRQ_RQ}.   
\begin{algorithm} \caption{Residual Quantization} \label{alg:MultiLayer_RRQ_RQ}
\begin{algorithmic}[0]
    \INPUT  Training set $\mathrm{F}$ (whitened), $\#$ of layers $L$, $\#$ of codewords $m$ (assuming equal for all layers)
    \OUTPUT Codebooks $\mathrm{C}^{[l]}$'s and codes $\mathrm{X}^{[l]}$'s, for $l = 1, \cdots, L$
\end{algorithmic}

\begin{algorithmic}[1]
\State $\mathrm{F}^{[0]} \gets \mathrm{F}$ 
\State $\hat{\mathrm{F}} \gets \mathrm{0}$  \Comment all-zero matrix of $(n \times N)$
\State $\hat{\mathcal{D}}^{[0]} \gets 1$  \Comment Normalized distortion on the training set.
\For{$l = 1, \cdots, L$} 
\State $\mathrm{C}^{[l]}, \mathrm{X}^{[l]} \gets \text{K-means} (\mathrm{F}^{[l-1]},m)$ \Comment Solved using Algorithm \ref{alg:SingleLayer_Kmeans}.
\State $\hat{\mathrm{F}}^{[l-1]} \gets \mathrm{C}^{[l]} \mathrm{X}^{[l]}$
\State $\hat{\mathrm{F}} \gets \hat{\mathrm{F}} + \hat{\mathrm{F}}^{[l-1]}$  \Comment Current approximation
\State $\mathrm{F}^{[l]} \gets \mathrm{F} - \hat{\mathrm{F}}$   \Comment Residual (input to the next stage's quantizer)
\State $\hat{\mathcal{D}}^{[l]} \gets \frac{||\mathrm{F}^{[l]}||_{\mathcal{F}}^2}{||\mathrm{F}||_{\mathcal{F}}^2}$  \Comment Current (normalized) train-set distortion
\EndFor
\end{algorithmic}
\end{algorithm}

We saw earlier in section \ref{sec:SingleLayer_Synthesis} that the K-means formulation lacks structure and does not benefit from any prior on the joint description of sources. In particular, we saw e.g., in Fig. \ref{fig:SingleLayer_VRKmeans_DimDEMO} or Table \ref{Table:SingleLayer_VRKmeans_RD} that this becomes critical at high dimensions and makes encoding impossible with K-means at those limits.  

Similar to the PQ family, the RQ is based on the standard K-means algorithm. It is of no surprise, therefore, that RQ suffers from the limitations of K-means in high dimensions. It turns out that this is a severe limitation for RQ, making its design to only several layers. For example, we quote the following paragraph from \cite{RQ:480761} about different RQ-based methods:

``Nearly all RVQ literature considers the two-stage case and then inductively reasons that two-stage results can be generalized to RVQ’s with many stages. Although reasonable, there are problems that arise when this approach is adopted; design methods developed for two-stage RVQ’s may not be practical nor have satisfactory generalizations to many-stage RVQ’s due to unforeseen difficulties... There may exist a subset of the inputs of a RVQ stage where the decoded representation for each point in the subset degrades with the additional stage.''

\subsection{RRQ algorithm} \label{subsec:MultiLayer_RRQ_RRQ}
The ``unforeseen difficulties'' mentioned above, we argue, is due largely to the K-means formulation rather than the residual structure of RQ. To address these issues, we propose to replace the K-means in RQ with what we developed earlier in section \ref{subsec:SingleLayer_Synthesis_VRKmeans} and referred to as the VR-Kmeans algorithm. 

The resulting procedure becomes what we term as the Regularized Residual Quantization (RRQ), which is stated in Algorithm \ref{alg:MultiLayer_RRQ_RRQ}.
\begin{algorithm} \caption{Regularized Residual Quantization} \label{alg:MultiLayer_RRQ_RRQ}
\begin{algorithmic}[0]
    \INPUT  Training set $\mathrm{F}$, $\#$ of layers $L$, $\#$ of codewords $m$, target rate per layer $R$, thresholding hyper-parameter $\gamma'$, optimization hyper-parameter $\lambda$, Newton step $\eta$
    \OUTPUT Codebooks $\mathrm{C}^{[l]}$'s and codes $\mathrm{X}^{[l]}$'s, for $l = 1, \cdots, L$
\end{algorithmic}
\begin{algorithmic}[1]
\State $\mathrm{F}^{[0]} \gets \mathrm{F}$ 
\State $\hat{\mathrm{F}} \gets \mathrm{0}$  \Comment all-zero matrix of $(n \times N)$
\For{$l = 1, \cdots, L$} 
\State Estimate the covariance matrix of the whitened input $\mathrm{F}^{[l-1]}$ as $\text{diag} \big([\sigma_1^2, \cdots, \sigma_n^2]^T \big)$
\State $\mathrm{S}, \mathcal{A}_{\gamma} \gets \text{Rev-WFiller}([\sigma_1^2, \cdots, \sigma_n^2], R, \gamma') $   \Comment Solved using Algorithm \ref{alg:SingleLayer_WFiller}
\State $\mathrm{C}^{[l]}, \mathrm{X}^{[l]} \gets \text{VRK-means} (\mathrm{F}^{[l-1]},m, \mathrm{S},\mathcal{A}_{\gamma}, \lambda, \eta)$ \Comment Solved using Algorithm \ref{alg:SingleLayer_VR-Kmeans}.
\State $\hat{\mathrm{F}}^{[l-1]} \gets \mathrm{C}^{[l]} \mathrm{X}^{[l]}$
\State $\hat{\mathrm{F}} \gets \hat{\mathrm{F}} + \hat{\mathrm{F}}^{[l-1]}$  \Comment Current approximation
\State $\mathrm{F}^{[l]} \gets \mathrm{F} - \hat{\mathrm{F}}$   \Comment Residual (input to the next stage's quantizer)
\EndFor
\end{algorithmic}
\end{algorithm}


 \begin{figure} 
   \begin{center} 
\includegraphics[width=0.8\textwidth]{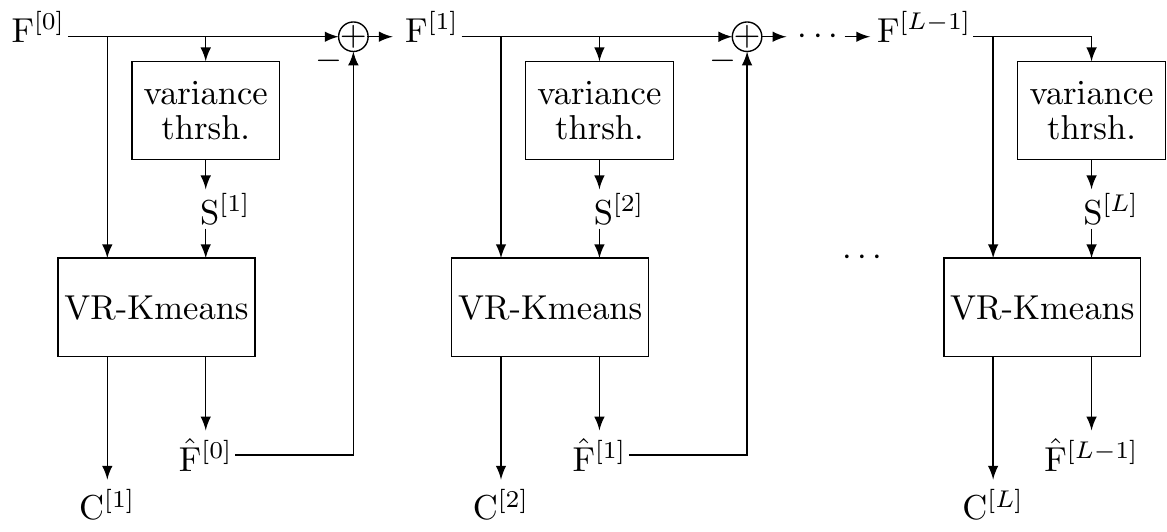} 
   \end{center}
   \caption{The RRQ algorithm based on the VR-Kmeans and Rev-WFiller.}
   \label{fig:MultiLayer_RRQ_diagram}
   \end{figure}

As the RRQ layers increase, the data approaches gradually to \textit{i.i.d.}. This means that the variance-decaying profile diminishes, and moreover, the inter-dependencies of dimensions not captured by the whitening will also reduce. Therefore, the hyper-parameters of Algorithm \ref{alg:MultiLayer_RRQ_RRQ} should also adapt accordingly. 

In particular, the regularization parameter $\lambda$ should increase as there is less and less data-particularities at higher layers and more and more resemblance to the \textit{i.i.d.} structure of section \ref{subsubsec:SingleLayer_Synthesis_probabilist_iid}. In practice, we find it appropriate to set $\lambda \rightarrow \infty$ after the fifth layer. This means that it is safe to avoid the optimization procedures of VR-Kmeans and simply generate the codebooks of $\mathrm{C}^{[l]}$ with $l \geqslant 5$ from the appropriate distributions. This, in fact, significantly speeds up the training procedure, making the RRQ a fast learning algorithm.
\subsection{Rate-distortion behavior} \label{subsec:MultiLayer_RRQ_DR}
\begin{figure}  [!h]

   \begin{center} 
   \includegraphics[width=0.99\textwidth]{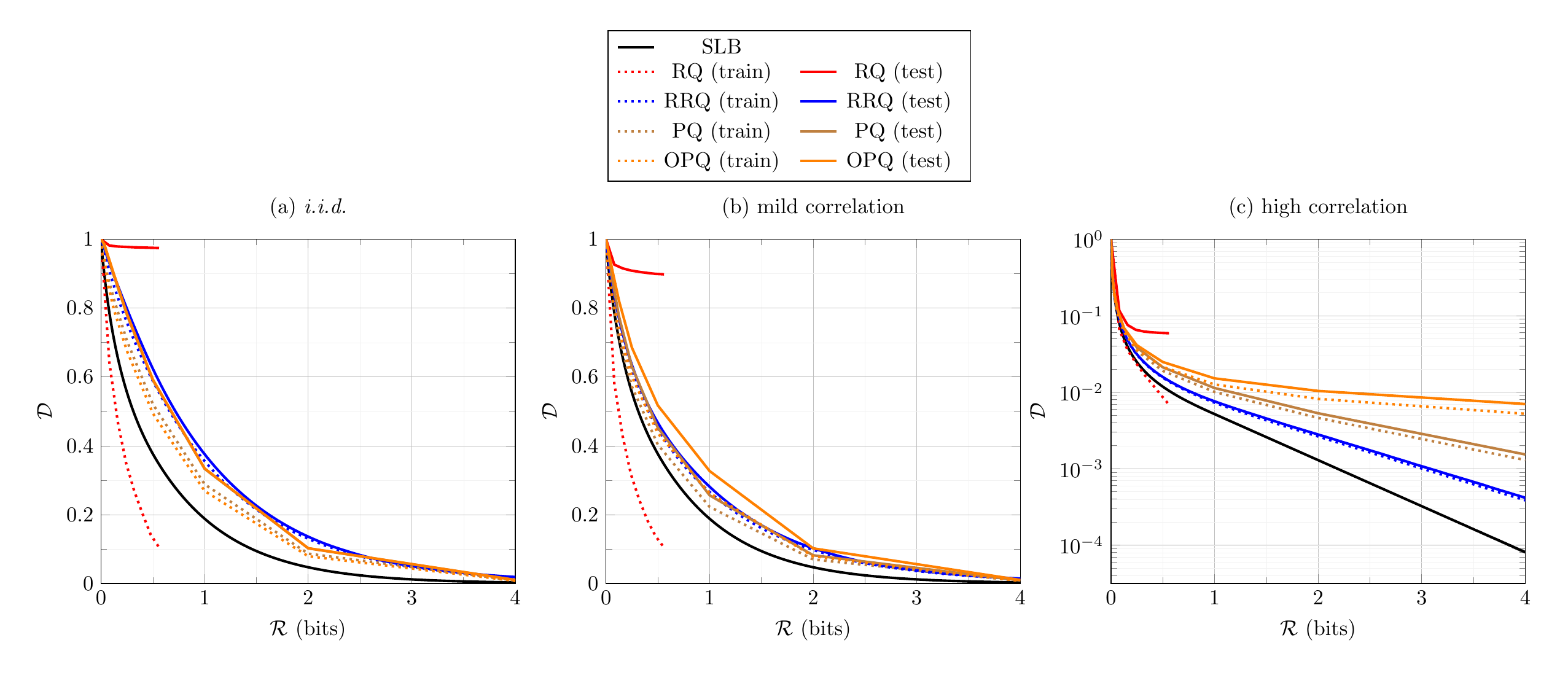}

   \end{center}
\caption{Distortion-rate curves for the RRQ (based on VR-Kmeans) and the RQ (based on K-means) for AR(1) Gaussian sources with $\rho = 0,0.5,0.99$, corresponding to (a) \textit{i.i.d.}, (b) mildly correlated and (c) highly correlated sources. Results are averaged over 5 independent experiments.}    
   \label{fig:MultiLayer_RRQ_DR}
   \end{figure}

We now experiment with the RRQ to see how it can target arbitrarily high rates. Fig. \ref{fig:MultiLayer_RRQ_DR} compares the distortion-rate curves of the RRQ with RQ, on the same data as is in the experiments of Fig. \ref{fig:SingleLayer_VRKmeans_RateLimit}.

Clearly, the RQ based on the standard K-means is quickly over-fitting.\footnote{This is a demonstration of what we quoted from \cite{RQ:480761} in section \ref{subsubsec:MultiLayer_RRQ_SOTA_RQ}, regarding the limitation of RQ to usually 2 layers.} Its regularized version, however, reduces the train-test gap, as prescribed by $\mu$.

We also experiment with PQ and OPQ codes. It is interesting to see that as the data becomes more correlated, they suffer more from the independence assumption they impose on the sub-codebooks. On the other hand, as the rate increases, i.e., as the number of partitions becomes larger, the disparity between them also increases for PQ and OPQ. This, however, is not an issue with the RRQ, as we see a noticable performance improvement w.r.t. PQ.

Later in chapter \ref{chapter:ImCompression}, we will successfully use the RRQ for image compression and apply it on whitened images, as will be clarified in section \ref{sec:ImCompression_RRQ}.
\section{Multi-Layer Sparse Ternary Codes} \label{sec:MultiLayer_MLSTC}
In the previous section, we mentioned the limitations of the single-layer structures in terms of their operational rate. Inspired by the successive refinement framework in information theory and the RQ from signal processing, we then extended them to multiple layers. In particular, we proposed the RRQ algorithm as the straightforward, residual-based development of the VR-Kmeans and solved the rate limitation issues, targeting arbitrarily high rates with reasonable distortion.

While the K-means, the VR-Kmeans and hence the RQ and the RRQ are synthesis structures, on the other hand, it is of no surprise that similar concepts are valid for analysis models as well. In fact, we have already seen in section \ref{subsubsec:SingleLayer_Analysis_Reconstruction_DSW18} that for the analysis-based STC framework, as the rate increases, its allocation to different dimensions deviates from optimality, an inevitable phenomenon due to design constraints which leads to poor performance for high rates as seen in Fig. \ref{fig:SingleLayer_STC_DR}.

How do we circumvent the issue with the analysis model and in particular the STC? It is very natural to ask wheather similar solutions to synthesis models exist also for the STC. In fact, while the RQ solution of section \ref{sec:MultiLayer_RRQ} has a synthesis model shape, the theoretical framework of additive successive refinement does not have a constraint on the encoding-decoding construction. Therefore, we utilize this straightforward idea next in section \ref{subsec:MultiLayer_MLSTC_Alg} and validate it in practice in the experiments of section \ref{subsec:MultiLayer_MLSTC_RD}.

\subsection{ML-STC algorithm} \label{subsec:MultiLayer_MLSTC_Alg}
We are not aware of any existing solution within the analysis-shaped encoding schemes that takes advantage of successive refinement ideas.  However, we now use this idea to develop the Multi-Layer STC (ML-STC) from the single-layer structure of STC developed in section \ref{subsubsec:SingleLayer_Analysis_Reconstruction_DSW18}. 

\begin{figure} 
   \begin{center} 
\includegraphics[width=1.1\textwidth]{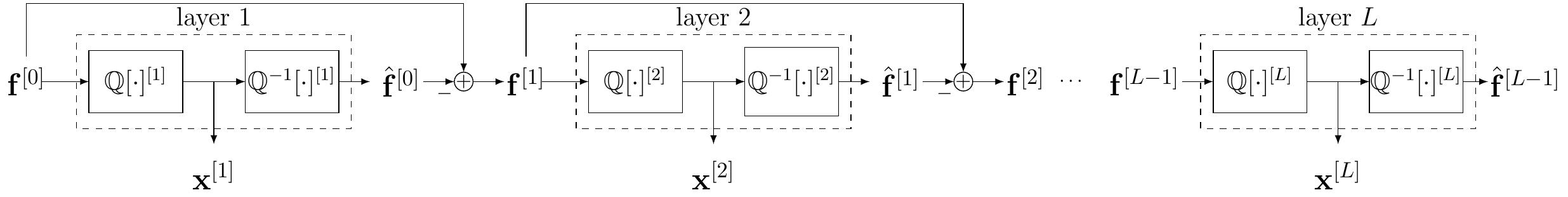} 
   \end{center}
   \caption{ML-STC architecture.}
   \label{fig:MultiLayer_MLSTC_Schematic}
   \end{figure}

This idea, identical to the residual encoding of the previous section, is simply put in the recursion rule of Eq. \ref{eq:MultiLayer_MLSTC_Recursion}, and is illustrated in Fig. \ref{fig:MultiLayer_MLSTC_Schematic}:

\begin{equation} \label{eq:MultiLayer_MLSTC_Recursion}
\begin{aligned}
\mathbf{x}^{[l]} &= \mathbb{Q}\big[\mathbf{f}^{[l-1]} \big] = \phi_{\lambda_X}^{[l]}(\mathrm{A}^{[l]} \mathbf{f}^{[l-1]}) \odot \boldsymbol{\beta}^{[l]}, \\
\hat{\mathbf{f}}^{[l-1]} &= \mathbb{Q}^{-1}\big[\mathbf{x}^{[l]} \big] = \mathrm{A}^{{[l]}^T}  \mathbf{x}^{[l]},\\
\mathbf{f}^{[l]} &= \mathbf{f}^{[l-1]} - \hat{\mathbf{f}}^{[l-1]},  
\end{aligned} 
\end{equation}
where the superscripts depict the index of the layer $l = 1, \cdots, L$. The input to the algorithm at layer $l$ is $\mathbf{f}^{[l-1]}$, which is the residual of the approximation from layer $l-1$ and is initialized as $\mathbf{f}^{[0]} = \mathbf{f}$. The rest of the procedure is the same as the single-layer case.

We saw earlier in section \ref{subsubsec:SingleLayer_Analysis_Reconstruction_RD} that, in order not to deviate from optimality, the STC should be operated only at lower rates. Therefore, we choose the ternarization threshold $\lambda$ in all the $\phi_{\lambda_X}^{[l]}(\cdot)$'s to be very high, such that they produce very sparse $\mathbf{x}^{[l]}$'s.

Algorithm \ref{alg:MultiLayer_STC_MLSTC} summarizes the ML-STC.

\begin{algorithm} [H] \caption{ML-STC} \label{alg:MultiLayer_STC_MLSTC}
\begin{algorithmic}[0]
    \INPUT  Training set $\mathrm{F}$ (whitened), $\#$ of layers $L$ and threshold $\lambda$ (assuming equal for all layers)
    \OUTPUT Projections  $\mathrm{A}^{[l]}$'s, weighting-vectors $\boldsymbol{\beta}^{[l]}$'s and codes $\mathrm{X}^{[l]}$'s, for $l = 1, \cdots, L$
\end{algorithmic}
\begin{algorithmic}[1]
\State $\mathrm{F}^{[0]} \gets \mathrm{F}$ 
\State $\hat{\mathrm{F}} \gets \mathrm{0}$  \Comment all-zero matrix of $(n \times N)$
\For{$l = 1, \cdots, L$} 
\State Estimate the covariance matrix $\mathrm{C}_F^{[l-1]}$ from input samples $\mathrm{F}^{[l-1]}$ 
\State $\mathrm{U} \mathrm{\Sigma} \mathrm{U}^T  \gets \text{EIG}(\mathrm{C}_F^{[l-1]})$  \Comment Eigen-value decomposition
\State $\mathrm{A}^{[l]} \gets \mathrm{U}^T$ 
\For{$j = 1,\cdots, n$} \Comment $\boldsymbol{\beta}^{[l]} = [\beta_1^{[l]}, \cdots, \beta_n^{[l]}]^T$.
\State $\beta_j^{[l]} \gets  \frac{\sigma_j \exp\Big(\frac{-\lambda^2}{2\sigma_j^2}\Big)}{\sqrt{2\pi} \mathcal{Q}\Big(\frac{\lambda}{\sigma_j}\Big)} $  \Comment $\mathrm{\Sigma} = \text{diag}([\sigma_1^2, \cdots, \sigma_n^2]^T)$
\EndFor
\State $\mathrm{X}^{[l]} \gets \phi_{\lambda} \big( \mathrm{A}^{[l]} \mathrm{F}^{[l-1]} \big) \odot \big( \boldsymbol{\beta}^{[l]} \mathbf{1}_N^T \big)$  \Comment Matrix form of Eq. \ref{eq:SingleLayer_STC_EncodingWeighted}
\State $\hat{\mathrm{F}}^{[l-1]} \gets \mathrm{A}^{{[l]}^T} \mathrm{X}^{[l]}$ 
\State $\hat{\mathrm{F}} \gets \hat{\mathrm{F}} + \hat{\mathrm{F}}^{[l-1]}$  \Comment Current approximation
\State $\mathrm{F}^{[l]} \gets \mathrm{F} - \hat{\mathrm{F}}$   \Comment Residual (input to the next stage's quantizer)
\EndFor
\end{algorithmic}
\end{algorithm}

Although the main building-blocks of training of this algorithm are the eigenvalue decomposition (using both eigenvectors and eigenvalues) of the covariance matrices, this algorithm does not rely solely on the second-order statistical properties of the data. This is because the PCA is applied successively and on the residuals of encoding, rather than once and on the input data. In fact, as the layers increase, the residuals, and hence the input to the next layers, i.e., $\mathrm{F}^{[l]}$'s become closer and closer to the white Gaussian noise.

We evaluate the performance of this algorithm in section \ref{subsec:MultiLayer_MLSTC_RD} and section \ref{sec:MultiLayer_STNets}, and will utilize it in practice, later in chapter \ref{chapter:Search}.

\subsection{ML-STC-Procrustean algorithm} \label{subsec:MultiLayer_MLSTC-Procrustean_Alg}
We saw earlier in section \ref{subsubsec:SingleLayer_Analysis_Reconstruction_Procrustean} that by abandoning some of the assumptions of the linear decoding of section \ref{subsubsec:SingleLayer_Analysis_Reconstruction_DSW18} and relying more on the training data instead of the priors, we can formulate the learning procedure as in Eq. \ref{eq:SingleLayer_STC_Procrustean_main} and solve it using an alternating minimization approach and by using the orthogonal Procrustes framework. This led us then to the Algorithm \ref{alg:SingleLayer_ProcrusteanSTC}, which we dubbed the \texttt{Procrustean-STC}.

Within the multi-layer case, exactly similar to the ML-STC framework, we extend the \texttt{Procrustean-STC} algorithm by training over the residuals of the previous encoding. We term this new algorithm as ML-STC-Procrustean and summarize it as in Algorithm \ref{alg:MultiLayer_STC_MLSTC_Procrustean}.
\begin{algorithm} [H] \caption{ML-STC-Procrustean} \label{alg:MultiLayer_STC_MLSTC_Procrustean}
\begin{algorithmic}[0]
    \INPUT  Training set $\mathrm{F}$ (whitened), $\#$ of layers $L$ and threshold $\lambda$ (assuming equal for all layers)
    \OUTPUT Projections  $\mathrm{A}^{[l]}$'s, weighting-vectors $\boldsymbol{\beta}^{[l]}$'s and codes $\mathrm{X}^{[l]}$'s, for $l = 1, \cdots, L$
\end{algorithmic}
\begin{algorithmic}[1]
\State $\mathrm{F}^{[0]} \gets \mathrm{F}$ 
\State $\hat{\mathrm{F}} \gets \mathrm{0}$  \Comment all-zero matrix of $(n \times N)$
\For{$l = 1, \cdots, L$} 
\State $\mathrm{A}^{[l]},  \boldsymbol{\beta}^{[l]}, \mathrm{X}^{[l]}  \gets \texttt{Procrustean-STC} \big( \mathrm{F}^{[l-1]}, \lambda \big)$ \Comment From Algorithm \ref{alg:SingleLayer_ProcrusteanSTC}. 
\State $\hat{\mathrm{F}}^{[l-1]} \gets \mathrm{A}^{{[l]}^T} \mathrm{X}^{[l]}$ 
\State $\hat{\mathrm{F}} \gets \hat{\mathrm{F}} + \hat{\mathrm{F}}^{[l-1]}$  \Comment Current approximation
\State $\mathrm{F}^{[l]} \gets \mathrm{F} - \hat{\mathrm{F}}$   \Comment Residual (input to the next stage's quantizer)
\EndFor
\end{algorithmic}
\end{algorithm}

We will evaluate this algorithm in section \ref{sec:MultiLayer_STNets}.
\subsection{Rate-distortion behavior} \label{subsec:MultiLayer_MLSTC_RD}
Let us see how the basic ML-STC of Algorithm \ref{alg:MultiLayer_STC_MLSTC} performs at higher rates. We take the same experimental setup as in Fig. \ref{fig:SingleLayer_STC_DR}, where our performance was limited to very low rates. 

\begin{figure} [H]  
   \begin{center} 
\includegraphics[width=0.99\textwidth]{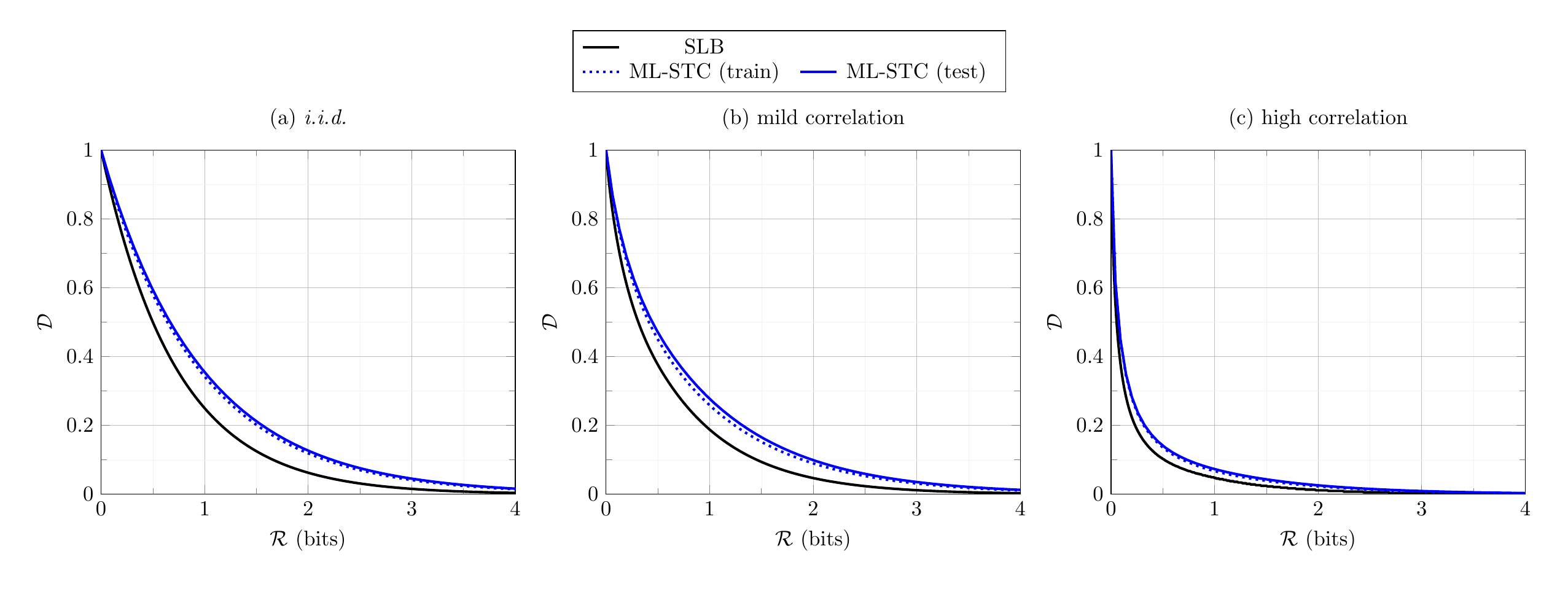} 
   \end{center}
   \caption{Distortion-rate curves of  ML-STC under AR(1) Gaussian source with varying correlation factors: (a) $\rho=0$, (b) $\rho=0.5$ and (c) $\rho=0.90$.}
   \label{fig:MultiLayer_MLSTC_DR}
   \end{figure}

This is illustrated in Fig. \ref{fig:MultiLayer_MLSTC_DR}, where we see that the ML-STC covers the entire rate-regime and provides very reasonable distortion.

We next extend our multi-layer structures to a neural network, where we use the multi-layer algorithms developed in this section, i.e., the ML-STC and the ML-STC-Procrustean to pre-train neural structures.
\section{Sparse Ternary Networks} \label{sec:MultiLayer_STNets}
Our multi-layer solutions so far, i.e., the synthesis RRQ and the analysis ML-STC were designed to minimize only the residual error from a previous stage of encoding. In other words, we were only encoding the causal residual errors and not the anti-causal residuals. In other words, at layer $l$ of encoding, we did not make any effort to make the encoding of layer $l+1$ optimal. 

A natural question to ask is whether this strategy is jointly optimal? While the framework of additive successive refinement guarantees optimality of our simple residual-based structure for Gaussian \textit{i.i.d.} data and for the asymptotic case of $n \rightarrow \infty$, in practice, for arbitrary data, even if we perform whitening, the prerequisites of these theorems do not exactly hold. Therefore, while with proper whitening we might expect not to be very far from optimality, it still makes sense to reduce the gap by considering solutions that target joint optimality.

Let us first take the synthesis case and in particular the RQ framework. There has been some efforts in the literature (e.g., \cite{RQ:212286,RQ:476315}, and a review in \cite{RQ:480761}) to try to achieve some sort of joint-optimality. They initially train the codebooks with only the causal residuals and then update each layer sequentially by fixing other codebooks. These methods, however, are essentially based on heuristics and are very hard to implement. Moreover, they show only little performance improvements. In fact, it is not straightforward to jointly train synthesis dictionaries.

On the other hand, for the analysis models, the situation seems more favorable. As we will show next, the analysis models may benefit from the back-propagation algorithm to jointly tune the learned transforms.
\subsection{Building up a neural network architecture} \label{subsec:MultiLayer_STNets_Network}
While the ML-STC is based on layer-by-layer learning of the residuals, its structure bears some resemblance to the neural network architectures. Essentially, similar to neural structures, it is based on repeatedly performing projections followed by applying non-linearities. It is, therefore, somehow expected to think of the back-propagation optimization paradigm of neural networks to apply to the ML-STC as well.

A technical issue, however, seems to hinder the application of this idea in practice: For the back-propagation technique to work, all operations, including the non-linearities should be differentiable. Otherwise, the errors from the latter layers will be zeroed-out and not propagated to the initial layers. In fact, the ternarizing operator, as well as the binarizing sign function, or in general all quantizer functions are intrinsically non-differentiable.

Take the case of the sign function. This is used to produce binary codes and is very popular in many applications like binary hashing and image compression. Recently, motivated by the success of deep learning, both of these domains have adopted neural networks in their design architectures. A central issue in these designs remains the way the non-differentiability of the sign function is handled. 

Most methods simply ignore the quantization part in the pipeline and just train the network without the non-differentiable sign function. This is later applied during the test time without being optimized. 

Some other methods, e.g., \cite{ImComp:raiko-binffnn-2015,ImComp:TodericiOHVMBCS15} approximate the binarizing non-linearity with an additive noise process which is differentiable. However, this approximation is not exact and the additive noise is a function of the quantization argument.

Some methods gradually anneal the differentiable approximation function to converge to the sign function during training iterations. For example, \cite{DeepHash:Cao2017HashNetDL} parametrizes the $\tanh(\cdot)$ nonlinearity as $y = \tanh(\beta x)$ and gradually increases the value of $\beta$ to approximate more closely the sign function, i.e., $\lim_{\beta \rightarrow \infty} \tanh(\beta x) = \text{sign}(x)$.

Now let us consider the more general case of the sign function that we used in our STC paradigm.
\subsubsection{Ternarizing operator as non-linearity} \label{subsubsec:MultiLayer_STNets_Network_Ternarization} 
The ternarizing operator $\phi_{\lambda}(\cdot)$ used at the core of STC is also non-differentiable. However, it has a favorable property w.r.t. the sign function, as we will explain next.

\begin{figure} 
   \begin{center} 
\includegraphics[width=0.99\textwidth]{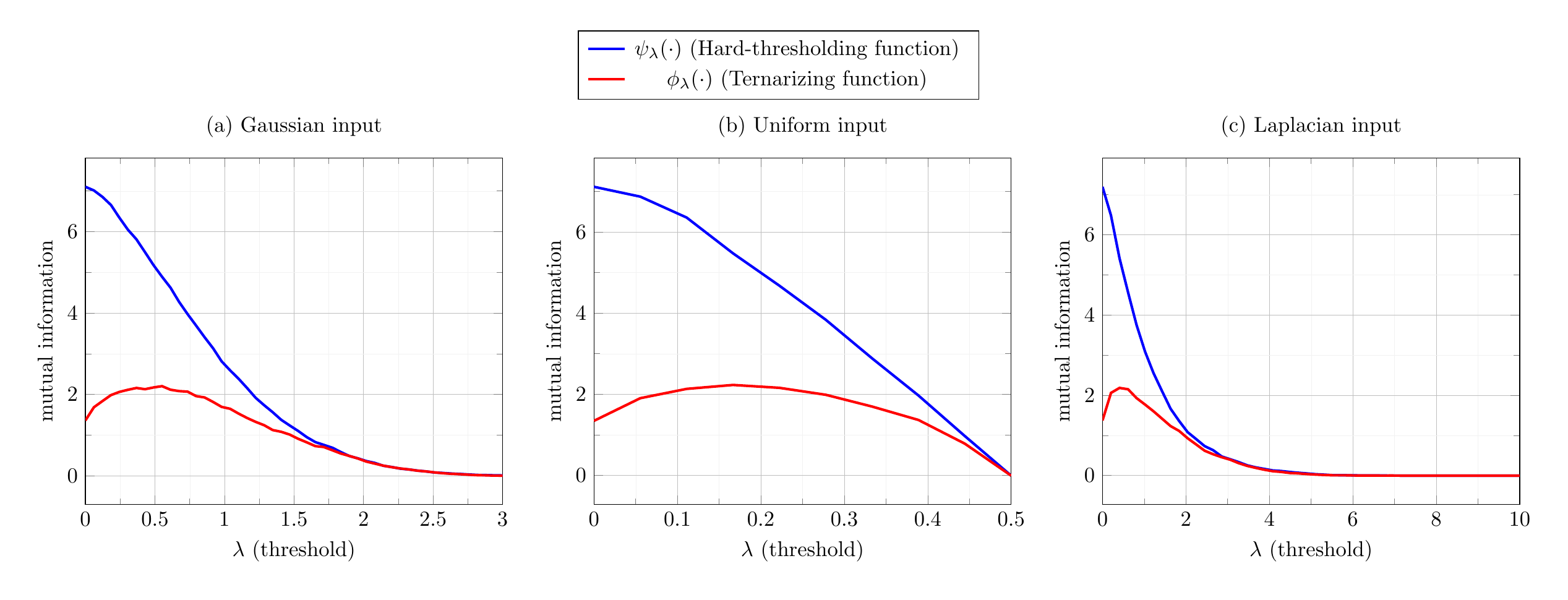} 
   \end{center}
   \caption{Mutual information between input and output of the ternarizing and hard-thresholding functions for different values of $\lambda$ and under (a) Gaussian (b) Uniform and (c) Laplacian input sources. The sign function corresponds to $\lambda=0$.}
   \label{fig:MultiLayer_MLSTC_MIEstimation}
   \end{figure}

Fig. \ref{fig:MultiLayer_MLSTC_MIEstimation} sketeches the mutual information between the input and output of the ternarizing function, i.e., $I(F;\phi_{\lambda}(F))$, as well as the hard-thresholding function, i.e., $I(F;\psi_{\lambda}(F))$ , w.r.t. different values of $\lambda$.\footnote{These values are not analytically calculated. Rather, they are estimated from the Python toolbox presented in \cite{ITEstimation:szabo14information}.}

As is seen from this figure, as $\lambda$ is increased, and hence, as the code becomes sparser, the mutual information between the input and output of these two functions coincide. This means that most of the information content will be present in the position of the non-zeros, rather than their values. 

This has a very important practical consequence for the training of neural network structures based on the ternarizing non-linearity. In fact, if operated at highly sparse regimes, instead of the \textit{non-differentiable ternarizing function}, one can use the \textit{differentiable hard-thresholding function} during the training phase and while applying the back-propagation technique, without losing any information content.

After the training phase is terminated, the hard-thresholding function is replaced by putting the ternarizing function back into the network. This, however, needs the weighting vector $\boldsymbol{\beta}$ to be adjusted. We use the same procedure as in Eq. \ref{eq:SingleLayer_STC_beta*-per-dim} to adjust $\boldsymbol{\beta}$. 

Notice that under whatever technique, this procedure is not possible for the binarizing sign function without incurring a significant approximation error. This is clear from Fig. \ref{fig:MultiLayer_MLSTC_MIEstimation}, where for all the 3 shown sources, there is a significant gap between the information contents of the differentiable and non-differentiable non-linearities.

\subsubsection{Bias terms} \label{subsubsec:MultiLayer_STNets_Network_Bias}
Our solutions so far consisted of linear projections followed by non-linearity for the encoding and linear projections for decoding. Neural networks, however, use affine transformations, i.e., projections plus a bias term followed by non-linearity. Thanks to the back-propagation, this is very efficiently done in neural networks since the elements leading to gradient calculation for the projection matrix at a certain layer are readily useful for gradient calculation of the bias term at that layer as well.

Now that we plan to use the back-propagation paradigm in our networks, we can simply extend our model by replacing the linear terms with affine terms. So our multi-layer model generalizes as in Eq. \ref{eq:MultiLayer_STNets_RecursionBias}: 

\begin{equation} \label{eq:MultiLayer_STNets_RecursionBias}
\begin{aligned}
\mathbf{x}^{[l]} &= \mathbb{Q}\big[\mathbf{f}^{[l-1]} \big] = \phi_{\lambda_X}^{[l]} \big(\mathrm{A}^{[l]} \mathbf{f}^{[l-1]} + \boldsymbol{\mu}^{[l]} \big) \odot \boldsymbol{\beta}^{[l]}, \\
\hat{\mathbf{f}}^{[l-1]} &= \mathbb{Q}^{-1}\big[\mathbf{x}^{[l]} \big] = \mathrm{A}^{{[l]}^T}  \mathbf{x}^{[l]} + \boldsymbol{\eta}^{[l]},\\
\mathbf{f}^{[l]} &= \mathbf{f}^{[l-1]} - \hat{\mathbf{f}}^{[l-1]},  
\end{aligned} 
\end{equation}
where $\boldsymbol{\mu}^{[l]}$ and $\boldsymbol{\eta}^{[l]}$ are the newly added parameters to the network. 

This means that, for any stage $l$, we will have to train 4 parameters, i.e.: $\big\{ \mathrm{A}^{[l]}, \boldsymbol{\beta}^{[l]},  \boldsymbol{\mu}^{[l]}, \boldsymbol{\eta}^{[l]}  \big\}$.

\subsubsection{STNets architecture} \label{subsubsec:MultiLayer_STNets_Network_Architecture} 

\begin{figure} [H]  
   \begin{center} 
\includegraphics[width=1.1\textwidth]{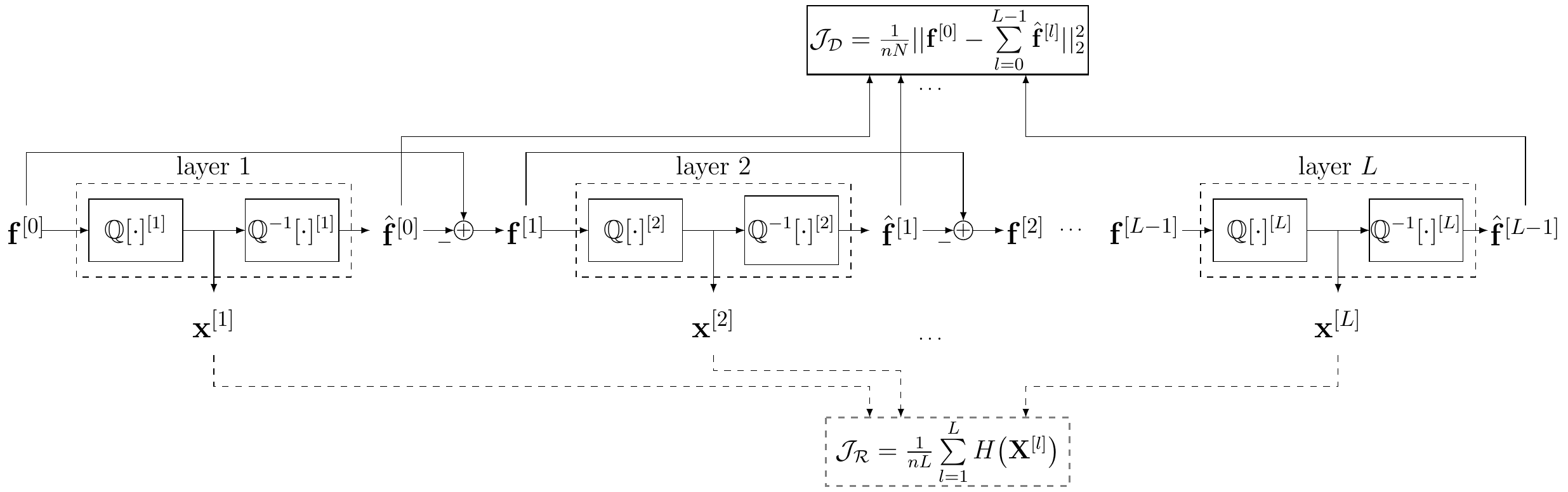} 
   \end{center}
   \caption{STNets architecture.}
   \label{fig:STNets_Schematic}
   \end{figure}

Fig. \ref{fig:STNets_Schematic} sketches the architecture of our proposed $L$-layer network based on ternary encoding.\footnote{From the cognitive sciences and the connectionist points of view, we do not claim any similarity or compliance of our proposed network with any living being intelligent structure, as all our developments are driven by purely computational reasons. Yet we adopt the term ``neural network'' for our proposed structure, since it is a construct with multiple layers of processing based on simple core units and for which the back-propagation technique is applicable.} We term this structure as the Sparse Ternary Networks (STNets).

Similar to the ML-STC structure, it consists of $L$ layers of encoding and decoding based on the STC, where each layer encodes the residual error of the decoding of the previous layer. 

As for the application of the back-propagation, the minimization of the distortion cost function $\mathcal{J}_{\mathcal{D}}$ is performed, not only from the last layer, but also from all intermediate layers.

Apart from the distortion cost, we can think of minimization of a rate cost function $\mathcal{J}_{\mathcal{R}}$ as well, based on the ternary entropy of Eq. \ref{eq:SingleLayer_STC_Rate}. Since this is also differentiable, $\mathcal{J}_{\mathcal{R}}$ can be added to a final loss along with the distortion loss as:
\begin{equation}  \label{eq:MultiLayer_STNets_GeneralLoss}
\mathcal{J} = \mathcal{J}_{\mathcal{D}} + \mu \mathcal{J}_{\mathcal{R}}, 
\end{equation} 
and back-propagated through, where $\mu$ is a regularization weight . This is summarized in Eq. \ref{eq:MultiLayer_STNets_Losses}:

\begin{equation} \label{eq:MultiLayer_STNets_Losses}
\begin{aligned}
\mathcal{J}_{\mathcal{D}} &= \frac{1}{nN}|| \mathbf{f}^{[0]} - \sum\limits_{l = 0}^{L-1} \hat{\mathbf{f}}^{[l]}  ||_2^2, \\
\mathcal{J}_{\mathcal{R}} &= \frac{1}{nL} \sum\limits_{l = 1}^{L} H \big(\mathbf{X}^{[l]} \big) \\
                          &= \frac{1}{nL} \sum\limits_{l = 1}^{L} \Big[ \sum_{j = 1}^n \Big(2\alpha_j^{[l]} \log_2(\alpha_j^{[l]}) + (1 - 2\alpha_j^{[l]}) \log_2 (1 - 2\alpha_j^{[l]}) \Big)  \Big].
\end{aligned}
\end{equation}

We have not yet implemented the rate loss in our experiments and we only minimize the distortion loss, i.e., $\mathcal{J} = \mathcal{J}_{\mathcal{D}}$. However, rate is constrained implicitly from $\phi_{\lambda}(\cdot)$ non-linearity and for a fixed value of $\lambda$. Minimizing it directly by explicitly imposing it to the final loss may encourage more variance decaying projections and hence the concentration of non-zero activity of the codes within a smaller subset of the dimensions. 

\subsection{Training strategies for Sparse Ternary Networks} \label{subsec:MultiLayer_STNets_TrainingStrategies}
Now that we have built up our neural network architecture, it is important to think of strategies to train it.

In general, we propose to have a double-stage training for the STNets, i.e.,  a pre-training stage, and a fine-tuning stage. 

In the pre-training stage, the network parameters for each layer, i.e., $\mathrm{A}^{[l]}$'s and $\boldsymbol{\beta}^{[l]}$'s are trained layer-by-layer. Since we do not have any solution for the bias terms, i.e., $\boldsymbol{\mu}^{[l]}$'s and $\boldsymbol{\eta}^{[l]}$'s, they are initialized with zero values at this stage.

In the fine-tuning stage, the network is initialized with the results of the pre-training stage. The entire parameter set for each stage is then updated using the standard back-propagation algorithm. 

\begin{figure} [H]  
   \begin{center} 
\includegraphics[width=0.99\textwidth]{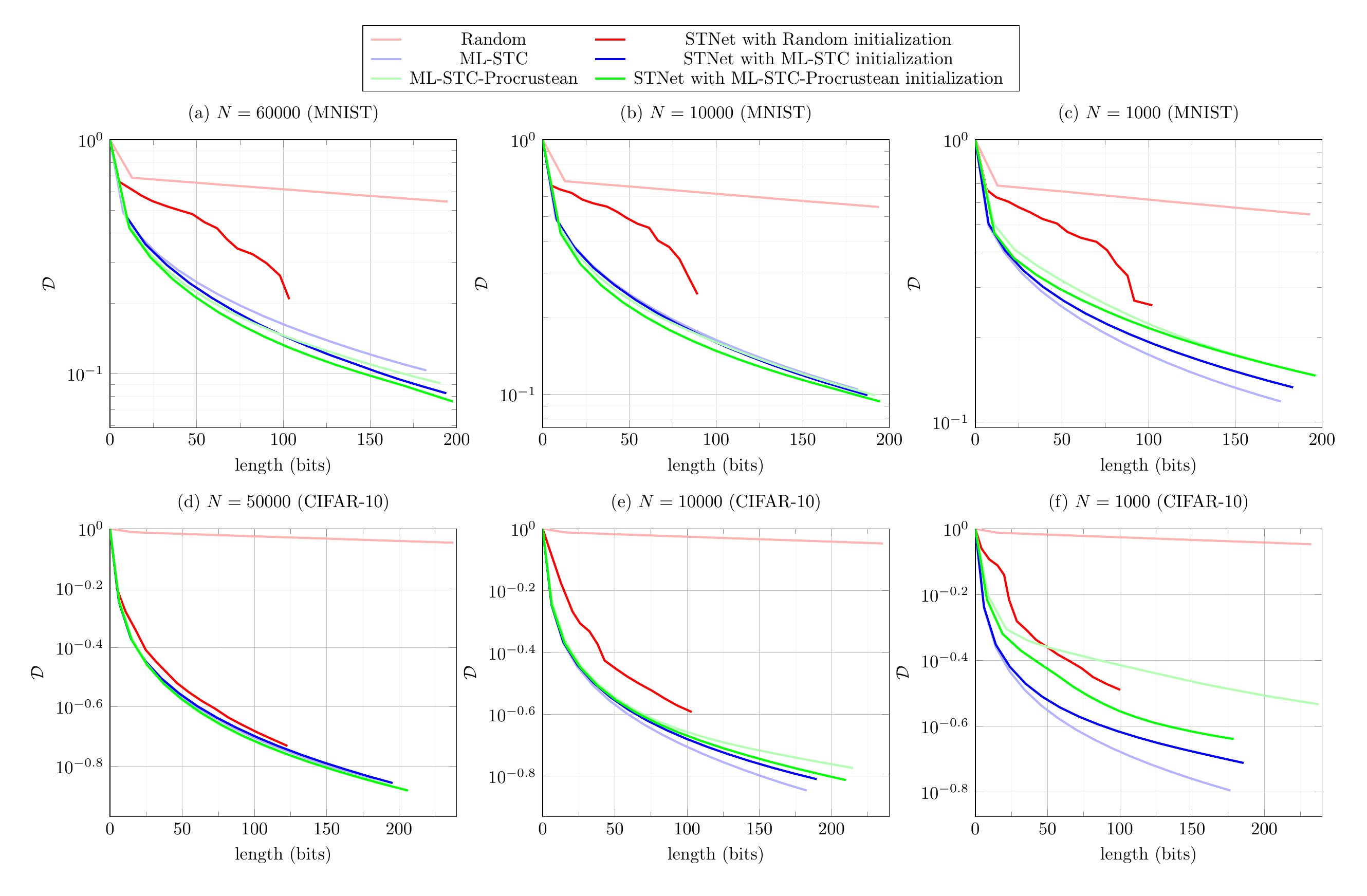} 
   \end{center}
   \caption{Distortion-rate curves of the STNets under 3 different sample regimes and for the MNIST and CIFAR-10 databases of digits on the test set. All networks have $L = 15$ layers. }
   \label{fig:MultiLayer_STNets_MNIST_DR}
   \end{figure}

Fig. \ref{fig:MultiLayer_STNets_MNIST_DR} demonstrates the results of this double-stage training under the STNets architecture and for the rate-distortion optimization of the standard MNIST and CIFAR-10 databases.

The 3 plots of this figure correspond to 3 different regimes of availability of training samples. For all experiments, 10 epochs of training with equal-sized mini-batches of the training set have been performed using the Adam \cite{Adam:kingma2014adam} optimization strategy for back-propagation with standard parameters. All networks had $L=15$ layers\footnote{Throughout this chapter we were assuming for the simplicity of representation, that the code-length is equal to the data dimension, i.e., $m=n$. However, in these experiments, we put $m=100$ for all MNIST experiments ($n=784$) and $m=500$ for all CIFAR-10 experiments ($n=3072$).} and the fine-tuning procedure was implemented in the PyTorch \cite{PyTorch:paszke2017automatic} package. 

From these experiments, the following somehow predictable, yet exciting observations can be made:

Firstly, the pre-training based on the ML-STC (Algorithm \ref{alg:MultiLayer_STC_MLSTC}) and ML-STC-Procrustean (Algorithm \ref{alg:MultiLayer_STC_MLSTC_Procrustean}) are very useful and significantly reduce the training time w.r.t. the random initializations. For this experiment, perhaps much longer training time should have been dedicated to the STNets with random initialization to catch up with pre-trained counterparts. In fact, in our other experiments, we notice that the random initialization is successful, only when the number of layers is small. A lot of engineering work has been dedicated within the deep learning communities to be able to increase the number of layers of randomly-initialized networks. Our proposal to initialize with pre-training obviates the need for such techniques.

Secondly, the higher the number of training samples available, the more the intricate models can be successful. In particular, the most powerful analysis-based model we have introduced in this thesis is the STNets pre-trained with the ML-STC-Procrustean, which outperforms others in Fig. \ref{fig:MultiLayer_STNets_MNIST_DR}-(a),(d), when the full training set is used. In this case, the second winner is the STNets pre-trained with ML-STC which outperforms the simpler ML-STC.

Conversely, the more the training samples are scarce, the more the algorithms that rely on priors rather than the observations tend to behave more successfully. So in Fig. \ref{fig:MultiLayer_STNets_MNIST_DR}-(c),(f) where only $N= 1000$ training samples are available, the basic ML-STC outperforms all other methods. Further learning with back-propagation or with the Procrustean approach over-fits to the training set. So in this case, very strong and careful regularizations like the dropout \cite{Drouput:Srivastava:2014} with a high probability of units turning off, or the weight-decay with a high constant should be used.   

\subsection{Discussions} \label{subsec:MultiLayer_STNets_Discussions}
In this chapter, we have been able to transform our layer-by-layer training paradigms of ML-STC and ML-STC-Procrustean into the STNets, which is essentially a neural network model. This has two important consequences.

First, we are able to utilize simple but powerful (analytical) solutions for smaller sub-problems within the bigger problem of training the whole model. For example, the PCA, although the solution to a very simple problem, but is in fact, optimal (under its own assumptions), implementable very efficiently, requiring the least number of training samples and has very clear and strong theoretical foundations. So within the setup of PCA, any other solution will be inferior to it. Our proposed ML-STC architecture breaks up the big problem into many such sub-problems and solves each of them efficiently (and for most parts analytically). 

This, however, was based on some assumptions. To compensate for making such restricting assumptions, we used the fine-tuning stage using the back-propagation. 

Second important consequence follows from the fact that our model is now a neural network structure that solves the whole problem jointly, rather than layer-by-layer. Provided that enough training data is available, this leads to better solutions as we saw in Fig. \ref{fig:MultiLayer_STNets_MNIST_DR}. Other than this fact, working with a neural structure results in two other important benefits.

Firstly, we may be able to benefit from a wealth of practical know-how developed within the deep learning communities. For example, we can use dropout \cite{Drouput:Srivastava:2014} and batch-normalization \cite{BNormalization:ioffe2015batch} for more successful training, while benefiting from very efficient and well-supported implementations of back-propagation like PyTorch \cite{PyTorch:paszke2017automatic} with ready-to-use options for Adam \cite{Adam:kingma2014adam} optimization.

Secondly, and left as future work, other than the distortion and rate cost functions of Eq. \ref{eq:MultiLayer_STNets_Losses}, we may be able to impose other cost functions for other tasks. For example, we may pre-train the network to optimize for rate-distortion behavior and then add a classification loss over some semantic labels.

Notice that our proposed network has several important particularities w.r.t. the neural structures from the literature.

First, the underlying non-linearity used is the ternarizing operator. Modern neural networks usually use the ReLu and its variants, or the more classical sigmoid or tanh functions. In order to produce discrete representations, to the best of our knowledge, they all use either the binarizing sign function or simply scalar quantization of the representations. While our motivations for adopting the ternarizing function and in general, ternary representation have originated from sparsifying transform learning, as explained in chapters \ref{chapter:SingleLayer} and \ref{chapter:Search}, within the back-propagation optimization paradigm, we stated the benefits of our choice of ternary w.r.t. the common binary non-linearities in section \ref{subsubsec:MultiLayer_STNets_Network_Ternarization}.

Second, the way the cost function of our STNets is formed is somewhat unusual. Typically, neural networks use the output of the last layer to form the loss function. STNets receives $L$ inputs from all the $L$ layers. This is somehow similar to the so-called ``skip-connections'' of \cite{ResNet:7780459} which has shown to be one of the breakthroughs in the development of deep learning. However, the reason behind the effectiveness of these skip-connections (that are fed to the next layers rather than to the cost function directly) is little known. For our network, this is explicitly required by the successive refinement idea for which clear understanding is known.

\section{Conclusions} \label{sec:MultiLayer_Conclusions}
Following the concluding argument of chapter \ref{chapter:SingleLayer}, which states that single-layer encoding structures are either very difficult to handle (e.g., the NP-hard sparse coding problem mentioned in chapter \ref{chapter:ModelingLit}), or are too limited in capacity, we extended them to multiple layers and hence designed high capacity models with straightforward encoding-decoding recipes. This was based on the successive refinement framework from information theory for which performance guarantees and theoretical understandings exist and leads to the favorable residual based additive structures.

We pursued this idea under the two family of algorithms developed in chapter \ref{chapter:SingleLayer}, i.e., the synthesis model and the analysis model. Our treatment of synthesis model extended the VR-Kmeans algorithm that we have developed in section \ref{subsec:SingleLayer_Synthesis_VRKmeans} to the multi-layer RRQ framework of section \ref{subsec:MultiLayer_RRQ_RRQ}, which solves the issue of over-fitting in RQ and enables it to train arbitrary number of layers of high-dimensional data, thanks to the regularization it provides, and achieves excellent rate-distortion performance. We will use RRQ in the compression of facial images in chapter \ref{chapter:ImCompression}.

Under the analysis model, we then applied the same idea of residual-based training to the STC framework, which we have introduced in section \ref{sec:SingleLayer_Analysis}, and developed the ML-STC framework in section \ref{sec:MultiLayer_MLSTC}. We saw that this extension lifts the rate-limitation issue of its single-layer predecessor and maintains a strong rate-distortion trade-off within the entire rate regime. We then introduced an extension of this algorithm, i.e., the  ML-STC-Procrustean in section \ref{subsec:MultiLayer_MLSTC-Procrustean_Alg}, which has a more data-oriented and less assumption-based learning paradigm. 

We then pursue another evolution of our algorithms, that of the neural networks. The ML-STC consists of repeated projections plus non-linearities, resembling standard neural network structures. We showed how to fine-tune its parameters with the standard recipes of neural network training.

This, however, faces an obstacle. Like any quantizing function, e.g., the sign function to generate binary codes, the ternarizing operator is non-differentiable. This being a severe limitation to compression networks, which hinders them from fully benefiting the back-propagation, is not a limitation for our ternarizing function. We studied the information preservation of this function and concluded that, in the very sparse regimes, it has the same behavior as its differentiable counterpart, i.e., the hard-thresholding function, since most of the information will be concentrated in the position of non-zeros, rather than their values. So we could replace the ternarizing operator with the hard-thresholding function during training, and put it back in place during test time, followed by re-weighting. 

Having resolved this obstacle, we considered several training strategies. We noticed that our neural structure, i.e., the STNets as we termed it, can be pre-trained with the ML-STC or the ML-STC-Procrustean, and fine-tuned with back-propagation. This way, simplifying assumptions, e.g., the perfect independence of dimensions after whitening, or not considering non-causal errors during layer-by-layer training will be lifted, as we use more data. From the other hand, instead of starting off with random values, as is common in standard neural networks, we reduce the training time by starting from analytically found solutions.

This leaves us with a large set of choices regarding the training strategy to pick up under different training sample regimes. When data is abundant, initialize the STNets with ML-STC-Procrustean and fine-tune with back-propagation. In case there is limitation, initialization with ML-STC would be preferred. If the limitation is severe, it would make more sense to suffice with the ML-STC and avoid back-propagation (or otherwise use stronger regularization). In the extreme case when data is too limited to even estimate the covariance matrix, one would initialize with random values, only taking into account the variance.

While we will show further applications for these algorithms in the next part of the thesis, a lot of future directions can be envisioned. An immediate direction would be to pre-train the network for rate-distortion optimality, as we did in this chapter and further tune the parameters with label-aware losses. One would expect to reduce the training time, thanks to this pre-training.


\part{Applications}
\chapter{Large-scale similarity search} \label{chapter:Search}
A fundamental task in a large variety of data-oriented domains is to search for similarities within some database and the queries provided by the users. Many applications are essentially built upon this idea. Content-based information retrieval systems, multimedia-based search engines, content identification systems, biometric-based authentication systems and many computer vision tasks can essentially be conceptualized as similarity search paradigms from semantically meaningful vectorial features.

The basic idea behind these applications is that semantic meanings like people's identities or contents in images can somehow be reflected into vectorial spaces, such that ``neighborhood'' in those spaces would imply semantic ``similarity''. So that if two feature vectors are close to each other under some sense, e.g., the Euclidean norm, their underlying content, e.g., topics for text documents, are also similar.

Being both practically and theoretically important, several different communities have studied different variations of this problem for their applications, including computer vision, machine learning, information theory, information forensics and theoretical computer science.

One example of such use case is in reverse image search engines\footnote{The term image search engine usually refers to a retrieval system, where the query is based on key-words, while ``reverse image search'' is when the query itself is an image.}, where large collections of images are stored in a database and users wish to find similar instances to their own images, e.g., in order to locate the content of their images, or finding higher resolutions of their query images.  

Other examples appear in the near-duplicate retrieval of texts, retrieval of genomic data, or in multi-class classification, where the instances per class are too few or disproportionate, the number of classes is very large and discriminative machine learning models are too expensive to run, leaving with the only option to do nearest neighbor-based classification.

While designing successful feature learning frameworks that can reflect semantic similarity into the vectorial neighborhood is already a very challenging task, how to actually perform the search on those feature vectors and the issues regarding this procedure is a central and challenging task on itself. This is the problem that we try to address in this chapter.

To get an idea about such challenges, imagine a reverse image search engine with one million images in its database. The image descriptor features are usually dense and high-dimensional vectors, around hundreds or thousands. The na\"{i}ve search strategy that involves exhaustively scanning the database requires one-million inner-products between vectors in say $\Re^{1000}$. Considering the fact that this search engine might receive thousands of such queries per second and should provide the search results within a fraction of a second, we can easily conclude that the exhaustive scan strategy is impossible to implement in practice for such systems. Therefore, the computational complexity is a central issue for large-scale search, particularly when part of the process should be performed on mobile platforms for some applications.

From the other hand, the vectorial features of the database may not fit into main memory and can fit only into disk storage. As a general rule, however, the larger the storage size of a memory device is, the slower are its connection to the processing units. This means that the similarity search computations should be performed only on compact representations of the feature vectors and I/O requests to disk storage should be avoided, as much as possible. Therefore, memory and representation compactness is another very central issue for large-scale systems.

In order to address this double challenge of complexity and memory, it makes much sense for many applications to think of approximate solutions. So instead of finding the exact closest item from the database, we search for a close enough item. This is justified since we gain computational speed-ups and use more compact representations. This leaves us with an important triple ``\textit{memory-complexity-performance}'' trade-off in design. In fact, trading performance might very well be acceptable for many applications. The underlying feature vectors are usually not perfect after all. So it does not make much sense to use-up resources in order to provide exact search results, which would be based on imperfect features.

To obtain better memory-complexity-performance trade-offs is a topic of much research in the domains like computer vision, machine learning, and information theory, where this concept appears. This chapter tries to conceptualize some aspects of this trade-off, as well as providing practical solutions for it. While some of our arguments have flavors from information theory, our practical examples are mostly computer vision applications, where we use image feature descriptors for search.

We formally define the similarity search problem in section \ref{sec:Search_PFormulation} and conceptualize the exact and approximative solutions for it. In section \ref{sec:Search_SOTA}, we provide a concise overview to the literature of similarity search, distinguishing between two general families of solutions. We then outline our proposed solution and sketch its general differences with these two families. In section \ref{sec:Search_IT}, we take an information-theoretic look to this problem and introduce the concept of ``coding gain'' for similarity search. This is where we show that the ternary encoding of the STC framework, which we introduced earlier in chapter \ref{chapter:SingleLayer}, has superior characteristics compared to the conventional binary hashing. Based on the ML-STC framework which we have introduced earlier in section \ref{sec:MultiLayer_MLSTC}, we then put all elements together and describe the complete architecture of our proposed similarity search strategy in section \ref{sec:Search_MLSTC}. We next perform practical experiments to compare our ML-STC with state-of-the-art solutions from the literature. Finally, we conclude this chapter in section \ref{sec:Search_Conclusions}.  
\section{Problem formulation} \label{sec:Search_PFormulation}
Consider a database $\mathrm{F} = [\mathbf{f}(1), \cdots, \mathbf{f}(N)]$ that consists of $N$ feature vectors $\mathbf{f}(i)$'s $\in \Re^n$, which are assumed to convey some semantic meaning within their neighborhood structure. Common candidates for these feature vectors for computer vision applications are conventional hand-crafted features like GIST \cite{GIST:Oliva2001}, aggregated features like triangulation embedding \cite{TEmbedding:6909819} or neural network-based representations like the neural codes \cite{NeuralCodes}. 

For a given query $\mathbf{q}$, the original task of Nearest Neighbor (NN) search requires finding a list of similar items to $\mathbf{q}$ within $\mathrm{F}$, i.e., 

\begin{equation} \label{eq:Search_PFormulation_ExactList}
\mathcal{L}^*(\mathbf{q}) = \{ 1 \leqslant i \leqslant N| d(\mathbf{f}(i) , \mathbf{q} ) \leqslant  \epsilon  \},
\end{equation}
where $\epsilon$ is a small constant, and the similarity is characterized by a distance measure $d(\cdot,\cdot): \Re^n \times \Re^n \rightarrow \Re^+$, as $d(\mathbf{q},\mathbf{f}(i))$. This is usually assumed to be the (squared) Euclidean distance $d_{\mathcal{E}}(\mathbf{q},\mathbf{f}(i)) = \frac{1}{n}||\mathbf{q} - \mathbf{f}(i)||_2^2$, that we have been using throughout the thesis. Note that other than Eq. \ref{eq:Search_PFormulation_ExactList}, the desired $\mathcal{L}^*(\mathbf{q}) $ may be formalized slightly differently, e.g., by sorting the items in the database w.r.t. their similarities to $\mathbf{q}$, and picking up the $T$ most similar ones.

The problem arises when either of $N$, the size or $n$, the dimensionality are high. In practice and for modern applications, it turns out that this is usually the case, both for $n$, which is typically around several hundred or thousands, and also for $N$, which can be as high as a couple of millions or billions. In fact the computational complexity, as well as the storage cost for such an operation is in $\mathcal{O}(nN)$ of floating-points. This means that finding the exact $\mathcal{L}(\mathbf{q})$ of Eq. \ref{eq:Search_PFormulation_ExactList} should be compromised for storage cost and computational complexity, but at the price of an approximative, inexact list $\hat{\mathcal{L}}(\mathbf{q})$. 

Addressing this problem is referred to as the Approximate Nearest Neighbor (ANN) search.
\subsection{Approximate Nearest Neighbor (ANN) search}   \label{subsec:Search_PFormulation_ANN}

We conceptualize the general idea of ANN search using the concept of encoder-decoder pair. Consider an encoder $\mathbb{Q}[\cdot] \colon \Re^n \to \mathcal{X}^m$ that provides codes in $\mathrm{X} = [\mathbf{x}_1, \cdots, \mathbf{x}_N]$, from the database items, i.e., $\mathbf{x}_i = \mathbb{Q}[\mathbf{f}_i]$.

When the query is presented, the approximative list $\hat{\mathcal{L}}(\mathbf{q})$ is computed, based only on $\mathrm{X}$ (and not $\mathrm{F}$). This process can be interpreted as a sort of decoding\footnote{Note that we have been using the concept of encoder-decoder throughout the thesis and within the context of rate-distortion theory. However, in that context, by decoder we meant an operation that reconstructs the input to the encoder, and we used $\mathbb{Q}^{-1}[\cdot]$ to denote this. Here, by decoder we mean an operation that returns a list of similar items and we denote it as $\mathbb{D}[\cdot]$.} based on some decoder $\mathbb{D}[\cdot, \cdot]$. However, depending on the encoding and the search method used, the construction of the decoder can be different. We will elaborate more on this in section \ref{sec:Search_SOTA}.

\begin{figure}  [!h]
   \begin{center} 
   \includegraphics[width=0.6\textwidth]{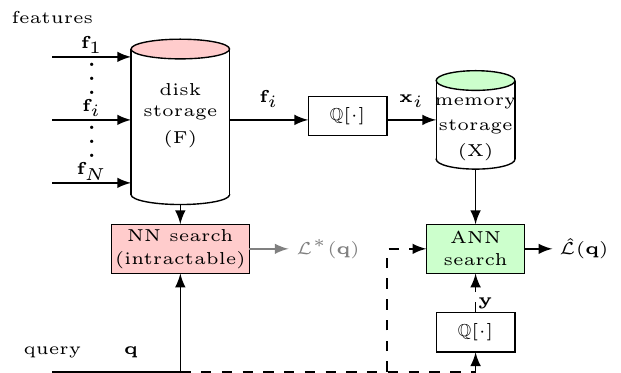}
   \end{center}

\caption{A generic pipeline for ANN search for similarity. The red-shaded areas are related to the original NN search problem and are intractable. ANN search, instead, provides an approximative $\hat{\mathcal{L}}(\mathbf{q})$ but with tractable encoder-decoder pair (green-shaded areas). Depending on the design, either the query or its encoded version is used for ANN search.}    
\label{fig:Search_PFormulation_ANNGeneric}
\end{figure}

A generic pipeline for ANN search problem is depicted in Fig. \ref{fig:Search_PFormulation_ANNGeneric}. 

A large amount of research, as well as the rest of this chapter,  are dedicated essentially to the design of good encoder-decoder pairs. How to evaluate the goodness of this procedure is described next.
\subsection{Evaluation protocol}   \label{subsec:Search_PFormulation_EvalProt}
Depending on the application, the quality of $\hat{\mathcal{L}}(\mathbf{q})$ returned by an ANN algorithm is evaluated mostly based on two measures: the mean average precision (\textit{mAP}), and the \textit{R-Recall@T}.

\textbf{mAP@T} is the mean of Average Precision over all queries, where the Average Precision is the area under the precision-recall curve. This is evaluated for a list size of T and w.r.t. the ground-truth and the returned list.

\textbf{R-Recall@T} indicates the frequency ratio of the presence of top R correct items retrieved by the algorithm in a list of size T, where the correct items are indicated by the ground-truth. Again, this is averaged for all queries presented.

A third measure is used mostly in biometric systems which is referred to as the probability of correct identification ($P_{id}$). In identification systems, usually the size of the returned list is restricted to $|\hat{\mathcal{L}}(\mathbf{q})| = 1$. So $P_{id}$ measures whether the one and the only related item from the database is in the result of the search. This, however, is essentially equivalent to \textit{1-Recall@-1}. 

These performance measures are calculated for varying memory or complexity budgets. The memory usage can be measured by the number of bits required to store a code vector $\mathbf{x}_i$ in the database. Another way to measure memory consumption is by the entropy of representation which measures, on the average, the number of bits to store a database item, if a perfect source-coder is used.

In practical settings, the computational complexity is measured simply as the run-time of finding $\hat{\mathcal{L}}(\mathbf{q})$, after $\mathbf{q}$ has been introduced. This, however, is both system- and implementation-dependent. So the algorithm-theoretic measures should be used in theoretical studies. One such measure is the complexity ratio, which is defined as the ratio of the big-O complexity of performing the search for one query over the complexity of performing the search exhaustively in the original domain, i.e.,  $\mathcal{O}(Nn)$. Since some of the computation is done in floating-point and some in fixed-point operations and they have different constants, for a fair comparison of the two algorithms, it is better to separate the complexity ratios of fixed- and floating-point operations.

The VQ-based family of methods, as we will see in section \ref{subsec:Search_SOTA_VQ}, is essentially based on reconstruction distortion. These algorithms, therefore, report distortion level for different bit-rates as well. In this thesis, in order for the distortion to be interpretable across all databases, we normalize the distortion by the database's norm as $\mathcal{D} = \frac{ ||\mathrm{F} - \hat{\mathrm{F}} ||_{\mathcal{F}^2}}{||\mathrm{F}||_{\mathcal{F}}^2}$. So when the description length is zero, the normalized distortion is always one.

\section{Literature review} \label{sec:Search_SOTA}
The literature of ANN search is cross-domain and very extensive. In fact, depending on the database size, dimensionality and available memory and computational resources, different flavors of this problem may take very different forms. As we pointed out also in chapter 1, we limit our attention to scenarios where both memory and complexity are of concern. In particular, we emphasize that we only focus on methods where both memory and computational complexities are \textbf{at most linear in $N$}, the database size, and \textbf{at most polynomial in $n$}, the dimensionality.

This excludes, e.g., graph-based methods, which are based on constructing similarity matrix of database items which is in $\mathcal{O}(N^2)$, or space partitioning and indexing structures like k-d trees whose memory usage grows exponentially with dimension. These two restrictions, in fact are necessary when dealing with million- or billion-scale databases of dimensionality around several hundred.

Within this regime, the literature of ANN search can roughly be categorized into two families: the binary hashing methods, and the VQ-based methods.

\subsection{Binary-hashing methods}   \label{subsec:Search_SOTA_Hashing}
Binary hashing methods aim at benefitting from the straightforward storage and processing facilities of binary data type by providing short binary codes both for the database items and the query. 

Both for the database items and the query, this is done simply by performing a projection step, followed by a binarization as:
\begin{equation}  \label{eq:Search_SOTA_Binarization}
\begin{aligned}
\mathbf{x}  &= \mathbb{Q}[\mathbf{f}] = \text{sign}(\mathrm{A}\mathbf{f}), \\
\mathbf{y}  &= \mathbb{Q}[\mathbf{q}] = \text{sign}(\mathrm{A}\mathbf{q}), 
\end{aligned}
\end{equation}
where $\mathrm{A}$ is an $(m \times n)$ projection matrix, usually with $m < n$, and the sign function is applied element-wise and produces the binary alphabets $\mathcal{X} = \{ +1, -1 \}$ and $\mathcal{Y} = \{ +1, -1 \}$.

Once the binary codes of the database are stored in memory, the binary hashing decoder $\mathbb{D}_{\text{BH}}[\cdot,\cdot] \colon \mathcal{X}^m \times \mathcal{X}^m \to \{ i | 1 \leqslant i \leqslant N \}$ exhaustively scans the database codes for items similar to the query code and provides the list as: 

\begin{equation}   \label{eq:Search_SOTA_HashingList}
\hat{\mathcal{L}}(\mathbf{q}) = \mathbb{D}_{\text{BH}}[\mathbf{y}, \mathrm{X}] = \{ 1 \leqslant i \leqslant N| d_H( \mathbf{x}_i , \mathbf{y} ) \leqslant \epsilon \}.
\end{equation}

The similarity is calculated using the Hamming distance\footnote{Assuming the query is a noisy version of one of the database items, it can easily be shown that the Hamming distance is the maximum-likelihood optimal decoder for binary data.}, which is simply equivalent to the \textsf{XOR} operation on binary strings, as $d_H(\mathbf{x},\mathbf{y}) = \frac{1}{m} \sum_{m'=1}^m \mathbf{x}_{m'} \oplus \mathbf{y}_{m'}$.  

The search procedure based on Hamming distance computation is very efficient in modern CPUs, thanks to the \textsf{POPCNT} instructions. However, notice that the search is still exhaustive.

The main challenge in the design of binary hashing is to preserve the neighborhood structure of the data in $\Re^n$ to within the space of binary codes. Inspired by distance preservation guarantees within random projections, e.g., the Johnson-Lindenstrauss lemma \cite{johnson1984extensions}, many methods have used random matrices for projections, e.g., the famous work of Sim-hash \cite{charikar2002similarity}. This has provided the possibility to provide probabilistic performance guarantees, e.g., \cite{LSH:4031381,andoni2009nearest,Dasgupta2011:LSH,Panigrahy:LSH}. More recently, however, within the same community, e.g., in \cite{andoni2014beyond}, it has been concluded that data-dependent hashing can be superior to random projections.

Beyond the random design, a lot of methods try to adapt the projection matrix $\mathrm{A}$ to the data and hence increase the coding efficiency. This has received an enormous amount of attention with solutions ranging from spectral methods, e.g., in \cite{SpectralHashing:NIPS2008} that tries to preserve neighborhood information by compressing the Laplacian of the similarity graph of the data, semi-supervised approaches like \cite{5539994} that use class labels to increase the search efficiency or the more recent deep learning-based methods like \cite{7301269,7298862} that learn multiple layers of projection to learn a compact binary code.

A very prominent effort to learn projection matrices is the ITQ \cite{ITQ:6296665} that tries to minimize the distortion error of the projected data with its binarized version. The idea is to use PCA to reduce the dimension and then iteratively use the orthogonal Procrustes problem to find the best rotation whose binarization incurs the least distortion. This has turned out to be very successful. In fact, recent methods based on deep learning, e.g., \cite{7298862}, hardly show a marginal advantage over ITQ and only at limited scenarios\footnote{Look, e.g., at fig. 4 or fig. 5 (a-c) of \cite{7298862} where only minimal performance gain over ITQ was reported and only for very low rates.}. This work is extended in Sparse Projections of \cite{7298954}, where they generalize ITQ for higher rates. 

While this is a very broad line of research, useful recourses reviewing the trend of such methods can be found in \cite{wang2014hashing,wang2016learning}.

\begin{figure}  [!h]
\begin{center} 
\subcaptionbox{Binary Hashing family\label{subfig:Search_SOTA_Hashing}} {\includegraphics[width=0.445\textwidth]{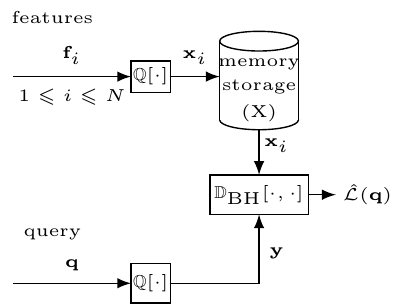}}
\subcaptionbox{Vector-Quantization family\label{subfig:Search_SOTA_VQ}} {\includegraphics[width=0.445\textwidth]{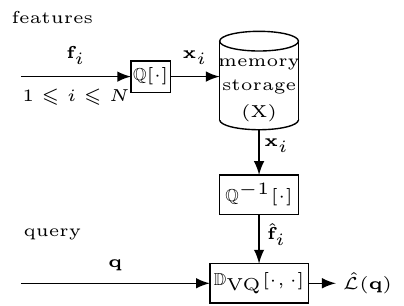}}
\end{center}
\vspace{-0.5cm}   
\caption[example]{ANN search pipelines for two families of methods.(a) Binary hashing methods only consider encoded query. (b) VQ-based methods perform the search only in the reconstruction domain.}
   \label{fig:Search_SOTA} 
   \end{figure}

Schematic diagram of hashing-based search is sketched in Fig. \ref{subfig:Search_SOTA_Hashing}. Note that, however the learning algorithm may be strong, the family of binary hashing methods suffers from a fundamental shortcoming. Once the query is encoded, i.e., $\mathbf{y} = \mathbb{Q}[\mathbf{q}]$, the original $\mathbf{q}$ is neglected in the decoding process. Although the original $\mathbf{q}$ is at hand, binary hashing fails to use it. Since the original feature vectors $\mathbf{f}_i$'s are not available anymore after encoding, this can only be achieved if $\mathbf{q}$ could somehow be matched with reconstructions of database codes, i.e., $\hat{\mathbf{f}}(i) = \mathbb{Q}^{-1}[\mathbf{x}(i)]$'s. Binary codes, however, as we have shown earlier in Fig. \ref{fig:SingleLayer_STC_DR}, suffer from poor rate-distortion performance. We will elaborate on this point later in section \ref{sec:Search_MLSTC} and during the experiments of section \ref{sec:Search_Experiments}.

\subsection{VQ-based methods}   \label{subsec:Search_SOTA_VQ}
The Vector Quantization family, unlike binary hashing, considers the reconstruction of the database codes for approximative search. This is sketched in Fig. \ref{subfig:Search_SOTA_VQ}.

Consider the setup we had at section \ref{sec:SingleLayer_General}, where the reconstruction decoder approximates $\mathbf{f}$ from $\mathbf{x}$ as $\hat{\mathbf{f}} = \mathbb{Q}^{-1} [\mathbf{x}]$. The VQ search decoder $\mathbb{D}_{\text{VQ}}[\cdot,\cdot] \colon \Re^n \times \Re^n \to \{ i | 1 \leqslant i \leqslant N \}$ is then a function of the reconstructions, i.e.:

\begin{equation}  \label{eq:Search_SOTA_VQList}
\hat{\mathcal{L}}(\mathbf{q}) = \mathbb{D}_{\text{VQ}}\big[ \mathbf{q}, \mathbb{Q}^{-1}[\mathrm{X}] \big] = \mathbb{D}_{\text{VQ}}\big[ \mathbf{q}, \hat{\mathrm{F}}\big] = \{ 1 \leqslant i \leqslant N | d_{\mathcal{E}}(\mathbf{q} , \hat{\mathbf{f}}_i) \leqslant \epsilon  \}.
\end{equation} 

Therefore, unlike the hashing methods, the search decoding is done entirely in the space of vectors, rather than in the space of codes.  

How is the search quality related to reconstruction quality? In fact, the fundamental idea behind all VQ-based ANN solutions is reflected in Eq. \ref{eq:Search_SOTA_VQIdea}:

\begin{equation}  \label{eq:Search_SOTA_VQIdea}
\begin{aligned}
||\mathbf{F} - \mathbf{Q}||_2^2        &\leqslant  || \mathbf{F} - \hat{\mathbf{F}}||_2^2  +  ||\hat{\mathbf{F}} - \mathbf{Q}||_2^2, \\
\frac{1}{n}\mathbb{E}\big[||\mathbf{F} - \mathbf{Q}||_2^2 \big] - \frac{1}{n}\mathbb{E}\big[||\hat{\mathbf{F}} - \mathbf{Q}||_2^2 \big]  &\leqslant  \frac{1}{n}\mathbb{E}\big[||\mathbf{F} - \hat{\mathbf{F}}||_2^2 \big],\\
d_{\mathcal{E}}(\mathbf{F},\mathbf{Q})  - d_{\mathcal{E}}(\hat{\mathbf{F}},\mathbf{Q})  &\leqslant d_{\mathcal{E}}(\mathbf{F},\hat{\mathbf{F}}) = \mathcal{D},
\end{aligned}
\end{equation}
where the first inequality is the triangle inequality between $\mathbf{q}$, $\mathbf{f}$ and $\hat{\mathbf{f}}$.

This means that, on the average, the approximation error of replacing the true distance between the query and a database item, i.e.,  $d_{\mathcal{E}}(\mathbf{f}_i,\mathbf{q})$, which is what the original NN search of Eq. \ref{eq:Search_PFormulation_ExactList} is based on with the distance that the VQ uses to provide $\hat{\mathcal{L}}(\mathbf{q})$, i.e.,   $d_{\mathcal{E}}(\hat{\mathbf{f}}_i,\mathbf{q})$, is upper bounded  by the VQ's average distortion $\mathcal{D}$. In other words, for a fixed rate-budget, improving the search quality is tantamount to reducing the distortion of VQ.

The above reasoning was first pointed out in \cite{PQ:5432202,PQ:5946540}, where the authors use the technique of Product Quantization (PQ) \cite{gersho1991vector} to realize this idea. Ever since, a lot of attention has been paid to the VQ-based approach, providing solutions that suggest more advanced encodings.\footnote{Remember that within the context of rate-distortion and independent from the similarity search problem, we have already reviewed VQ-based encodings including PQ earlier in section \ref{subsec:MultiLayer_RRQ_SOTA}.} Examples of such methods are the OPQ (\cite{OPQ:6619223,OPQ:Nowrouzi6619232}), the AQ (\cite{AQ:babenko2014additive}), the LOPQ (\cite{LOPQ:6909695}), the CQ (\cite{CQ:zhang2014composite,CQ:8357922}), the SCQ (\cite{SCQ:7299085}) and the RQ (\cite{RQ:chen2010approximate}).

Another important contribution of \cite{PQ:5432202} was to further provide a practical solution to perform the search in the reconstruction domain using Look-Up-Tables (LUTs) and within the PQ construction. Once the query is received, its distances to the database's sub-codebooks are computed and stored in LUTs, based on which the query's distance to all database items is virtually (and approximatively) constructed and the final list is returned.

This, however, has two issues: First, the search is still exhaustive within the items of the database. The search is then limited to million-scale data. When targeting billions of items, other measures should be taken. This is done usually using the inverted-file structures like the IVFADC in \cite{PQ:5432202}, or the IMI of \cite{IMI:6915715}. The idea is to perform a coarse quantization and apply the PQ on the residuals of this quantization.\footnote{The IVFADC is equivalent to performing RQ (with $L = 2$) + PQ, and the IMI is equivalent to PQ (with $p=2$) + RQ (with $L = 2$) + PQ.} The query then finds its nearest neighbors only on more likely partitions and hence the search becomes non-exhaustive. 

The idea of database partitioning based on VQ, however, is not noise-robust. In fact, when an item is clustered into some partition, depending on the spatial formation of the Voronoi-cells, a small amount of noise may be enough to cluster the noisy version into another partition, potentially even a partition with a very far centroid. This means that, in order to guarantee not to miss the true centroid, a large portion of adjacent centroids should also be checked, which is particularly critical in high dimensions. This makes the search non-efficient and these methods have to become highly engineered and implementation-oriented. Moreover, the inverted-file structures are like pre-processing to PQ, and not a native part of it, which makes the overall pipeline non-straightforward. 

The second shortcoming of search based on VQ is the fact that it does not directly benefit from fast search in a discretized encoded domain and the entire search happens in the floating-point real domain. The distance calculations of VQ (mostly due to the LUT part) is more than around 5 times slower than binary hashing. 

\subsection{Binary hashing vs. VQ}   \label{subsec:Search_SOTA_HashingVsPQ}
Here we summarize several important highlights about the hashing-based and VQ-based methods.

\textbf{Binary hashing family:}
\begin{itemize}
\item Very efficient decoding based on Hamming-distance, thanks to the \textsf{POCNT} instructions.
\item Poor rate-distortion performance, intrinsic to binary encoding.
\item Due to poor rate-distortion performance, not able to benefit from the query $\mathbf{q}$, and only using the encoded $\mathbf{y} = \mathbb{Q}[\mathbf{q}]$ in search decoding. 
\item Poor coding-gain, as we will see later in section \ref{sec:Search_IT}.
\item Exhaustive search in nature.
\end{itemize}

\textbf{VQ family:}
\begin{itemize}
\item Very good rate-distortion performance.
\item Not benefiting from a possible fast search in the encoded domain.
\item Slow distance computation due to LUTs. 
\end{itemize}

Within the VQ family, in particular, we can mention the following properties for the \textbf{PQ-based methods:}
\begin{itemize}
\item Simple and straightforward distance formation as the sum of distances of sub-codebooks (advantage over residual-based methods).
\item Exhaustive in nature and hence should be preceded by a sort of residual encoding (disadvantage w.r.t. residual-based methods).
\item Rate-specific design, i.e., having to retrain all sub-codebooks for any targeted rate (disadvantage w.r.t. residual-based methods).
\item Not flexible design, e.g., the number of codewords is bounded around $m=256$, and the sub-codebook divisions are also limited.
\end{itemize}

So how do the two families compare overall? While the binary hashing methods may be more favorable for speed and the VQ family more efficient for memory storage, one can conclude that the VQ family is more successful overall (see, e.g., the recent survey \cite{ANN:7915742}). In fact, the VQ-based methods constitute the core of the recent FAISS \cite{FAISS:JDH17}, a library for fast search developed by Facebook AI Research. 

How can one benefit from both of these strategies? In particular, how can we benefit from fast search in the encoded domain, and at the same time have optimal rate-distortion performances? Before proposing our solution, we mention the one, and to the best of our knowledge the only such attempt, i.e., the Polysemous codes \cite{douze2016polysemous}. The idea is to reflect the distances of the learned codewords into their enumerations, i.e., if two codewords are near, e.g., in $\Re^{\frac{n}{p}}$ space, they should get close enumerations as well. This is enforced by a simulated-annealing type of optimization after the codebooks are trained. The benefit of such re-enumeration is that similar to binary codes, one can consider the encoding of database items into PQ sub-codebooks like $\log_2{(m)}$-ary codes and perform a fast Hamming-distance-based search to filter out a lot of database candidates. The remaining list of database candidates is further refined using usual PQ-based search. As a result, the authors show that by pruning around $95\%$ of the database using fast search from codes and further refinement using PQ/OPQ, they get performances on par with OP/OPQ.

We next provide our general strategy for ANN search.
\subsection{Our proposed framework}   \label{subsec:Search_SOTA_Proposed}
This thesis advocates a third type of solution for the problem of ANN search by proposing a middle ground between the two families: 1) We first find an initial list $\hat{\mathcal{L}}_1(\mathbf{q})$, entirely within the fixed-point encoded domain. 2) This list is then refined in the original $\Re^n$ from the reconstructed vectors to give $\hat{\mathcal{L}}_2(\mathbf{q})$. This second ``list-refinement'' stage, unlike the vector compression solution, is non-exhaustive and is performed only within $|\hat{\mathcal{L}}_1(\mathbf{q})|$ items, far smaller than the original $N$. As will be shown in the experiments, while the second stage adds only a small overhead to the computational cost, it significantly improves performance and requires the same storage.

The realization of this idea requires that the encoding scheme possess three important characteristics: (\textit{i}) They should preserve as much amount of information about the original data in the codes as possible. (\textit{ii}) Their decoding should be efficient. (\textit{iii}) They should maintain good rate-distortion behavior. Except for the second point, it cannot be achieved for binary codes as we have seen before. Therefore, instead of binary codes, we use the framework of Sparse Ternary Codes (STC) that we have developed earlier in section \ref{sec:SingleLayer_Analysis}, and in particular, its multi-layer variant, i.e., the ML-STC of section \ref{sec:MultiLayer_MLSTC}.

\begin{figure}
 \begin{center} 
 \includegraphics[width=0.70\textwidth]{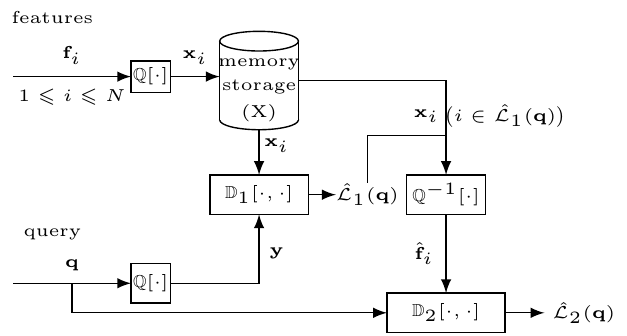}
 \end{center}
\caption{Proposed framework: An initial search is performed on the codes to prune the majority of database items. The remaining short-list is further refined by reconstructing from those codes and direct comparison with the query.}    
\label{fig:Search_SOTA_Proposed}
\end{figure}
Fig. \ref{fig:Search_SOTA_Proposed} sketches the general pipeline for similarity search based on this idea and Eq. \ref{eq:Search_SOTA_Proposed_Lists} describes the construction of the lists:
\begin{equation}  \label{eq:Search_SOTA_Proposed_Lists}
\begin{aligned}
\hat{\mathcal{L}}_1(\mathbf{q}) &= \mathbb{D}_{\text{1}}[\mathbf{y}, \mathrm{X}] = \texttt{STC-Fast-Decoder}(\mathbf{y}, \mathrm{X}), \\
\hat{\mathcal{L}}_2(\mathbf{q}) &= \mathbb{D}_{\text{2}}\big[ \mathbf{q}, \hat{\mathbf{f}}_{\{ i \in \hat{\mathcal{L}}_1(\mathbf{q}) \}}  \big] = \{ i \in \hat{\mathcal{L}}_1(\mathbf{q}) | d_{\mathcal{E}}(\mathbf{q} , \hat{\mathbf{f}}_i) \leqslant \epsilon_2  \}.
\end{aligned}
\end{equation}

The fast decoding of $\texttt{STC-Fast-Decoder}(\mathbf{y}, \mathrm{X})$ will be described later in Algorithm \ref{alg:Search_FastDecoder} and other details of the whole procedure will be clarified in section \ref{sec:Search_MLSTC}. 

We next describe the motivations behind choosing a ternary encoding and in particular our STC framework.

\section{An information-theoretic picture} \label{sec:Search_IT}
In this section, we try to analyze the problem from an information-theoretic perspective. We aim at designing ``good codes'' from the viewpoint of information preservation. Through this end, we introduce a simple and intuitive measure to characterize the goodness of a code and we term it the ``coding gain for similarity search''. This brings us to several conclusions for code design, e.g., while the binary encoding can be interpreted as one extreme to the more general ternary encoding where sparsity is minimal, we will show that the other side of the spectrum where sparsity is high, is more advantageous from the viewpoint of coding gain. 

For such information-theoretic arguments, it is inevitable to make some assumptions and limit our focus to more particular cases. Therefore, throughout this section, the underlying setup is the identification problem that we describe next. 
\subsection{Identification systems} \label{subsec:Search_IT_CI}
A special case of the ANN search problem is the identification problem. While the general ANN concept is used mostly in retrieval systems and computer vision problems, where similarity is defined in a broad sense and hence a list of database items is required, in biometric applications, the similarity is usually restricted to different instances of the same underlying identity. This means that the query should find exactly one item from the database. In other words, the list size is usually restricted to $|\mathcal{L}(\mathbf{q})| = 1$. 

\begin{figure}
 \begin{center} 
 \includegraphics[width=0.70\textwidth]{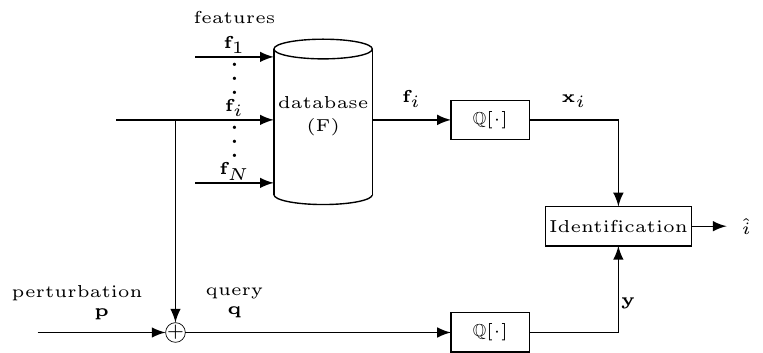}
 \end{center}
\caption{Schema of an identification system. The query can be modeled as a noisy version of one of the database items.}    
\label{fig:Search_IT_IdentificationDiagram}
\end{figure}

This problem appears mainly in biometric systems and content identification systems. In biometric systems, a person is identified based on his/her biometric features, e.g., iris scan or fingerprint. In content identification, multimedia contents are checked if they contain pieces of registered multimedia like music or video clips. Examples of such use cases appear in copyright protection or copy detection. Fig. \ref{fig:Search_IT_IdentificationDiagram} sketches a generic identification system. 

A simplifying element w.r.t. the more general ANN search problem is the fact that the query may be assumed to be a noisy (additive) version of some database item $\mathbf{f}_i$. This, along with other simplifying assumptions like simple distributions for the data, make the information-theoretic analyses feasible. This was started from \cite{Frans:1228096}, where the authors defined the identification capacity as the exponent of the number of database items $N$ that could reliably be identified in an asymptotic case, where the feature dimension $n  \rightarrow  \infty$. The authors modeled the degradation as noisy communication channels while they considered the vectors as random channel codes. The authors then characterized the identification capacity as $I(F;Q)$, i.e., the mutual information between the enrolled items $F$ and the noisy queries $Q$. In this setup, however, the increase in $n$ leads to an exponential increase in $N \simeq 2^{nI(F;Q)}$. This incurs infeasible search/memory complexities, making the system impractical. So similar to the general ANN search, identification systems seek fast and compact setups. Therefore, subsequent works attempted at decreasing these complexities. For example, \cite{Frans:5205870} considered a two-stage clustering-based system to speed-up the search, while \cite{Tuncel:4839033} considered the compression of vectors before enrollment and studied the achievable storage and identification rates.

How could these information-theoretical arguments be useful for practical ANN search?  In fact, similar to the classical Shannon theory setup, these methods consider asymptotic cases, where the aim is to identify as many numbers as possible under the identification criterion $P_{id} \geqslant 1 - \epsilon$. It can be argued, however, that the number of items $N$ is fixed in practice and can be well below the amount that the identification capacity can accommodate, i.e., $N \ll 2^{nI(F;Q)}$. Moreover, this analysis is focused on achieving a negligible probability of the error event, i.e., $\mathbb{P}[\hat{i} \neq i] \rightarrow 0$. In practice, we might think of $|\mathcal{L}(\mathbf{q})| > 1$, or even non-negligible $\mathbb{P}[\hat{i} \neq i]$.

Now consider the case, where instead of being interested in identifying as many numbers of items as possible, we are given a fixed number of messages or items. Instead, we want to minimize the decoding complexity by encoding the data into a less-entropic space. This is our motivation for the definition of the concept of coding gain that we define next.
\subsection{Coding gain for similarity search} \label{subsec:Search_IT_CodingGain}
Which encoding scheme to choose to gain memory and computational advantages over the exhaustive search by possibly compromising the performance in terms of identification accuracy? 

Towards this end, to quantify the efficiency of a coding scheme $\mathbb{Q}[\cdot]$ for ANN search, the coding gain is defined as the ratio of mutual information between the encoded version of a database item and its noisy query, and the entropy of the encoded representations, i.e.: 

\begin{equation} \label{eq:Search_IT_CodingGain}
g_{\mathbf{F}}(\mathbb{Q}) = \frac{I(\mathbf{X};\mathbf{Y})}{H(\mathbf{X})},
\end{equation}
where have that $\mathbf{X} = \mathbb{Q}[\mathbf{F}]$ and $\mathbf{Y} = \mathbb{Q}[\mathbf{Q}]$.\footnote{Note that the encoding of the database items and the query need not have the same parameters. In fact, for our ternary design, we adapt the encoding to the degradation statistics.}

Mutual information in the definition of Eq. \ref{eq:Search_IT_CodingGain} takes into account all the channel transition probabilities. In fact, independent of the decoding algorithm used, it indicates the search performance. 

From the other hand, the entropy in the denominator characterizes both memory and complexity. Obviously, the cost of the database storage is directly linked to $H(\mathbf{X})$. One can use source coding to store the encoded data close to the entropy. As for the search complexity, since the effective space size is $|\mathcal{X}^n| \approx 2^{H(\mathbf{X})}$, a lower entropic space also implies a lower search complexity. This means that independent of the decoding algorithm, searching in a less entropic space is faster.

Under the following model, we will next analyze the behavior of both binary and ternary encoding and derive the coding gains w.r.t. each of them.

\textbf{Signal model:} 
Our signal model is the same as the setup of section \ref{subsec:SingleLayer_Analysis_IT}, i.e., we have \textit{i.i.d.} Gaussian data for $\mathbf{F}$, and since we use unit-norm projections, the data in the projected domain, i.e., $\tilde{\mathbf{F}} = [\tilde{F}_1, \cdots, \tilde{F}_m]^T$ follows the same distribution, i.e., with elements $\tilde{F} \sim \mathcal{N}(0,\sigma^2)$ and the corresponding noisy projected data is distributed as $\tilde{Q} \sim \mathcal{N}(0, \sigma^2 + \sigma_P^2)$.

\subsubsection{Binary encoding} \label{subsec:Search_IT_Binary}
While we can model the perturbation between $\mathbf{F}$ and $\mathbf{Q}$ as a Gaussian channel, the bit-flipping between $X_b =  \text{sign} (\tilde{F})$ and $Y_b = \text{sign} (\tilde{Q})$ can be considered a Binary Symmetric Channel (BSC). In \cite{Voloshynovskiy2010:ITW}, this was derived as a BSC with the probability of bit flipping $P_b = \mathbb{E}_{p(\tilde{q})}[\mathcal{Q}(\frac{|\tilde{q}|}{\sigma_P})] =  \frac{1}{\pi} \arccos(\rho)$, where $\mathcal{Q}(u) =  \int_u^{\infty}  \frac{1}{\sqrt{2\pi}}e^{\frac{-u'}{2}} du'$ is the Q-function.

The entropy of $\mathbf{X}_b$ with length $m$ is simply $H(\mathbf{X}_b) = m$ for the equiprobable bits. The mutual information for a BSC is given as $I(X_b,Y_b) = 1 - H_2(P_b)$, where the binary entropy is defined as $H_2(p) = -p\log_2{p} - (1-p) \log_2{(1-p)}$. 

\subsubsection{Ternary encoding} \label{subsec:Search_IT_Ternary}
Remember we have derived the information measures for the ternary codes, earlier in section \ref{subsec:SingleLayer_Analysis_IT}. In particular, the entropy was given in Eq. \ref{eq:SingleLayer_STC_H-ternary}, while the mutual information was decomposed in Eq. \ref{eq:SingleLayer_STC_I} for which the joint entropy was derived in Eq. \ref{eq:SingleLayer_STC_HJoint}. This calculation required the computation of the transition probabilities of the ternary channel of Eq. \ref{eq:SingleLayer_STC_transition}, which was a function of the threshold values $\lambda_X$ and $\lambda_Y$.\footnote{Note that by putting $\lambda_X = \lambda_Y = 0$, we obtain the same entropy and mutual information as the binary case.}

\subsubsection{Coding gain comparison of binary and ternary encoding} \label{subsubsec:Search_IT_Comparison}
\begin{figure}  [!h]
\vspace{-0.75cm}   
   \begin{center}
   \begin{tabular}{c}
    \includegraphics[width=0.80\textwidth]{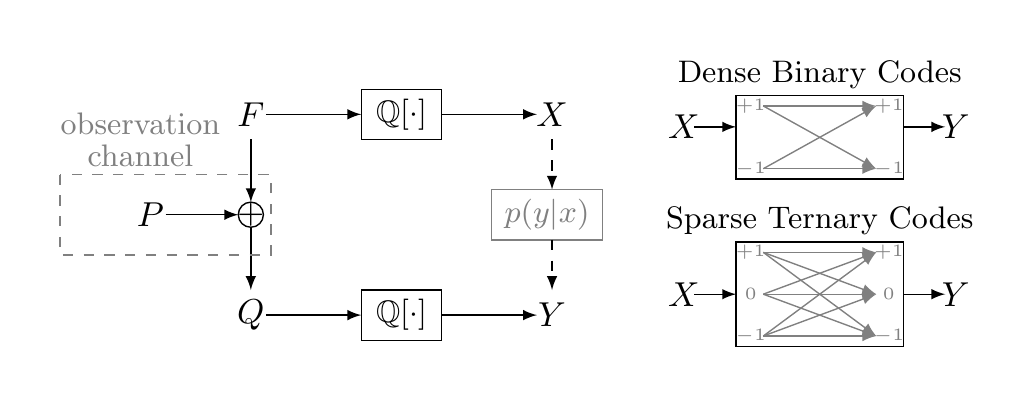}
   \end{tabular}
   \end{center}
   \vspace{-0.25cm}
\caption{The channel characterized by $p(y|x)$ models the degradation between the data and its corresponding query in the encoded domain. A BSC models the perturbation in binary codes, while a noise-adaptive ternary channel models the STC.}
{ \label{fig:Search_IT_Channels}}
   \end{figure}

As was pointed out earlier, while the degradation between the data and query can be modeled as a noisy communication channel, e.g., the Gaussian channel, in the encoded domain, this degradation can be modeled as another channel, albeit with lower capacity. This idea is illustrated in Fig. \ref{fig:Search_IT_Channels}, where binary and ternary channels model the degradations in the encoded domain.

\begin{figure*}  
   \begin{center} 

\includegraphics[width=1.0\textwidth]{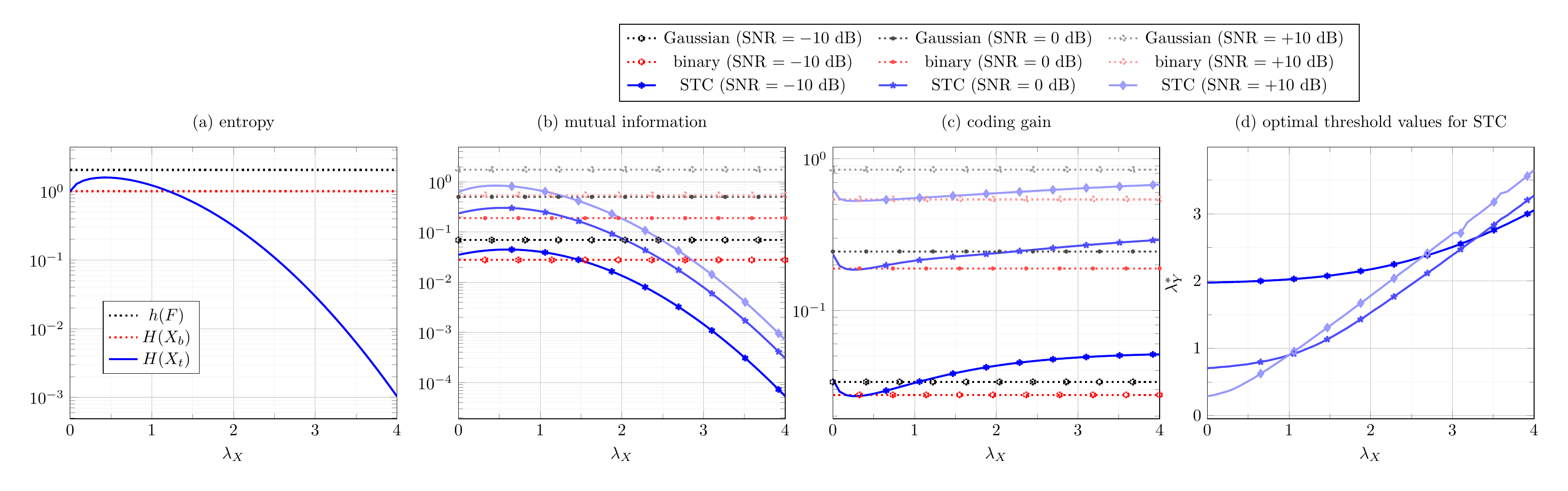}
   \end{center}
\vspace{-0.5cm}    
   \caption[example]{Coding gain comparison for ternary and binary encoding. Although the $I(X_t;Y_t)$ curve has a concave shape with respect to $\lambda_X$, which means that its value decreases as the code becomes sparser, $H(X_t)$ decreases with a faster rate for increasing $\lambda_X$. This means that the coding gain increases as the code become sparser and then saturates at $\frac{I(X_t;Y_t)}{H(X_t)} = 1 - \frac{H(X_t|Y_t)}{H(X_t)} \leqslant 1$. }
   \label{fig:Search_IT_CGainComparison}
   \end{figure*}

We are now ready to compare the coding gain of Eq. \ref{eq:Search_IT_CodingGain} for the binary and ternary encodings. This is calculated for three different values of SNR $= 10log_{10}\frac{\sigma_F^2}{\sigma_P^2}$  by varying $\sigma_P^2$ and the results are shown in Fig. \ref{fig:Search_IT_CGainComparison}.

For the ternary case, for every value of $\lambda_X$, we find the optimal $\lambda_Y^*$ that maximizes $I(X_t;Y_t)$ using a simple grid-search since we do not have a closed-form expression for that. In practice, usually $\sigma_P^2$ is fixed by the problem, $\lambda_X$ is chosen by memory constraints and $\lambda_Y^*$ may be computed using cross-validation.

As is seen from Fig. \ref{fig:Search_IT_CGainComparison}, the proper choice of thresholds leads to interesting regimes, where for the same entropy and hence the same number of bits, the ternary code preserves more mutual information compared to binary codes. 


\subsection{Identification performance of binary and ternary codes} \label{subsec:Search_IT_Identification}
As a simple experiment to compare the performances of the binary encoding and STC, we consider the identification of synthetic data by comparing the probability of correct identification for different pairs of memory and complexity ratio. 

For the ternary case, we use the sub-optimal but fast decoder of Algorithm \ref{alg:Search_FastDecoder}, which will be explained in the next section.

Memory usage is measured by entropy of a coded block, i.e., $m_bH(X_b)$ for the binary and $m_tH(X_t)$ for ternary. We measure the complexity as the ratio of the big-O complexity of a search algorithm to the big-O complexity of performing the original exhaustive scan. This is measured as $\frac{Nm_b}{Nn}$ for the binary and $\frac{4\alpha_X \alpha_Y Nm_t}{Nn}$ for the STC with fast decoder of Algorithm \ref{alg:Search_FastDecoder}.

\begin{figure} 
   \begin{center} 
   \includegraphics[width=0.85\textwidth]{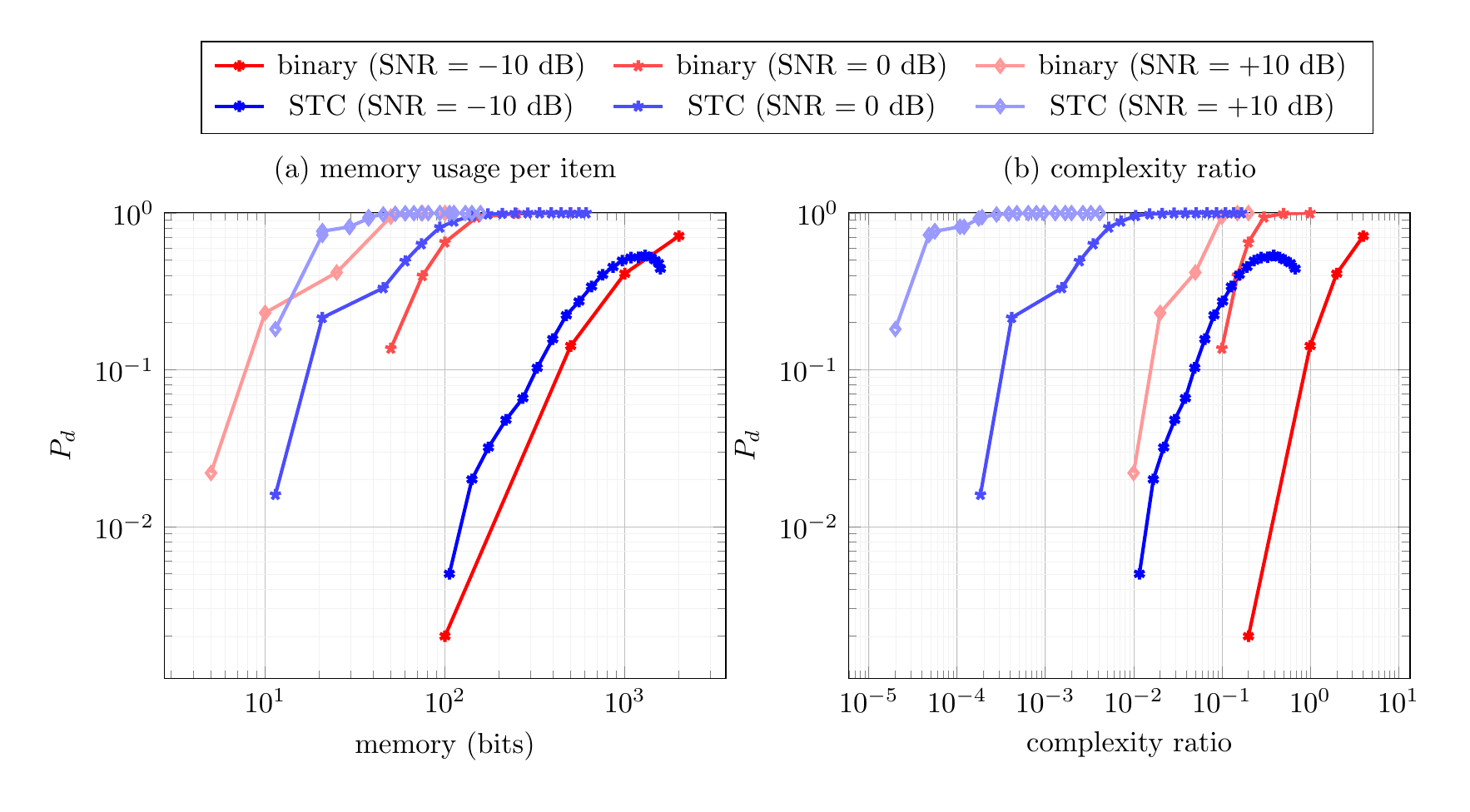}
   \end{center}
\vspace{-0.5cm}   
\caption[example]{Performance-memory-complexity profile for identification of $N = 1$ Mio synthetic data with $n = 500$. The sub-linear fast decoder of Algorithm \ref{alg:Search_FastDecoder} was used for STC. }
   \label{fig:Search_IT_IdentificationPerformance} 
   \end{figure}

We kept the complexity of the floating-point projection stage the same for both cases in each experiment. The results are depicted in Fig. \ref{fig:Search_IT_IdentificationPerformance}. For equal memory usage, a large gap is observed between the complexity ratios of the two counterparts. Furthermore, usually much better performance is achieved for the STC.

%
%

\section{Similarity search using ML-STC} \label{sec:Search_MLSTC}
Up to now, we have defined the similarity search problem and have seen how the two families of solutions, namely the binary hashing and the VQ try to address the triple trade-offs of memory, complexity, and performance for this problem. We then outlined the general schema of our proposed solution for the ANN search which consists of two stages of decoding: a fast decoding in the space of codes and a complementary decoding based on reconstruction to refine the approximative list of the first stage. We further saw that the type of encoding proposed by the STC framework is significantly advantageous w.r.t. the binary encoding, as measured by the coding gain.

How do we actually realize this double-stage idea for similarity search? Section \ref{subsec:Search_MLSTC_FastDecoding} describes the initial fast-decoding idea and section \ref{subsec:Search_MLSTC_Architecture} presents the overall similarity search architecture that we propose.

As for the encoding and the underlying structure, we use the framework of ML-STC\footnote{Note that among rate-distortion-based frameworks, other than the ML-STC, we have developed the RRQ (section \ref{subsec:MultiLayer_RRQ_RRQ}), the ML-STC-Procrustean (section \ref{subsec:MultiLayer_MLSTC-Procrustean_Alg}) and the STNets (section \ref{sec:MultiLayer_STNets}), as well. The RRQ is VQ-based and generally, does not benefit from fast initial decoding in the space of codes. However, the ML-STC-Procrustean and the STNets both benefit from the same encoding and have superior rate-distortion performance w.r.t. the ML-STC, yet we use the latter in our experiments for the sake of simplicity.} that we have developed earlier in section \ref{sec:MultiLayer_MLSTC} and for which we have studied the rate-distortion performance.

\subsection{Fast decoding of STC} \label{subsec:Search_MLSTC_FastDecoding}
We have seen earlier in section \ref{subsubsec:SingleLayer_Analysis_IT_MLDecoding} that, having observed the output of a ternary channel, how we can estimate the likelihood of different candidate STC inputs to this channel. This can be used for search when the encoded query should find its nearest neighbors from the database.  While this rule was maximum-likelihood optimal (Eq. \ref{eq:SingleLayer_Analysis_MLDecoder}), it had to scan all the database items exhaustively to find the match. More precisely, it had to check all the 9 transition probabilities of $\mathrm{P}_t$. 

However, we saw in section \ref{subsubsec:Search_IT_Comparison} that higher coding gains are achieved when sparsity is higher. So almost for all items, most elements should be 0's and a lot of the transitions will happen within the 0's. This means that they are less informative and we can neglect them without losing much information. On the other hand, for complexity, this provides a significant saving since we only compute transitions to and from $\pm1$'s which happen less often and are more informative. So out of the 9 transitions in $\mathrm{P}_t$, we only consider 4 and neglect the other 5.

This means that the search will no longer be exhaustive since a query with a $+1$ at a certain position will focus on finding items that happen to have a $+1$ or $-1$ at that certain position and neglect a lot of other items that have a $0$ at that location.

Let us first concretize this idea given a single-layer code and then extend it to multiple layers.
\subsubsection{Single-layer} \label{subsubsec:Search_MLSTC_FastDecoding_Single}
Given the single-layer STC's for all database items, i.e., $\mathrm{X} = [\mathbf{x}_1, \cdots, \mathbf{x}_N]$, the idea of fast decoding can be realized, e.g.,  using LUT's in the form of inverted files where the query code, depending on its non-zero activity, votes for items read from the corresponding LUT's. Algorithm \ref{alg:Search_MLSTC_LUT} describes how such LUT's can be constructed.

\begin{algorithm} \caption{LUT's for STC} \label{alg:Search_MLSTC_LUT}
\begin{algorithmic}[0]
    \INPUT STC's: $\mathrm{X} = [\mathbf{x}_1, \cdots, \mathbf{x}_N]$
    \OUTPUT Look-Up-Tables: $\textsf{LUT}^+$ and $\textsf{LUT}^-$
\end{algorithmic}
\begin{algorithmic}[1] 
\State Initialize empty lists $\textsf{LUT}^+ = \lbrace \textsf{LUT}^+_1, \cdots,\textsf{LUT}^+_m\rbrace$ and $\textsf{LUT}^- = \lbrace \textsf{LUT}^-_1, \cdots,\textsf{LUT}^-_m\rbrace$.
\For{$i = 1,\cdots,N$} 
\For{$m' = 1, \cdots, m'$}
\If {$\mathrm{X}(m',i) = +1$}
\State $\textsf{LUT}^+_{m'} \leftarrow \lbrace \textsf{LUT}^+_{m'}, i \rbrace$
\ElsIf {$\mathrm{X}(m',i) = -1$}
\State $\textsf{LUT}^-_j{m'} \leftarrow \lbrace \textsf{LUT}^-_{m'}, i \rbrace$
\EndIf
\EndFor
\EndFor
\end{algorithmic}
\end{algorithm}

Once the LUT's are constructed, the encoded query votes positively for items from the database with similar activity and votes negatively for items with dissimilar activity. Algorithm \ref{alg:Search_FastDecoder} details this procedure, where the voting vector $\mathbf{v}_{\mathbf{q}}$ of size $N$ is initialized with zeros. Whenever an item $i$ from the database gets a match with the query, its current value of $\mathbf{v}_{\mathbf{q}}(i)$ is increased by a constant factor $\nu^+$. On the other hand, the mismatch values, i.e., the cases where the sign of active coefficients is different, will be penalized in the voting procedure by a negative constant $\nu^-$.

After all the relevant items are counted, the initial approximative list $\hat{\mathcal{L}}_1(\mathbf{q})$ is returned by finding all items whose votes are higher than a certain threshold, or by picking the $|\hat{\mathcal{L}}_1(\mathbf{q})|$ top elements with highest votes. 

\begin{algorithm} \caption{\texttt{STC-Fast-Decoder}} \label{alg:Search_FastDecoder}
\begin{algorithmic}[0]
    \INPUT STC encoded query $\mathbf{y}$, $\textsf{LUT}^+$ and $\textsf{LUT}^-$ (from Algorithm \ref{alg:Search_MLSTC_LUT}), \\voting constants $\nu^+$ (to encourage match) and $\nu^-$ (negative value to penalize mismatch)
    \OUTPUT The initial approximative list $\hat{\mathcal{L}}_1(\mathbf{q})$ (or the voting vector $\mathbf{v}_{\mathbf{q}}$)
\end{algorithmic}
\begin{algorithmic}[1]
\State Initialize a voting vector $\mathbf{v}_{\mathbf{q}}$ ($N \times 1$) with zeros.
\State List the indices of all $+1$ elements of $\mathbf{y}$ in $\mathcal{L}^+$ and all $-1$ elements in $\mathcal{L}^-$.
\For {all $i^{++} \in \mathcal{L}^+$ and $i^{--} \in \mathcal{L}^-$} \Comment Encouraging sign matches.
\State $\mathbf{v}_{\mathbf{q}}(\textsf{LUT}^+_{i^{++}}) \leftarrow \mathbf{v}_{\mathbf{q}}(\textsf{LUT}^+_{i^{++}}) + \nu^+$ 
\State $\mathbf{v}_{\mathbf{q}}(\textsf{LUT}^-_{i^{--}}) \leftarrow \mathbf{v}_{\mathbf{q}}(\textsf{LUT}^-_{i^{--}}) + \nu^+$ 
\EndFor
\For {all $i^{-+} \in \mathcal{L}^+$ and $i^{+-} \in \mathcal{L}^-$}  \Comment Penalizing sign mismatches.
\State $\mathbf{v}_{\mathbf{q}}(\textsf{LUT}^-_{i^{-+}}) \leftarrow \mathbf{v}_{\mathbf{q}}(\textsf{LUT}^-_{i^{-+}}) + \nu^-$  
\State $\mathbf{v}_{\mathbf{q}}(\textsf{LUT}^+_{i^{+-}}) \leftarrow \mathbf{v}_{\mathbf{q}}(\textsf{LUT}^+_{i^{+-}}) + \nu^-$
\EndFor
\State Report all $i$'s ($1 \leqslant i \leqslant N$) with $\mathbf{v}_{\mathbf{q}}(i)$ bigger than some threshold as $\hat{\mathcal{L}}_1(\mathbf{q})$. \Comment Or take the top $|\hat{\mathcal{L}}_1(\mathbf{q})|$ such values.
\end{algorithmic}
\end{algorithm}

The voting hyper-parameters $\nu^+$ and $\nu^-$ should be chosen as positive and negative constants, respectively. In setups where the query noise is low, the event of sign mismatch can be very rare, so the magnitude of $\nu^-$ should be chosen much larger than $\nu^+$. These values can be set using cross-validation.

\subsubsection{Multi-layer} \label{subsubsec:Search_MLSTC_FastDecoding_Multi}
Now consider the case where we have multi-layer STC's, as in the ML-STC encoding scheme of section \ref{sec:MultiLayer_MLSTC}. Suppose we have $L$ layers of database codes as $\mathrm{X}^{[l]} = [\mathbf{x}^{[l]}_1, \cdots, \mathbf{x}^{[l]}_N]$ for $1 \leqslant l \leqslant L$, and similarly for the query we have $\mathbf{y}^{[1]}, \cdots, \mathbf{y}^{[l]}, \cdots, \mathbf{y}^{[L]}$.

How do we effectively use the multi-layer codes in the initial fast-decoding procedure? The idea is to aggregate the voting vectors from different layers together. However, different layers do not have the same significance in terms of first, their overall impact in distance approximation and second the robustness against query noise. So we aggregate their votes with different weights $\omega^{[1]}, \cdots, \omega^{[l]}, \cdots, \omega^{[L]}$.

The impact of different layers in overal distance approximation can be measured from their relative reduction of distortion. So on the average, this can be measured as $\frac{\mathcal{D}_{l}}{\mathcal{D}_{l-1}}$ for $l = 1, \cdots, L$, where $\mathcal{D}_{l}$ is the normalized average distortion at layer $l$ and $\mathcal{D}_0 = 1$.

The characterization of the robustness of different layers to noise, however, requires full knowledge of noise and the degradation model, which is not known in practice. Intuitively speaking, this should follow a basic inequality in information theory often referred to as the source-channel separation theorem (Theorem 21 of \cite{shannon2001mathematical}) or the information transmission inequality (\cite{berger1998lossy}). According to this inequality, for a source with the distortion-rate function $D(R)$ to be transmitted through a channel with capacity $C$, one can achieve the distortions higher than the distortion-rate function, only if the rate is lower than the capacity. In other words, only $D \geqslant D(R)\big\rvert_{R=C}$ is achievable.

This has an important practical consequence for our multi-layer encoding. It can be concluded that only the first several layers of codes should be decoded and the codes corresponding to layers higher than some $l'$ should entirely be ignored in the fast decoding.\footnote{Note, however, that, this does not say anything about the rate-distortion encoding of the database items. Obviously, all layers should be used for the reconstruction within the list-refinement procedure that we explain next.} Therefore, as a heuristic rule, we choose the voting weights as $\omega^{[l]} = \frac{\mathcal{D}_{l}}{\mathcal{D}_{l-1}}$ for $l \leqslant l'$, and $\omega^{[l]} = 0$, for $l > l'$. In practice, we find $l' = 4$ or $l' = 5$ to be appropriate.

Algorithm \ref{alg:Search_MLAggregation} discusses the aggregation of multi-layer codes based on the fast decoding of Algorithm \ref{alg:Search_FastDecoder} and using the voting weights discussed above.

\begin{algorithm} \caption{Aggregation of votes for ML-STC} \label{alg:Search_MLAggregation}
\begin{algorithmic}[0]
    \INPUT for $l = 1, \cdots, L$: query code $\mathbf{y}^{[l]}$, database $\mathrm{X}^{[l]}$ (or the corresponding LUT's), voting weight $\omega^{[l]}$  
    \OUTPUT Initial list $\hat{\mathcal{L}}_1(\mathbf{q})$
\end{algorithmic}
\begin{algorithmic}[1]
\State Initialize a voting vector $\mathbf{v}(\mathbf{q})$ ($N \times 1$) with zeros.
\For {$l = 1, \cdots, L$}
\State $\mathbf{v}_{\mathbf{q}}  \leftarrow  \mathbf{v}_{\mathbf{q}} + \omega^{[l]} \times \texttt{STC-Fast-Decoder}[\mathbf{y}^{[l]}, \mathrm{X}^{[l]}]$
\EndFor
\State Report all $i$'s ($1 \leqslant i \leqslant N$) with $\mathbf{v}_{\mathbf{q}}(i)$ bigger than some threshold as $\hat{\mathcal{L}}_1(\mathbf{q})$. \Comment Or take the top $|\hat{\mathcal{L}}_1(\mathbf{q})|$ such values.
\end{algorithmic}
\end{algorithm}

\subsection{Similarity search architecture}  \label{subsec:Search_MLSTC_Architecture} 
Here we summarize our proposed search architecture by putting together the different elements we have discussed so far.

\begin{figure}  [!h]

   \begin{center} 
   \includegraphics[width=0.99\textwidth]{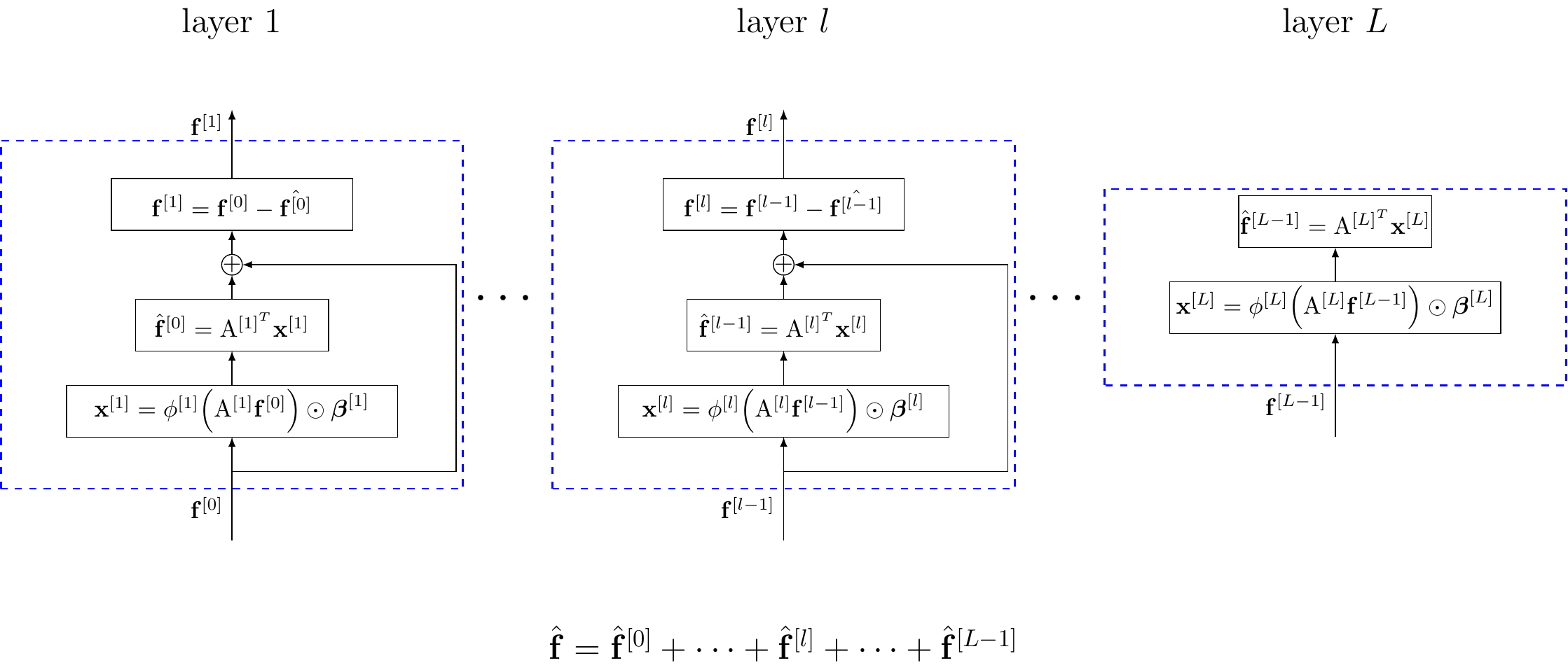}

   \end{center}
\caption{ML-STC architecture}    

   \label{fig:Search_MLSTC}
   \end{figure}

\textbf{Encoding:} We use the ML-STC encoding of section \ref{sec:MultiLayer_MLSTC}, which can be summarized as in Fig. \ref{fig:Search_MLSTC}. Note however that, as we saw earlier in section \ref{sec:Search_IT}, the encoding threshold $\lambda$ need not be the same for the database items and the query.

Once the database items are encoded, LUT's can be formed for all layers using the procedure of Algorithm \ref{alg:Search_FastDecoder}.

As we have outlined earlier in section \ref{subsec:Search_SOTA_Proposed} and sketched in Fig. \ref{fig:Search_SOTA_Proposed}, our decoding is a double-stage procedure consisting of the following:

\textbf{Initial fast decoding:} The initial decoding is performed on the encoded query and consists of a voting procedure that considers the sign matches and mismatches between the encoded query and the database. This procedure was described in Algorithm \ref{alg:Search_FastDecoder}. The votes of different layers are then aggregated as was described in Algorithm \ref{alg:Search_MLAggregation}, by taking into account their relative impact in distance approximation and noise robustness. This procedure produces an initial approximative list $\hat{\mathcal{L}}_1(\mathbf{q})$.  

\textbf{List refinement by reconstruction:} The initial list $\hat{\mathcal{L}}_1(\mathbf{q})$ is the result of a fast and sub-optimal decoding. Moreover, it only considers the encoded query $\mathbf{y}$ and not the original query vector $\mathbf{q}$. From the other hand, after the encoding, the original database vectors $\mathbf{f}_i$'s are not available anymore in the memory. So we refine $\hat{\mathcal{L}}_1(\mathbf{q})$ by comparing $\mathbf{q}$ with the reconstructions of the codes $\hat{\mathbf{f}}_i = \mathbb{Q}^{-1}[\mathbf{x}_i]$ using all the $L$ layers of ML-STC, i.e., $\hat{\mathbf{f}}_i = \hat{\mathbf{f}}^{[0]}_i + \cdots, \hat{\mathbf{f}}^{[L-1]}_i$. This list-refinement consists of re-ordering (and prunning) of $\hat{\mathcal{L}}_1(\mathbf{q})$ and produces $\hat{\mathcal{L}}_2(\mathbf{q})$, according to $\hat{\mathcal{L}}_2(\mathbf{q}) = \{ i \in \hat{\mathcal{L}}| d_{\mathcal{E}}( \hat{\mathbf{f}}_i , \mathbf{q} ) \leqslant \epsilon  \}$, or choosing the top $|\hat{\mathcal{L}}_2(\mathbf{q})|$ values. Notice that this stage is non-exhaustive and is performed only on a very small portion ($|\hat{\mathcal{L}}_1(\mathbf{q})| <5 \times 10^{-3} N$) of the reconstructed vectors. 

\section{Experiments} \label{sec:Search_Experiments}
Here we perform several experiments on our proposed similarity search pipeline. These are to validate the search performance for different memory budgets and by considering the complexity requirements into account. We next describe the databases used for the experiment, the baseline methods for comparison and the final results.

\subsection*{Public Databases} 
We use 4 databases in our experiments:

\textbf{AR(1)} is a toy database we have used earlier in chapter \ref{chapter:SingleLayer} and chapter \ref{chapter:MultiLayer}. This is used only for assessment of rate-distortion behavior. The data is synthetically generated from a Gaussian distribution correlated with an AR(1) process. We consider 3 different correlation factors, $\rho = 0, 0.5, 0.9$ to simulate different scenarios. We train the algorithms on a limited sample size and test them on a separate set. An important motivation behind using this database is the fact that theoretical lower-bound of rate-distortion performance (for the asymptotic case of $n \rightarrow \infty$) is known for this data.

\textbf{MNIST} \cite{MNIST:726791} is extensively used for benchmarking learning algorithms and consists of $28\times28$ image pixels of $0$-$9$ digits. Its usage for search is limited to mid-scale scenarios. However, it is frequently reported among binary hashing literature. The database comes with a training set on which we both train the algorithms and index for search. The test set is used as the query and to report the rate-distortion performance.

The following databases are extensively used for benchmarking ANN search methods:

\textbf{SIFT-1M} \cite{PQ:5432202} is a large collection of local SIFT descriptors. The database comes with a ``training set'' on which the parameters of the algorithms are tuned, a separate ``main set'', which is indexed for search and also a ``query set''. The rate-distortion is reported on the main set.

\textbf{GIST-1M} \cite{PQ:5432202} is a large-scale database consisting of global GIST image descriptors. The set splits are the same as SIFT-1M.

Table \ref{table:Search_Exp_databases} reports the size of the databases used.
\begin{table}[]
\centering
\resizebox{0.80\textwidth}{!}{
\begin{tabular}{c|c|c|c|c}

                & dimension (n) & train-set size & index-set size & \# queries       \\ \hline
AR(1)-synthetic & $512$          & $10,000$          & -               & $10,000$     \\ \hline
MNIST           & $784$           & $60,000$          & same as train   & $10,000$     \\ \hline
SIFT-1M         & $128$           & $100,000$         & $1,000,000$       & $10,000$     \\ \hline
GIST-1M         & $960$           & $500,000$         & $1,000,000$       & $1,000$      \\ \hline
\end{tabular}
}
\caption{Databases used for similarity search and rate-distortion performance}
\label{table:Search_Exp_databases}
\end{table}

\subsection*{Baseline methods}
We compare our proposed approach with the state-of-the-art methods from both families of vector compressors and binary hashing. The choice of these methods is based on their importance and popularity and availability of their public codes.

We do not report results from billion-scale solutions like the Inverted Multi-Index \cite{IMI:6915715} as they are rather pre-indexing of databases and can be used equally in conjunction with any of these methods.\footnote{Furthermore, we were limited by our computational resources to run billion-scale experiments.} Although we do not experiment with the recent Polysemous codes \cite{douze2016polysemous}, it is easy to make a comparison since their performance is upper bounded (and actually a bit worse) than PQ and they prune the database using fast search to around $0.05N$, which is much higher than our $|\hat{\mathcal{L}}(\mathbf{q})| < 0.001N$.

\textit{$\text{STC}^{[L]}$-initial} is a search based on an initial decoding of an L-layer STC, which is purely performed on the space of codes. \textit{$\text{STC}^{[L]}$-refined} improves upon the initial decoding by search in the reconstructed short-listed candidates. 

Besides for ternary encoding, the idea of list-refinement can also be applied for the binary hashing methods that are not designed for direct distortion minimization in the original space of vectors but only within the space of binary codes. So we can consider, e.g., \textit{$\text{ITQ}$-refined} as a variant of ITQ based on list-refinement.\footnote{Up to the best of our knowledge, this is the first work that considers list-refinement along with rate-distortion characterization of binary hashing.} This extension, in fact, largely increases the performance of ITQ. We apply the same procedure for Sim-Hash and Sparse Projections, observing the same behavior.

For these binary methods, the optimal reconstruction consists of pseudo-inversion of the original projection matrix followed by a scalar weighting which we learn from the training set. So for a binary code $\mathbf{x} = \text{sign}(\mathrm{A}\mathbf{f})$, the reconstruction of the original vector is $\hat{\mathbf{f}} = \beta \mathrm{A}^{\dagger}\mathbf{x}$ and the optimal weighting is formulated and derived as:
\begin{equation*}
\beta^* = \underset{\beta}{\text{argmin}} || \mathrm{F} - \beta \mathrm{A}^{\dagger}\mathrm{X}||_{\mathcal{F}}^2 = \frac{\text{trace}[\mathrm{A}^{\dagger}\mathrm{X}\mathrm{F}^T]}{\text{trace}[(\mathrm{A}^{\dagger} \mathrm{X})(\mathrm{A}^{\dagger}\mathrm{X})^T]} ,
\end{equation*}
where $\mathrm{A}^{\dagger}$ is the Moore-Penrose pseudo-inverse of $\mathrm{A}$.

We report the results for different entropies of the short codes, i.e., the number of bits required to represent the codes on the average. For binary hashing methods, this is simply the length of the codes. For vector compression methods, this is the summation of $\log_2(\cdot)$ of the number of centroids for each sub-band. For the STC, this is the summation of the ternary entropies of each dimension of the code as in Eq. \ref{eq:SingleLayer_STC_Rate}, summed over all layers.

We also detail the computational complexities of different stages of the algorithms with big-O notation for comparison. We do not report run-time results as this is implementation-dependant and involves a lot of implementation issues out of the scope of the thesis.

\subsection*{Results}

\begin{figure*}  [!h]
   \begin{center} 
\includegraphics[width=0.95\textwidth]{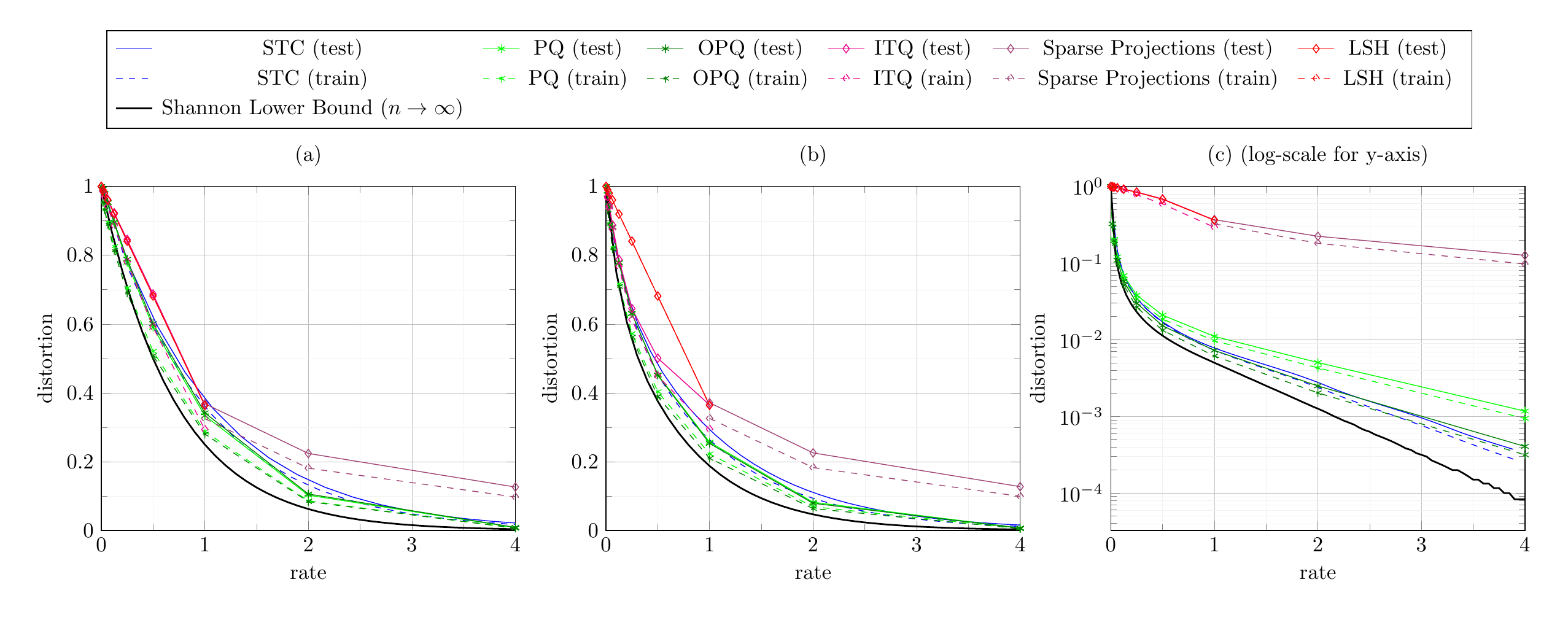}
\end{center}
\vspace{-0.5cm}    
   \caption{Distortion vs. rate on AR(1)-synthetic set for 3 different correlation levels. (a) \textit{i.i.d.} ($\rho = 0$),(b) mid-correlation ($\rho = 0.5$), (c) high-correlation ($\rho = 0.99$) }
   \label{fig:Search_Exp_SyntheticAR}
   \end{figure*}

\textbf{Rate-distortion behavior:} Fig. \ref{fig:Search_Exp_SyntheticAR} demonstrates the distortion-rate curves for AR(1)-synthetic dataset. As was also predictable from Fig. \ref{fig:SingleLayer_STC_DR}, binary methods have inferior rate-distortion performance. STC, on the other hand, has a performance on par with PQ and OPQ and for a wider operational regime. Note that we have introduced two more advanced versions of STC in multi-layers in chapter \ref{chapter:MultiLayer}, which have better rate-distortion performances than the ML-STC. In this chapter, however, we do not report them for the sake of simplicity.

\begin{figure*}  [!h]
   \begin{center} 

\includegraphics[width=1\textwidth]{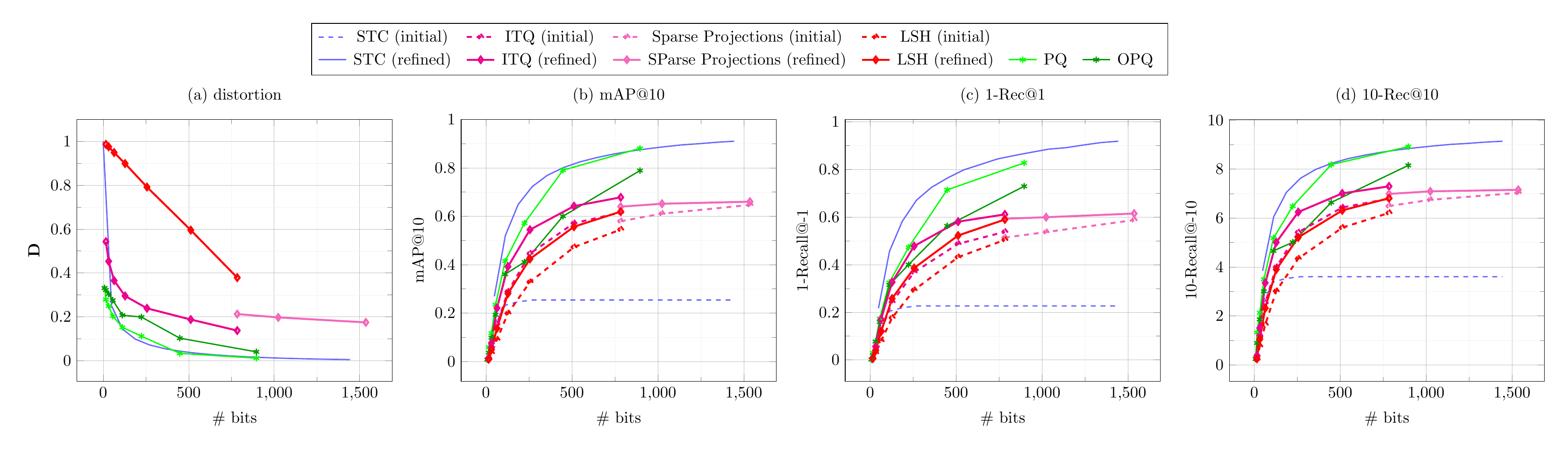}

   \end{center}
\vspace{-0.5cm}    
   \caption{ MNIST: The STC has $L = 16$ layers and a sparsity level $\alpha \approx 0.005$}
   \label{fig:Search_Exp_MNIST}
   \end{figure*}

\begin{figure}  [!h]
   \begin{center} 
\includegraphics[width=1\textwidth]{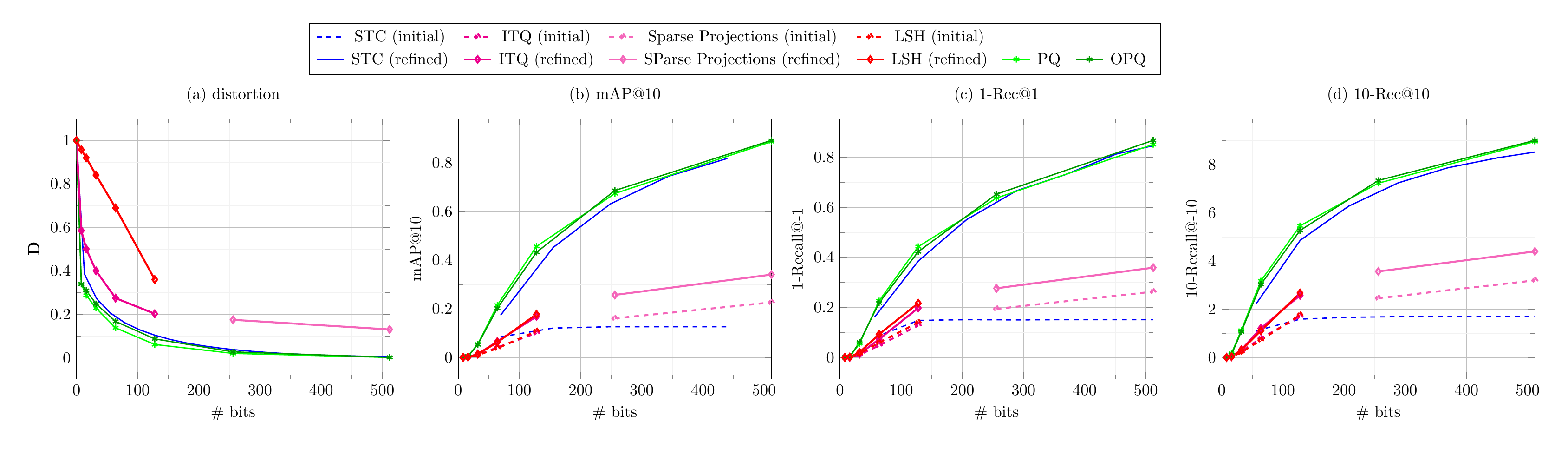} 

   \end{center}
\vspace{-0.5cm}    
   \caption{SIFT-1M: The STC has $L = 10$ layers and a sparsity level $\alpha \approx 0.04$}
   \label{fig:Search_Exp_SIFT1M}
   \end{figure}

\begin{figure}  [!h]
   \begin{center} 
\includegraphics[width=1\textwidth]{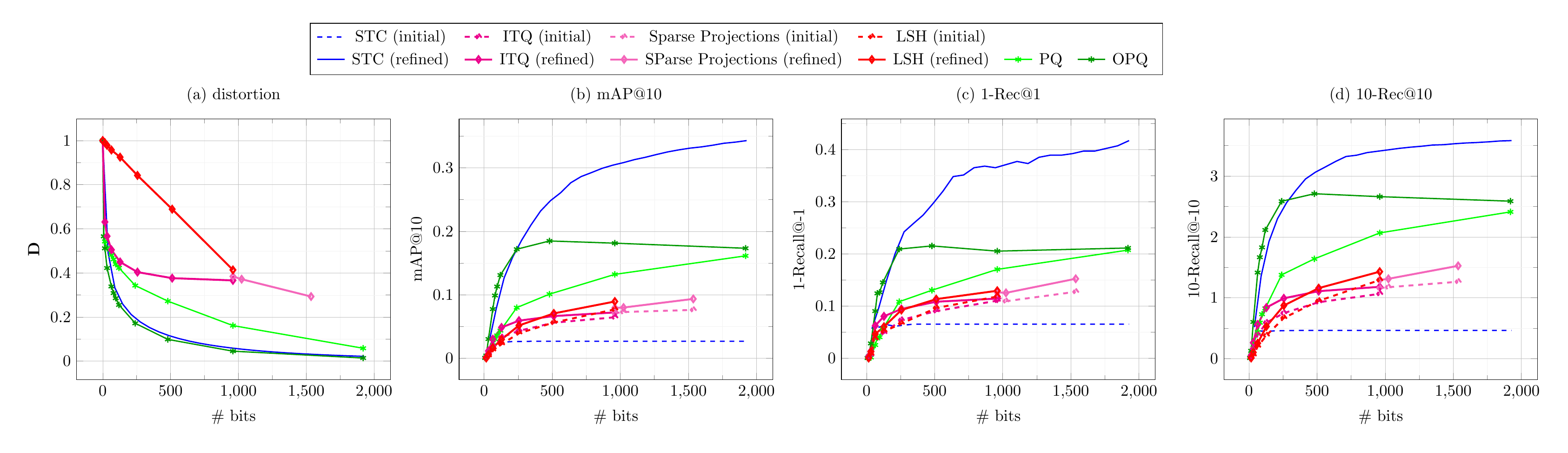} 

   \end{center}
\vspace{-0.5cm}    
   \caption{GIST-1M:The STC has $L = 26$ layers and a sparsity level $\alpha \approx 0.005$}
   \label{fig:Search_Exp_GIST1M}
   \end{figure}

\textbf{ANN search:} Figures \ref{fig:Search_Exp_MNIST}, \ref{fig:Search_Exp_SIFT1M} and \ref{fig:Search_Exp_GIST1M} demonstrate the results of ANN search on MNIST, GIST1M and SIFT1M, respectively. For these databases the initial short-list size was chosen as $\hat{\mathcal{L}} = 256, 1024, 512$, out of $N = 60000,1000000,1000000$, for the 3 experiments, respectively.

The search performance of STC is much higher than binary hashing and is on par with PQ/OPQ, while having considerably less complexity. It is seen that while STC largely outperforms PQ/OPQ on high-dimensional GIST-1M, it is slightly inferior on lower-dimensional SIFT-1M. This, in fact, is due to the fact that our assumption requires larger dimensions.
\subsection*{Computational complexities}
The computational complexities of the algorithms are detailed in Table \ref{table:Search_Exp_complexities}. The STC-refined has less Hamming distance computation than binary hashing, similar amount of Euclidean distance computation with PQ/OPQ and no look-up-table read as in PQ/OPQ, which is their main computation bottleneck. 

\begin{table}[]
\centering
\resizebox{0.95\textwidth}{!}{
\begin{tabular}{c|c|c|c}
\hline
            & Hamming distance computation & Euclidean distance computation & LUT read \\ \hline
$\text{STC}^{[L]}$-initial     & $4\alpha_X \alpha_Y NmL$ ($ \approx 6.8\times 10^6$)        & -                              & -        \\ \hline
$\text{STC}^{[L]}$-refined    & $4\alpha_X \alpha_Y NmL$  ($ \approx 6.8\times 10^6$)          & $ n|\mathcal{L}|$ ($\approx 5 \times 10^5$)             & -        \\ \hline
ITQ-initial                      & $mN$  ($\approx 5.12\times 10^8$)                         & -                              & -        \\ \hline
ITQ-refined                   & $mN$  ($\approx 5.12\times 10^8$)                         & $ n|\mathcal{L}|$  ($\approx 5 \times 10^5$)               & -        \\ \hline
PQ/OPQ                        & -                                                  & $ p\frac{n}{p}k$   ($\approx 2.5 \times 10^5$)                & $Npk$  ($\approx 1.6 \times 10^{10}$)        \\ \hline
\end{tabular}
}
\caption{Computational complexities: Values in parentheses correspond to operating values at $\approx 256$-bit codes on the GIST-1M experiment.}
\label{table:Search_Exp_complexities}
\end{table}

\section{Conclusions} \label{sec:Search_Conclusions}
Fast similarity search in large-scale databases is performed either using binary hashing that benefits from an efficient binary search, or Vector Quantization (VQ) that has excellent rate-distortion performance in approximating the vectorial distances within compact representations. While, from one hand, the first family of methods suffers from poor rate-distortion performance and restricting the search within the codes, and from the other hand, the second family cannot benefit from efficient search within the codes; in this chapter, we proposed a hybrid solution using the concept of list-refinement. This is based on pruning the majority of database candidates using an initial fast search within the space of ternary codes and then refining the result based on reconstruction from the codes. 

We showed that this strategy is best achieved using ternary encoding. So we first focused on coding efficiency as measured by the coding gain, an information-theoretic measure we introduced to encompass the triple trade-off of memory-complexity-performance. In particular, we concluded that our proposed Sparse Ternary Codes (STC), have higher coding gain than dense binary codes, provided that the sparsity is higher than a certain limit.

We then proposed a decoding scheme that trades off the preservation of mutual information with a significant complexity speed-up which relies on the sparsity of the ternary codes. This was considered for single-layer, as well as multi-layer codes.

The second crucial element was the rate-distortion optimality which we have studied earlier in chapter \ref{chapter:MultiLayer}. This was based on both sparsity (to minimize the rate-allocation sub-optimality), as well as multiple-layered design, for which we picked our ML-STC algorithm.

Putting these two aspects together, we sketched the entire search pipeline together and performed our experiments on mid- and large-scale setups where we showed we achieve performance, as well as, or better than the VQ-based family. However, this comes with a complexity much smaller than that of VQ codes, since most of the database items are pruned out using fixed-point efficient ternary search, rather than heavy LUTs. Moreover, thanks to its residual structure, the ML-STC codes are incremental, i.e., longer codes build on top of shorter codes. This is in contrast with the PQ-based methods that should entirely be redesigned, once a different code-length is targeted.

This leaves us with much promise for future research. Note that the ML-STC, as we have shown earlier in chapter \ref{chapter:MultiLayer}, can be extended to a neural network architecture, i.e., the STNets of section \ref{sec:MultiLayer_STNets}. While we have shown its superior rate-distortion performance in setups where training data is abundant (e.g., see Fig. \ref{fig:MultiLayer_STNets_MNIST_DR}- (a) and (d)), we can benefit further from its neural structure by defining mixed objective functions. So instead of crafted features, one can start directly from images and perhaps add label-aware loss functions to the rate-distortion objective. Note that this is not possible with VQ family since they do not have a neural structure. Moreover, for binary codes, it has been concluded that their performance is limited.
\chapter{Learning to compress images} \label{chapter:ImCompression}
In this chapter we focus our attention to images and the task of image compression. In particular, we take interest in learned scenarios were particularities of the given data may help achieve superior performance, as compared to the data-agnostic and more classical solutions.

Section \ref{sec:ImCompression_Why} tries to answer the question why learning from the data can be beneficial for image compression. We draw into attention four motivations for such an effort and then take a brief look at the recent literature of learning-based image compression.

In section \ref{sec:ImCompression_STNets}, we pick a very challenging scenario, i.e., the compression of natural color images with very high-resolution (around 2.5 megapixels) that do not possess a particular structure. Without performing all the necessary components within a typical image compression pipeline such as chrominance sub-sampling, entropy coding, bit-plane coding or careful rate-allocation, training our ML-STC-Procrustean on only 200 such images is enough to compete with the highly engineered JPEG2000 in terms of PSNR.

Under a more particular setup, in section \ref{sec:ImCompression_RRQ} we take the case of facial images as an example of a domain-specific scenario. We pick the RRQ framework that we developed in section \ref{sec:MultiLayer_RRQ} and apply it to compression of such images and show improvements over JPEG2000. 

While throughout this thesis, we were focusing more on achieving better rate-distortion trade-offs, we show in section \ref{subsec:ImCompression_RRQ_Denoising} that the very task of rate-distortion optimization can be helpful also to solve inverse problems. In particular, we show that compressibility under some learned model can be used as an effective prior for the task of image de-noising. More studies in this direction will be put as promising directions for future work that we briefly investigate in chapter \ref{chapter:CompPrior}.
\section{Why to learn to compress?} \label{sec:ImCompression_Why}
A very recent line of work has appeared within the machine learning community to try to target the task of lossy image compression using learning-based concepts. However, it is important to ask that after decades of research resulting in a large record of publications, as well as a lot of engineering work for image compression carried out by entities like the Joint Photographic Expert Group (JPEG) and having set multiple ISO standards and widely used codecs, why should we think about changing the entire paradigm of image compression and opt for more data-dependent, and possibly end-to-end approaches? One can think of 4 main reasons for such efforts:

\subsection*{i. Capturing more redundancies within the data:} 
The off-the-shelf solutions for image compression are designed for general natural images. In fact, the core image decomposition mechanisms used in these codes, i.e., the 2D-DCT transform in JPEG and the wavelet transform used in JPEG2000 are analytically-constructed transforms not learned from the data and believed to provide sparse and compressible representations on any image in general. Moreover, what if the images that we deal with have some particular structure typical for images originating from the same source? How can we benefit from this extra redundancy in the data, e.g., for satellite images and when we know that all of them look similar in some way?

This is achievable if we learn those particularities and structures directly from the data. This is where the machine learning algorithms can potentially become useful.
\subsection*{ii. Joint optimization of the compression pipeline:} 
Image compression is a very detailed task, consisting of multiple stages. This goes beyond learning a good transform or a sparsifying dictionary and consists of stages like discretization and entropy coding. While traditional methods solve these steps independently, perhaps the presence of training data may facilitate a joint optimization of the entire pipeline.

\subsection*{iii. Vector Quantization is superior to Scalar Quantization:}
Independent from the particularities within the data for which statistical learning should help with better representations, within a certain rate-regime, Vector Quantization (VQ) should have, at least in theory and for very high dimensions and low compression rates, a superior rate-distortion behavior as compared to the Scalar Quantization (SQ) techniques. This is a fundamental concept and is the important motivation behind the study of rate-distortion theory. 

To see this, let us take the extreme case that the data is \textit{i.i.d.} Gaussian, i.e., $F \sim \mathcal{N}(0,1)$. This means that there is no pattern or particularity to learn from this data. The best SQ for such a data at $\mathcal{R} = 1$ bits corresponds to a binary quantizer whose non-zero quantization bins are at $\pm \sqrt{\frac{2}{\pi}}$ (according to Eq. \ref{eq:SingleLayer_STC_beta*-per-dim}). This quantizer achieves an average distortion around $\mathcal{D} \simeq 0.36$ (according to Eq. \ref{eq:SingleLayer_STC_Dist-per-dim}).

If we consider the joint description of a large number of such data, i.e., $\mathbf{F} \sim \mathcal{N}(\mathbf{0},\mathrm{I}_n)$, however, the rate-distortion theory promises a much more favorable performance, i.e., $\mathcal{D} = 2^{-2 \times 1} = 0.25$, for the same amount of rate (which is achieved in asymptotics). This, in fact, is related to the geometry of the $\Re^n$ space, rather than statistical properties (see e.g., \cite{CoverThomas200607}).

Image compression codecs, however, use SQ in the transform domain. Learning-based algorithms, on the other hand, may be able to benefit from such joint descriptions.

\subsection*{iv. Compression as unsupervised representation learning:}
Beyond the task of image compression itself, another motivation for learning to compress is for the more general task of representation learning. During the last couple of years, a consensus has been reached within the machine learning communities that unsupervised learning is promising, and it has been shown multiple times that it can be useful for the task of supervised learning (e.g., see \cite{UnsupLearning:35536}). An important instance of unsupervised learning, in fact, is the task of compression which is closely connected or in a sense equivalent to probabilistic distribution estimation, which is the core problem in machine learning.

To see this, consider the general objective function for rate-distortion optimization, which minimizes the rate of the (latent) representation $\mathbb{Q}[\mathbf{F}]$, subject to a constraint on the distortion, i.e., $\mathcal{D}( \mathbf{F}, \mathbb{Q}^{-1}[\mathbb{Q}[\mathbf{F}]]) \leqslant D$. The rate minimization, however, is equivalent to entropy maximization, i.e., :
\begin{equation} \label{ImCompression:GeneralObj}
\begin{aligned}
&\underset{{\mathbb{Q}[\cdot], \mathbb{Q}^{-1}[\cdot]}}{\text{maximize }}  \log \big( p(\mathbf{\mathbb{Q}[\mathbf{f}]}) \big), \\
&\text{   s.t. } \mathcal{D} \leqslant D.
\end{aligned}
\end{equation}

So instead of maximizing the log-probability of the data, i.e., $\log \big( p(\mathbf{f}) \big)$, by learning to compress, we maximize a function of it, i.e., $\log \big( p(\mathbf{\mathbb{Q}[\mathbf{f}]}) \big)$. This, however, should not deviate from the original data, to the extent guaranteed by the data-fidelity constraint $\mathcal{D}$.

To see an example of this, take the case of generative modeling whose aim is to maximize $\mathbb{E} \{ \log p(f) \} = -H(\mathbf{F})$. This can be decomposed, in particular through the framework of Variational Auto Encoders (VAE's) of \cite{VAE:KingmaW13} as:\footnote{Notice that in our terminology, $\mathbf{f}$ is the data and $\mathbf{x}$ is the latent representation.}
\begin{subequations}
\begin{align*}
\log p(f) &= \underbrace{D_{KL}(q(x) || p(x|f))}_{\geqslant 0, \rightarrow 0} + \underset{ x \sim q} {\mathbb{E}} \{ \log p(f|x) \} - D_{KL}(q(x) || p(x))  \\
\log p(f) &\leqslant \underbrace{\underset{ x \sim q} {\mathbb{E}} \{ \log p(f|x) \}}_{-d(F,\hat{F})} - \underbrace{D_{KL}(q(x) || p(x))}_{\text{extra bits due to latent encoding}} \\
\end{align*}
\end{subequations}

Notice that the maximization of $\log p(f)$ in this formulation is conceptually similar to the general rate-distortion optimization, e.g., our Eq. \ref{eq:MultiLayer_STNets_GeneralLoss} and Eq. \ref{eq:MultiLayer_STNets_Losses}.\footnote{A similar argument has been provided in \cite{IMComp:Balle17a}.}

This, along with other similar examples, is an important motivation to study compression as representation learning.
\subsection{Autoencoders for compression: recent literature} \label{subsec:ImCompression_Why_SOTA}
Only very recently, the task of learning-based lossy image compression has appeared as an active area in machine learning \cite{IMComp:Toderici2015VariableRI,ImComp:toderici2017full,IMComp:Balle17a,IMCom:TheisSCH17,IMComp:pmlr-v70-rippel17a}. These methods are essentially, different variations of the autoencoder neural structures. This perhaps comes after \cite{hinton2006reducing} showed that it is possible to reduce the dimensionality of representations using autoencoders. However, there are two fundamental issues with autoencoders. 

First, they are essentially built to target one single operational rate, while a practical compression scheme requires multiple such rates, starting from very low to potentially being lossless. For example, \cite{IMCom:TheisSCH17,IMComp:Balle17a,IMComp:pmlr-v70-rippel17a} are methods that are designed to operate only at one rate. So one network is trained for one quantization step corresponding to a certain rate. An important question then is whether one can learn one single network for all layers and only change the step size at test time; a question that has recently been answered positively in \cite{ImComp:8462263}.

The second issue with applying autoencoders for compression is that dimensionality reduction is not exactly equivalent to compression. While the former tries to achieve useful representations in bottlenecked dimensions, the latter further requires that the representations be discretized and whose size be measurable in bits. Applying a discretizing function within a neural network pipeline, however, makes the flow of back-propagation to stop, since the differential of such functions are zero, a phenomenon we have mentioned earlier in section \ref{subsec:MultiLayer_STNets_Network}, and for which we proposed our solution in section \ref{subsubsec:MultiLayer_STNets_Network_Ternarization}.

A lot of methods, e.g., \cite{IMComp:Balle17a,ImComp:liu2018deep} simply remove the quantizer during the training phase and put it back in place during test time. Other methods use the soft approximation of the quantization function, e.g., tanh in \cite{ImComp:toderici2017full} and might use a stochastic regularization to minimize the approximation. Inspired by dithering in quantization literature, \cite{IMComp:Balle17a} approximates quantization as an additive noise. Another idea is to perform a kind of annealing during optimization iterations, e.g., as in \cite{ImComp:agustsson2017soft}, that starting from a soft alternative, gradually approach a hard quantizer.

Another important issue is with the resolution of the images being trained. In fact, a lot of these solutions can compress only tiny images, e.g., \cite{ImComp:TodericiOHVMBCS15} works only on $32 \times 32$ images. Increasing the image size, in practice, is done using patch-wise division of the images which neglects the global structure of the images.


\section{Compression of natural images using ML-STC} \label{sec:ImCompression_STNets}

Image compression is a very tailored pipeline requiring a great deal of engineering work. As an example, the lossy JPEG2000 codec \cite{iso2001information}, first converts the RGB channels to the YCbCr representation. The Cb and the Cr channels are very redundant and hence can be down-sampled without any noticeable quality loss. These channels are then transformed to the wavelet domain, scalar-quantized carefully and entropy coded with a very detailed and multi-staged mechanism, taking into account bit-planes of representations and their statistical properties.

In this chapter, we are not going into any of these details. Instead, we take raw images and apply our algorithms we developed earlier in chapter \ref{chapter:MultiLayer}. This, however, requires us to perform a whitening step on images as we discuss next.

\subsection{A simple image whitening technique} \label{subsec:ImCompression_STNets_Whitening}
Our algorithms require the input data to be somehow whitened, or at least de-correlated to some extent. For the synthetic sources like the AR(1) that we have experimented with, or for low dimensional image thumbnails like the MNIST or CIFAR databases, this could have easily been achieved using the simple PCA, as we did.

For high-resolution images, to maintain the global structure of the image, as we will elaborate more in section \ref{subsec:ImCompression_STNets_Patching}, the preference is to avoid using patches, or at least use very large patches. Applying PCA on high-resolution images, however, is hopeless. First, the complexity of PCA is in $\mathcal{O}(n^3)$ and it becomes exceedingly difficult to apply PCA on large images. More importantly, the estimation of covariance matrix has $\frac{n^2+n}{2}$ free parameters, which requires a significant number of images to be estimated appropriately.

So what do we do to capture the global structure of the images and whiten them with reasonable complexity? Here we propose a very basic, but fast and effective way as the following.

\begin{figure}   
\begin{center}
\includegraphics[width=0.5\textwidth]{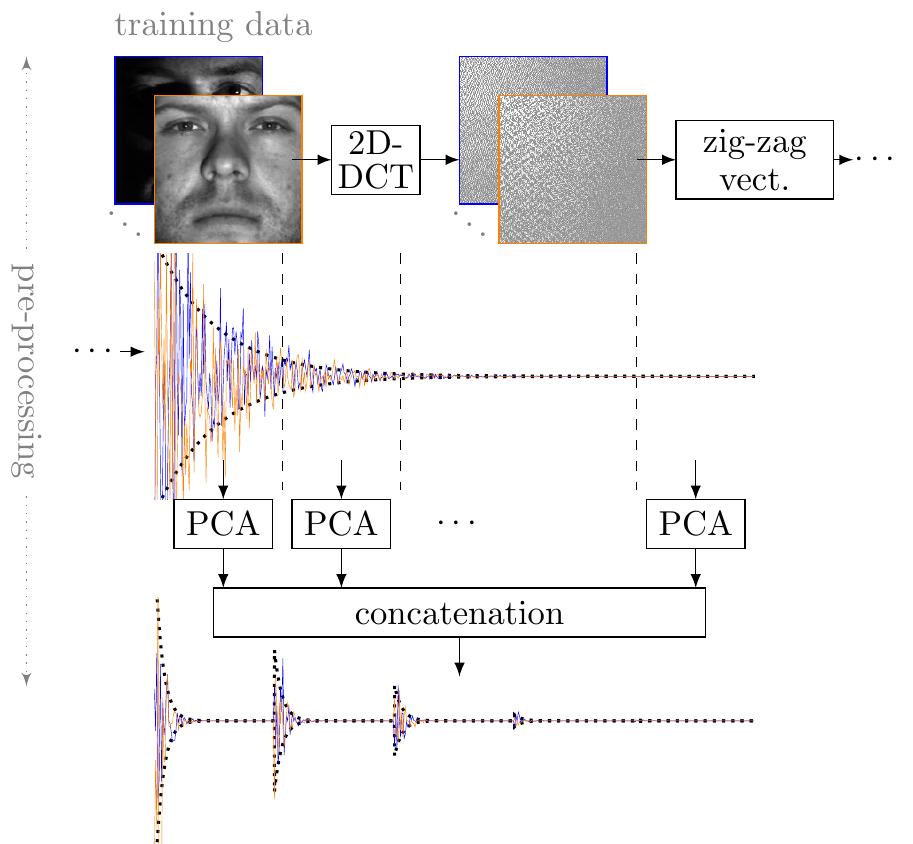}
 \caption{Image whitening using the 2D-DCT and the PCA transform.}
 \label{fig:ImCompression_RRQ_PreProcessing}
\end{center}
\end{figure}

We first apply the 2D-DCT transform on the global image (preferably without dividing it into patches). The cost of this step is at $\mathcal{O}(n \log{n})$, hence very fast and scales well with dimension. We then vectorize this 2D matrix (or 3D tensor if we consider images of the training set) using zig-zag scanning. While the 2D-DCT has some decorrelating properties for natural images, it is data independent and leaves some correlation in the transform domain. We capture (most of) this correlation using the PCA and using the given training set. However, instead of the full-frame data, assuming that the correlations after the 2D-DCT happen mostly at adjacent dimensions, we apply the PCA multiple times and in sub-band divisions of the whole frequency range. This is equivalent to assuming that only similar frequencies are correlated and farther frequency bands can be assumed to be orthogonal to each other. Fig. \ref{fig:ImCompression_RRQ_PreProcessing} sketches this pre-processing stage. 

Assuming we have $p$ such frequency sub-bands, the complexity of this procedure is $\mathcal{O} \big( n \log{n} + p \times (\frac{n}{p}) ^3 \big) = \mathcal{O}(\frac{n^3}{p^2})$, much smaller than the original $\mathcal{O}(n^3)$ of the PCA ($\frac{n}{p}$ is chosen almost constant in practice). More importantly, the covariance estimation which originally has $\frac{n^2 + n}{2}$ parameters is now estimated with only $\frac{n^2 + np}{2p}$ parameters, reducing the chance of over-fitting.

We can interpret this procedure as a sort of regularized PCA estimation where (similar to PQ coding to K-means), instead of the full covariance matrix, the elements of $\frac{n}{p}$ sub-matrices are estimated only and the rest is set to zero. This is possible only after the 2D-DCT, which already decorelates the image to some extent.

\subsection{Compression of high-resolution natural images} \label{subsec:ImCompression_STNets_Exps}
Here we perform a simple experiment on a database\footnote{This database can be downloaded from \href{http://www.compression.cc/challenge/}{http://www.compression.cc/challenge/. We do not perform validation for hyper-parameter selection of our algorithm and take the validation set with 41 images, instead of the larger test set.} We use only 200 out of around 500 images in the training set.} of RGB high-resolution images (around 2.5 megapixels) with varying sizes. This is very challenging since from one hand, we want to keep the global structure and hence choose larger patches. From the other hand, the redundancy introduced because of zero-padding is more obvious for larger patches and hence the coding efficiency is less. Another side of the story is the trade-off between the patch-size and the availability of enough training samples. 

We take 200 images from the train set, divide the images into $64 \times 64 \times 3$ patches (without overlap, but with zero-padding) and perform the whitening described in section \ref{subsec:ImCompression_STNets_Whitening} with $p = 192$ sub-bands. This makes a total of $N = 118,258$ training samples with dimension $n= 64 \times 64 \times 3 = 12,288$. 

Since this data is highly variance-decaying (a ratio of around $1.5e+9$ between the largest and smallest values), it makes sense to perform rate-allocation before projections. So at each stage, before learning the projections, we perform a similar rate-allocation as in Algorithm \ref{alg:SingleLayer_WFiller}. This way, we considerably economize on the size of the projections. So instead of $n = 12,288$, on the average across all layers, the effective dimensionality is around $n'= 1,500$. 

We apply the ML-STC-Procrustean algorithm of section \ref{subsec:MultiLayer_MLSTC-Procrustean_Alg} to this data. We had $L=20$ layers with ternary codes of $k = 20$ non-zeros in each layer. 

\begin{figure}   
   \begin{center} 
\includegraphics[width=0.99\textwidth]{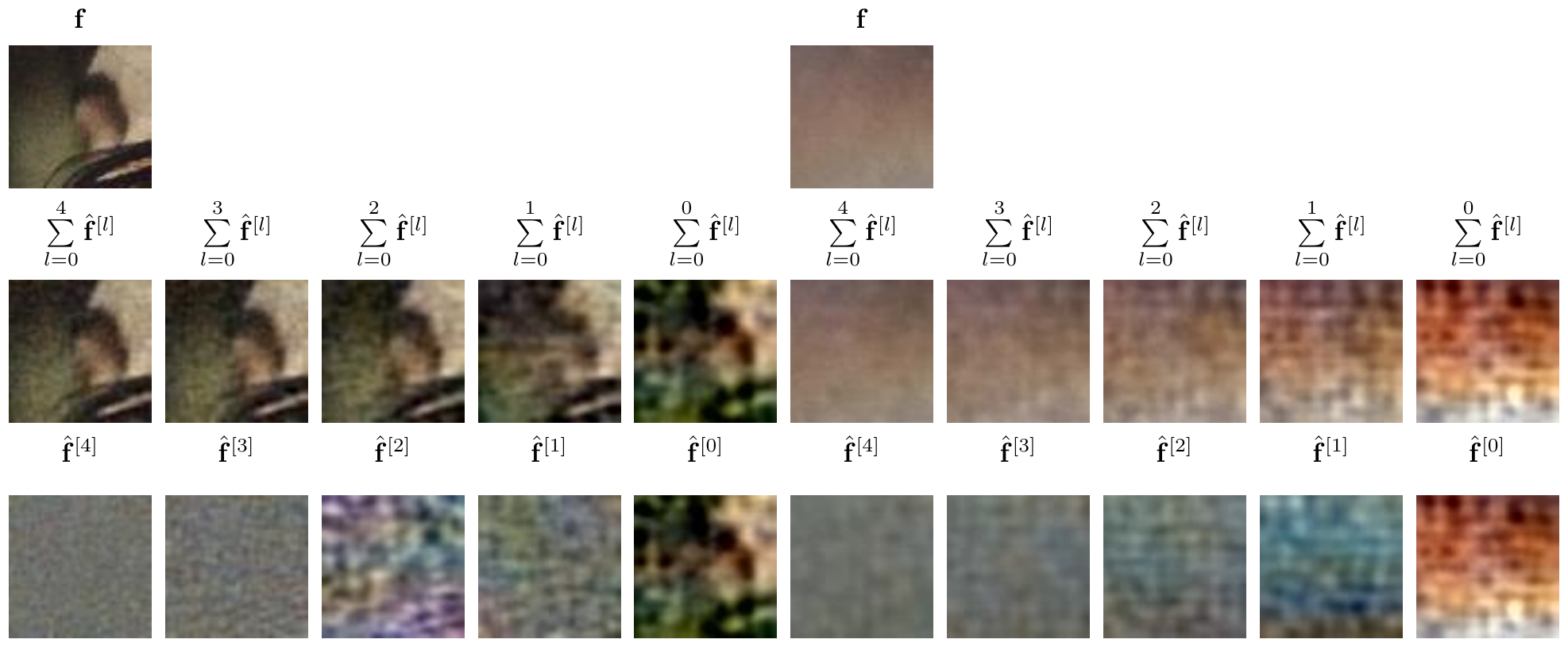} 
   \end{center}
   \caption{Successive approximation of matches at 4 different levels. Two different patch samples at left and right. The lower row is the $\hat{\mathbf{f}}^{[l]}$'s for $1 \leqslant l \leqslant 4$. The middle row is the cumulative summation of $\hat{\mathbf{f}}^{[l]}$'s and the first row is the input patch. }
   \label{fig:ImCompression_STNets_PatchesSR}
   \end{figure}

To see a better picture of our residual structure and how different layers contribute to the successive approximation of patches, Fig. \ref{fig:ImCompression_STNets_PatchesSR} shows two sample patches that get successively approximated in 4 layers. It is clear that most of the correlated structures happen in the initial layers. As more layers are incorporated, the data becomes less structured.

\begin{figure}  
   \begin{center} 
\subcaptionbox{Original area} {\includegraphics[width=0.2\textwidth]{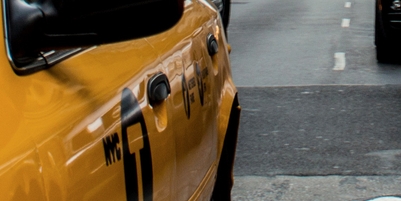}} 
\subcaptionbox{Ours} {\includegraphics[width=0.2\textwidth]{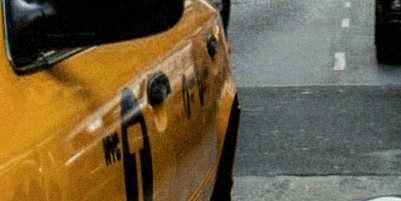}} 
\subcaptionbox{JPEG2000} {\includegraphics[width=0.2\textwidth]{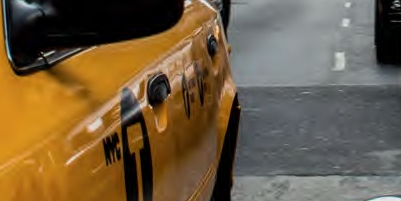}} 
\subcaptionbox{JPEG} {\includegraphics[width=0.2\textwidth]{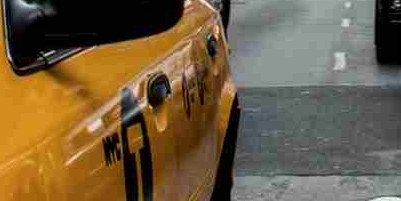}} 

\subcaptionbox{Original area} {\includegraphics[width=0.2\textwidth]{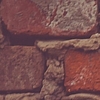}} 
\subcaptionbox{Ours} {\includegraphics[width=0.2\textwidth]{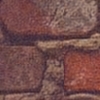}} 
\subcaptionbox{JPEG2000} {\includegraphics[width=0.2\textwidth]{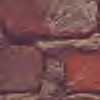}} 
\subcaptionbox{JPEG} {\includegraphics[width=0.2\textwidth]{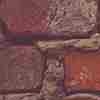}}

\end{center}
\vspace{-0.5cm}    
   \caption{Compression of high-resolution images with ML-STC-Procrustean. }
   \label{fig:ImCompression_STNets_Samples}
   \end{figure}

Fig. \ref{fig:ImCompression_STNets_Samples} shows two zoomed areas of compressed images under our algorithm, as compared with JPEG and JPEG2000.


In terms of compression artifacts, our algorithm is patch-based like JPEG. However, since we use $64 \times 64$ patches instead of $8 \times 8$ in JPEG, the blocking artifact is considerably reduced. JPEG2000 produces its own artifacts, which are due to its underlying wavelet decomposition. Our algorithm, overall, seems to be more successful in compressing textures, rather than edges. This, perhaps is due to the simple pre-processing that we use that can be improved in many ways.

\begin{figure}  
   \begin{center} 
\includegraphics[width=0.6\textwidth]{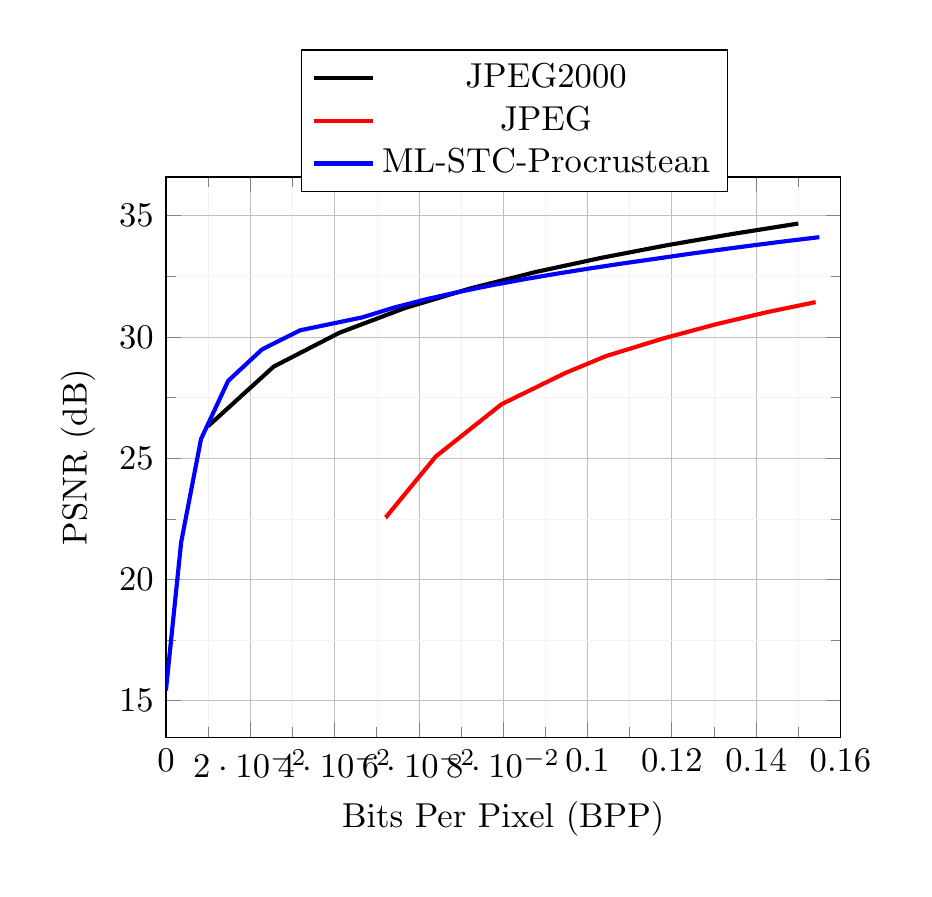}
\end{center}
\vspace{-0.5cm}    
   \caption{PSNR vs. BPP for compression of high-resolution color images.}
   \label{fig:ImCompression_STNets_PSNR}
   \end{figure}

Finally, in terms of MSE, Fig. \ref{fig:ImCompression_STNets_PSNR} compares the results of the compression of these images under the JPEG, JPEG2000 and our ML-STC-Procrustean.

Our results largely outperform JPEG but outperforms JPEG2000 only in rates smaller than $BPP < 0.08$. It should, however, be mentioned that, in this experiment, we ignored two crucial steps in image compression pipelines:
 
First, we do not perform any entropy coding. When implemented, we expect to gain considerable bit saving in this respect. Second, we did not perform chrominance sub-sampling, which halves the dimension of the images without any considerable performance loss. 

Apart from these steps, this experiment can be improved in many ways:

First, due to limitations in computational resources, we only used 200 images and without overlapping the patches. In case more training data is incorporated, we can perform back-propagation to fine tune these results, as was done in section \ref{subsec:MultiLayer_STNets_TrainingStrategies} and Fig. \ref{fig:MultiLayer_STNets_MNIST_DR}. 

Second, after the DCT stage, apart from the projection matrices (and bias terms), also the whitening sub-band matrices can also be tuned jointly with other parameters.

Third, the rate-allocation can be improved as well. In order to reduce complexity, we chose a very harsh regularization for rate-allocation and without any (cross-) validation. This can as well be learned from the data.

Finally, and perhaps more importantly, the current experiment was based on patch-wise dividing of the images. This is extremely inefficient, as we will show next in section \ref{subsec:ImCompression_STNets_Patching}, and can be avoided using pyramidal decompositions for images.

It would be useful to see how our compression pipeline compares with other learning-based methods from the deep learning literature. As far as the learning algorithm itself is concerned, this, however, is not straightforward to achieve, since the reported results, other than the learning algorithm, are due to the entire pipeline which consists of many other blocks that cannot be fixed due to different engineering parameters concerned.

However, generally speaking, and by comparing different results from the recent literature, it can be concluded that for compression of high-resolution RGB images, and as far as PSNR is concerned (rather than perceptual measures), most algorithms fall below the performance JPEG2000, sometimes by large margins. In fact, only a few methods get similar or slightly better performance than JPEG2000.\footnote{As another indicator, refer to \href{http://www.compression.cc/results/?sort=psnr}{http://www.compression.cc/results/?sort=psnr} for a comparison of different algorithms on the same database. The results are reported at around BPP $ = 0.05$, with the best PSNR achieved as $30.89$ dB. Our result from Fig. \ref{fig:ImCompression_STNets_PSNR} show very similar performance at that rate.}

Moreover, our training procedure (with all its 20 operational points) took overall less than one hour to terminate, while using 6 threads of CPU from an average desktop. Deep-learning based methods from the literature typically require several days of training on highly parallelized GPUs and to produce only 1 operational point.
%

%
To motivate future work to abandon patch-wise processing of images, let us next quantify this effect with a simple experiment.
\subsection{Future work: to avoid image patching} \label{subsec:ImCompression_STNets_Patching}
A typical high-resolution image may have around several million triple-channel pixels. For an average computational resource, however, this million-scale dimensionality is impossible to handle directly, for many training algorithms. Moreover, this dimensionality requires billion-scale sample sizes to be able to perform learning without over-fitting.

One very common remedy is to divide images into rectangular boxes known as patches. This, however, eliminates the global structure of the image, as each vectorized patch is then considered an independent data-point in the space of $\Re^{\frac{n}{p}}$, where $n$ is the original image size and $p$ is the number of patches.

Here we try to measure this amount of information loss on the overall quality of image coding. For this, we contrast the JPEG with the JPEG2000 coding schemes in the following way: 

JPEG first divides the images into $8 \times 8$ patches, performs the compression on each one of them and finally tries to somehow compensate the loss of global information by taking the spatial position of the patches into account through an entropy coding. JPEG2000, on the other hand, applies the transform on the full-frame image and then performs the coding.

We perform the following experiments: We first compress full images using the JPEG and the JPEG2000 coding schemes and measure their compression performance in PSNR. 

\begin{figure}  [!h]
   \begin{center} 
\subcaptionbox{Original image} {\includegraphics[width=0.3\textwidth]{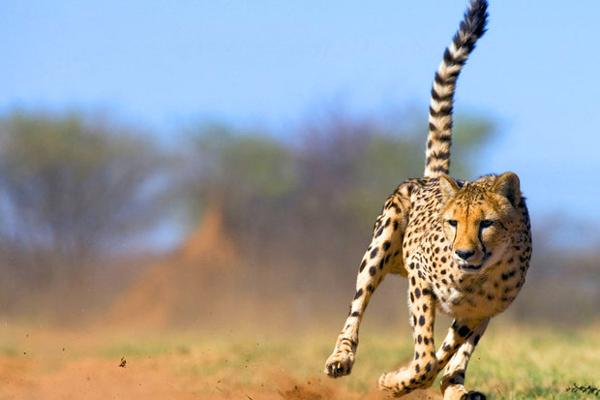}} 
\subcaptionbox{Original- \\JPEG2000 compressed} {\includegraphics[width=0.3\textwidth]{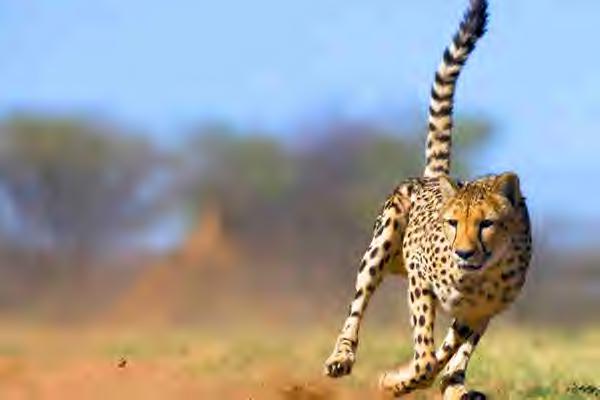}} 
\subcaptionbox{Original- \\JPEG compressed} {\includegraphics[width=0.3\textwidth]{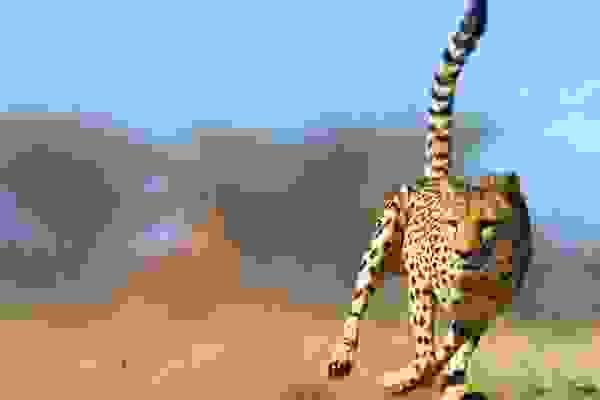}} 

\subcaptionbox{Shuffled image} {\includegraphics[width=0.3\textwidth]{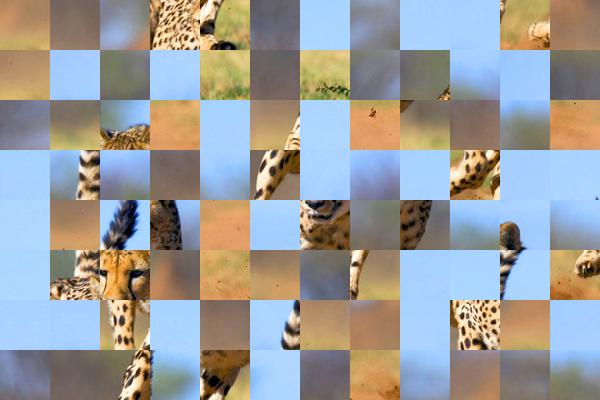}} 
\subcaptionbox{Shuffled- \\JPEG2000 compressed} {\includegraphics[width=0.3\textwidth]{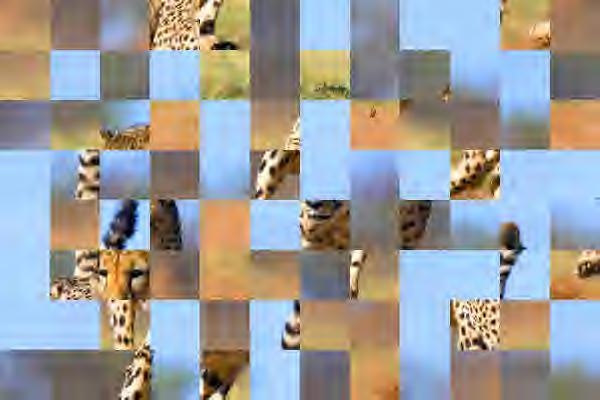}} 
\subcaptionbox{Shuffled- \\JPEG compressed} {\includegraphics[width=0.3\textwidth]{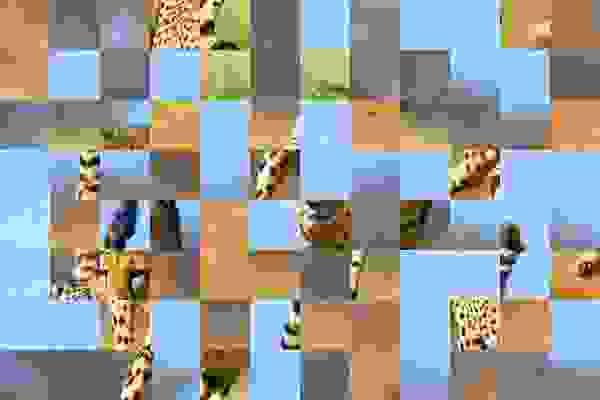}}

\end{center}
\vspace{-0.5cm}    
   \caption{The importance of preservation of spatial structure of images for compression. A lot of the advantage of JPEG2000 over JPEG is thanks to its global encoding (rather than the basis vectors).}
   \label{fig:ImCompression_STNets_PatchShuffle}
   \end{figure}

\begin{figure}   [!h]
   \begin{center} 
\includegraphics[width=1\textwidth]{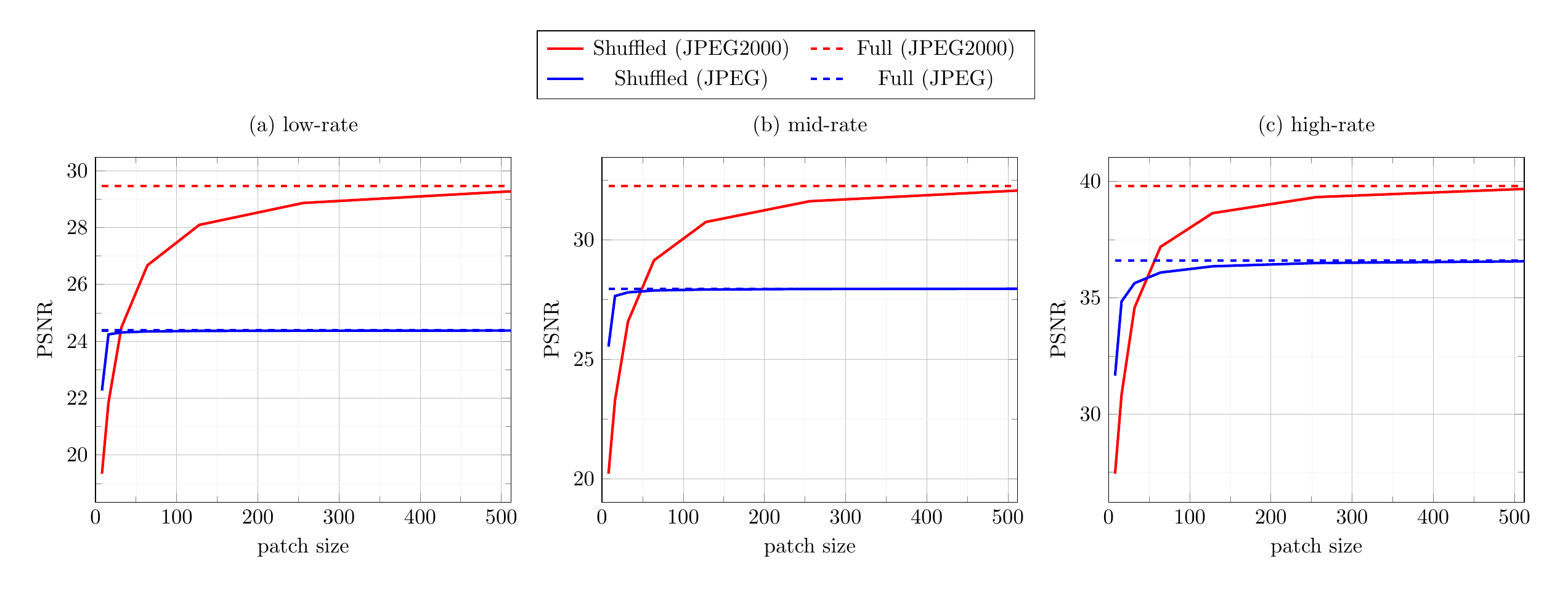}   
   \end{center}
\vspace{-0.5cm}    
   \caption{Effect of patch-based division of images on compression quality w.r.t. patch size for (a) low-rate (JPEG QF $=5$), (b) mid-rate (JPEG QF $=10$) and (c) high-rate (JPEG QF $=70$) compression. The results averaged over 10 randomly selected $1024 \times 1024$ images.}
   \label{fig:ImCompression_STNets_PatchSize}
   \end{figure}

Next, we divide images into patches of varying size, then randomly shuffle them and compress them with JPEG and JPEG2000. Fig. \ref{fig:ImCompression_STNets_PatchShuffle} visualizes this idea and Fig. \ref{fig:ImCompression_STNets_PatchSize} demonstrates the results of these experiments for 3 different compression ratios. 

JPEG2000 compresses the global image and hence its performance significantly degrades when compressing randomly shuffled images. JPEG, on the other hand, fails to capture the global information and hence its performance is almost intact for randomly shuffled images.

The result of this experiment reveals an important fact about the effect of dividing images into patches with various sizes: That it does significantly (at least for common patch sizes) degrade the performance of image coding, and perhaps many other tasks. This is in contrast to the common understanding that the superiority of JPEG2000 w.r.t. JPEG is due mostly to the fact that DWT is a better transform for natural images than DCT. 

While this might sound an obvious observation, it is at least a highly neglected one. Notice that in convolutional networks, in general, the overlap in patches helps to keep more of the global structure. When it comes to truly high-resolution images, however, due to computational bottlenecks like the limit in memory of GPUs, this is avoided, and images are still divided into patches. 

Therefore, other alternatives should be found that avoid image patching. One such possibility is the different pyramidal decompositions for images. 

While we did not experiment with pyramids, we keep this idea as a future objective to follow.


\section{Compression of facial images using RRQ} \label{sec:ImCompression_RRQ}
Let us now pick a different scenario than natural images and investigate whether learning to compress is possible when there is more structured similarity between images than the general class of natural images. This can be the case for domain-specific images with lots of applications, e.g., in astronomical, biomedical or satellite images, where huge collections of such images are to be stored, transmitted and processed.  

Here we take the particular example of facial images. So suppose we are given a collection of face photographs, and we want to benefit from the extra redundancy present in this collection to further compress them.

Several works address this problem from within the image processing community. In particular, \cite{Facial:4286990} was an early attempt based on VQ. \cite{Facial:Bryt2008270} learns the dictionaries based on the K-SVD \cite{1710377} while \cite{Facial:6844846} uses a tree-based wavelet transform. A codec is proposed in \cite{Facial:5740941} by using the Iteration Tuned and Aligned Dictionary (ITAD). In spite of their high compression performance, the problem with most of these approaches is that they rely very much on the alignment of images and they are less likely to generalize once the imaging setup is changed a bit. Some of them require the detection of facial features (sometimes manually) and then alignment by geometrical transformation into some canonical form and also a background removal stage. 

We target this problem using the RRQ framework of section \ref{subsec:MultiLayer_RRQ_RRQ}. Next in section \ref{subsec:ImCompression_RRQ_Exp}, we target the compression task. Later in section \ref{subsec:ImCompression_RRQ_Denoising}, we go one step further and investigate whether capturing this redundancy using compression can be beneficial for the task of image denoising.

\subsection{Compression experiments} \label{subsec:ImCompression_RRQ_Exp}
Here we perform a simple compression experiment on facial images. We use the \textit{CroppedYale-B} database \cite{CYale:GeBeKr01} which contains $2414$ images of size $192 \times 168$ from 38 subjects. Each subject has between 57 to 64 acquisitions with extreme illumination changes. We choose half of the images for each subject randomly for training and the rest for testing. 

We first perform whitening using the procedure described in section \ref{subsec:ImCompression_STNets_Whitening}, and by choosing $p=96$ sub-bands and without image patching in order to maintain the global structure of the images. This results in two equal-sized matrices of $1207$ items with dimensionality $32256 = 192 \times 168$, as train and test sets. After the whitening, these matrices have a very sharp variance-decaying profile. 

Encoding such high-dimensional data with such little amount of available training data is impossible without the careful regularization of the RRQ.\footnote{From the experiments of Fig. \ref{fig:MultiLayer_STNets_MNIST_DR}, it is very much expected that neural architectures with random initializations (without pre-training) will quickly over-fit in this setup. Otherwise, they have to divide images into very small patches to get a bigger set. This, however, is eliminating the global structure as we have shown in Fig. \ref{fig:ImCompression_STNets_PatchSize}.} 

\begin{figure}  [!h]
   \begin{center} 
\subcaptionbox{D-R curve (normalized, log-scale)\label{subfig:ImCompression_RRQ_CYaleCompression_DR}} {\includegraphics[width=0.4\textwidth]{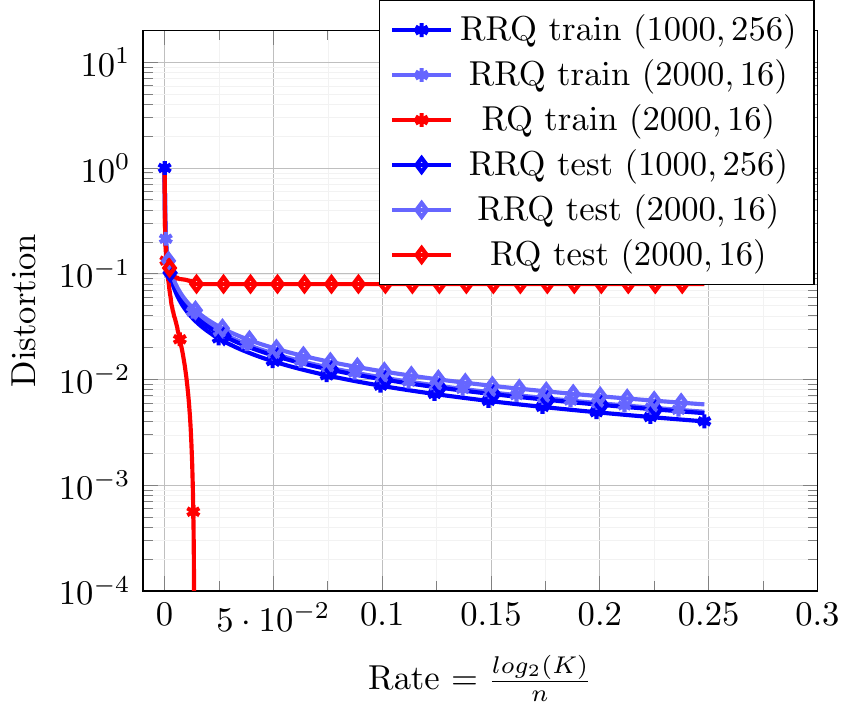}} 
\subcaptionbox{Image compression \label{subfig:ImCompression_RRQ_CYaleCompression_PSNR}} {\includegraphics[width=0.4\textwidth]{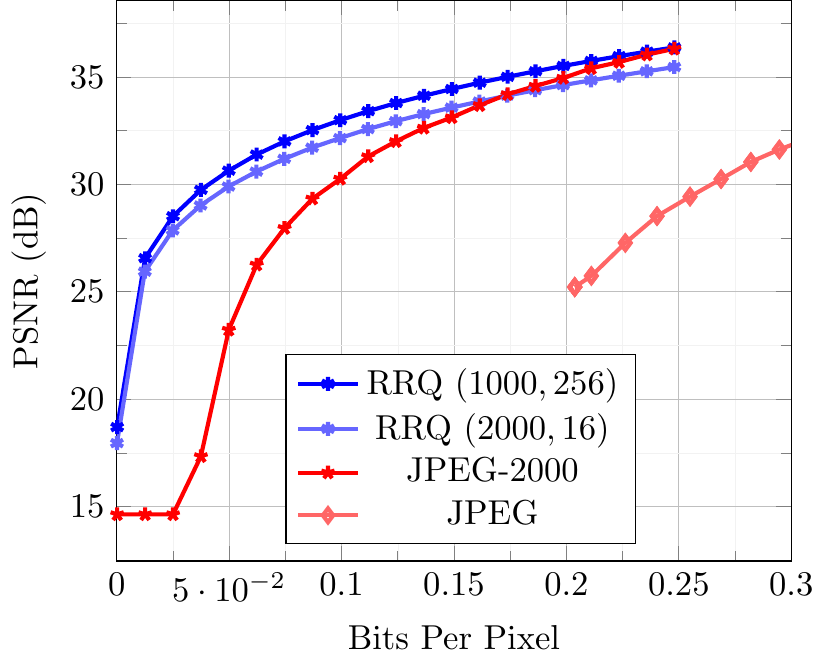}}
\end{center}
\vspace{-0.5cm}    
   \caption{Compression of the \textit{CroppedYale-B} set of facial images: (a) Distortion-rate performance of RQ and RRQ on the train and test sets and for different $(L,m)$ pairs. (b) Comparisson of PSNR vs. bpp for RRQ, JPEG and JPEG2000. Results averaged over 20 randomly chosen images from the test set.}
   \label{fig:ImCompression_RRQ_CYaleCompression}
   \end{figure}

Fig. \ref{subfig:ImCompression_RRQ_CYaleCompression_DR} shows the distortion-rate curves of RRQ in comparison with the RQ which quickly over-trains at such training regime. Fig. \ref{subfig:ImCompression_RRQ_CYaleCompression_PSNR} compares the RRQ with the JPEG and JPEG2000 codes and in terms of PSNR w.r.t. bits per pixel.

The RQ and RRQ were both experimented on two value-pairs of $L$, the number of layers and $m$, the number of codewords per layer. It should be noted that we did not perform any entropy coding over the codes. Further compression improvement can be achieved by entropy coding over the tree-like structure of the codebooks.
\subsection{Compression for denoising} \label{subsec:ImCompression_RRQ_Denoising}
Having learned useful structures from a collection of images for compression, can we go one step further and use these structure to perform other tasks, e.g., image denoising?

While this was not the main focus of this thesis, we elaborate more on this question in the next chapter and leave further research as an exciting direction for future research.

In this section, however, without proposing an algorithm, we perform a simple experiment to validate this idea. 

So suppose we have trained our algorithm for compression of facial images, e.g., the RRQ and as we did in section \ref{subsec:ImCompression_RRQ_Exp}. In the test time, however, the images happen to be contaminated with noise, perhaps since the acquisition systems are different.

What happens if we feed the noisy images to the compression network that is trained with clean data? It turns out that the network reconstructs the input image with the knowledge from clean images and hence denoises them to some extent, as well.

\begin{figure}
 \hspace{0.5cm}
\begin{minipage}[c]{0.58\textwidth}
\captionsetup[subfigure]{labelformat=empty}
\centering
  \subcaptionbox{} {\includegraphics[width=0.22\textwidth]{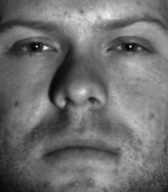}} 
    \subcaptionbox{ $\sigma_P^2 = 0.3$ } {\includegraphics[width=0.22\textwidth]{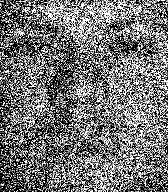}}
    \subcaptionbox{$23.49$ dB} {\includegraphics[width=0.22\textwidth]{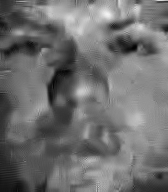}}
   \subcaptionbox{$24.73$ dB} {\includegraphics[width=0.22\textwidth]{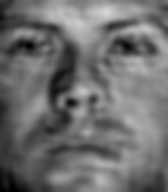}}  

      \subcaptionbox{} {\includegraphics[width=0.22\textwidth]{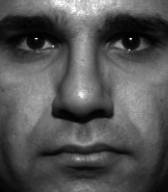}} 
    \subcaptionbox{ $\sigma_P^2 = 0.15$ } {\includegraphics[width=0.22\textwidth]{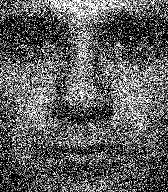}}
    \subcaptionbox{$25.82$ dB} {\includegraphics[width=0.22\textwidth]{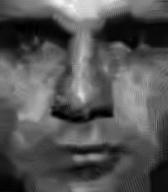}}
   \subcaptionbox{$25.77$ dB} {\includegraphics[width=0.22\textwidth]{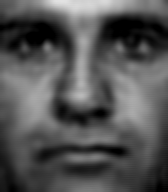}}
   \caption{Samples of image denoising. Order of columns: original image, noisy (noise variance), BM3D (PSNR) and RRQ (PSNR).}
    \label{fig:ImCompression_RRQ_DenoisingSamples}
 \end{minipage}
  \begin{minipage}[c]{0.4\textwidth}
      \includegraphics[width=0.99\textwidth]{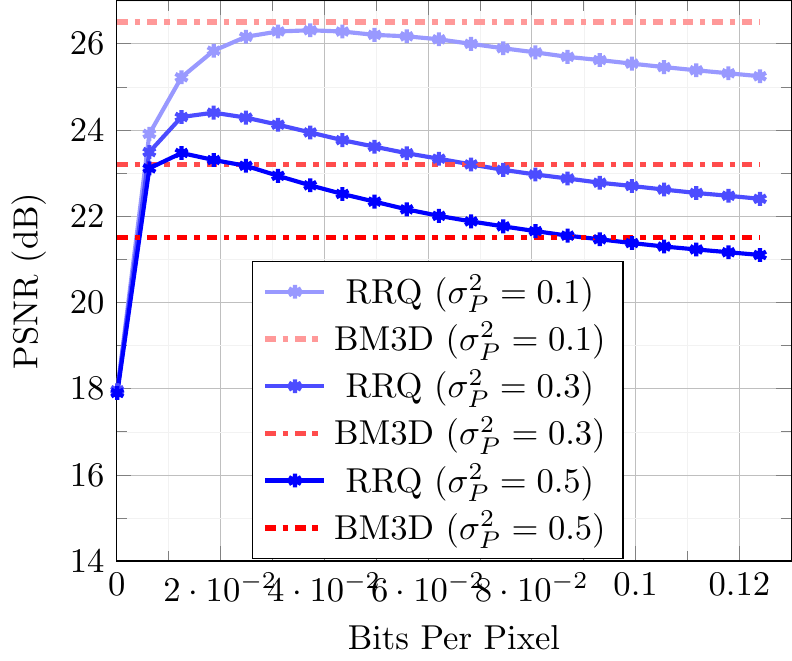}
        \caption{The performance of denoising for different levels of reconstruction}
        \label{fig:ImCompression_RRQ_DenoisingCurves}
 \end{minipage}
 \end{figure}

This experiment is demonstrated in Fig. \ref{fig:ImCompression_RRQ_DenoisingSamples}, where samples of clean noisy and denoised images are shown.\footnote{Regarding the noise variances, note that the pixel values are normalized within $[0,1]$, and not $[0,255]$.} Surprisingly, at the very highly noisy regimes, this idea turns out to be more effective than the BM3D \cite{BM3D:dabov2007image}, which is by far one of the most successful denoising algorithms to date. 

As can be seen from the samples, although BM3D produces very smooth images, it degrades the details of faces since it does not have the prior of clean facial images.  Our compression network, on the other hand, injects these priors effectively while constructing the noisy images. Therefore, the borders of faces are well preserved.


Fig. \ref{fig:ImCompression_RRQ_DenoisingCurves} shows the average denoising performance of our network for 20 randomly selected images from the test set, under 3 different noise levels, and across all layers. This is again compared with the BM3D, showing improvements at very noisy regimes.

An interesting phenomenon is observed during denoising and across different rates. In fact, as more layers are invoked in reconstruction, the denoising performance is increased, but up to a critical point after which the performance starts to degrade. Moreover, this critical point is happening in lower rates for more noisy inputs. So after this critical point, the decoder tries to reconstruct noise more than the data. This phenomenon, however, is expected according to Shannon's source-channel coding separation theorem \cite{shannon2001mathematical}, which states that for a source with distortion-rate function $D(R)$ to be transmitted through a channel with capacity $C$, one can achieve the distortions higher than the distortion-rate function, only if the rate is lower than the capacity. In other words, the lower bound of distortion saturates at $D \geqslant D(R)\big\rvert_{R=C}$. In our case, this critical point is the rate whose corresponding distortion equals the noise variance.

\section{Conclusions} \label{sec:ImCompression_Conclusions}
We first argued that learning to compress images is useful and promising. This concerns imroving the performance of lossy compression in general and in domain-specific cases, as well as for representation learning. 

We then targeted the very challenging task of lossy compression of natural RGB images in high resolutions. After proposing a simple image whitening procedure, we chose our ML-STC-Procrustean algorithm and trained it on 200 high-resolution images. We showed that even without applying several important elements of a compression pipeline, e.g., entropy coding, bit-plane coding or chrominance sub-sampling, our algorithm achieves comparable PSNR performance to the JPEG2000 and much better compression ratio compared to the JPEG. 

We then studied the effect of block-wise dividing the images into patches and measured how it can affect the performance of a compression algorithm. We concluded that this is crucial and can perhaps be more important than finding a better data-adaptive transform. Therefore, as an important future direction, we address this issue by avoiding any block-wise division, perhaps using pyramidal decompositions.

In another scenario, we targeted compression of facial images as an example of domain-specific scenarios where there is significant structured redundancy in the data. We simply applied our RRQ algorithm to the facial images and concluded that it is possible to benefit from such extra redundancy in the images, as we outperformed the JPEG2000, and even without entropy coding. 

Finally, we took this trained network, but went one step further from compression and considered reconstructing noisy images with this network that was trained on clean facial images. We noticed that these noisy images get denoised as they are reconstructed. Surprisingly enough, this technique, for highly noisy regimes can outperform the famous BM3D, which is not trained on a collection of clean images. This leaves us with a very promising direction for future research, some of which we outline next in chapter \ref{chapter:CompPrior}.


\chapter{Future works: beyond compression} \label{chapter:CompPrior}
Suppose we have trained a compression network on a set of images. As we have argued earlier in Eq. \ref{ImCompression:GeneralObj}, this means that, in an indirect way, i.e., through the latent compact codes, we are maximizing the log-probability of the data.

Can this density estimation be useful to perform other tasks than compression itself? While we leave the rigorous treatment of this idea as an interesting future direction, here we demonstrate some of our preliminary results. 

Among the many possibilities for the realization of this idea, next in section \ref{secCompPrior_InverseProblems}, we consider compressibility as an efficient prior to solve inverse problems. We experiment with the tasks of image denoising and compressive sensing and demonstrate some promising results.


\section{Inverse problems} \label{secCompPrior_InverseProblems}
Consider the general inverse problem of recovering $\mathbf{f}$ from the degraded observation $\mathbf{q} = \mathrm{T}\mathbf{f} + \mathbf{p}$, where $\mathrm{T} \in \Re^{l \times n}$ models the acquisition process, e.g., sampling or blurring, and $\mathbf{p}$ is an additive white Gaussian noise as $\mathbf{P} \sim \mathcal{N}(\mathbf{0}, \sigma_p^2 \mathrm{I}_l)$. 

What happens if we try to inject compressibility as a prior into the Maximum A Posteriori formulation of Eq. \ref{eq:ModelingLit_MAP}? In other words, what happens if, instead of common priors like sparsity or smoothness, we penalize or encourage solutions based on the quality of their compression under some model?

In very general terms, this idea has been formulated already. However, it has not received much attention in the literature, perhaps since the discrete nature of compression is, in general, an obstacle for optimization.

Within the existing works, similar to this ideas is to perform the model selection with minimum description length principle and has appeared perhaps first in \cite{rissanen1978modeling}. In \cite{482110}, a data denoiser is proposed that feeds the noisy data to a compressor with a distortion equal to noise variance and takes the output as a denoised version. The work of \cite{Moulin:857789} analyzes the problem from an information-theoretic perspective linking the recovery problem as rate-distortion optimization. In \cite{Moulin:923281}, an image denoising scheme is proposed, where the compressor is considered as a wavelet-thresholding operator. The more recent work of \cite{4840520} revisits this old idea by modeling the wavelet coefficients of noiseless data. 

These works, however, focus only on particular and in fact very basic compression schemes to be suitable for joint optimization with the recovery/enhancement task. Moreover, none of these works focus on learning-based scenarios.

Unlike these works, here we propose to consider compression as a black-box operation and decouple compression and image enhancement from each other. Assuming that a good compressor already exists, e.g., it has been trained on a set of clean data, we consider the compression encoder-decoder pair as a black-box function that can be evaluated as many times as we need. We pursue this idea in an iterative algorithm that we introduce next. 

\subsection{An iterative algorithm} \label{subsec:CompPrior_InverseProblems_iterative}
For an encoder-decoder pair $\mathbb{Q}[\cdot]$-$\mathbb{Q}^{-1}[\cdot]$ operating at rate $R$, let us denote for simplicity $h_R(\mathbf{f}) = \mathbb{Q}^{-1}\big[\mathbb{Q}[\mathbf{f}]\big]$, as a function that compresses and then decompresses its input. Suppose that this pair is already trained on a set $\mathrm{F} = [\mathbf{f}_1, \cdots, \mathbf{f}_N]$ using, e.g., the RRQ of section \ref{sec:MultiLayer_RRQ}, the ML-STC of section \ref{sec:MultiLayer_MLSTC}, or is otherwise any compression framework like the JPEG or JPEG2000 for natural images.

For the task of solving inverse problems, we are not particularly interested in knowing the internal structure of the encoder-decoder, or this is entirely unknown to us, as is possible for many scenarios. However, since the encoder-decoder is provided, we can run it as many times as we want.\footnote{The idea of optimization involving  black-box functions is somehow similar to the recent work of ``image rendition'' in \cite{rendition_milanfar}, where the process of image degradation is assumed to be unknown.} In other words, $h_R(\mathbf{f})$ is a black-box function that we can evaluate multiple times, and we do not know anything about its internal structure, except for one important property: Since $h_R(\cdot)$ involves a quantizer function, e.g., the ternary operator or the sign function, its derivative w.r.t. $\mathbf{f}$ is either undefined at the discontinuities or equals zero. In other words, $h_R(\mathbf{f})' = \mathbf{0}$. 

How do we recover $\mathbf{f}$ from an observation $\mathbf{q}$? We formulate our MAP-like estimation as:
\begin{equation}  \label{eq:CompPrior_MAP}
\hat{\mathbf{f}} = \underset{{\mathbf{f}}}{\text{argmin }}\frac{1}{2}|| \mathrm{T} \mathbf{f} - \mathbf{q}||_2^2  + \frac{\mu}{2}|| \mathbf{f} - h_R(\mathbf{f}) ||_2^2,
\end{equation}
where we added the compressibility prior with a regularization constant $\mu$, such that to penalize large values of $\mathbf{f} - h_R(\mathbf{f})$.  

Since $h_R(\cdot)$ is unknown, Eq. \ref{eq:CompPrior_MAP} cannot be solved in closed-form. However, since all elements of its objective function are differentiable and particularly the fact that $h_R(\mathbf{f})' = \mathbf{0}$, we can use first-order routines to solve it.\footnote{Note that while the zero gradient of the compression objective due to the non-smooth quantization function was a prime challenge in training networks for compression, in this formulation, however, it is in our favor since it drastically simplifies the solution of the MAP estimation.} 

In particular, a simple gradient descent leads to the following iterative solution of Eq. \ref{eq:CompPrior_MAP} as:
\begin{equation}  \label{eq:CompPrior_GD}
\mathbf{f}^{(t+1)} = \mathbf{f}^{(t)} - \tau \Bigg[  \mathrm{T}^T \big[ \mathrm{T} \mathbf{f}^{(t)} - \mathbf{q} \big] + \mu \big[ \mathbf{f}^{(t)} - h_R(\mathbf{f}^{(t)}) \big]  \Bigg],
\end{equation}
where $\tau$ is a step size which we keep as constant for simplicity.

Let us next see how the iterative algorithm of Eq. \ref{eq:CompPrior_GD} can be useful for solving inverse problems. We provide two examples from image denoising, and noisy compressive sampling on AR(1) sources.
\subsection{Image denoising using JPEG2000} \label{subsecCompPrior_InverseProblems_denoising}
Instead of learned image compression networks, for the moment, suppose that we suffice to an off-the-shelf image codec for solving inverse problems. 

In particular, let us see if we can denoise images using the simple JPEG2000. The idea is then to use the iterative algorithm of Eq. \ref{eq:CompPrior_GD} for a simple denoising, i.e., we let $\mathrm{T} = \mathrm{I}_n$, and we consider the black-box function $h_R(\cdot)$ as the compression followed by de-compression using JPEG2000, for an appropriate rate $R$. So let us take an image, say the image ``\textit{man}'', contaminate it with additive noise and see if we can denoise.

This, however, may not be a sufficient regularization since JPEG2000 is not particularly trained to compress only clean images. In fact, the only compressibility prior used in JPEG2000 is the sparsity under wavelet decomposition, which is very basic. Note that the very highly engineered pipeline of JPEG2000 does not help with regularization since, e.g., the entropy coding or bit-plane coding stages do not reduce the actual entropy of representation, but only reduce bit-length of the file. In other words, the effective space size $|\mathcal{F}^n| \simeq 2^{h_R(\mathbb{Q}[\mathbf{F}])}$ provided by JPEG2000 is only bounded by wavelet thresholding. However, it should be noted that the discretization is key, as simple non-quantized wavelet thresholding will not work under our iterative solution. 

Therefore, we should not expect powerful denoising only using JPEG2000. As a simple additional regularization for this toy experiment, along with the compressibility prior of Eq. \ref{eq:CompPrior_MAP}, we use the differentiable prior $\frac{\mu'}{2}|| \nabla \mathsf{I}||_2^2$, where $\nabla \mathsf{I}$ is the first-order derivative of the image pixels along the $x$ and $y$ axes (Sobolev prior). The optimization iterations on the image $\mathsf{I}$ are then followed as:
\begin{equation*}  
\mathsf{I}^{(t+1)} = \mathsf{I}^{(t)} - \tau \Bigg[ \mathsf{I}^{(t)} - \mathsf{I}^{(0)}  + \mu \big[ \mathsf{I}^{(t)} - h_R(\mathsf{I}^{(t)}) \big] + \mu' \nabla^2 \mathsf{I}^{(t)} \Bigg],
\end{equation*}
where the algorithm is initialized with $\mathsf{I}^{(0)}$ as the noisy image.

\begin{figure}   [!h]
   \begin{center} 
\includegraphics[width=1\textwidth]{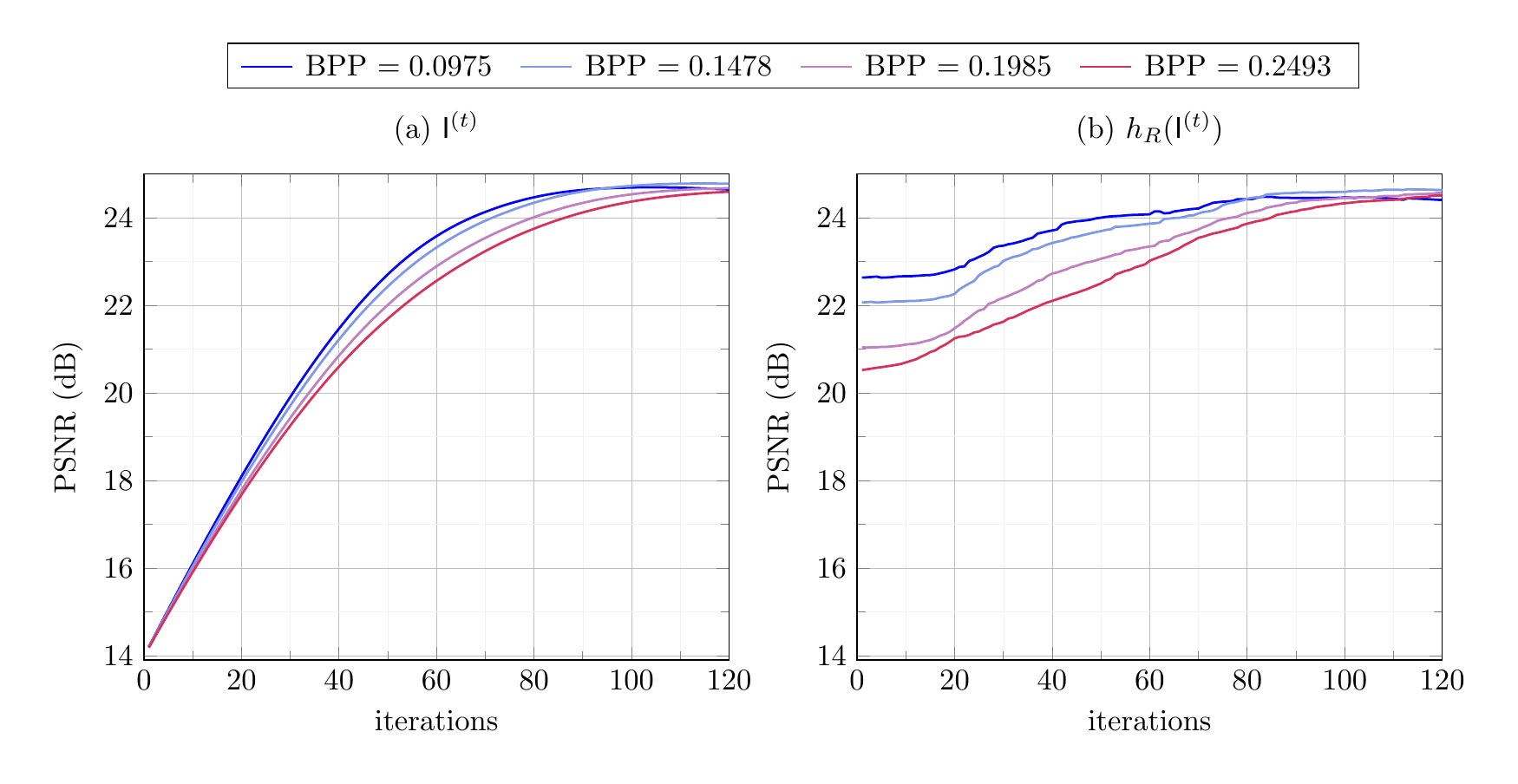}   
   \end{center}
\vspace{-0.5cm}    
   \caption{PSNR vs. iterations for the image denoising of Eq. \ref{eq:CompPrior_GD} (initialized with the noisy image).}
   \label{fig:CompPrior_Denoising1_iterations}
   \end{figure}

Fig \ref{fig:CompPrior_Denoising1_iterations} shows the evolution of PSNR during optimization iterations in denoising of the image ``\textit{man}'' contaminated with additive noise of $\sigma_P^2 = 0.2$. The results are shown both for $\mathsf{I}^{(t)}$ and $h_R(\mathsf{I}^{(t)})$, and for different rates of JPEG2000 compression.

Surprisingly, this very simple denoising algorithm seems to be quite effective as the PSNR is increasing during iterations. Note that this algorithm is very different from simply applying JPEG2000 on the noisy image and hoping for effective denoising, as this would simply correspond to $h_R(\mathsf{I}^{(0)})$, which is not very satisfactory.  

\begin{figure}  [h]
   \begin{center} 
\subcaptionbox{Original} {\includegraphics[width=0.3\textwidth]{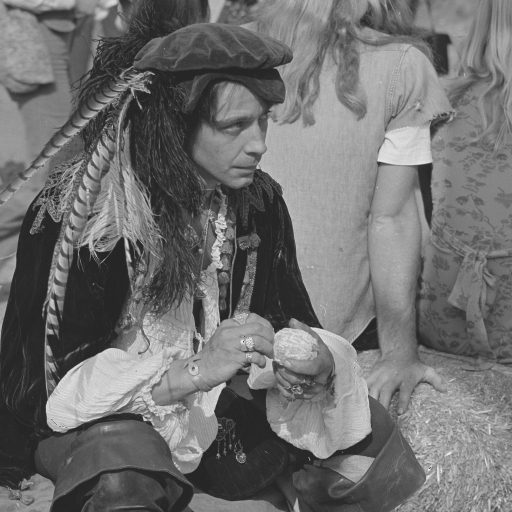}} 
\subcaptionbox{Noisy\\$\text{PSNR}=13.98 \text{ dB}$} {\includegraphics[width=0.3\textwidth]{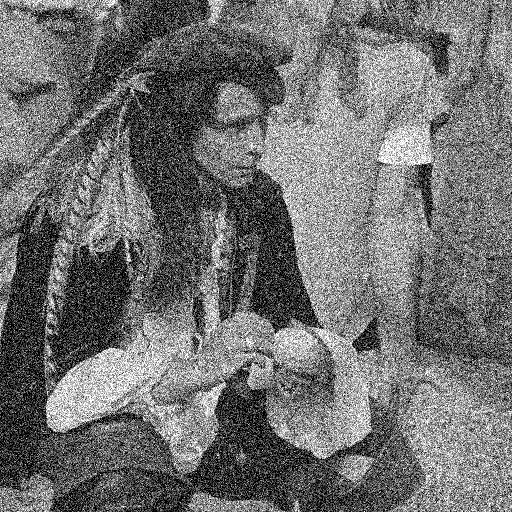}}
 
\subcaptionbox{Denoised (JPEG2000 prior)\\$\text{PSNR}=23.15 \text{ dB}$} {\includegraphics[width=0.3\textwidth]{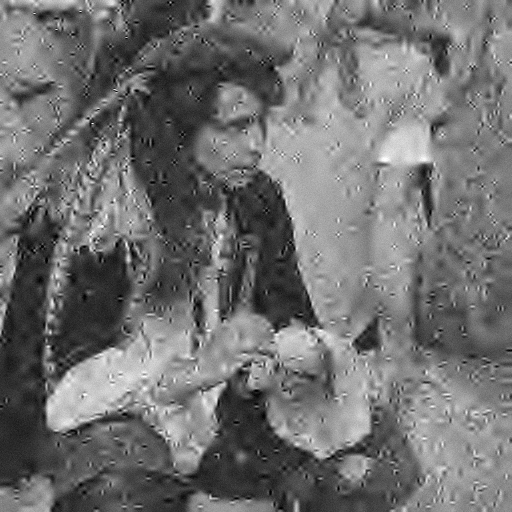}} 
\subcaptionbox{Denoised (Sobelov prior)\\$\text{PSNR}=22.09 \text{ dB}$} {\includegraphics[width=0.3\textwidth]{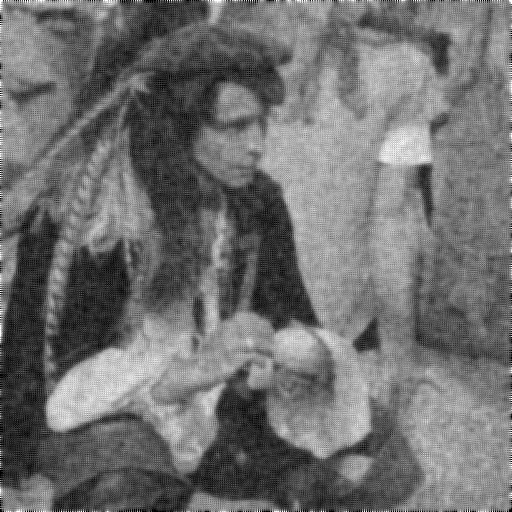}}
\subcaptionbox{Denoised (both priors)\\$\text{PSNR}=24.78 \text{ dB}$} {\includegraphics[width=0.3\textwidth]{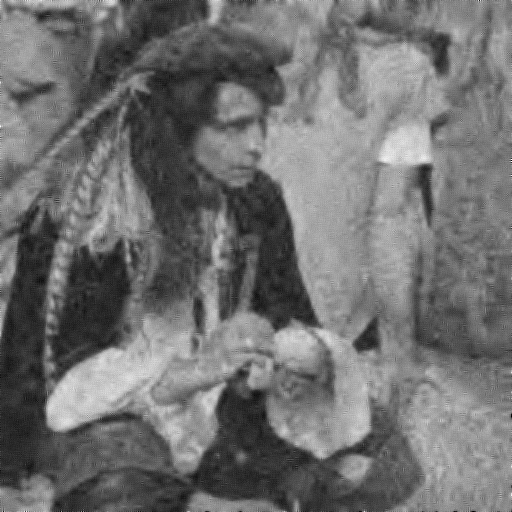}}  

\end{center}
\vspace{-0.5cm}    
   \caption{Image denoising with Eq. \ref{eq:CompPrior_GD} when initialized with the noisy image ($\sigma_P^2=0.2$). (a) Original image, (b) Noisy image ($\sigma_P^2 = 0.2$), (c) Only the JPEG2000 prior used ($\mu' = 0$), (d) Only the Sobolev prior used ($\mu = 0$, (e) Both priors used together.}
   \label{fig:CompPrior_Denoising1_samples}
   \end{figure}

To determine the effect of each regularization, we show the results for the JPEG2000 prior alone ($\mu =100, \mu' = 0$), the Sobolev prior alone ($\mu = 0, \mu' = 300$), as well as their joint application ($\mu = 5, \mu'= 150$). Fig. \ref{fig:CompPrior_Denoising1_samples} shows the images denoised with these priors.

As was expected, however, since JPEG2000 does not have strong priors about the set of clean images, the performance is limited. In fact, BM3D denoises this image with PSNR$= 26.73\text{ dB}$, which is around 2 dB higher than our result. 

\begin{figure}  [h]
   \begin{center} 
\subcaptionbox{Noisy\\$\text{PSNR}=3.10 \text{ dB}$} {\includegraphics[width=0.3\textwidth]{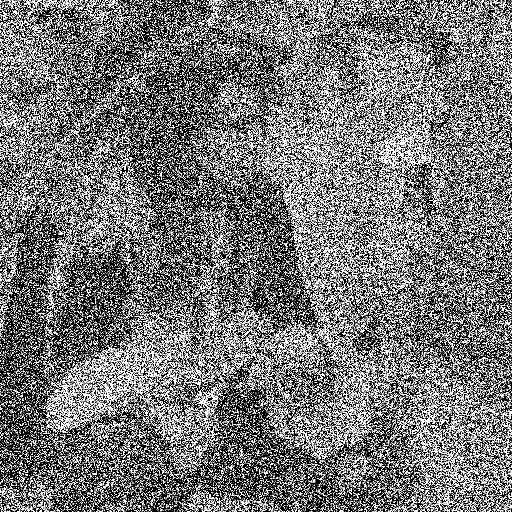}} 
\subcaptionbox{Denoised (BM3D)\\$\text{PSNR}=20.25 \text{ dB}$} {\includegraphics[width=0.3\textwidth]{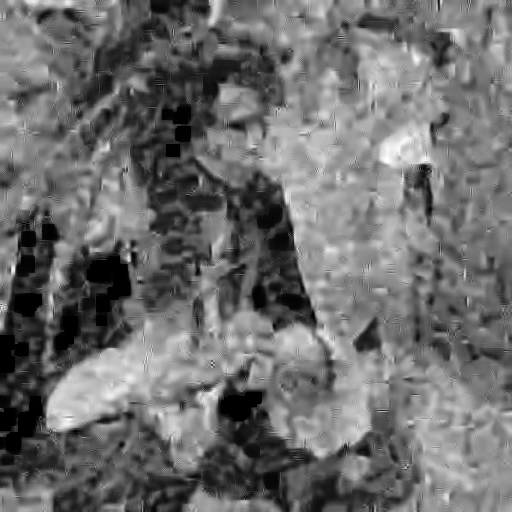}}
 \subcaptionbox{Denoised (proposed)\\$\text{PSNR}=21.35 \text{ dB}$} {\includegraphics[width=0.3\textwidth]{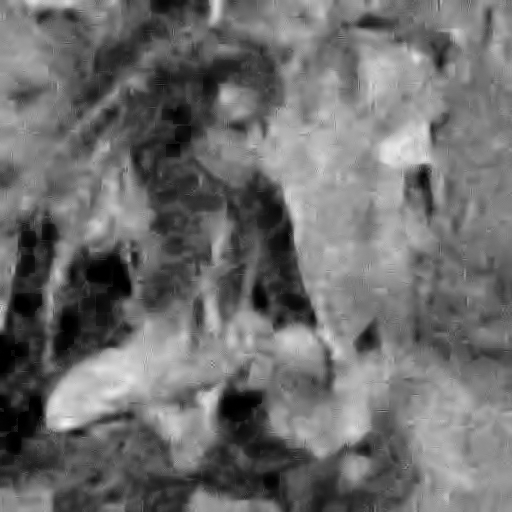}} 
\end{center}
\vspace{-0.5cm}    
   \caption{Image denoising in highly noisy regimes ($\sigma_P^2 = 0.7$). (a) Noisy image, (b) Denoised with BM3D, (c) Denoised with Eq. \ref{eq:CompPrior_GD} when initialized with BM3D. }
   \label{fig:CompPrior_Denoising2_samples}
   \end{figure}

But what if instead of the noisy image, we initialize our algorithm with the result of a more intricate denoiser like BM3D? It turns out that while this is not very effective for low-noise regimes and under JPEG2000 (since the denoised images are already smooth), it can be a very effective denoiser for highly noisy regimes, as can be seen from Fig. \ref{fig:CompPrior_Denoising2_samples}, where the algorithm improves on BM3D about $1.1 \text{ dB}$. Note that a lot of the artifacts of BM3D have been removed.

In fact, the idea of our iterative solution can be extended far beyond the JPEG2000 as the core compressor and for a large variety of inverse problems. Particularly, we hypothesize that the use of learned networks for compression can be very promising for our algorithm in Eq. \ref{eq:CompPrior_GD}.

While we leave this for future research, we next provide another motivational example: that of noisy compressive sensing.
\subsection{Compressive sensing of auto-regressive sources} \label{subsecCompPrior_InverseProblems_CS}
We now consider the problem of noisy compressive sensing as another instance of inverse problems that can be addressed with the idea of ``compressibility as a prior''. Let us now take the case of auto-regressive sources and in particular AR(1) sources.

So suppose we have trained a network on such data. As an example, let us pick our ML-STC of section \ref{subsec:MultiLayer_MLSTC_Alg} and train it on AR(1) source, as we did in Fig. \ref{fig:MultiLayer_MLSTC_DR}. 

Now that this network is trained, our objective is to recover the under-sampled and noisy measurements of the instances from the test set, i.e., to recover $\mathbf{f}$ from the unders-sampled and noisy $ \mathbf{q} = \mathrm{T} \mathbf{f} + \mathbf{p}$, where $\mathrm{T} \in \Re^{l \times n}$ is a fat matrix (i.e., $l < n$) with Gaussian random values and $\mathbf{p}$ is white Gaussian noise with variance $\sigma_P^2$.

We apply the iterative algorithm of Eq. \ref{eq:CompPrior_GD} to recover the data. This algorithm is initialized with $\mathbf{f}^{(0)} = (\mathrm{T}^T\mathrm{T})^{-1}\mathrm{T}^T \mathbf{q}$, i.e., the Moore-Penrose pseudo-inverse of $\mathrm{T}$ to back-project $\mathbf{q}$ to $\mathbf{f}$. Note that this is the optimal reconstruction without regularization, i.e., for $\mu = 0$. 

\begin{figure}  [!h]
   \begin{center} 
\includegraphics[width=0.8\textwidth]{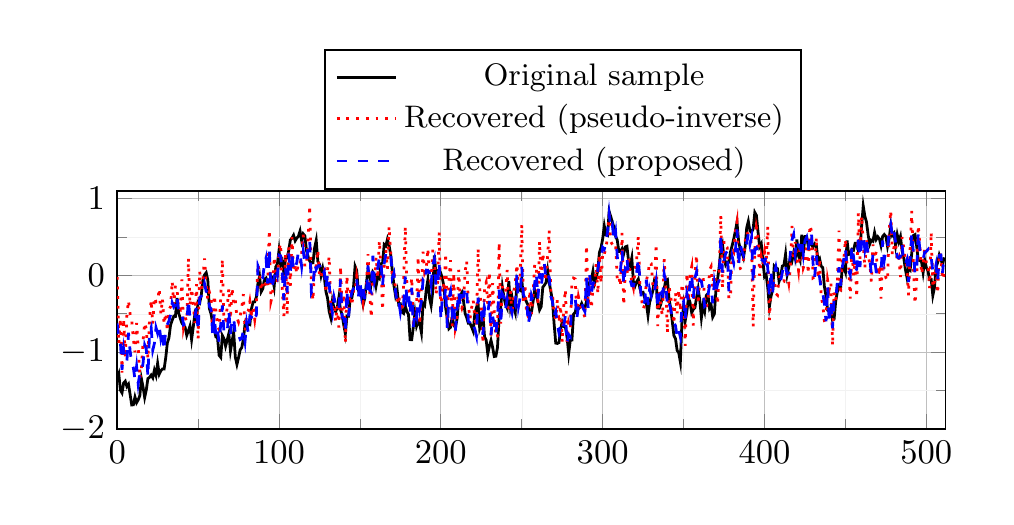} 
\end{center}
\vspace{-0.5cm}    
   \caption{Noisy compressive sensing of AR(1) data with $n=512$ with the iterative algorithm of Eq. \ref{eq:CompPrior_GD}. The compressability prior is imposed using the ML-STC and with $\mu = 100$. The observation is sampled with $l=256$ and contaminated with noise of variance $\sigma_P^2 = 1$.}
   \label{fig:CompPrior_InverseProblems_CS_samples}
   \end{figure}

Fig. \ref{fig:CompPrior_InverseProblems_CS_samples} shows a sample of $\mathbf{f}$ from the test set, as well as its recovery using pseudo-inversion and our Eq. \ref{eq:CompPrior_GD}. It is clear that our regularization produces AR(1) solutions while the solution returned by pseudo-inversion does not have such structures.

\begin{figure}  [!h]
   \begin{center} 
\includegraphics[width=0.8\textwidth]{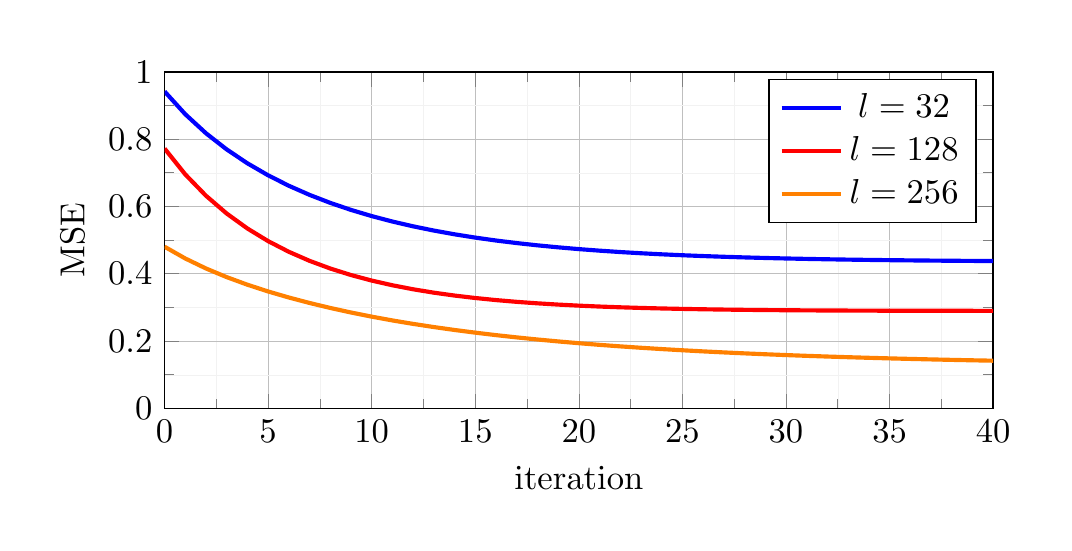}
\end{center}
\vspace{-0.5cm}    
   \caption{Noisy compressive sensing of AR(1) data with the iterative algorithm of Eq. \ref{eq:CompPrior_GD}. The compressability prior is imposed using the ML-STC and with $\mu = 100$.}
   \label{fig:CompPrior_InverseProblems_CS_results}
   \end{figure}

Fig. \ref{fig:CompPrior_InverseProblems_CS_results} shows the evolution of MSE during iterations for 3 different sub-samplings in $l=32,128,256$, 
and with noise variance $\sigma_P^2 = 1$, while the data dimension was $n=512$, and the correlation factor for the source was $\rho=0.99$.

This simple idea seems to be very efficient, and the reconstruction error is decreasing for all values of $l$. We do not provide any convergence guarantee, but we notice that when initialized other than with the pseudo-inverse, e.g., with $\mathbf{f}^{(0)} =  \mathrm{T}^T \mathbf{q}$, the convergence takes several more iterations, but it always converges. This, however, depends on $\mu$, and the step-size $\tau$ and the operating rate of the compressor.

\section{Conclusions} \label{sec:CompPrior_Conclusions}
This chapter presented some of our ongoing works, as well as some promising directions for future research based on learned compression.

Among various possibilities, we focused on one instance application that learning to compress can be useful beyond the task of compression itself. In particular, we investigated compressibility as a prior to solve inverse problems and imposed it as a regularization to the data-fidelity term. This was achieved using a black-box interpretation of the compression-decompression network, which led to a simple gradient descent on the objective.

We saw two variants of this problem. First, we considered image denoising, where the JPEG2000 codec was used to apply the compressibility prior. We saw that this can be useful to denoise natural images in highly noisy regimes. In particular, we improved on BM3D to more than 1 decibel of PSNR in these regimes.

The second variant of inverse problems that we investigated was the noisy compressive sensing. We experimented with auto-regressive sources, where we trained an ML-STC on a set of clean AR(1) examples. This was used in our iterative algorithm to recover under-sampled and noisy measurements of such data. We saw that it is possible to significantly improve upon pseudo-inversion using the proposed algorithm.

These algorithms can be improved in many ways. For image denoising, instead of JPEG2000, emphasis should be put on learned compression schemes, where training is done on clean images. This can be extended to other inverse tasks, e.g., image super-resolution, image inpainting or deblurring.

Our results on compressive sensing were promising but preliminary. We did not provide any recovery or convergence guarantees. We leave this, and a lot of other interesting possibilities for future investigation.

\setcounter{secnumdepth}{-1}
\chapter{Concluding remarks} 
This thesis studies several aspects of data and signal processing, addresses some issues and proposes several solutions for them. The focus was on vectorial data and in particular images and image descriptors and the considerations and issues when dealing with them in large-scale scenarios. The similarity search and compression aspects were highlighted, where efficient and useful representations for data are to be learned from its different examples.

Our central consideration when modeling the data and learning the representations was compactness. This led us to carefully study and optimize the rate-distortion behavior of the proposed data models. We then argued that this optimization is useful beyond compression itself and can help a couple of other tasks in signal and image processing.

In the first part of the thesis, we started with reviewing a diverse range of methods and ideas from the literature of signal processing and machine learning. We interpreted all attempts under the general framework of Bayesian modeling and categorized them under two rough divisions of basic and composite models. We discussed their properties and benefits and their behavior under different sample regimes. This helped us in setting up our main strategy in design and development of the models as to start with basic ones and under simple assumption-based setups, where data-distribution is assumed and then gradually lift the assumptions and rely more on the data samples, as the models evolve from basic to composite ones. 


This was pursued in the second part of the thesis, where we developed the algorithmic infrastructure based on the two families of basic prior models, i.e., synthesis and analysis models. The development of synthesis models first led us to the VR-Kmeans algorithm, which regularizes the standard K-means by imposing a rate-allocation criterion in learning the codebooks. We showed that this is very helpful in avoiding over-fitting, particularly in high-dimensional settings.

The criterion of having discrete representations led us to the development of the analysis model as the Sparse Ternary Codes (STC) framework, which is inspired by $\ell_0$ regularization of the data in the projected domain. The information-theoretic properties of this encoding were studied, and several possibilities for reconstruction of these codes were investigated. 

Studying the rate-distortion properties of these two frameworks led us to conclude that the basic analysis or synthesis modeling is not enough. In particular, we noticed that we cannot operate at high rates and hence high-fidelity representations while having fast encoding-decoding procedures based on these two basic models. We then concluded that composite solutions based on these basic models should be preferred.

This transition from basic to composite modeling was based on the framework of successive refinement from information theory. In particular, we chose the additive residual based encoding to enhance our basic models. The VR-Kmeans was developed to the RRQ framework which maintains a very good rate-distortion performance at arbitrary rates and for any number of layers, without getting over-fitted.

Based on the two practical decoding schemes proposed for reconstruction of STC, its multi-layer extension was the ML-STC and the ML-STC-Procrustean. The former being more assumption-based, the latter model relies more on data samples and learns multiple layers of optimal transformations from the data using a Procrustean approach.

These composite models, however, are trained layer-by-layer and without assuming the anti-causal errors between layers. For the analysis-based models, i.e., the ML-STC and the ML-STC-Procrustean, another prominent evolution is still possible. These models can be considered as neural network structures and can hence benefit from the back-propagation technique to jointly optimize all layers. This was hindered by a technial issue regarding the non-differentiability of discretizing ternary operator. Thanks to the favorable information preservation properties of ternary encoding, however,the issue was solved using a smooth surrogate without causing approximation errors.

This made the evolution of our models complete, which we termed as STNets, a neural structure with discrete representations that is pre-trained layer-by-layer and using our sample-efficient and theoretically-sound recipes and is then fine-tuned using the back-propagation technique that benefits from the practical insights and infrastructures available in deep-learning communities. We then have a range of possibilities for training to choose from, w.r.t. sample-size and computational budget available. 

The third part of the thesis considers the applications of these algorithms in three directions. First, the problem of fast similarity search was addressed where we categorized the efforts in the literature under two families of binary hashing that benefits from fast search in the space of codes but degrades search quality and the VQ-based methods that have high search performance but do not benefit from fast search within the codes. We proposed that a third approach is possible using our ML-STC model that performs a double-stage procedure to search. The first stage is performing a very efficient search in the space of ternary codes, which we showed that possess superior coding gain w.r.t. the binary codes. While this stage prunes the majority of database candidates, the second stage refines the list of candidates by reconstructing the candidate codes and performing a floating-point search on them. This is also very efficient thanks to the rate-distortion behavior that we have optimized. We performed million-scale experiments on public databases showing superior performance.

The second application we addressed was the learned compression of images. We first argued that it can be beneficial under many scenarios to give up on the traditional image compression codecs and develop data-adaptive solutions instead. Two cases for learned image compression were considered: We first experimented with high-resolution and color natural images and showed that even when skipping several procedures in the compression pipeline, we can achieve results comparable to the JPEG2000 codec when training is done on 200 such images using the ML-STC-Procrustean. We then argued the necessity of avoiding block-wise division for images by measuring the loss in coding efficiency due to breaking the global spatial structure of the images into patches.

We then experimented with domain-specific images where a lot of redundancy is shared between the images of interest. In particular, we showed that we can surpass JPEG2000 in compression of facial images using a simple RRQ and without performing entropy coding. As a further step beyond compression, we tested whether this trained model can be beneficial for other tasks and in particular image denoising. Interestingly enough, we showed that by merely compressing and decompressing noisy facial images, we can outperform the BM3D in denoising of very highly noisy images, since the network has seen samples of clean images and can inject priors about face structure during reconstruction.

This led us to investigate more in this direction. So the third application of the thesis was to consider the usefulness of compression beyond itself. We presented several of our ongoing research and showed that it is possible to regularize inverse problems with compressibility as a prior, i.e., to encourage solutions that are better compressible under the (trained) model. This was achieved using an iterative algorithm that considers the compression network as a black-box that can be evaluated multiple times during iterations. In particular, we showed that highly noisy images denoised by BM3D can be improved up to more than one decibel in PSNR using our proposed algorithm when the compression engine is the JPEG2000. We then targetted the noisy compressive sampling of auto-regressive sources and showed that we can significantly improve upon pseudo-inversion.

\subsubsection*{Future works} 

The thesis leaves a lot of directions for further research. Instead of the ML-STC that we used for simplicity, the STNets framework can further enhance the rate-distortion behavior by benefitting from more training examples. Moreover, instead of using image descriptors, the STNets framework can very well be used to train useful features directly from raw pixels.

The image compression pipeline proposed can be improved in many ways. We did not implement several important steps like entropy coding. As was concluded, patch-based encoding of images should be replaced with pyramidal decompositions. This way, the loss of global content of images, as well as the problem with variable size images will be resolved. Our proposed whitening procedure can also be improved in many ways. For example, the global 2D-DCT can be followed by local processing in order to make the independence assumption of sub-bands more realistic. Finally, the training can benefit from larger samples.

Our presented results for solving inverse problems with compressibility prior were preliminary and ongoing. Many aspects should be considered, e.g., how much we can gain with learned compression instead of JPEG2000 for image denoising, convergence guarantees, momentum-based solvers instead of simple gradient descent, recovery guarantees for compressive sensing, and also studying inverse problems other than denoising and compressive sensing.

Beyond these aspects, we can imagine other directions that can benefit from the frameworks developed in this thesis. An important step is to model the sequential codes from different layers of the ML-STC using sequence modelers like the Recurrent Neural Networks (RNNs). This can bring a lot of possibilities, e.g, tasks like generative modeling or image classification can be addressed by adding a parallel RNN to the STNets and forming a composite cost function to be optimized jointly.



\begin{spacing}{0.9}


\bibliographystyle{unsrt} 
\cleardoublepage
\bibliography{references} 



\end{spacing}


\begin{appendices} 

\end{appendices}

\printthesisindex 

\end{document}